\documentclass[runningheads]{llncs}
\usepackage{graphicx}
\usepackage{comment}
\usepackage{amsmath,amssymb} %
\usepackage{color}

\usepackage{array}
\usepackage{booktabs}
\usepackage{xspace}
\usepackage{caption}
\usepackage[english]{babel}

\usepackage{hyperref}
\hypersetup{
    colorlinks=true,
    linkcolor=blue,
    filecolor=cyan,      
    urlcolor=magenta,
}

\makeatletter
\DeclareRobustCommand\onedot{\futurelet\@let@token\@onedot}
\def\@onedot{\ifx\@let@token.\else.\null\fi\xspace}

\def\eg{\emph{e.g}\onedot}

\def\etal{\emph{et al}\onedot}
\makeatother

\usepackage[width=122mm,left=12mm,paperwidth=146mm,height=193mm,top=12mm,paperheight=217mm]{geometry}

\newcommand{\ve}[1]{\ensuremath{\mathbf{#1}}} %
\newcommand{\ma}[1]{\ensuremath{\mathsf{#1}}} %
\newcommand{\set}[1]{\ensuremath{\mathcal{#1}}} %

\begin{document}
\pagestyle{headings}
\mainmatter
\def\ECCVSubNumber{4930}  %

\title{High Resolution Face Age Editing} %

\titlerunning{High Resolution Face Age Editing}
\author{Xu Yao\inst{1,2},  
Gilles Puy\inst{3}, 
Alasdair Newson\inst{1},  
Yann Gousseau\inst{1},  
Pierre Hellier \inst{2}
}
\authorrunning{X. Yao, G. Puy, A. Newson, Y. Gousseau and P. Hellier}
\institute{LTCI, T\'el\'ecom Paris, Institut polytechnique de Paris, France
\and InterDigital R\&I, 975 avenue des Champs Blancs, Cesson-S\'evign\'e, France 
\and Valeo.ai, 15 rue de la Baume, Paris, France}
\maketitle

\begin{abstract}
Face age editing has become a crucial task in film post-production, and is also becoming popular for general purpose photography. Recently, adversarial training has produced some of the most visually impressive results for image manipulation, including the face aging/de-aging task. In spite of considerable progress, current methods often present visual artifacts and can only deal with low-resolution images. In order to achieve aging/de-aging with the high quality and robustness necessary for wider use, these problems need to be addressed. This is the goal of the present work. We present an encoder-decoder architecture for face age editing. The core idea of our network is to create  both a latent space containing the face identity, and a feature modulation layer corresponding to the age of the individual. We then combine these two elements to produce an output image of the person with a desired target age. Our architecture is greatly simplified with respect to other approaches, and allows for continuous age editing on high resolution images in a single unified model. Source codes are available at \url{https://github.com/InterDigitalInc/HRFAE}. 

\keywords{High resolution, face aging, image-to-image translation.}
\end{abstract}

\section{Introduction}
\begin{figure*}
\begin{center}
\setlength{\tabcolsep}{2pt}
\begin{tabular}{ccccc}
25&35&45&55&65\\
\includegraphics[width=0.19\linewidth]{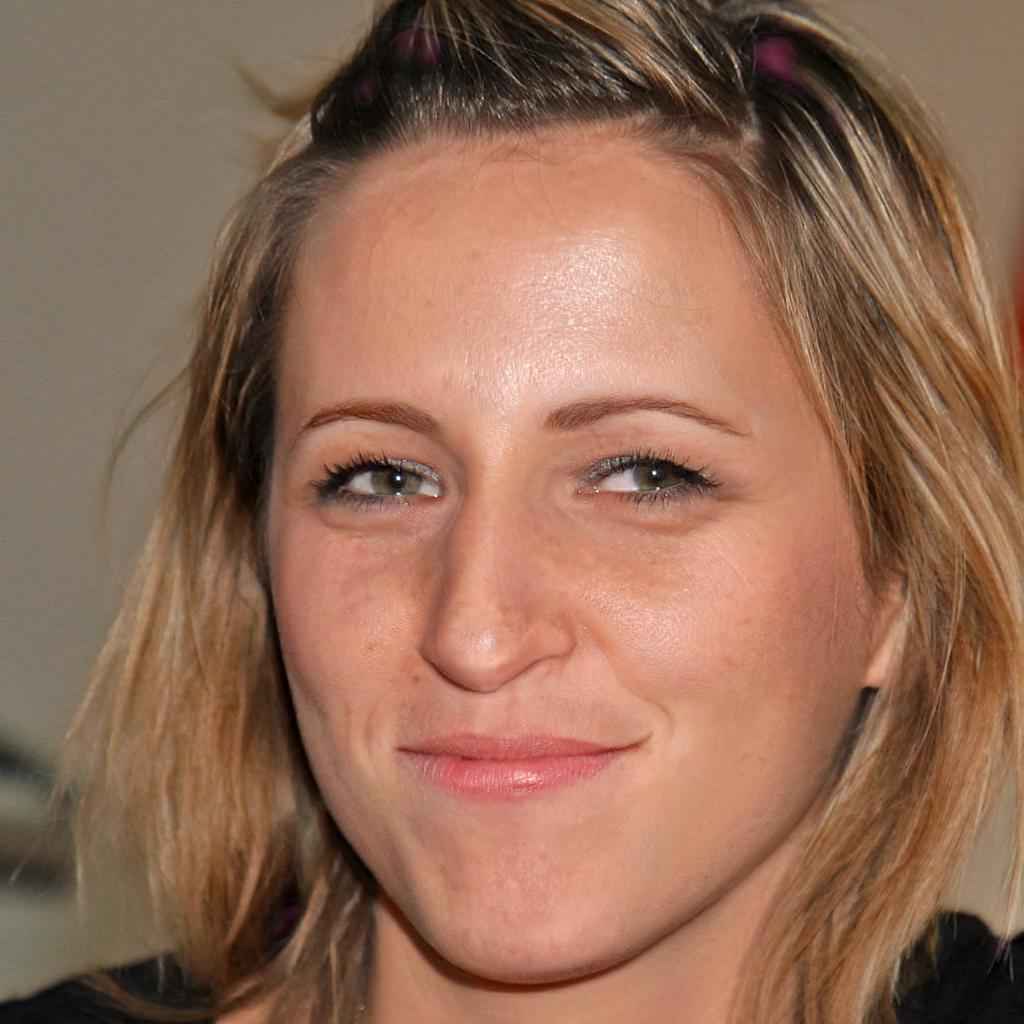}&
{\color{yellow}%
\setlength{\fboxsep}{0pt}%
\setlength{\fboxrule}{2pt}%
\fbox{\includegraphics[width=0.19\linewidth]{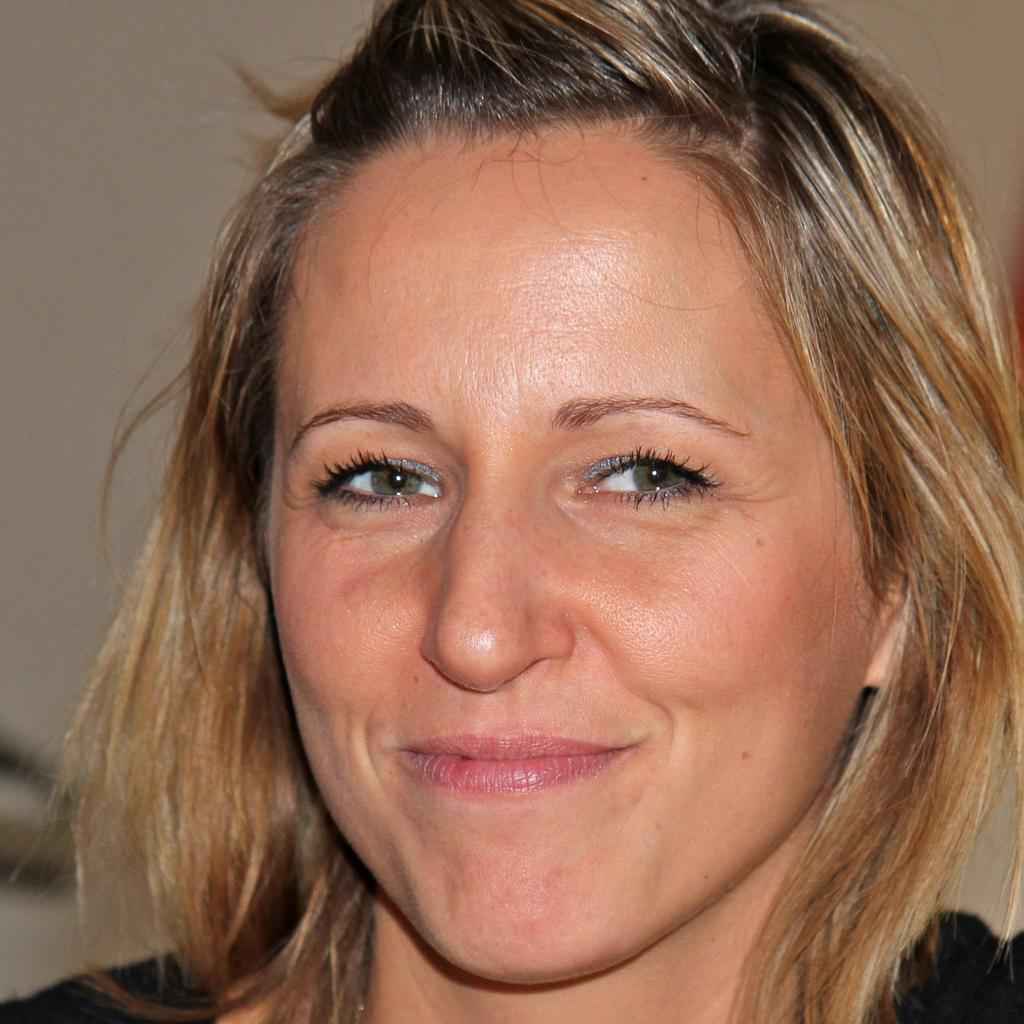}}}&
\includegraphics[width=0.19\linewidth]{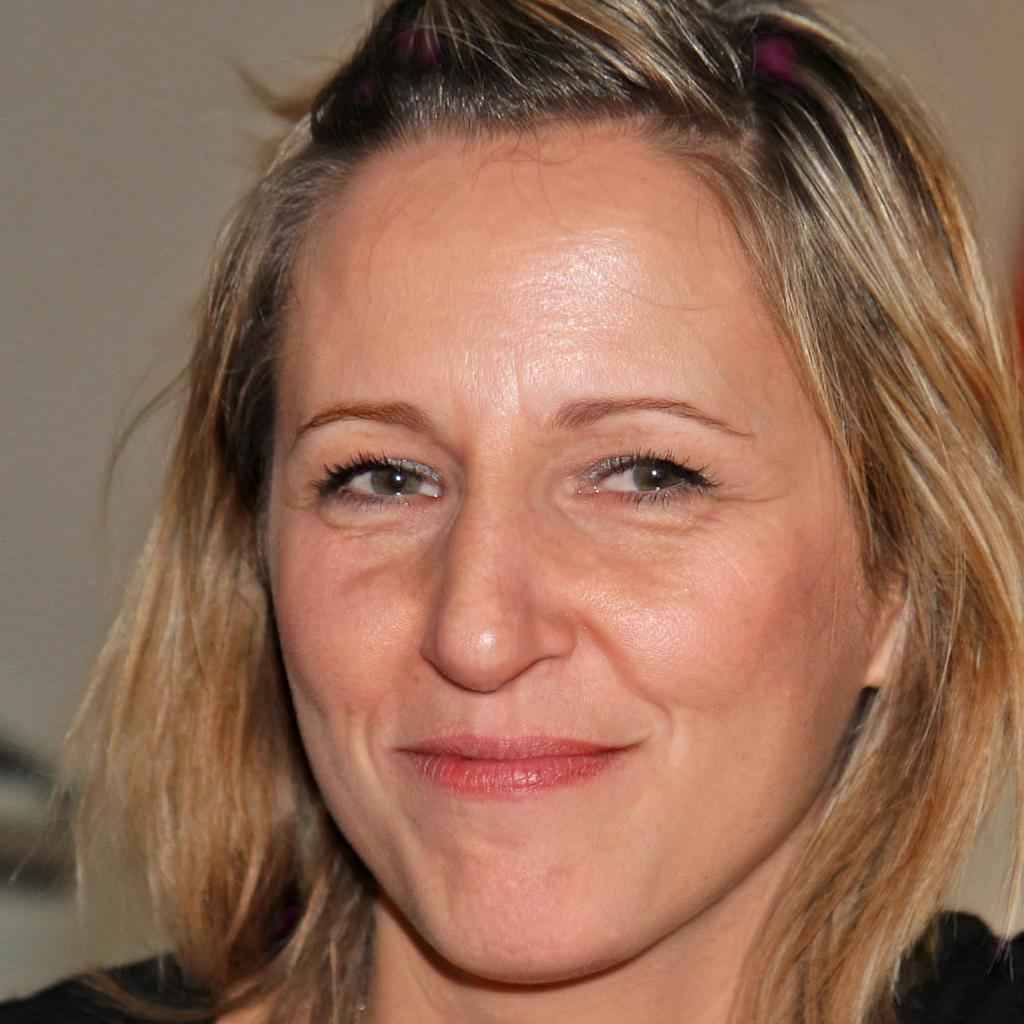}&
\includegraphics[width=0.19\linewidth]{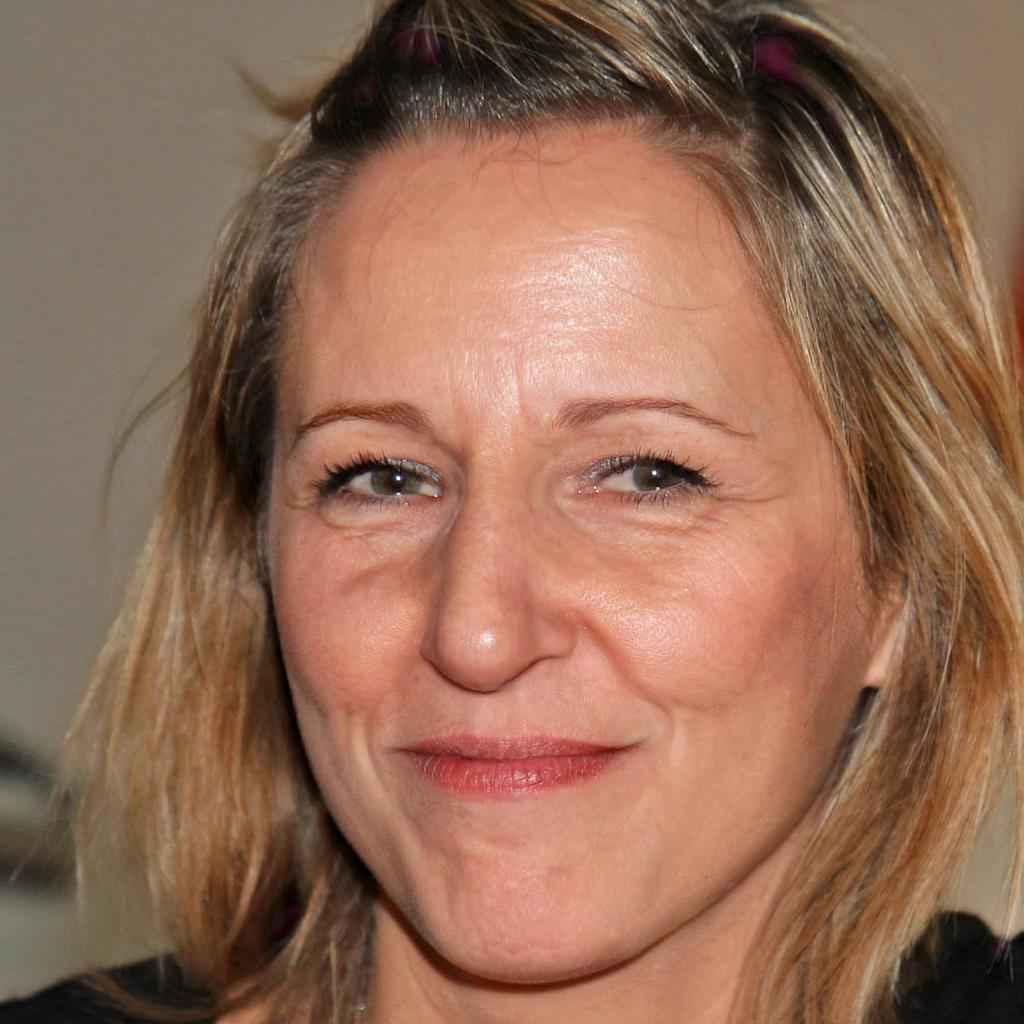}&
\includegraphics[width=0.19\linewidth]{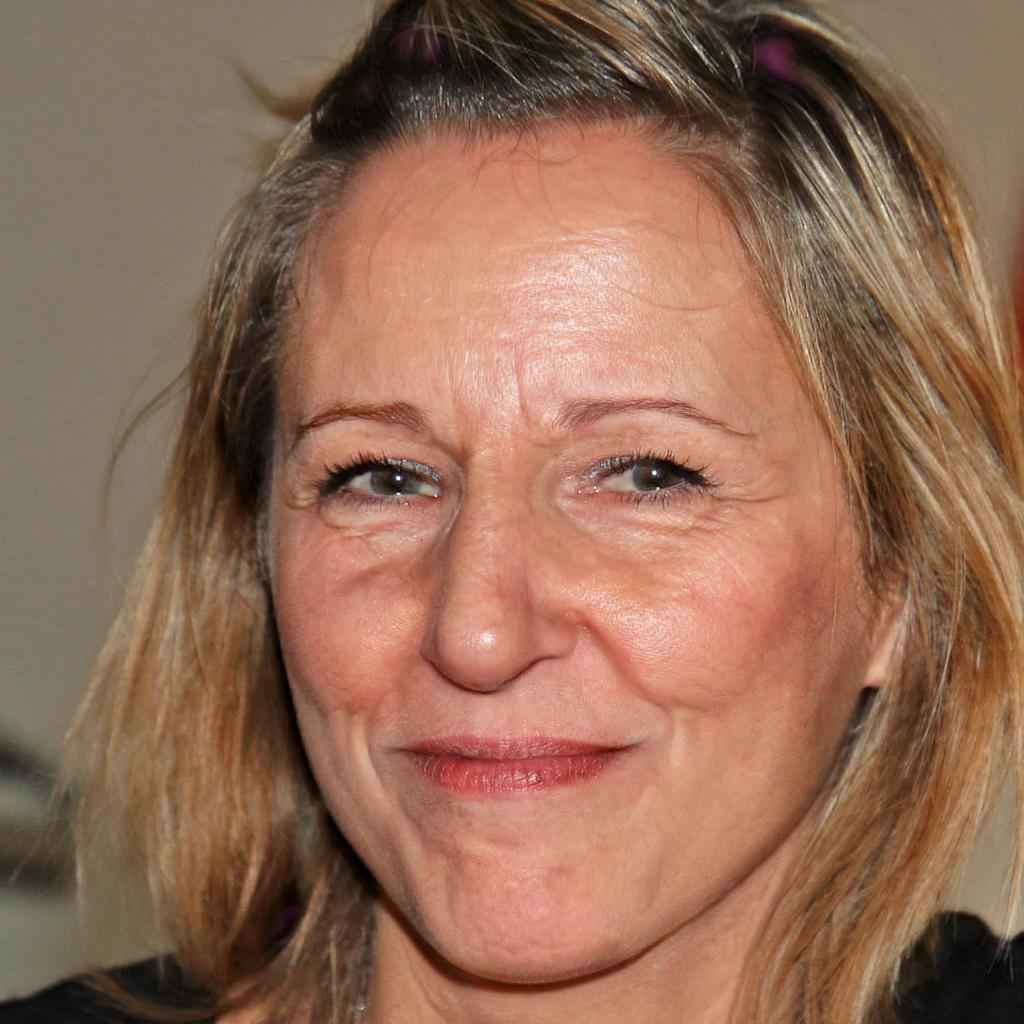}
\\
\includegraphics[width=0.19\linewidth]{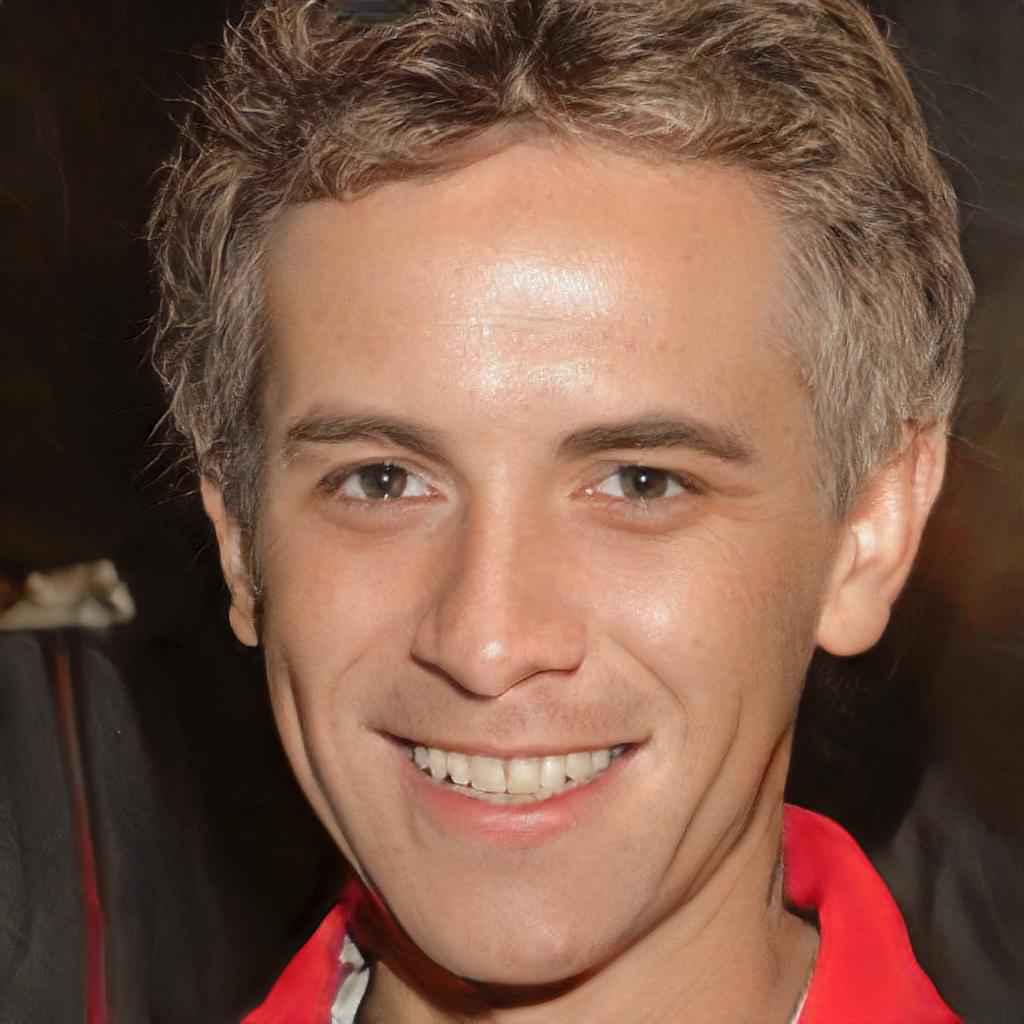}&
\includegraphics[width=0.19\linewidth]{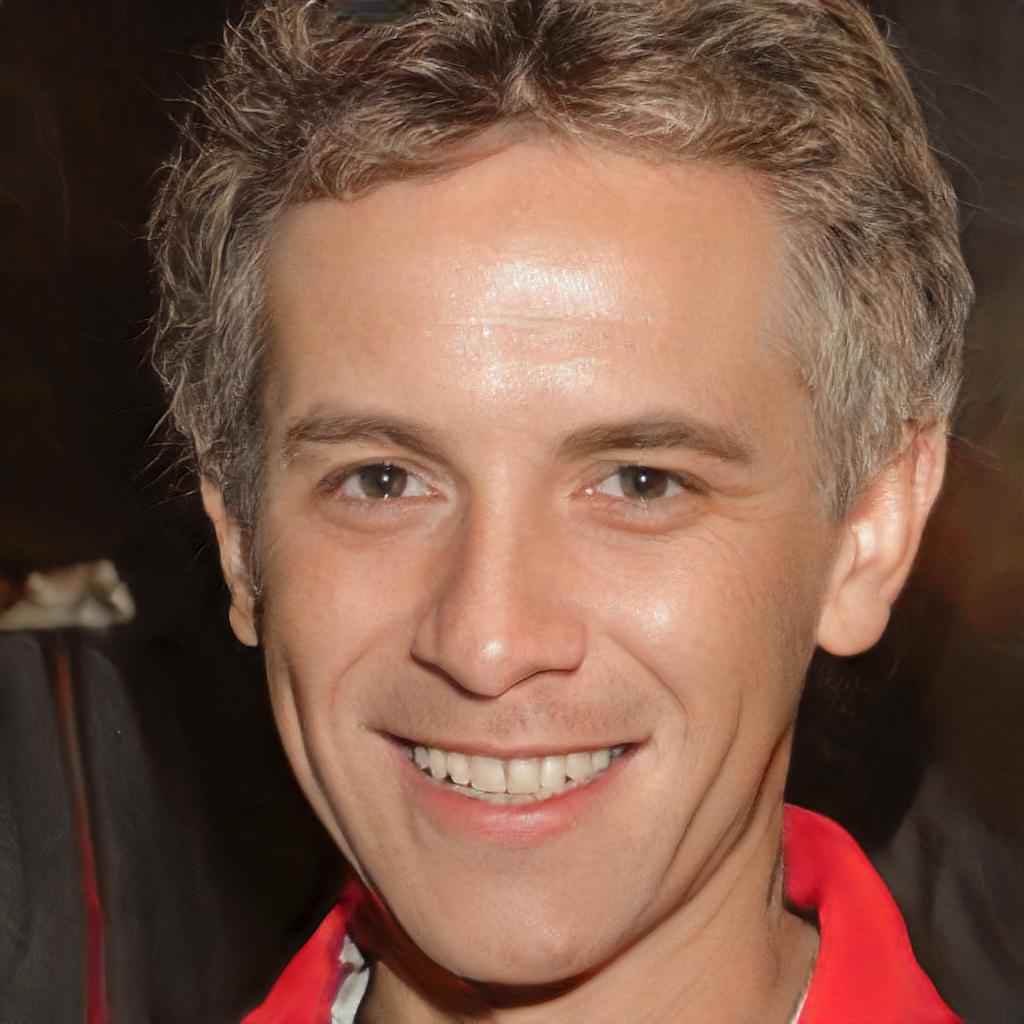}&
\includegraphics[width=0.19\linewidth]{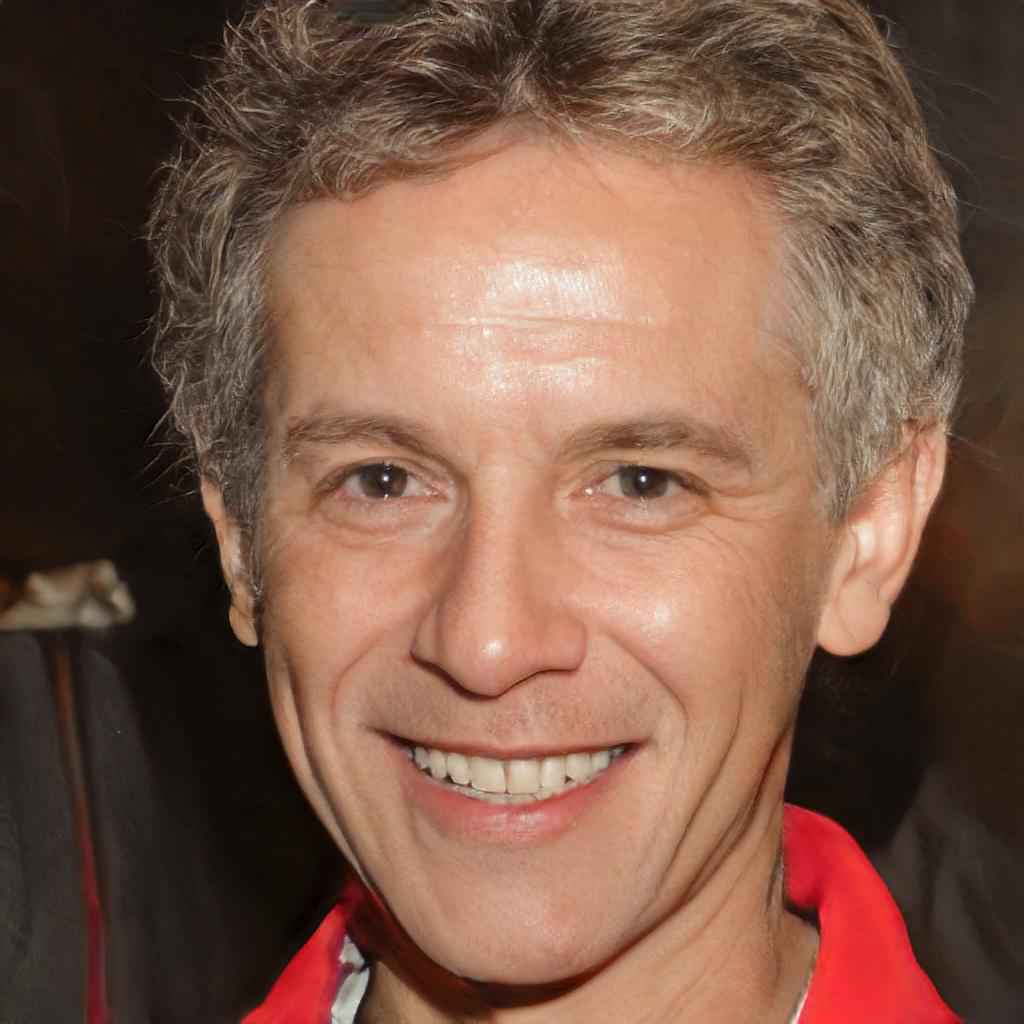}&
{\color{yellow}%
\setlength{\fboxsep}{0pt}%
\setlength{\fboxrule}{2pt}%
\fbox{\includegraphics[width=0.19\linewidth]{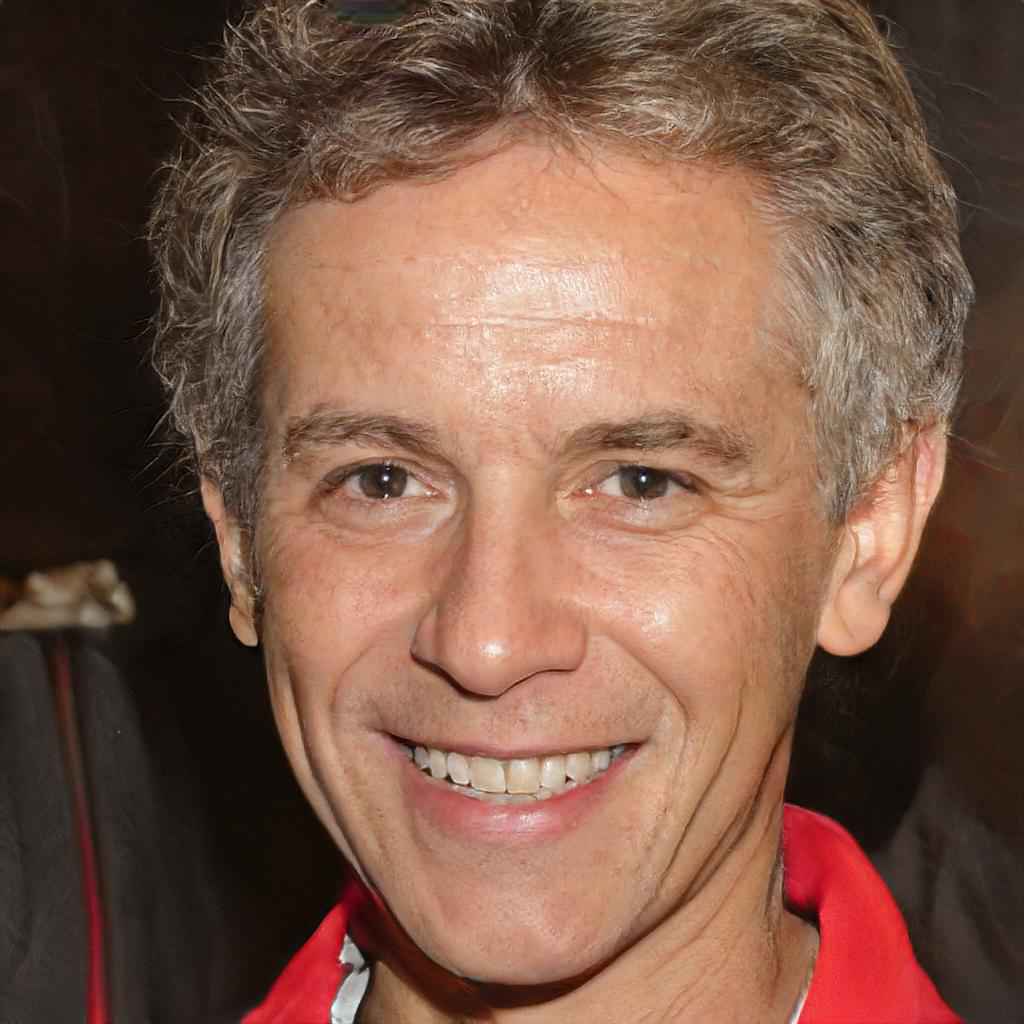}}}&
\includegraphics[width=0.19\linewidth]{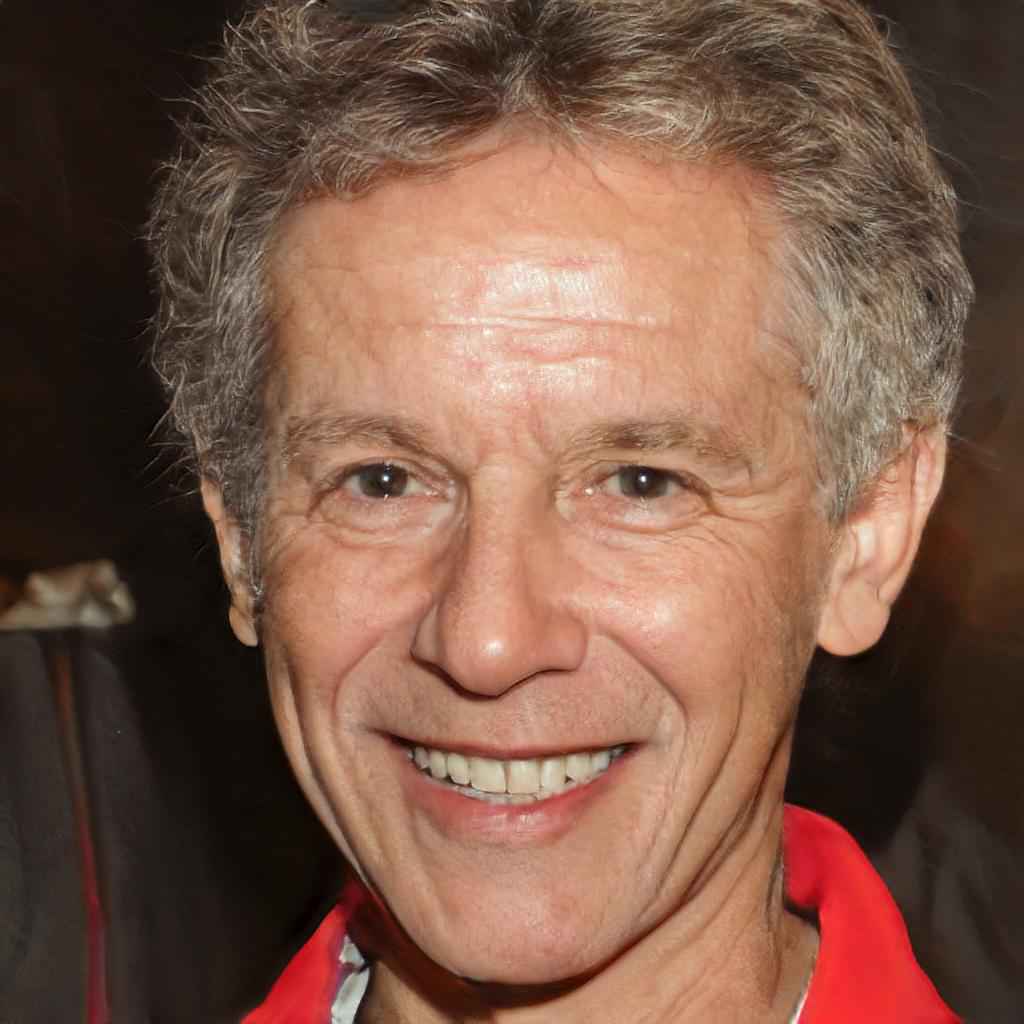}
\end{tabular}
\end{center}
\caption{\textbf{Age editing results on $1024\times 1024$ images}.We propose a single deep age transformer network able to perform both face aging and de-aging, producing high quality images that are sharp and with little artifacts. Using the face images indicated by a yellow frame as input, our network can output a photo-realistic image of the same person at any required target age in the range \{20, \ldots, 69\}.
}
\label{teaser}
\end{figure*}

Learning to manipulate face age is an important topic both
in industry and academia. In the movie post-production industry, many actors are retouched in some way, either
for beautification or texture editing. More specifically, synthetic aging or de-aging effects are usually generated by makeup or special visual
effects. Although impressive results can be obtained digitally, as in the recent Martin Scorcese's movie {\it The Irishman}, the underlying processes are extremely time consuming.  Thus, robust, high-quality algorithms for performing
automatic age modification are highly desirable. Nevertheless, editing faces is an intrinsically difficult task. Indeed, the human brain is particularly good at perceiving faces' attributes in order to detect, recognize or analyze them, for instance to infer identity or emotions. Consequently, even small artifacts are immediately perceived and ruin the perception of results. For this reason, our goal is to produce artifact-free, sharp and photorealistic results on high-resolution face images.

With the success of Generative Adversarial Networks (GANs)~\cite{Goodfellow2014} in high quality image generation, GAN-based models have been widely used for image-to-image translation~\cite{wang2018pix2pixHD,zhu2017unpaired}. 
Despite having set new standards for natural image synthesis, GANs are known to suffer from two major flaws : an abundance of small artifacts %
and strong instability of the training process. The latest face aging studies~\cite{he2019s2gan,Liu_2019_CVPR,song2018dual,wang2018face,zhang2017age} also adopt GAN-based models. Specifically, they divide face datasets into different age groups, feed young images into the generator, and rely on the discriminator to map output images to older age distributions. There are multiple limitations to this approach. Firstly, as can be expected, these approaches inherit the drawbacks of GAN-based methods - blurry background, small parasite structures, instability of training. %
Secondly, as the aging effect is generated by matching the output image distribution to the target group, these methods are limited to coarse aging/de-aging. To achieve fine-grained transformation, a separate model needs to be trained between each pair of ages.

In this work, we propose an encoder-decoder architecture for the problem of face age editing with high visual quality on high resolution images. 
In order to address the aforementioned limitations, namely the tendency to produce visual artifacts and training instability, we endeavour to keep the architecture as simple as possible. Firstly, we use a single network for both aging and de-aging. This is reasonable since the encoder part of our model is assumed to encode identity, emotion or details in the input image that are not related to age, so that the same latent space can be used for both tasks of aging and de-aging. Secondly, we rely on a {\it feature modulation layer}, that is compact, acts directly on the latent space and allows for continuous age transitions. Thirdly, unlike in competing methods where the discriminator used during adversarial training is conditioned on the target age, we use a discriminator which is not conditioned and concentrates solely on the photorealism of the output images to reduce editing artifacts. The discriminator can be considered as a regularizer which imposes photorealism other than a traditional discriminator trying to match two distributions. Thanks to this design, our model achieves efficient disentanglement of age attributes and face identity.
We present experimental results on high resolution images with qualitative and quantitative evaluations. In particular, these experiments provide clear evidence that the visual quality achieved by our results outperforms state of the art methods. Experiments on alternative datasets further illustrate the generalization capacity of the method.

\section{Related Works}
\label{sec:related}

\textbf{Face aging} \quad The survey work~\cite{fu2010age} gives an exhaustive overview of the traditional age synthesis algorithms. In this work, we are more interested in deep learning based methods, which have made impressive progress on face aging tasks during the last few years. A conditional GAN~\cite{mirza2014conditional} model is first introduced for face aging task by~\cite{antipov2017face,zhang2017age}. They encode the face image to the latent space, manipulate the latent code, and decode it to an aged face with the generator. However, the identity information is damaged during this process. This is further improved by~\cite{wang2018face,yang2018learning}, by adding an identity preserving term to the objective. Despite the improvement, their results are over-smoothed compared with the input images. To capture texture details, wavelet-based generative models are introduced by~\cite{li2019global,Liu_2019_CVPR}. Their complex models increase the training difficulty and still yield strong artifacts. All the aforementioned models only enable face aging from one age group to another, \eg, from 20s to 40s, lacking flexibility. Recently, ~\cite{he2019s2gan} proposed an encoder-decoder network, in which a personalized aging basis is synthesized and an age-specific transform is applied. Their model also relies on a conditional discriminator to distinguish aging patterns between age groups. Different from other methods, our model is designed for age editing with a random target age. Moreover, our approach produces much less artifacts, making age editing on images of high resolution ($1024 \times 1024$) possible.

\textbf{Image-to-image translation} \quad Face aging can be considered as an image-to-image translation problem, ie translating images between young age and old age domains. An optimization based method is proposed by~\cite{upchurch2017deep}, showing the possibility to use linear interpolation of deep features from pretrained convnets to transform images. GAN based methods~\cite{pix2pix2016,zhu2017unpaired,huang2018multimodal} further enable real-time translation, by training a feed forward generator. Existing image-to-image translation studies~\cite{chen2019semantic,choi2018stargan,lample2017fader,pumarola2018ganimation,qian2019make,xiao2018elegant} on face images also yield impressive results in manipulating facial attributes. Lample \etal~\cite{lample2017fader} design an autoencoder architecture to reconstruct images, and isolate single image characteristics in a latent component via a discriminator. These characteristics can then be modified directly in the latent space. Choi \etal~\cite{choi2018stargan} propose a method to perform image-to-image translation for multiple domains using only a single model. Pumarola \etal~\cite{pumarola2018ganimation} introduce an attention based model, which enables face animation by simple interpolation.

\textbf{High-resolution image synthesis} \quad 
In spite of the considerable progress of recent methods, manipulating/editing natural images of high resolution has not yet been achieved. Nevertheless, in another task - image generation, high quality results at high resolution are now available. Image generation at $1024 \times 1024$ resolution is first achieved by~\cite{karras2018progressive}, with a progressive growing of GAN architectures. The quality of their results is further improved by StyleGAN~\cite{karras2019style,karras2019analyzing}, which learns a separation of high-level attributes automatically during the training. Based on this work, Shen \etal~\cite{shen2019interpreting} propose an effective way to interpret the latent space learned by the generator and achieve high visual fidelity face manipulation on synthesized images. However, according to our experiments, only a fraction of natural images can be accurately reconstructed with a latent code, which makes this type of method impractical. In contrast, our proposed method achieves face age editing on $1024 \times 1024$ images, with great simplicity of architecture and loss design. The age editing is achieved only by an auxiliary modulating network, which could be potentially generalized to other face manipulation tasks.

\section{Method}
\begin{figure*}[t]
\begin{center}
\includegraphics[width=\linewidth]{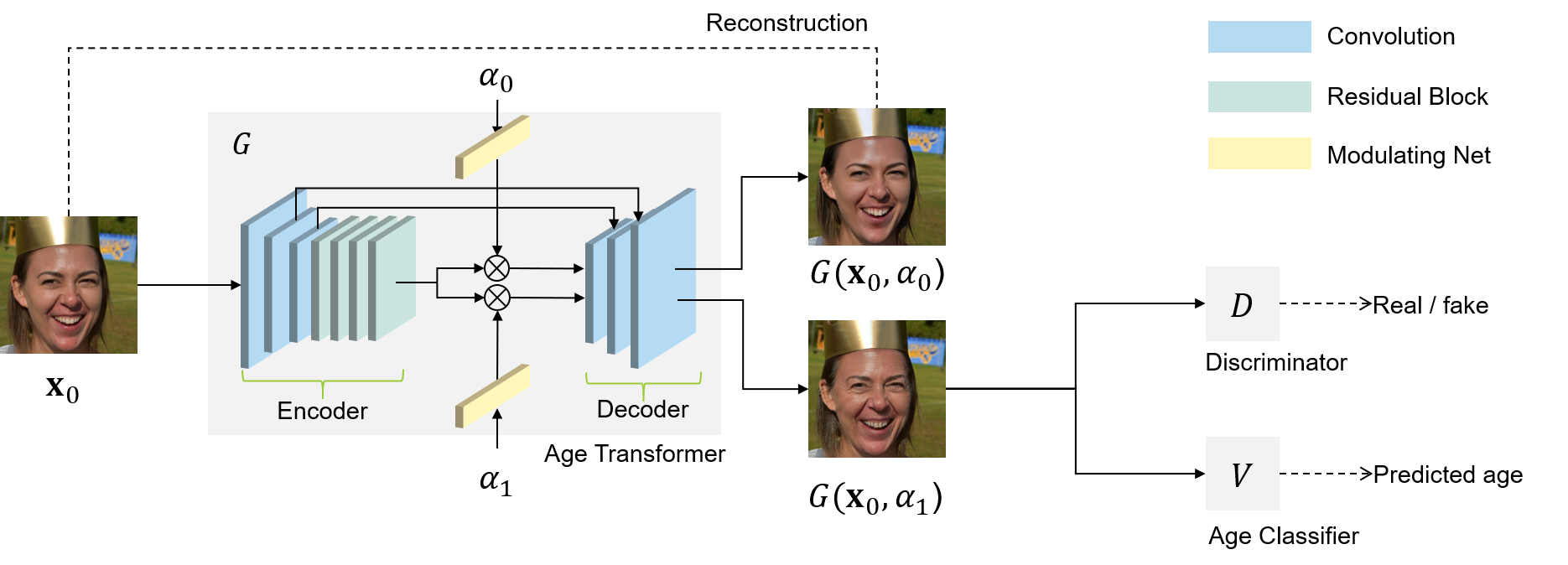}
\vspace{1pt}
\caption{\textbf{Training process}: each input image $\ve{x}_0$ is edited by the age transformer $G$ using the initial age $\alpha_0$ (reconstruction task) and the target age $\alpha_1$ (editing task). The reconstructed image $G(\ve{x}_0, \alpha_0)$ should be identical to the input image. The edited image $G(\ve{x}_0, \alpha_1)$ is further passed in a discriminator $D$ that ensures photorealism of the transformed image, and an age-classifier $V$ that ensures age-accurate transformation.
\textbf{The age transformer $G$} contains three sub networks: an encoder, a modulating network and a decoder. The encoder maps the input image $\ve{x}_0$ to an age-invariant deep feature space. The modulating network maps a target age $\alpha$ to a $128$-dimensional modulating vector. This vector is used to modulate each channel of the encoded features, hence applying the desired age transformation. The modulated features are finally passed in the decoder to obtain the transformed image. Two skip connections between the encoder and the decoder in order to preserve the age irrelevant details better.
}
\label{arch}
\end{center}
\end{figure*}

In this section, we present the face age editing problem and present our proposed model in detail. Figure \ref{arch} illustrates our proposed age transformer and training procedure. 

\subsection{Overview}

Let $\ve{x}_0$ be an image drawn randomly from a face dataset. We denote by $\alpha_0$ the age of the person in $\ve{x}_0$. Our goal is to transform $\ve{x}_0$ so that the person in this image looks like someone at $\alpha_1$ years old. We want the aged version of $\ve{x}_0$ to share many age-unrelated characteristics with $\ve{x}_0$: identity, emotion, haircut, background, etc. That is to say: the facial attributes not relevant to age, as well as the background, need to be preserved during age transformation. Therefore, we assume that a face aging model and a face de-aging model can share most of their parameters.
In this setting, we consider a single age transformer $G$ and assume that $G$ can transform any face image to any target age. The inputs of our model are the face image $\ve{x}_0$ and the target age $\alpha_1$. The output is denoted by $G(\ve{x}_0, \alpha_1)$, which depicts $\ve{x}_0$ at the target age $\alpha_1$.

\subsection{Age transformer}

The proposed age transformer shown in Figure \ref{arch} employs an auto-encoder architecture and is made of an encoder, a feature modulation block and a decoder. The encoder consists of three strided convolutional layers (the first one of stride 1, the other two of stride 2) and four residual blocks~\cite{he_resnet16}, while the decoder contains two nearest-neighbour upsampling layers and three convolutional layers, similar to the architecture used in~\cite{johnson2016perceptual,zhu2017unpaired}. The main difference compared to these works is our feature modulation block, in which the output features of the encoder are modulated by an age-specific vector (see details below). This idea is inspired by recent works on style transfer~\cite{dumoulin17,huang17} which show the possibility to represent different styles using the parameters of normalization layers.

\begin{itemize}

\item \textbf{Encoder} \quad The face image $\ve{x}_0$ is the input of the encoder. The output features are denoted by $\ma{C} \in \mathbb{R}^{n \times c}$, where $c=128$ is the number of channels and $n$ is the product of the two spatial dimensions. 

\item \textbf{Feature modulation for age selection} \quad  
The target age $\alpha_1$ is encoded as an one-hot vector, denoted by $\ve{z_1}$, and passed to the modulating network. This network consists of a single fully connected layer whith a sigmoid activation. It outputs a modulation vector $\ve{w} \in [0, 1]^c$, which is used to re-weight the features $\ma{C}$ before passing them into the decoder and obtaining the face image at the desired age. The modulated features are $\ma{C} \, {\rm diag}(\ve{w})$, where ${\rm diag}(\ve{w})$ is the diagonal matrix with diagonal $\ve{w}$.

\item \textbf{Decoder} \quad The decoder takes the modulated features $\ma{C} \, {\rm diag}(\ve{w})$ as input and two skip connections, used to preserve the finer details of the input image. The final output is denoted by
$
G(\ve{x}_0, \alpha_1)
$.

\end{itemize}

\subsection{Training}

As illustrated in Figure~\ref{arch}, we train our age transformer with an age classifier that ensures age-accurate transformation and a discriminator that preserves photorealism. 

The initial age $\alpha_0$ of $\ve{x}_0$ is easy to estimate using a pretrained age classifier, \eg,~\cite{rothe2015dex}. We thus do not use an age-annotated dataset for training. The original age range of the training dataset is denoted by $\set{Q} \subset \mathbb{N}$. At test time, the target age can be chosen as any age in $\set{Q}$. At training time, it would seem reasonable to chose any value in \set{Q} uniformly at random. However, we noticed that the artifacts appearing during large age transformations were better corrected when selecting a target age $\alpha_1$ far enough from $\alpha_0$ during training. We propose to sample $\alpha_1$ from the set $\set{Q}_{\alpha_0} = \{\alpha \in \set{Q}\ : \left|\alpha - \alpha_0\right| \geq \alpha_*\}$ at training time, where $\alpha_*$ is a predefined constant representing the minimum age transformation interval. We denote by $q(\alpha \vert \alpha_0)$ the uniform distribution over $\set{Q}_{\alpha_0}$.

\textbf{Classification loss} \quad To measure the age of $G(\ve{x}_0, \alpha_1)$, we use the same age classifier as the one used to estimate $\alpha_0$. %
During training, we freeze the weights of this classifier. The classifier, denoted by $V$, takes $G(\ve{x}_0, \alpha_1)$ as input and generates a discrete probability distribution %
over the set of ages $\{0, 1, \ldots, 100\}$. The classification loss satisfies
\begin{equation}
\label{eq:class_loss}
\mathcal{L}_{\rm class} = \mathbb{E}_{\ve{x}_0\sim p(\ve{x})} \mathbb{E}_{\alpha_1 \sim q(\alpha \vert \alpha_0)} \left[ \ell(\ve{z}_1, V(G(\ve{x}_0, \alpha_1))) \right]
\end{equation}
where $p(\ve{x})$ denotes the training image distribution over $\mathcal{X}$, $\ell$ denotes the categorical cross-entropy loss, and $\ve{z}_1$ is the one-hot vector encoding $\alpha_1$.

\textbf{Adversarial loss} \quad To enforce better photorealism of the modified images $G(\ve{x}_0, \alpha_1)$, we adopt an adversarial loss built using PatchGAN~\cite{pix2pix2016} with the LSGAN objective~\cite{mao2017least}. Unlike the latest works on face aging~\cite{he2019s2gan,Liu_2019_CVPR,song2018dual,wang2018face,zhang2017age}, our discriminator is used to distinguish between real and manipulated images without taking the age information into account. In our work, the aging and de-aging effects is obtained solely with the age classification loss. 

The  discriminator is denoted by $D$. The architecture of $D$ is the same as proposed in~\cite{pix2pix2016}. We use a patch size $142 \times 142$ for $1024 \times 1024$ images. The modified image $G(\ve{x}_0, \alpha_1)$ should be indistinguishable from real samples. Therefore, the losses we use are
\begin{align}
\label{eq:gan_loss}
\mathcal{L}_{\rm GAN}(G) =\ &\mathbb{E}_{\ve{x}_0\sim p(\ve{x})} \mathbb{E}_{\alpha_1 \sim q(\alpha \vert \alpha_0)} [(D(G(\ve{x}_0, \alpha_1))-1)^2],
\end{align}
when training $G$, and
\begin{align}
\mathcal{L}_{\rm GAN}(D) =\ &\mathbb{E}_{\ve{x}_0\sim p(\ve{x})}\mathbb{E}_{\alpha_1 \sim q(\alpha \vert \alpha_0)}[(D(G(\ve{x}_0, \alpha_1)))^2]\ + \nonumber \\ 
                             &\mathbb{E}_{\ve{y}\sim p(\ve{x})}[(D(\ve{y}) - 1)^2]
\end{align}
when training $D$. We apply $R_1$ regularization~\cite{Mescheder2018ICML} with $\gamma=10$ on the discriminator. 

\textbf{Reconstruction loss} \quad When the age transformer receives $\ve{x}_0$ and $\alpha_0$ as inputs, the generated output image $G(\ve{x}_0, \alpha_0)$ should be  identical to the input image. Hence, we minimize the following reconstruction loss:
\begin{equation}
\label{eq:recon_loss}
\mathcal{L}_{\rm recon} = \mathbb{E}_{\ve{x}_0\sim p(\ve{x})}[||G(\ve{x}_0, \alpha_0) - \ve{x}_0||_1].
\end{equation}

\textbf{Full loss} \quad We train the age transformer and the discriminator by minimizing the full objective 
\begin{equation}
\label{eq:total_loss}
\mathcal{L} =  \lambda_{\rm recon}\mathcal{L}_{\rm recon} + \lambda_{\rm class}\mathcal{L}_{\rm class} + \mathcal{L}_{\rm GAN}
\end{equation}
where $\lambda_{\rm recon}$ and $\lambda_{\rm class}$ are weights balancing the influence of each loss.

\section{Experiments}
In this section, we introduce our training setup and present the experimental results. We further evaluate the quality of our results using quantitative metrics.

\subsection{Data augmentation with synthetic images}

Our training dataset is built upon FFHQ~\cite{karras2019style}, a high resolution dataset which contains $70,000$ face images at $1024 \times 1024$ resolution. The dataset includes large variations in age, ethnicity, pose, lighting, and image background. However, the dataset contains only unlabeled raw images collected from Flickr.

To obtain the age information, we use an age classifier pretrained on IMDB-WIKI~\cite{rothe2015dex}. We observe that FFHQ contains much more samples of young faces than of old ones. This data imbalance is challenging since the aging and de-aging tasks would not be treated equally during training: most of faces being young, the age transformer would be trained to perform aging much more often than de-aging, failing to yield satisfying de-aging results. To compensate this imbalance in the age distribution, we propose to perform data augmentation using StyleGAN - a state-of-the-art high resolution image generation model~\cite{karras2019style}. We use the StyleGAN model pretrained on FFHQ to generate $300,000$ synthetic images. A quick visual inspection shows that most of the generated images have no significant artifacts and are nearly indistinguishable from real images by a human. Therefore, we use them for data augmentation to obtain a quasi-uniform age distribution over $\mathcal{Q}$: for any age bin with less than $1,000$ samples in the original FFHQ dataset, we complete this bin with some of the generated synthetic face images; for any age bin with more than $1,000$ samples, we select randomly $1,000$ face images from the original FFHQ dataset. The age-equalized dataset contains $47,990$ images over the range $\set{Q}=\{20, \ldots, 69\}$.

\subsection{Implementation details}

Our model is implemented in PyTorch~\cite{paszke2017automatic}. We take $95\%$ of the equalized dataset as our training set and the rest as test set. For the age transformer and the discriminator, spectral normalisation~\cite{miyato2018spectral} is applied on all the convolution layers except the last one of the age transformer. All the activation layers use Leaky ReLU~\cite{maas2013rectifier} with a negative slope of $0.2$. 

We consider age transformation only in the age range $\set{Q}=\{20, \ldots, 69\}$. The constant $\alpha_*$ is set to $25$. We have observed that the most significant artifacts appear when the gap between the source and target age is large. By choosing $\alpha_*$ large enough, we force the discriminator $D$ to suppress these artifacts during adversarial training. The weights $\lambda_{\rm recon}$ and $\lambda_{\rm class}$ are set to $10$ and $0.1$, respectively. We use Adam optimizer with a learning rate of $10^{-4}$. The age transformer $G$ is updated once after each discriminator update. Our model is trained for $20$ epochs to achieve face age editing on high resolution images. The first $10$ epochs are trained on $512 \times 512$ images with a batch size of $4$. 
The next $10$ epochs are trained on $1024 \times 1024$ images, for which we reduce the batch size to $2$, learning rate to $10^{-5}$ and $\lambda_{\rm recon}$ to $1$. 

\subsection{Qualitative evaluation}
\begin{figure*}[!t]
\begin{center}
\setlength{\tabcolsep}{1pt}
\begin{tabular}{ccccc}
25&35&45&55&65 
\\
{\color{yellow}%
\setlength{\fboxsep}{0pt}%
\setlength{\fboxrule}{2pt}%
\fbox{\includegraphics[width=0.18\linewidth]{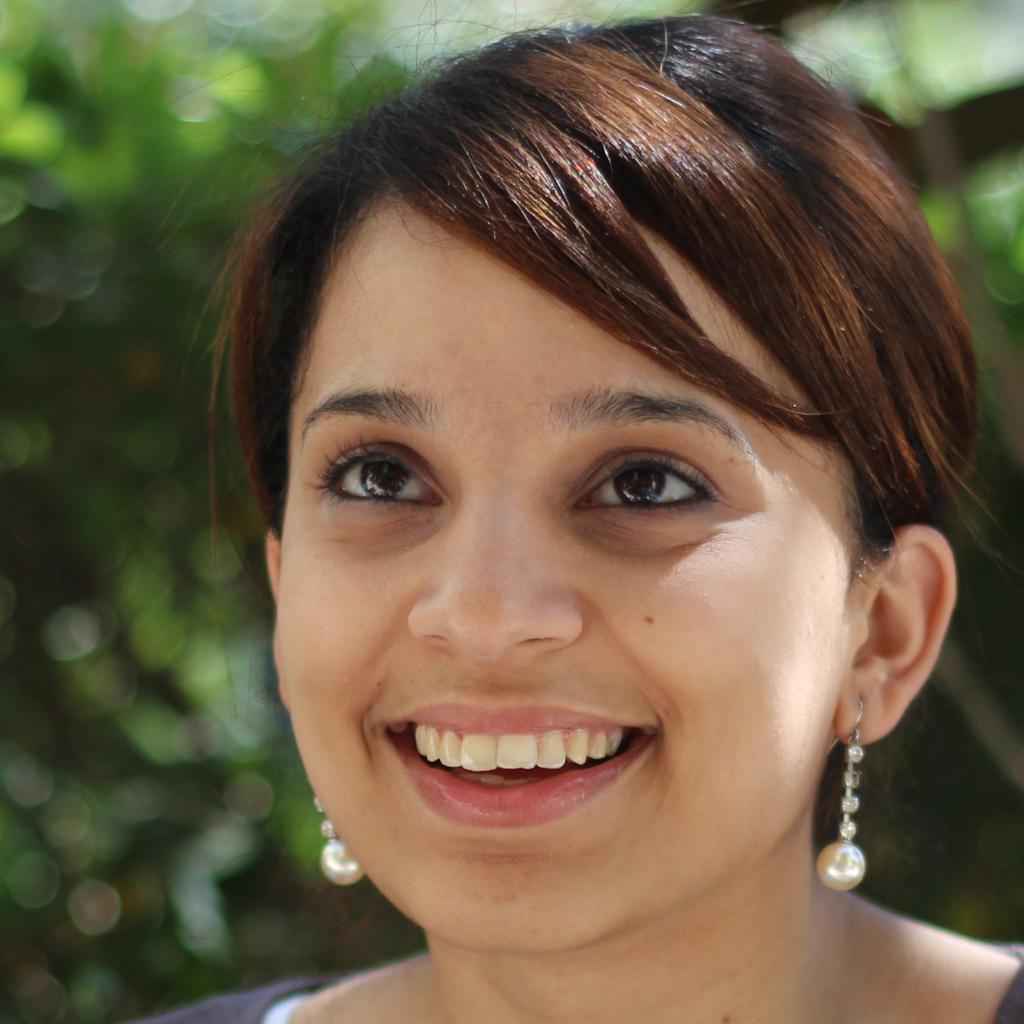}}} &
\includegraphics[width=0.18\linewidth]{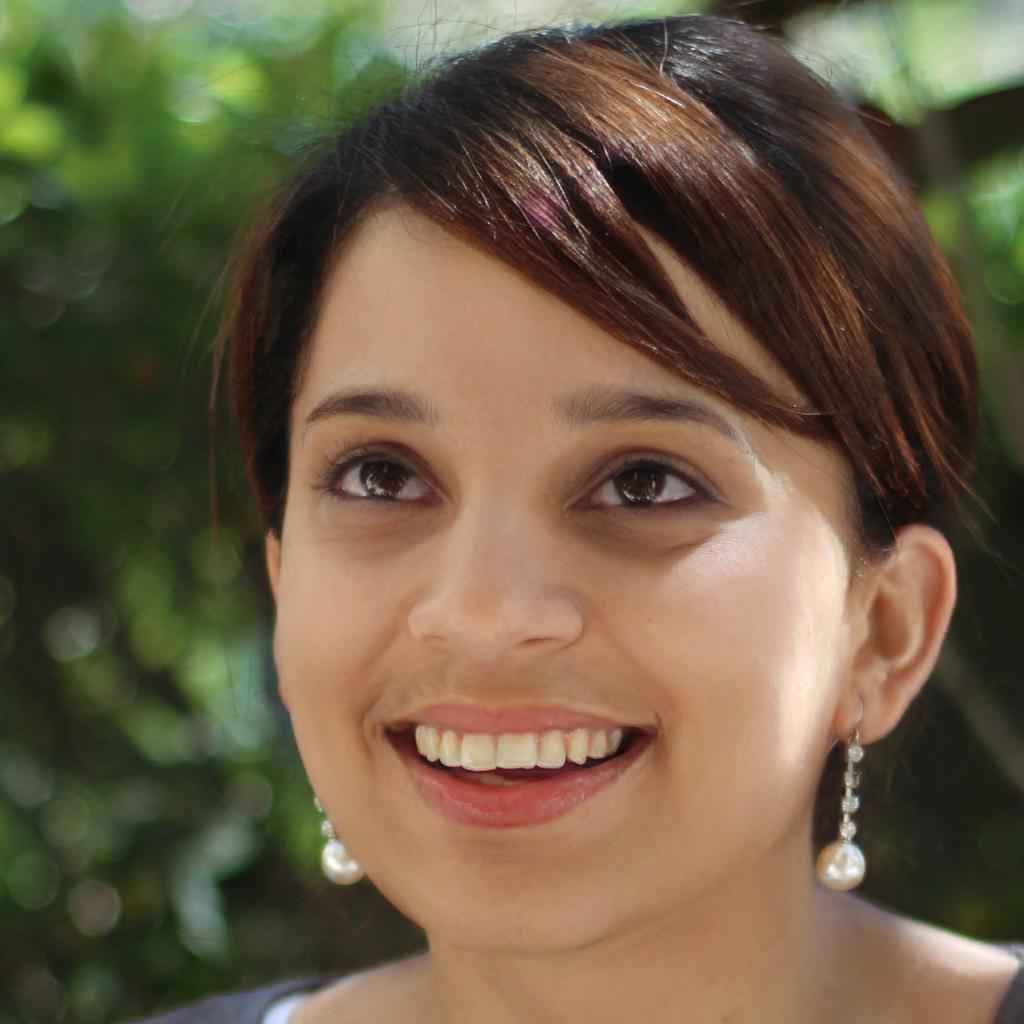} & 
\includegraphics[width=0.18\linewidth]{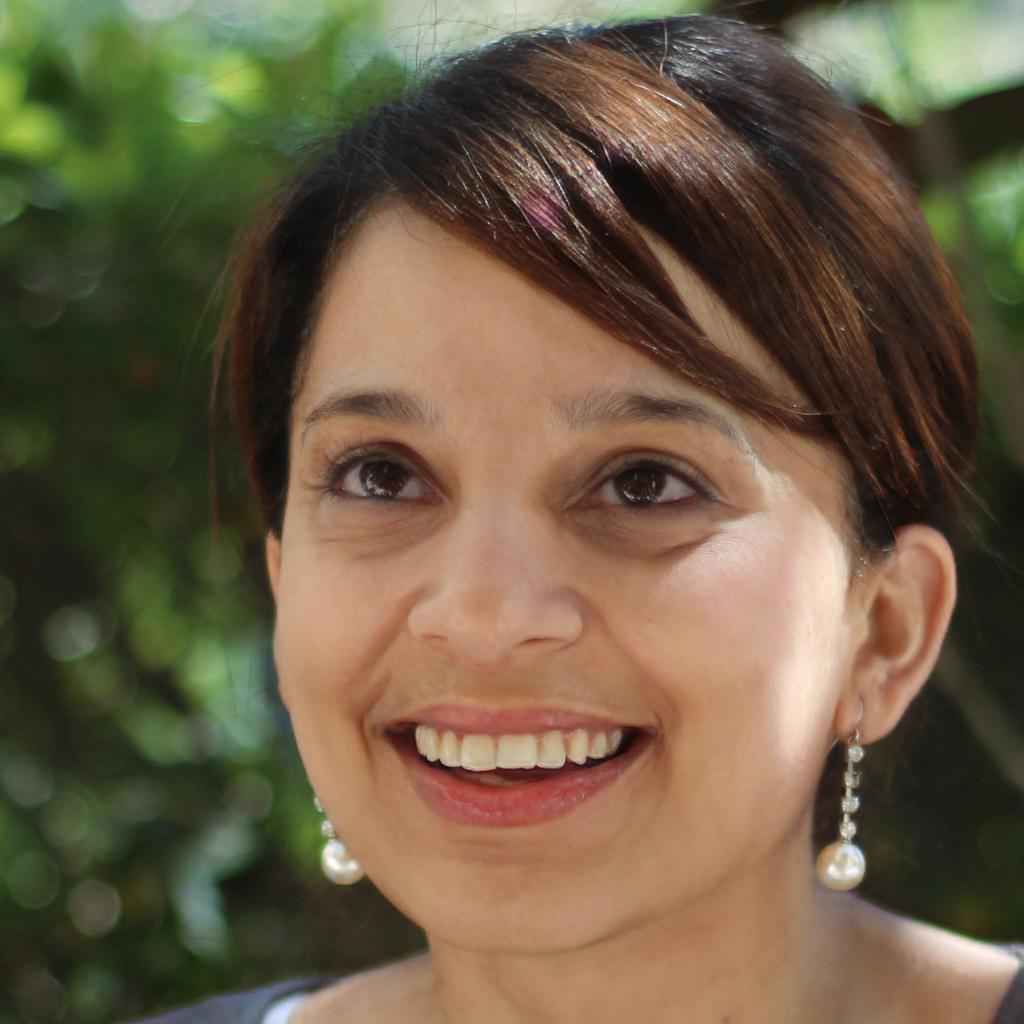} &
\includegraphics[width=0.18\linewidth]{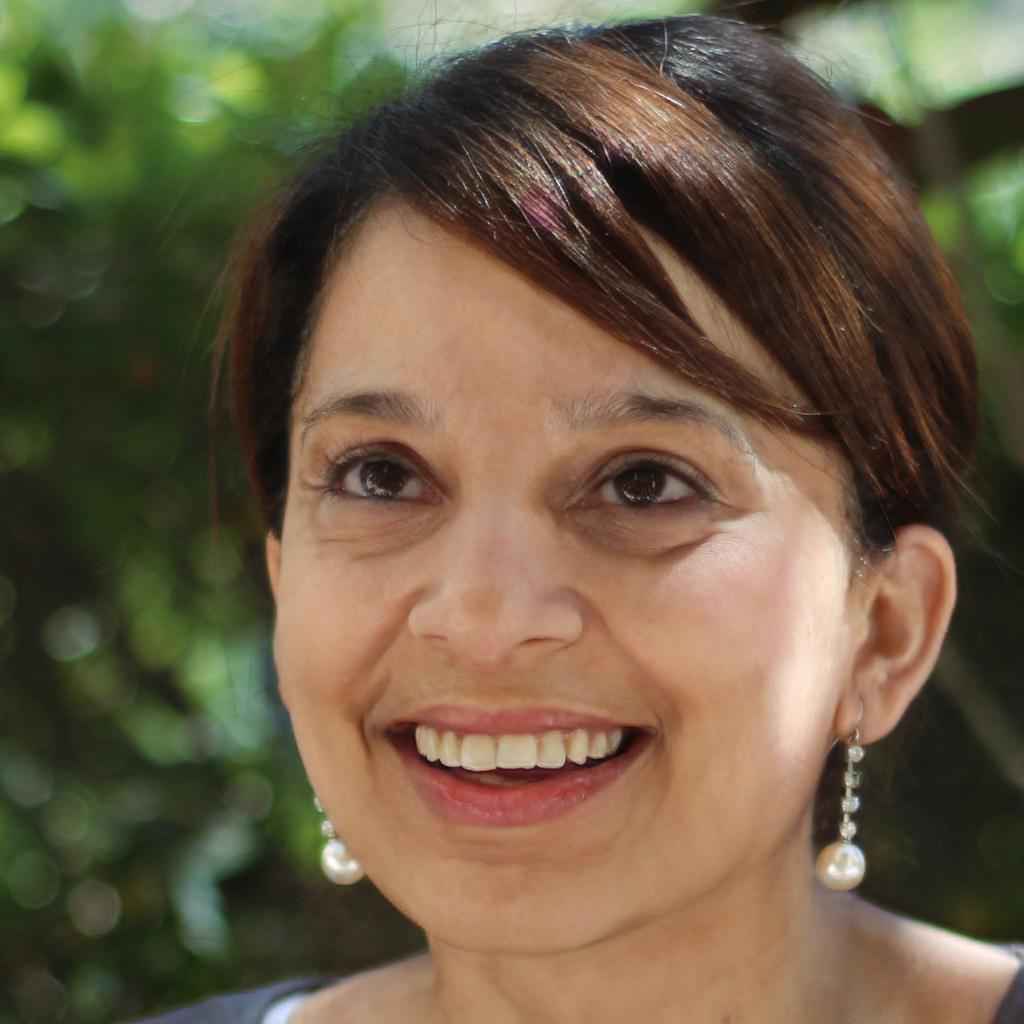} & 
\includegraphics[width=0.18\linewidth]{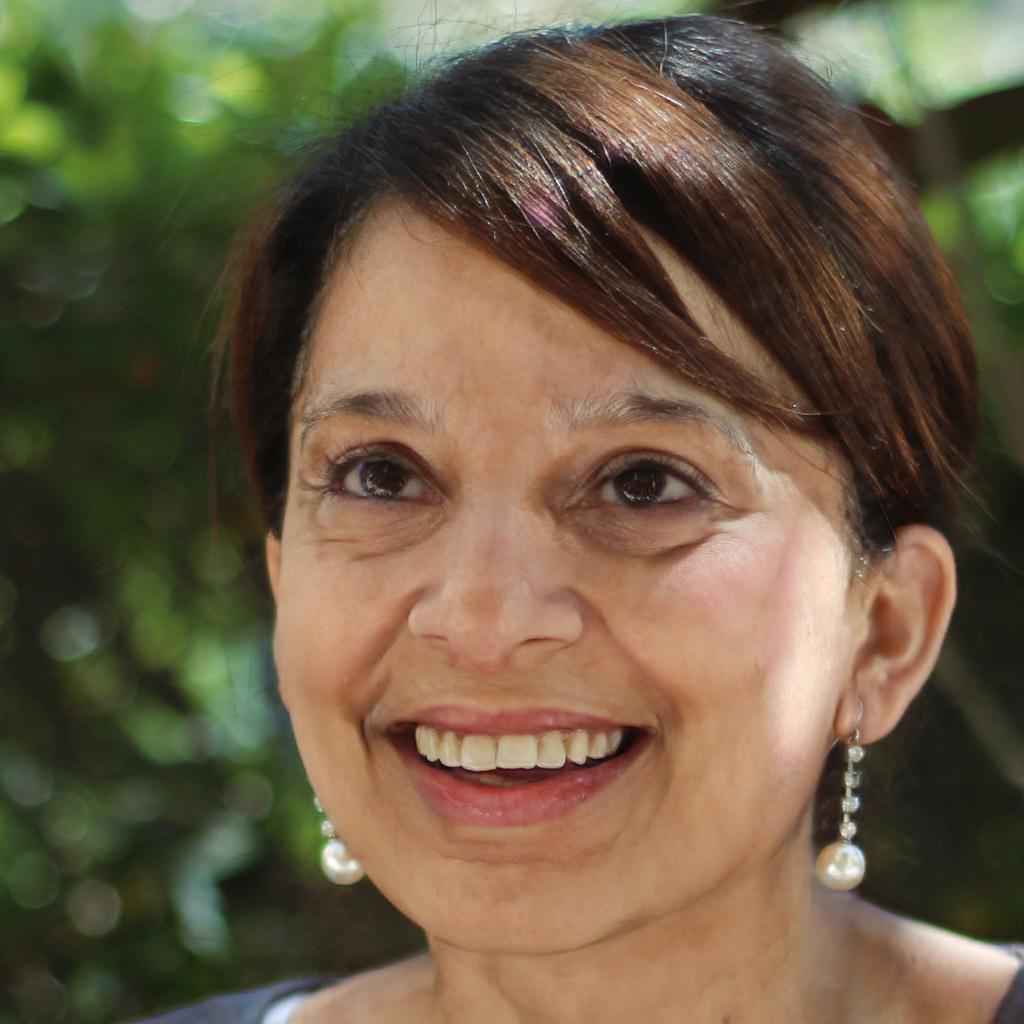} 
\\
\includegraphics[width=0.18\linewidth]{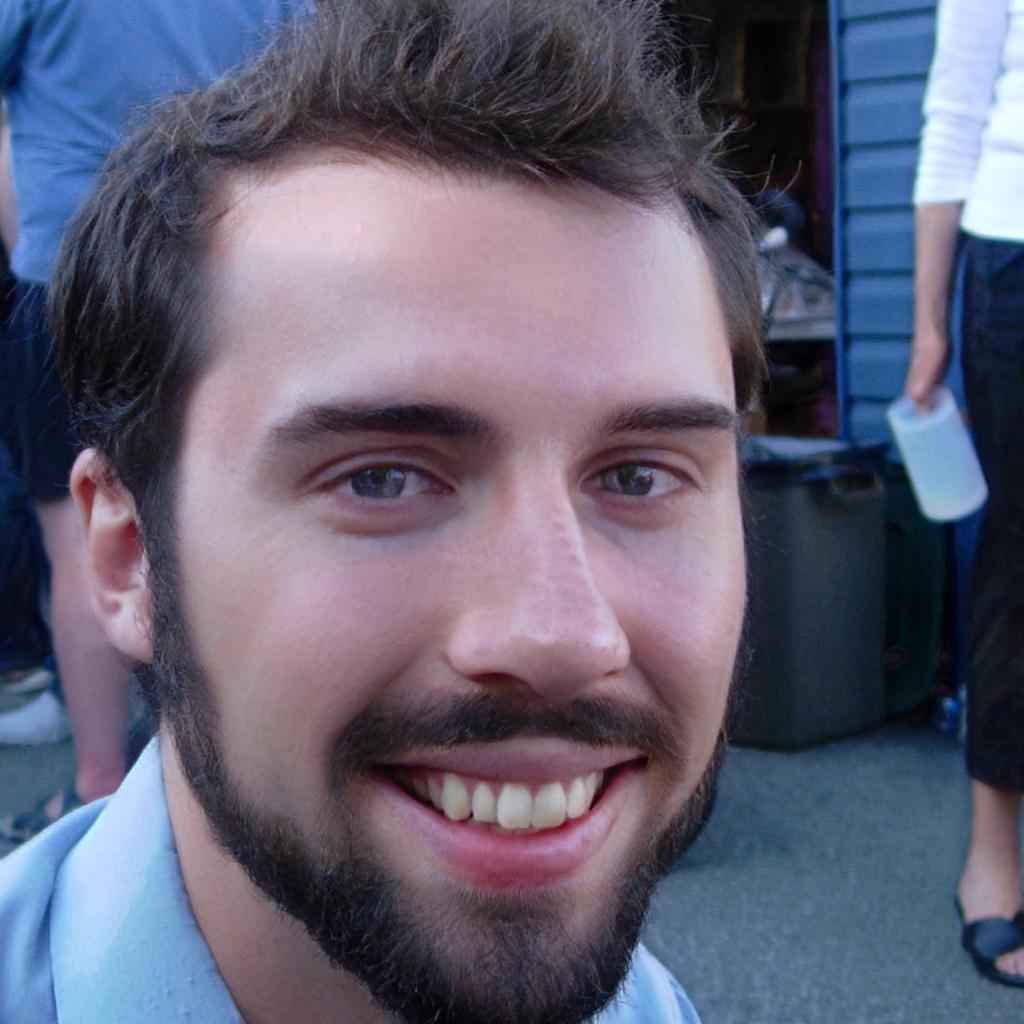} & 
{\color{yellow}%
\setlength{\fboxsep}{0pt}%
\setlength{\fboxrule}{2pt}%
\fbox{\includegraphics[width=0.18\linewidth]{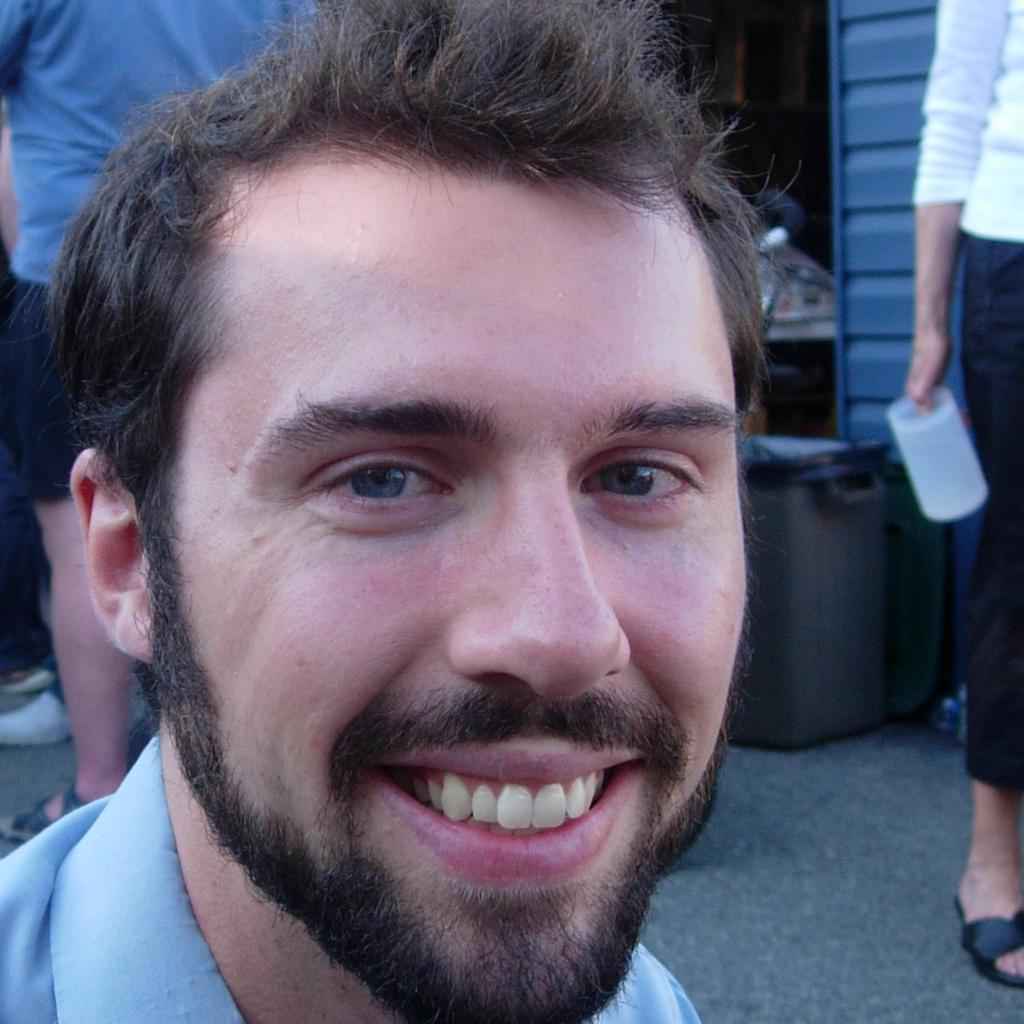}}} & 
\includegraphics[width=0.18\linewidth]{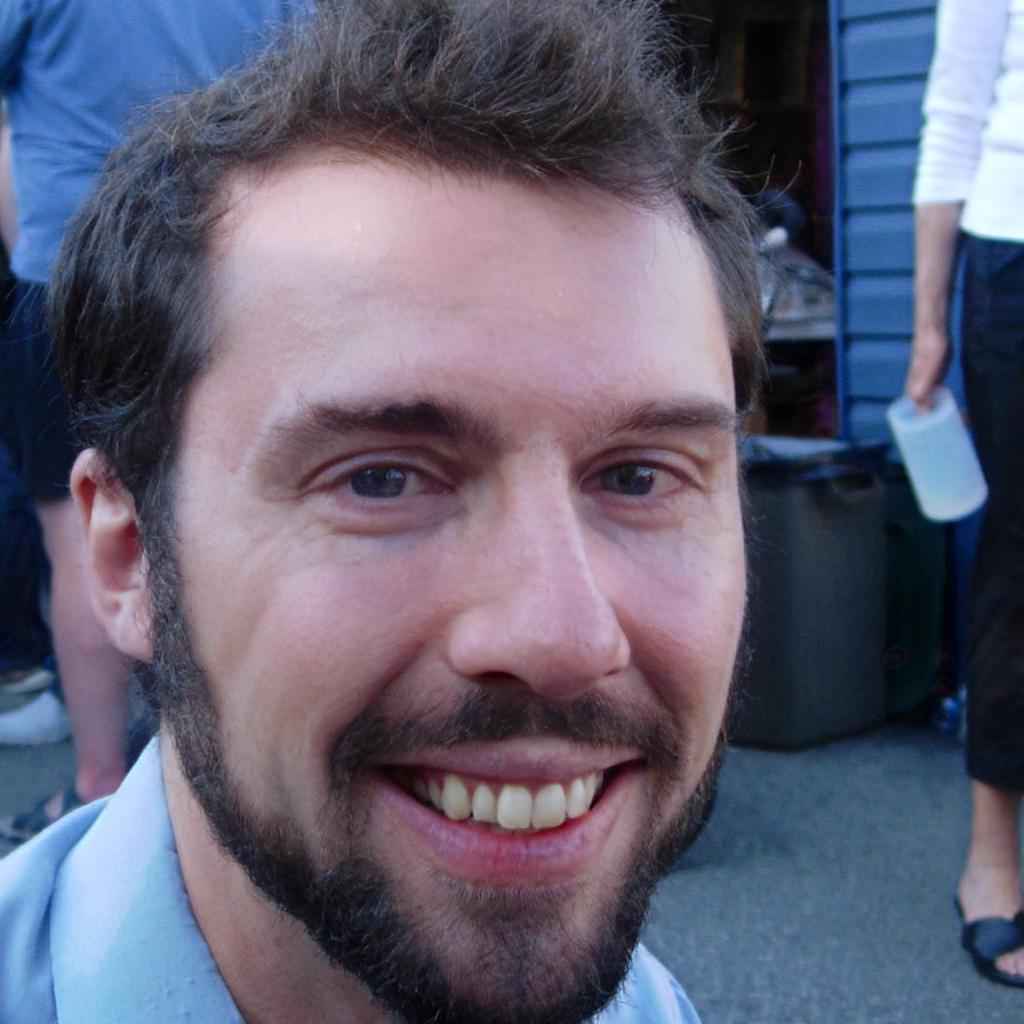} &
\includegraphics[width=0.18\linewidth]{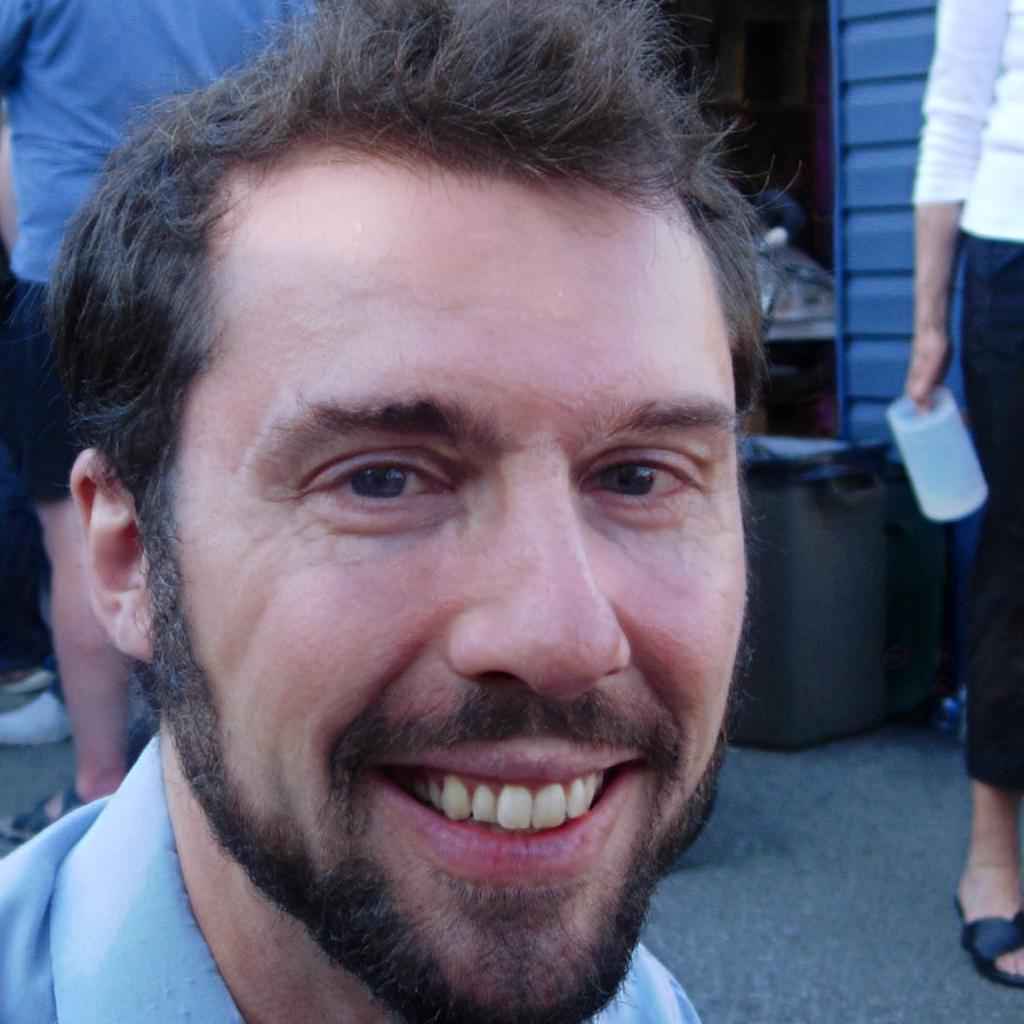} & 
\includegraphics[width=0.18\linewidth]{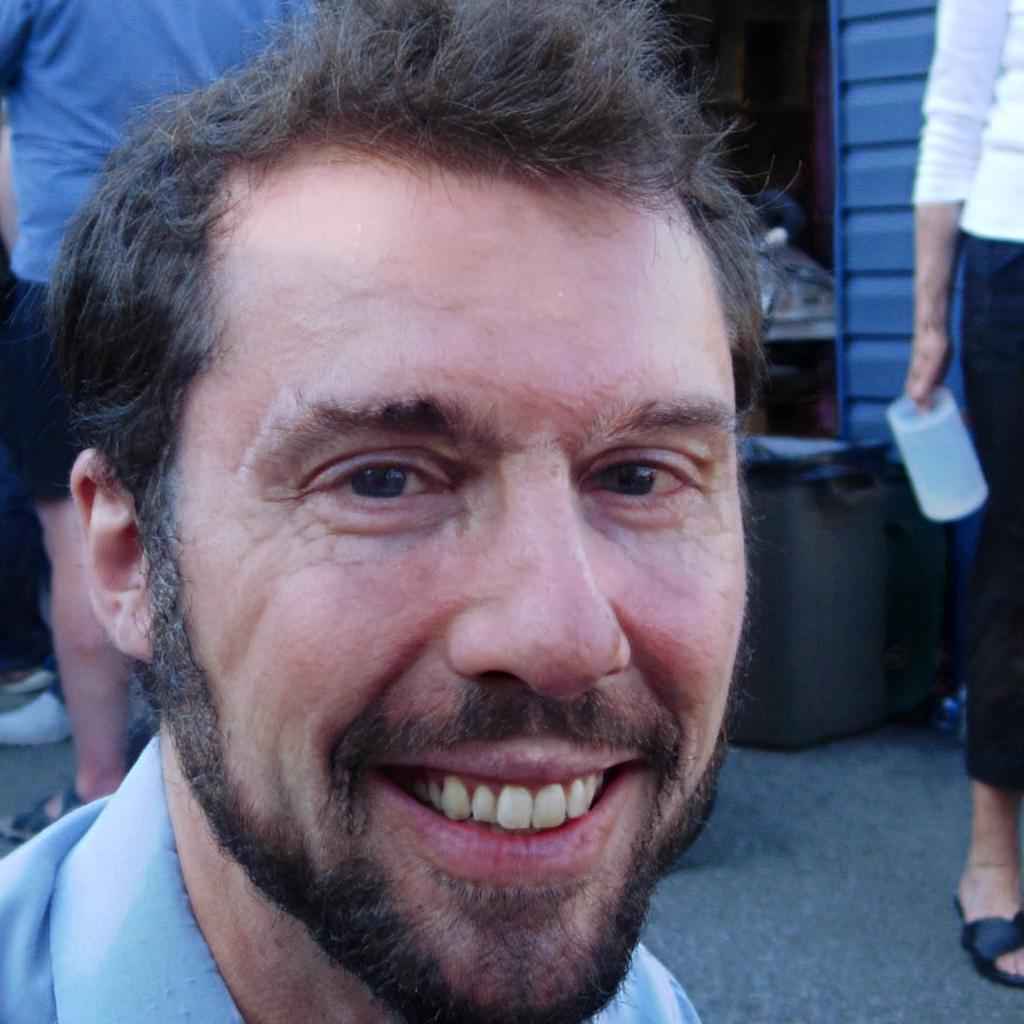} 
\\
\includegraphics[width=0.18\linewidth]{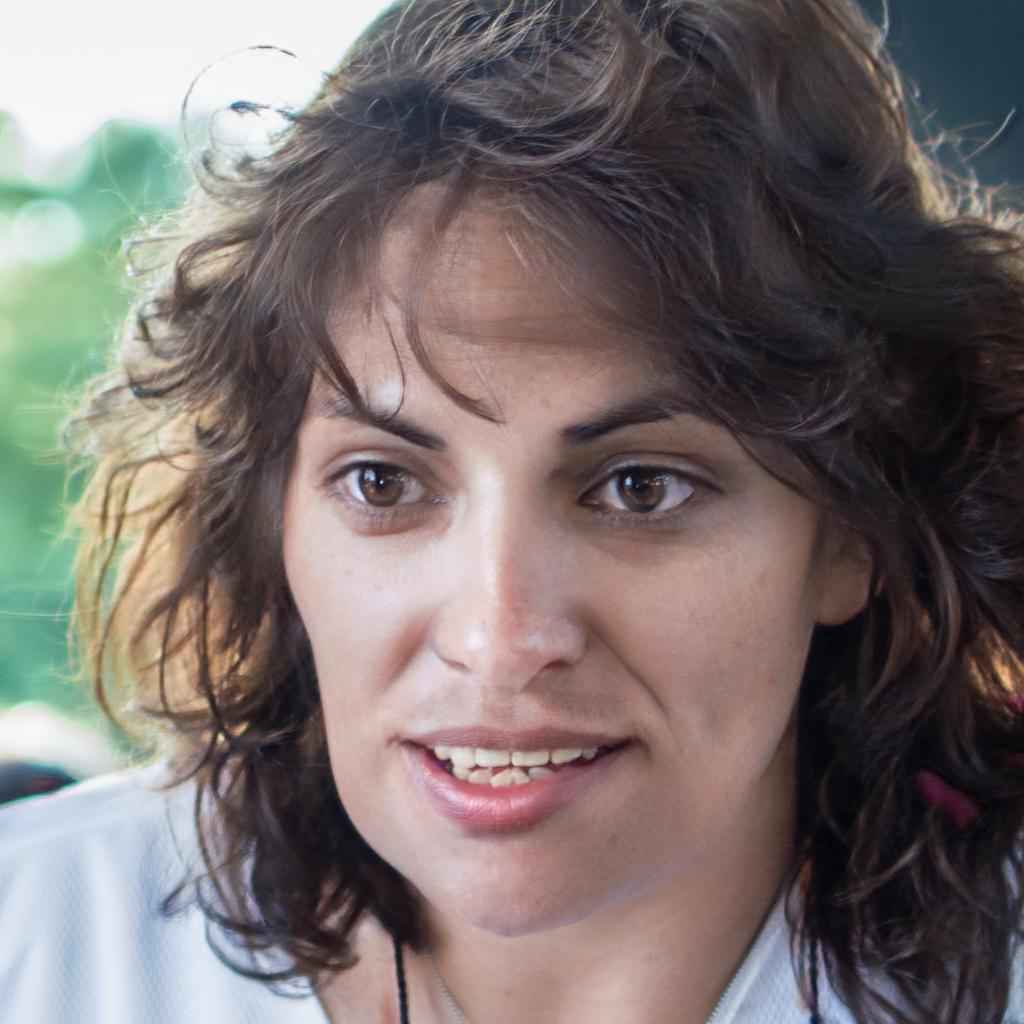} & 
\includegraphics[width=0.18\linewidth]{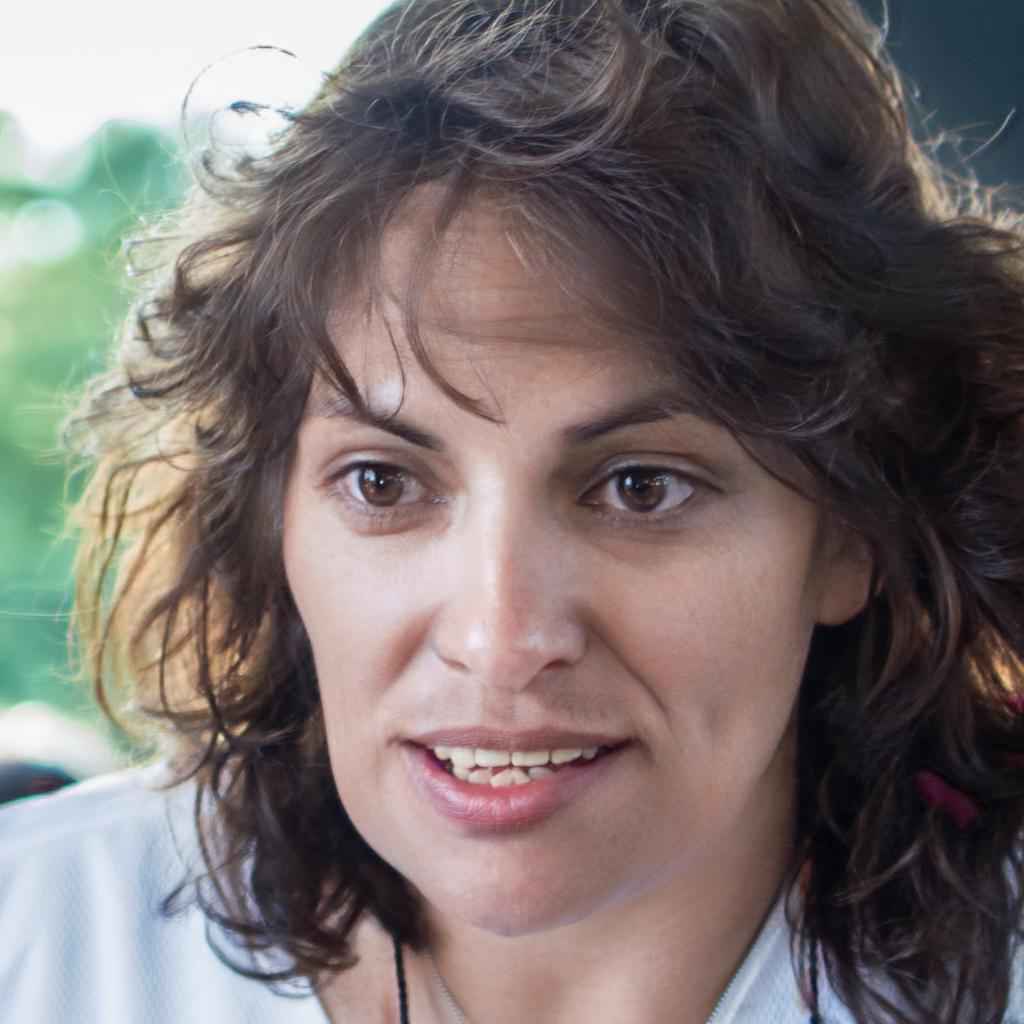} &
{\color{yellow}%
\setlength{\fboxsep}{0pt}%
\setlength{\fboxrule}{2pt}%
\fbox{\includegraphics[width=0.18\linewidth]{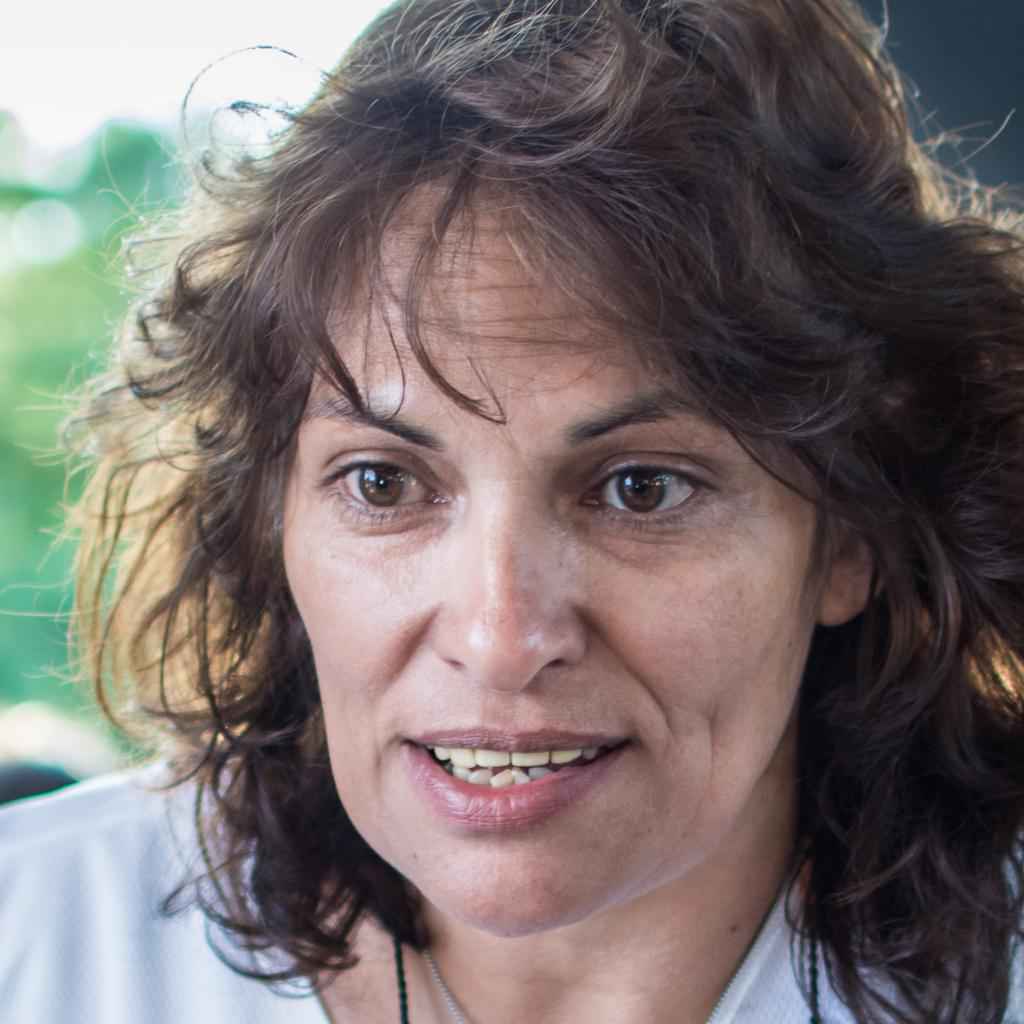}}} & 
\includegraphics[width=0.18\linewidth]{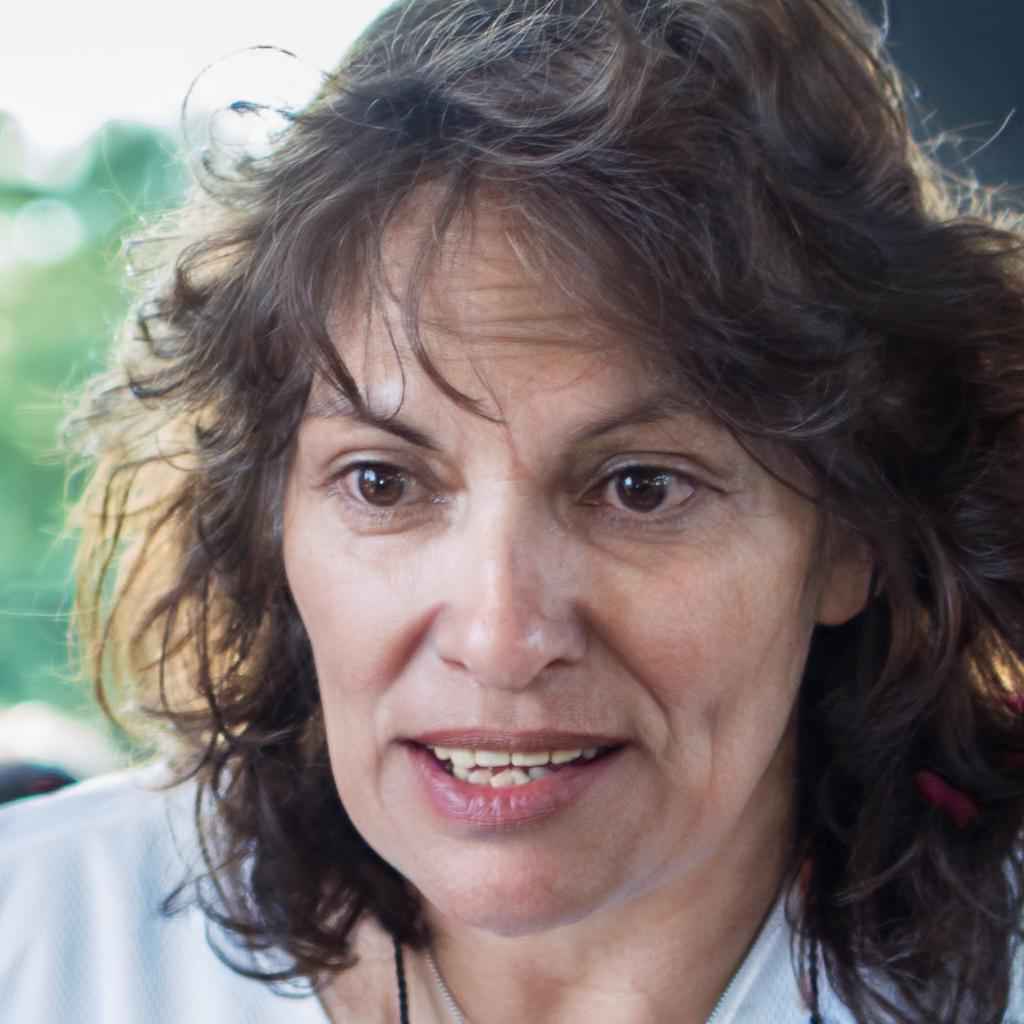} & 
\includegraphics[width=0.18\linewidth]{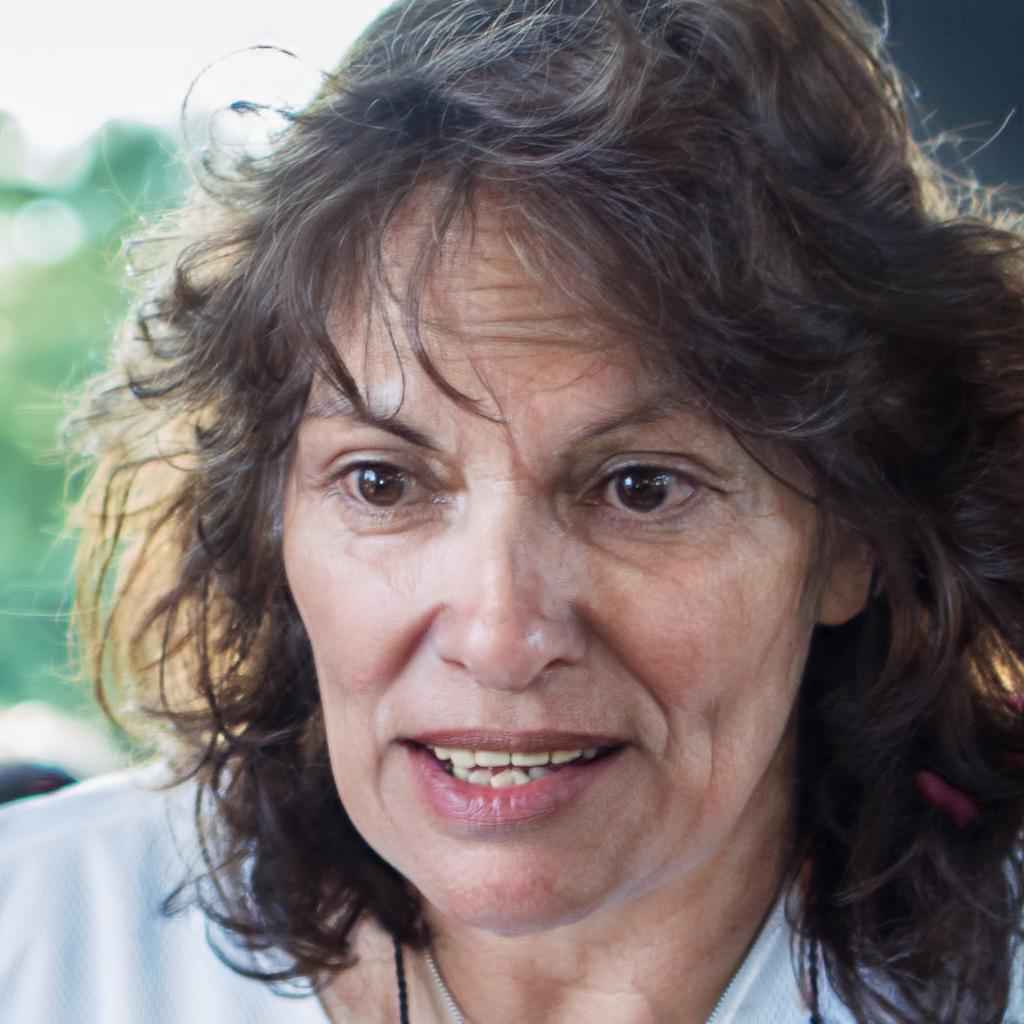} 
\\
\includegraphics[width=0.18\linewidth]{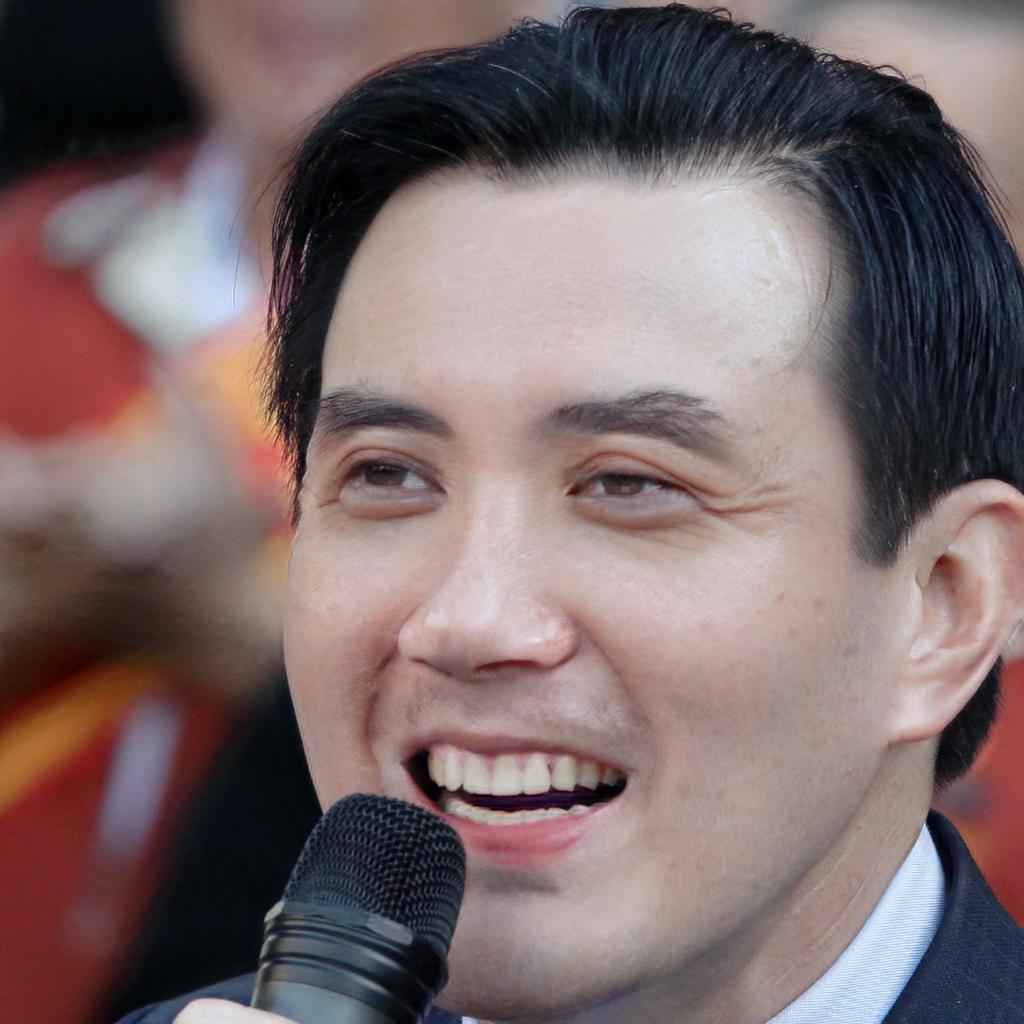} & 
\includegraphics[width=0.18\linewidth]{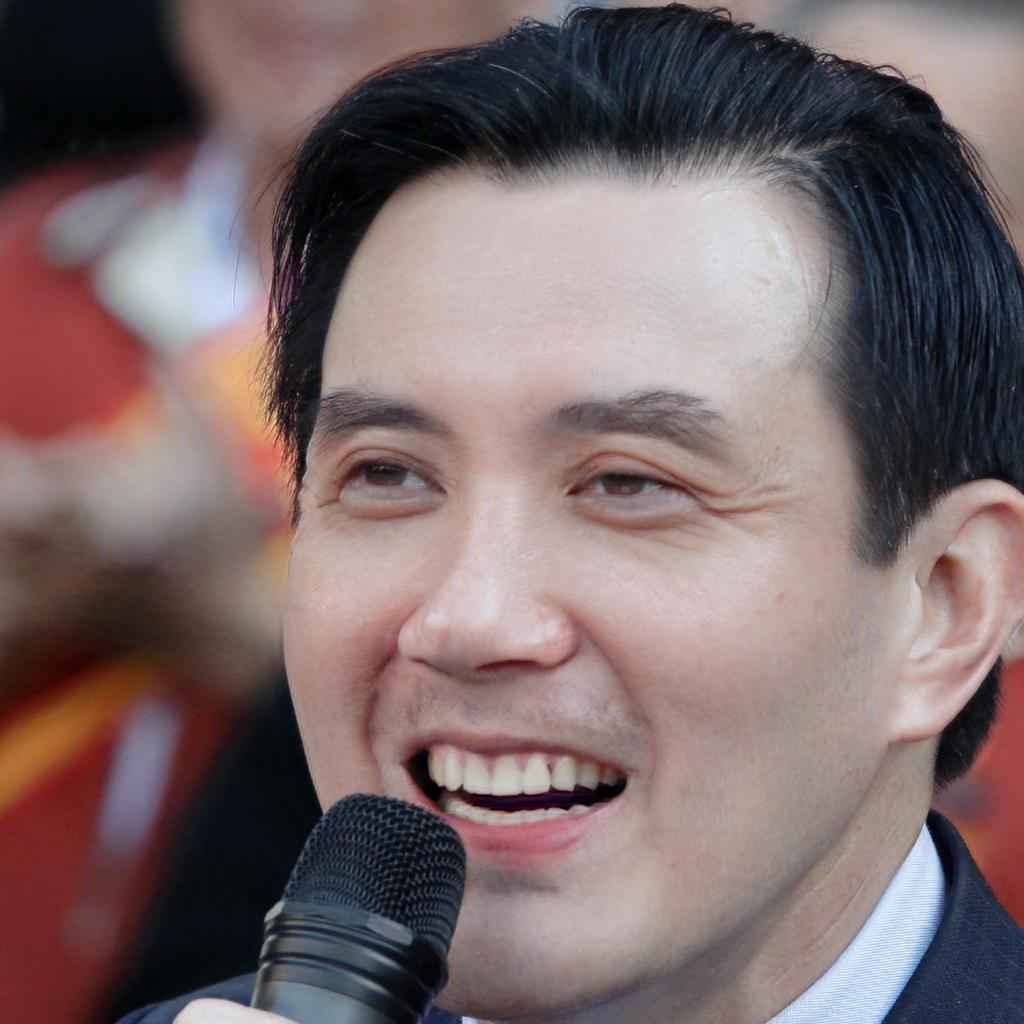} &
\includegraphics[width=0.18\linewidth]{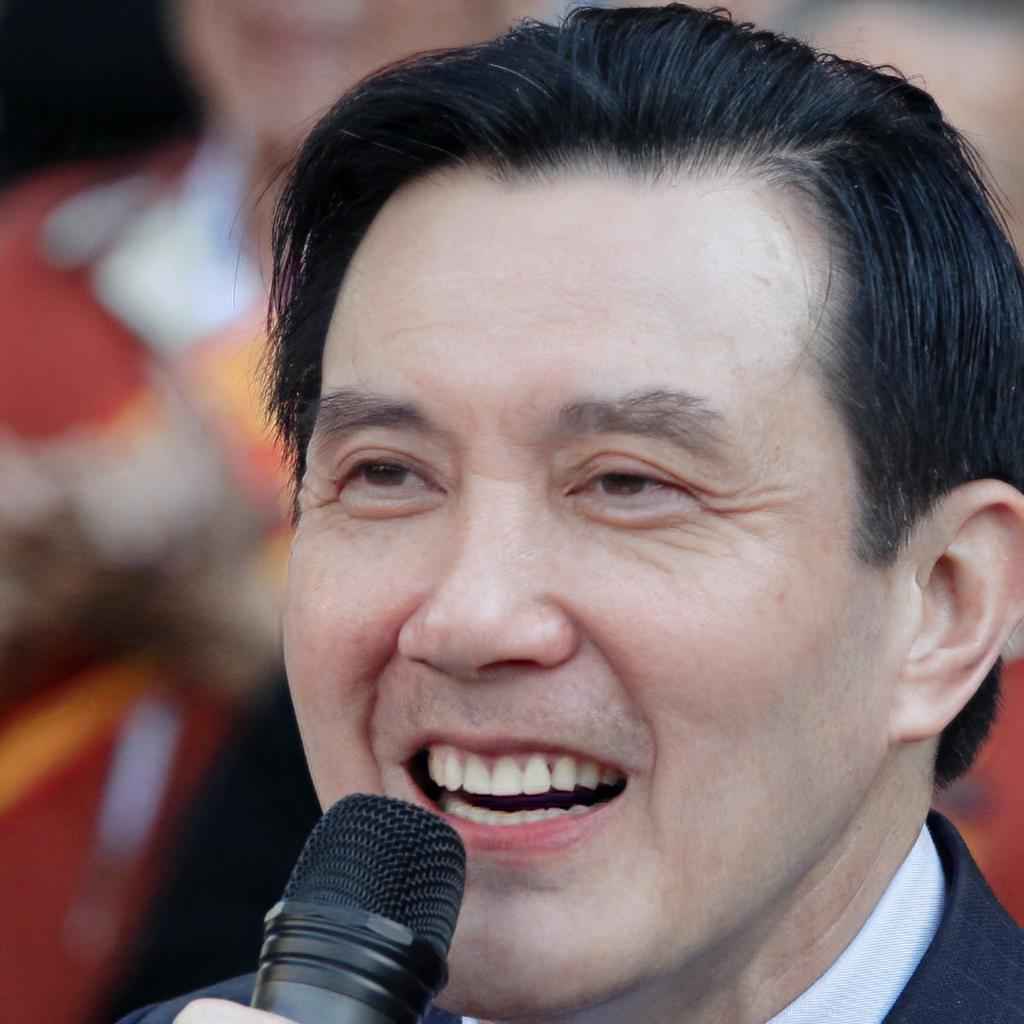} &
{\color{yellow}%
\setlength{\fboxsep}{0pt}%
\setlength{\fboxrule}{2pt}%
\fbox{\includegraphics[width=0.18\linewidth]{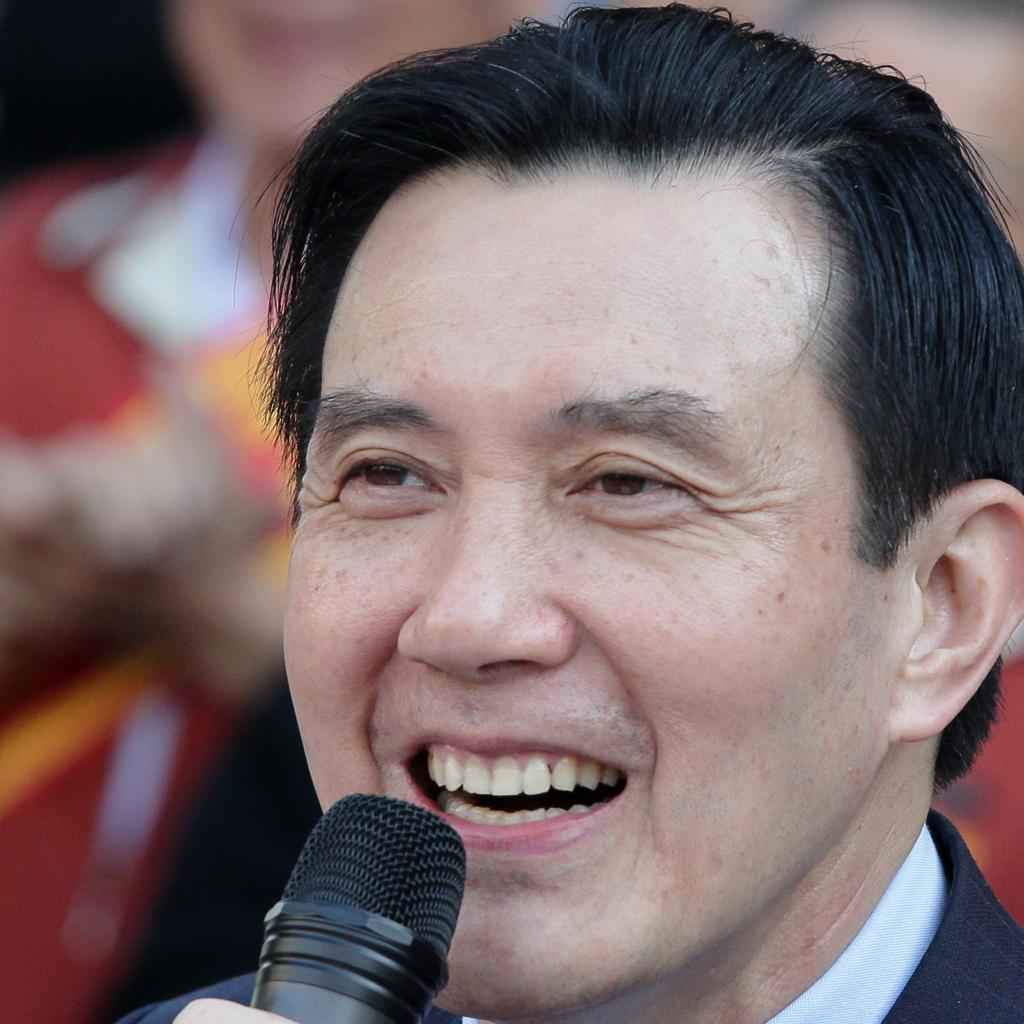}}} &  
\includegraphics[width=0.18\linewidth]{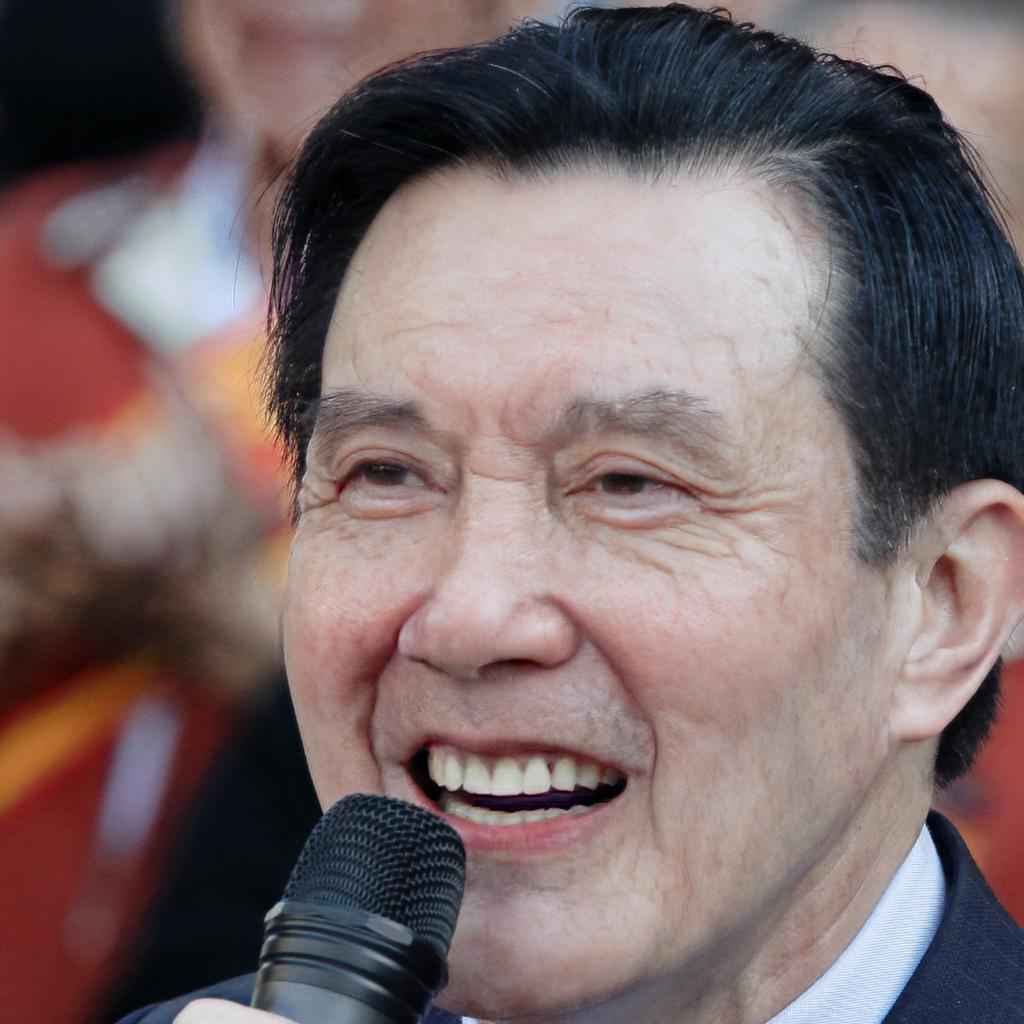} 
\\
\includegraphics[width=0.18\linewidth]{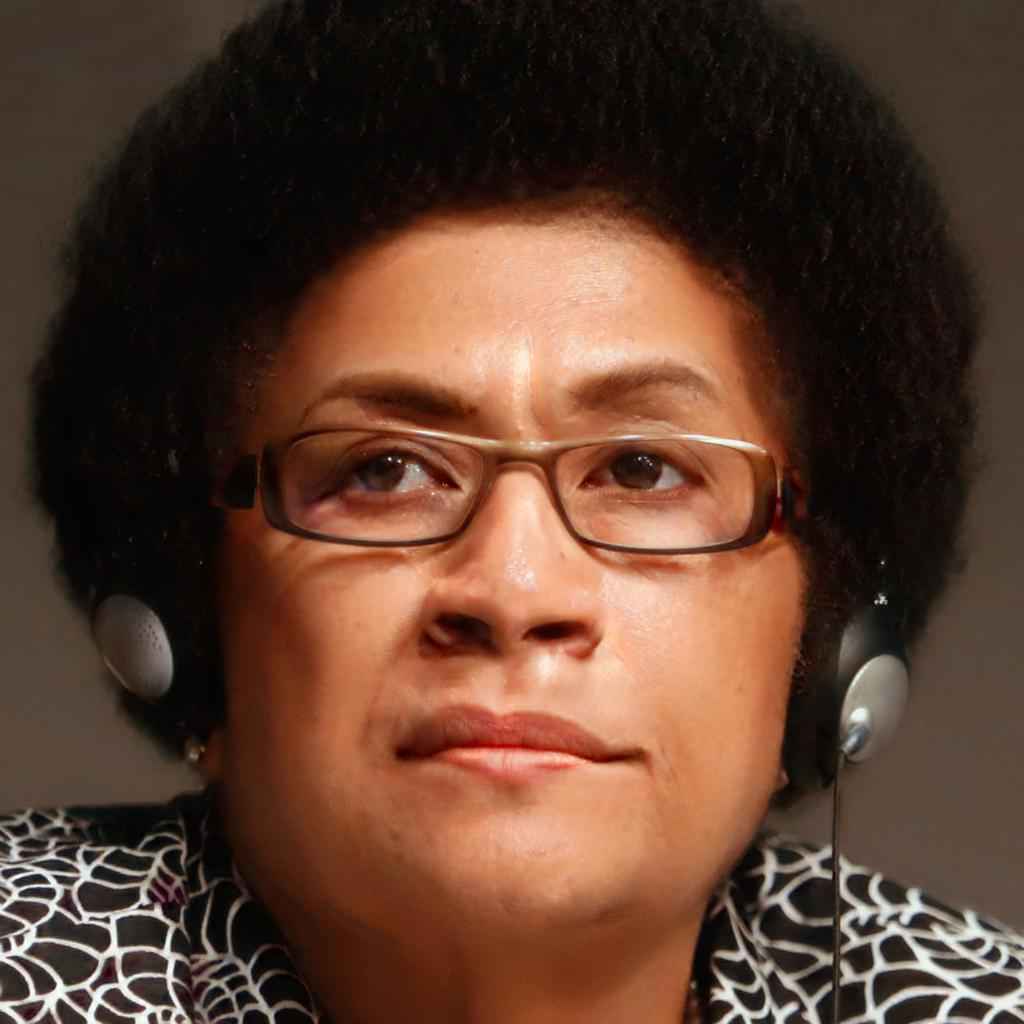} & 
\includegraphics[width=0.18\linewidth]{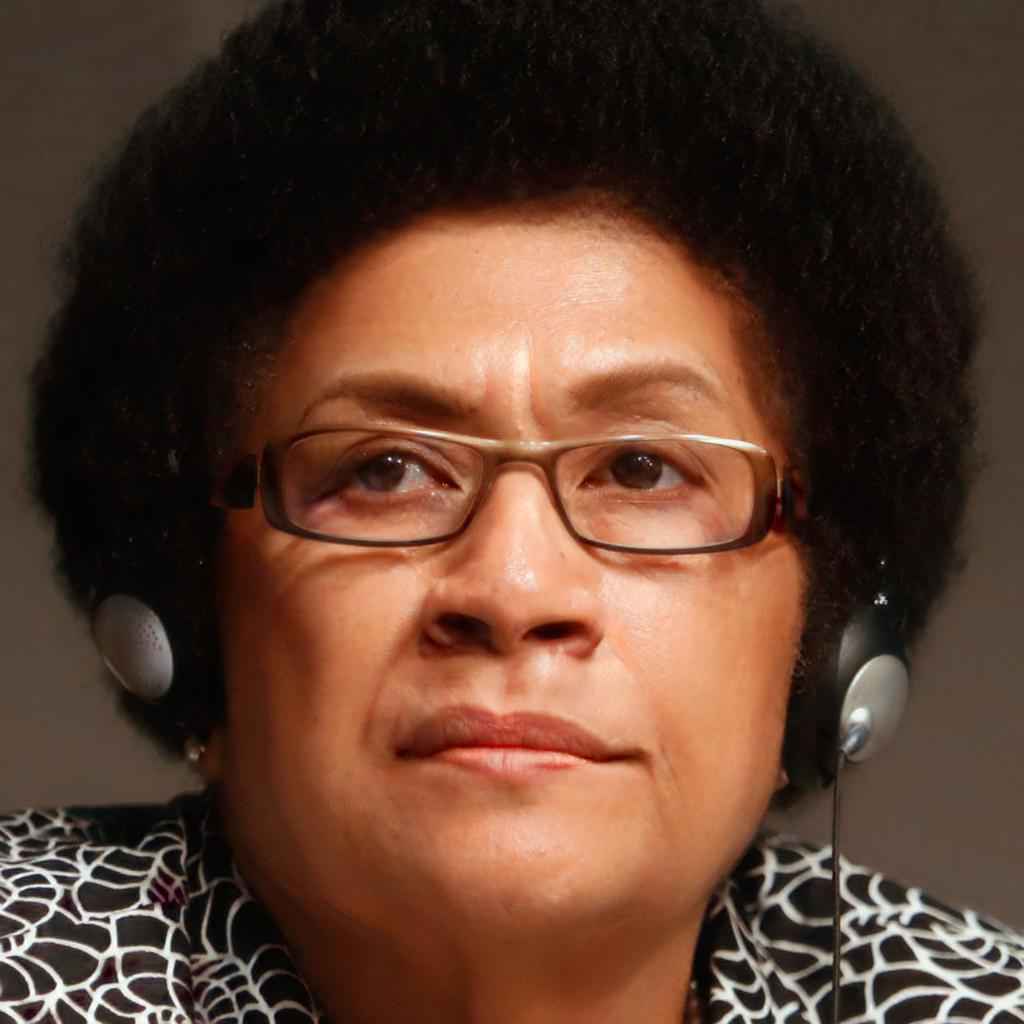} &
\includegraphics[width=0.18\linewidth]{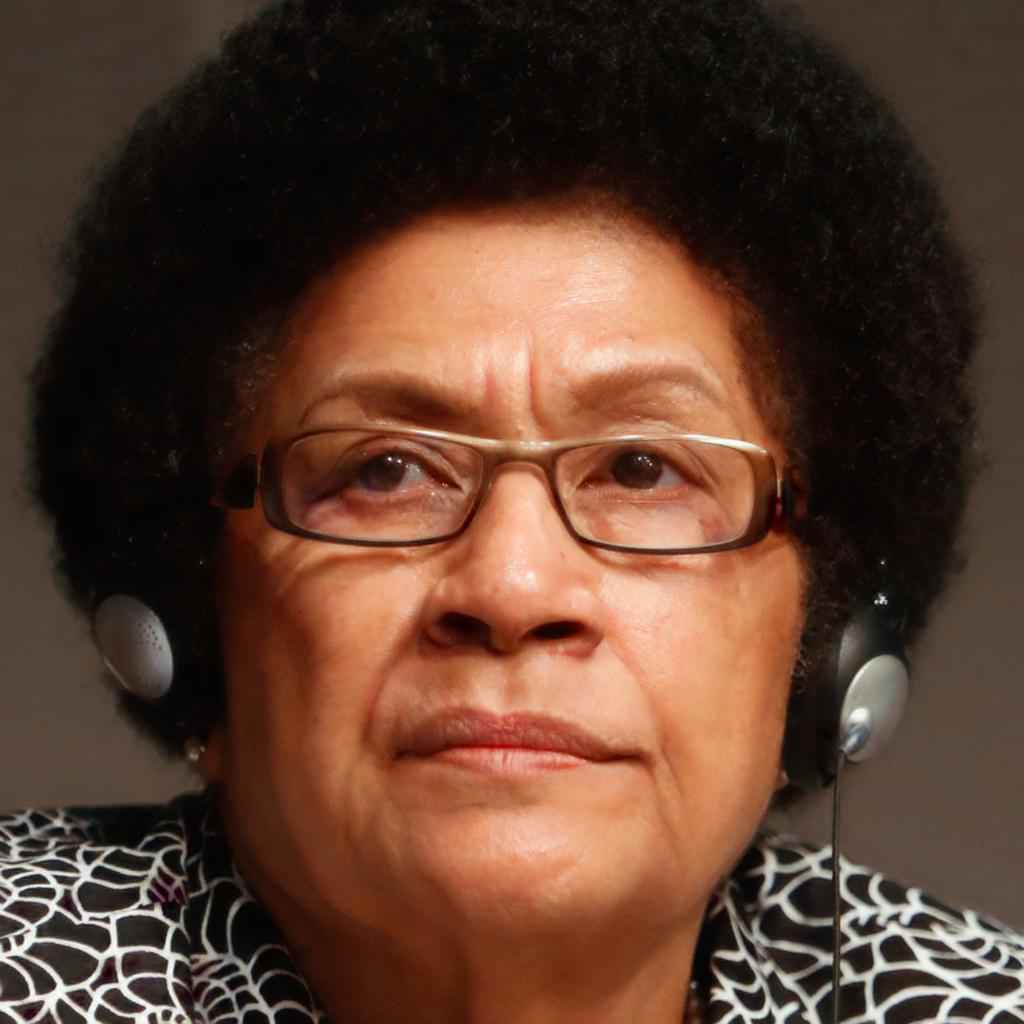} &
\includegraphics[width=0.18\linewidth]{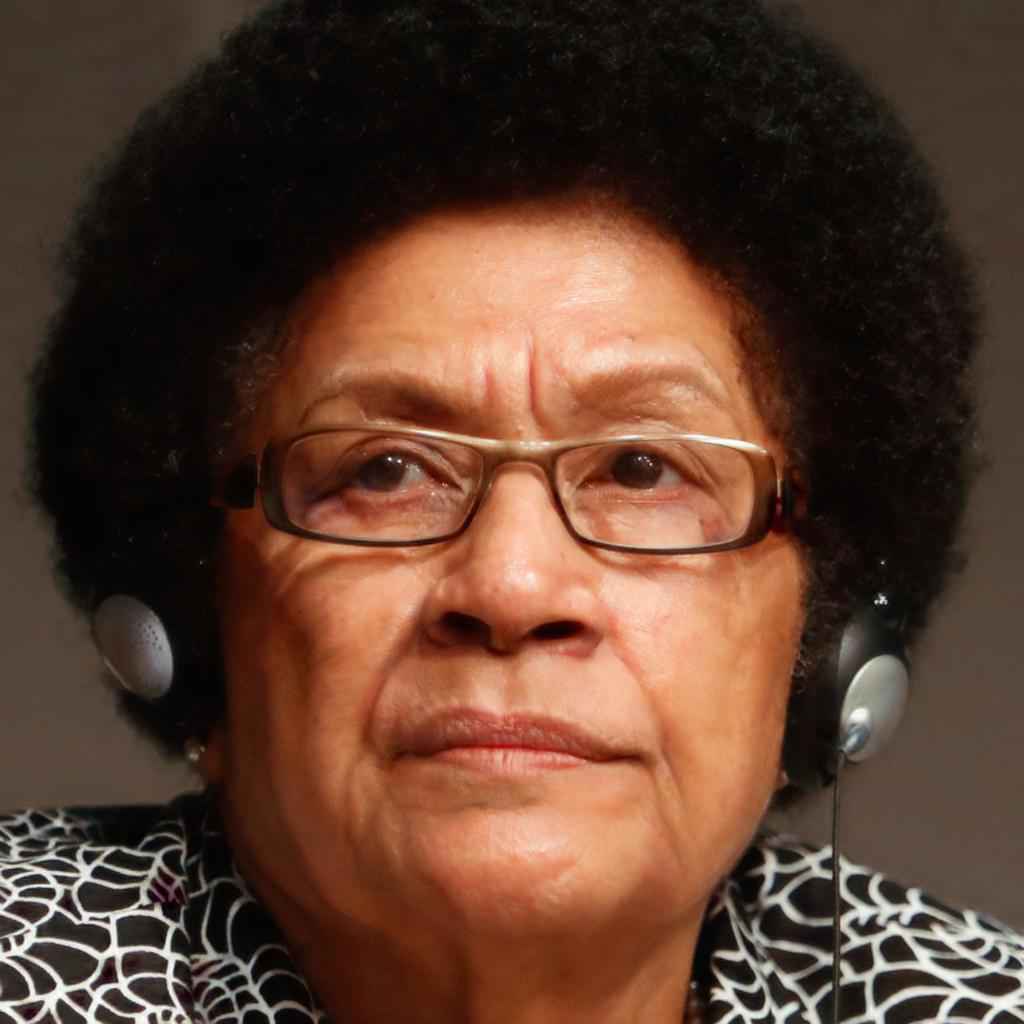} & 
{\color{yellow}%
\setlength{\fboxsep}{0pt}%
\setlength{\fboxrule}{2pt}%
\fbox{\includegraphics[width=0.18\linewidth]{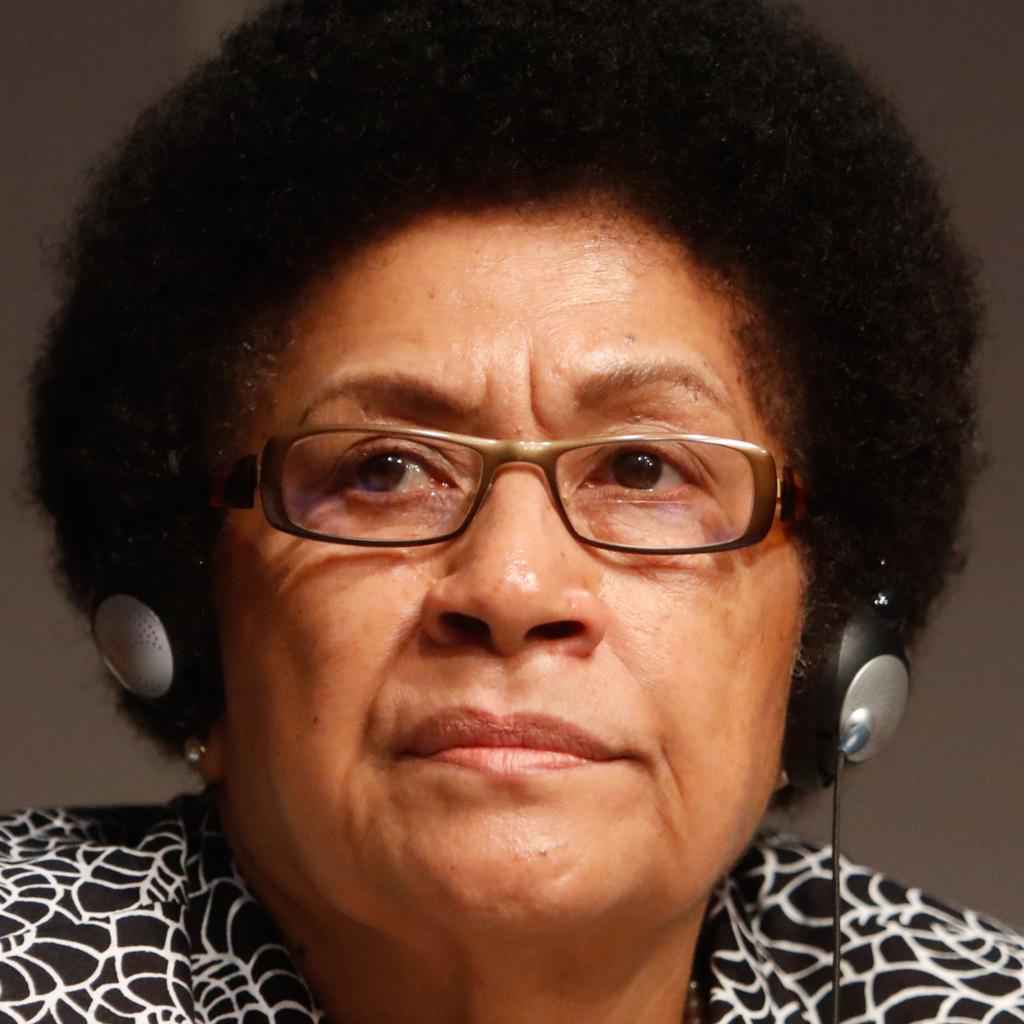}}} 
\\
\end{tabular}
\end{center}
\caption{\textbf{Age editing results on $ 1024 \times 1024$ images on FFHQ \cite{karras2019style}}. On each row, the yellow frame indicates the original image. Each column corresponds to a target age of: $25$, $35$, $45$, $55$, $65$. Our approach yields visually satisfying results without introducing significant artifacts. Only age relevant features are modified, while the identity, haircut, emotion and background are perfectly preserved. 
}
\label{1024_1}
\end{figure*}
\begin{figure*}[t]
\begin{center}
\scriptsize
\setlength{\tabcolsep}{1pt}
\begin{tabular}{ccccccccccc}
\includegraphics[width=0.085\linewidth]{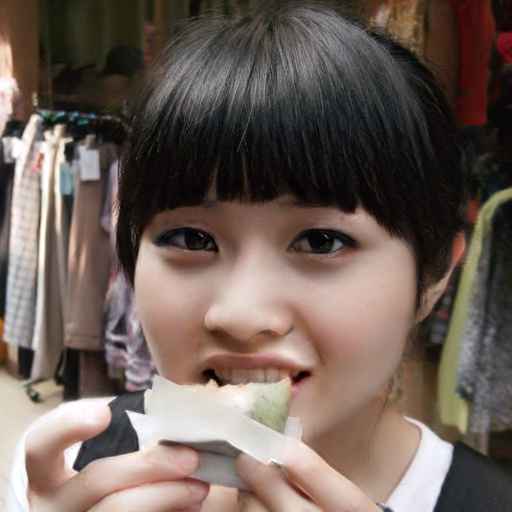} & 
\includegraphics[width=0.085\linewidth]{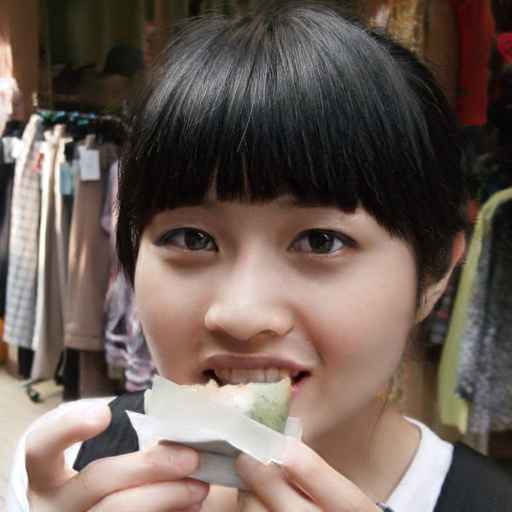} & 
\includegraphics[width=0.085\linewidth]{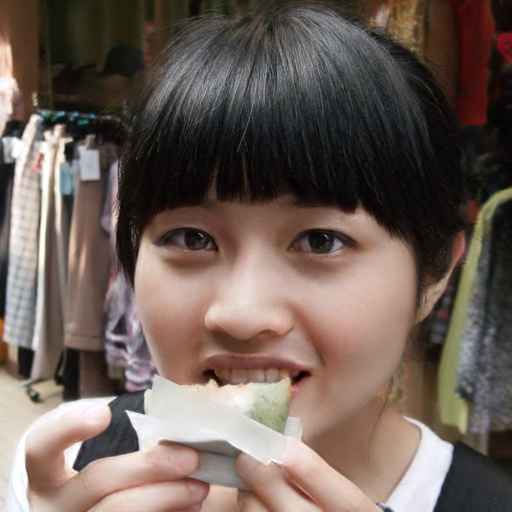} & 
\includegraphics[width=0.085\linewidth]{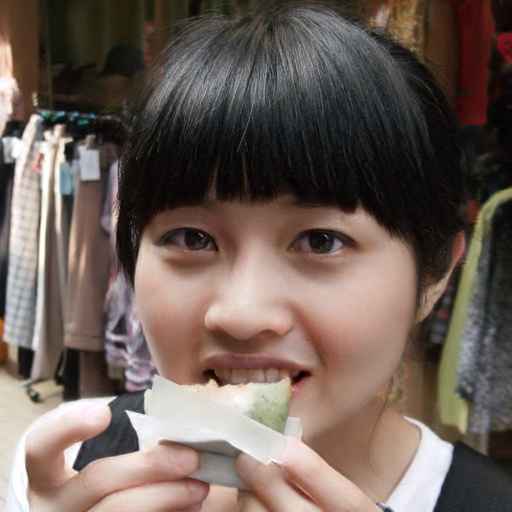} & 
\includegraphics[width=0.085\linewidth]{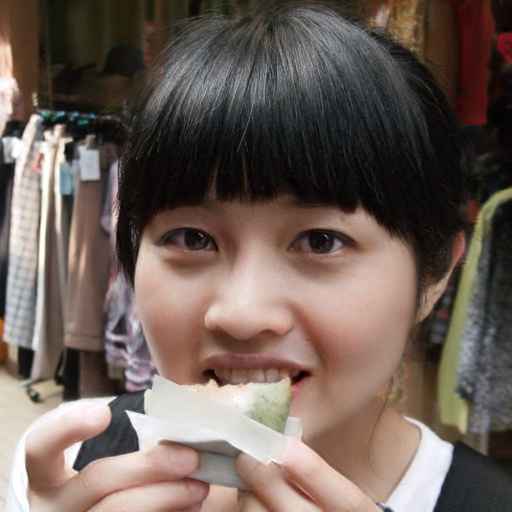} &
\includegraphics[width=0.085\linewidth]{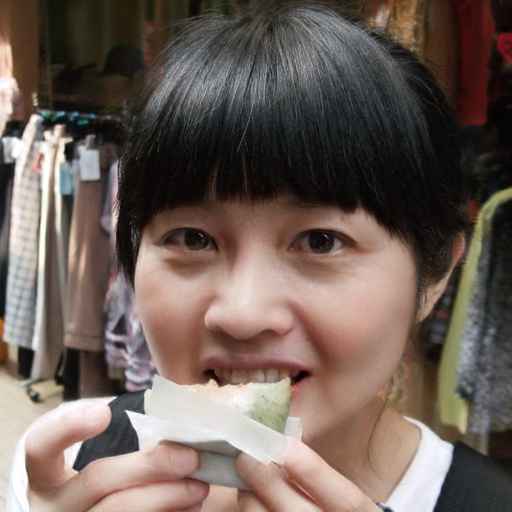} &
\includegraphics[width=0.085\linewidth]{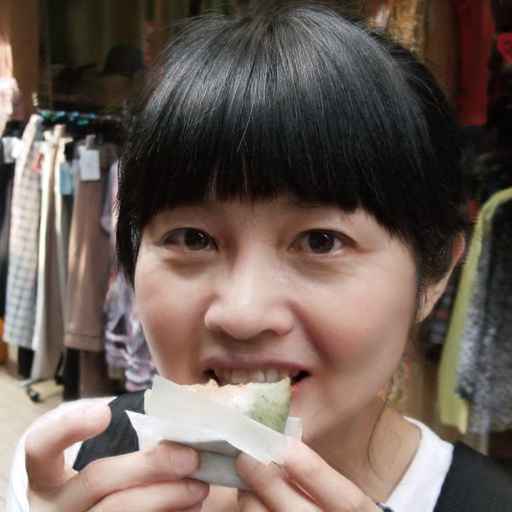} & 
\includegraphics[width=0.085\linewidth]{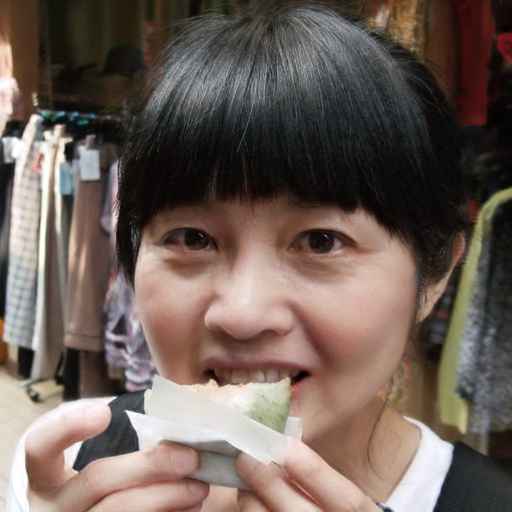} & 
\includegraphics[width=0.085\linewidth]{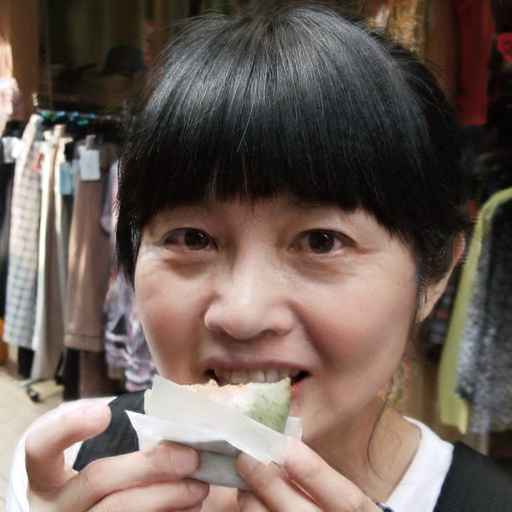} & 
\includegraphics[width=0.085\linewidth]{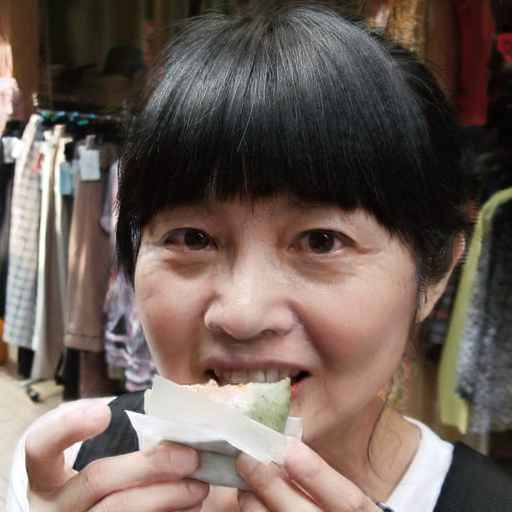} & 
\includegraphics[width=0.085\linewidth]{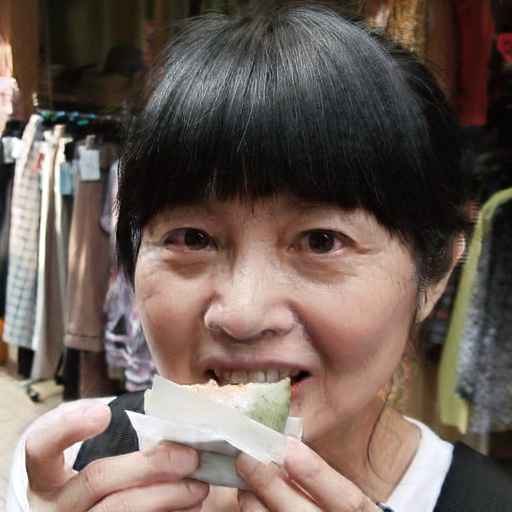} 
\\
\includegraphics[width=0.085\linewidth]{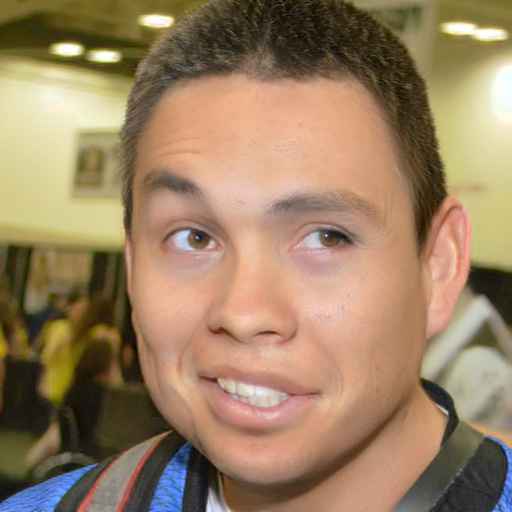} & 
\includegraphics[width=0.085\linewidth]{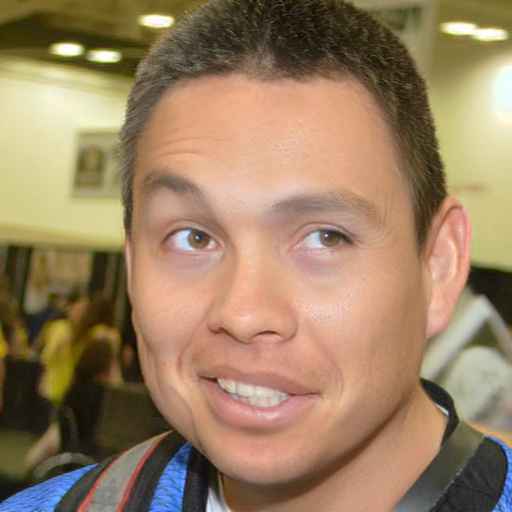} & 
\includegraphics[width=0.085\linewidth]{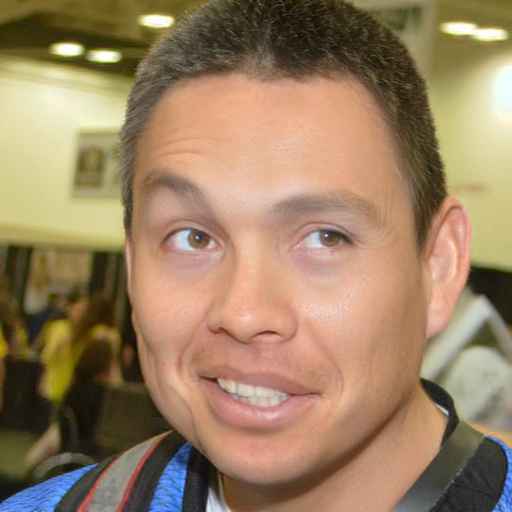} & 
\includegraphics[width=0.085\linewidth]{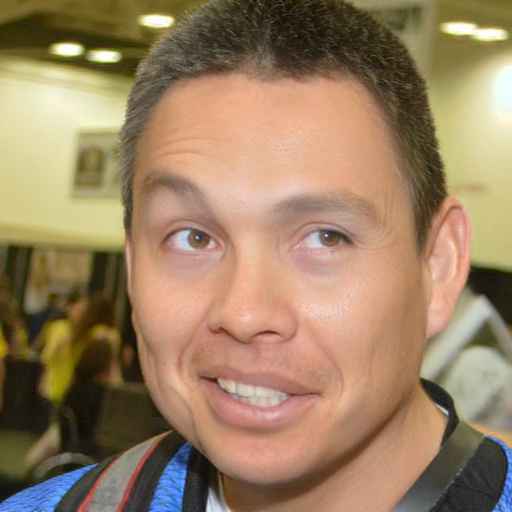} & 
\includegraphics[width=0.085\linewidth]{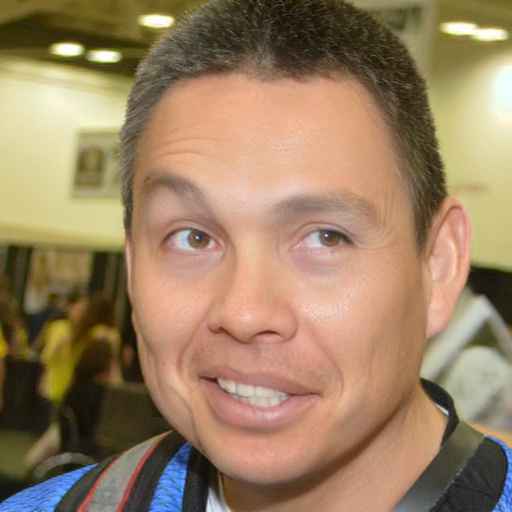} &
\includegraphics[width=0.085\linewidth]{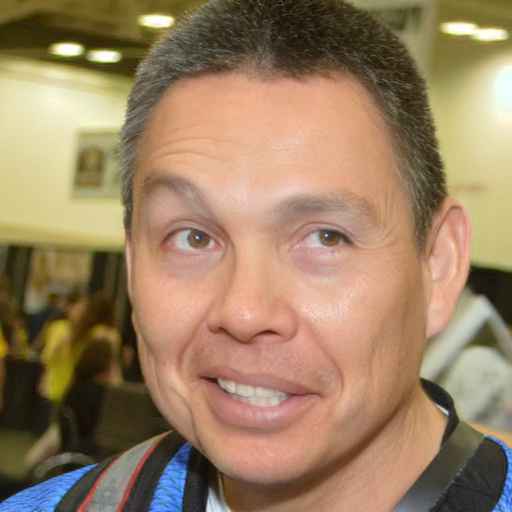} &
\includegraphics[width=0.085\linewidth]{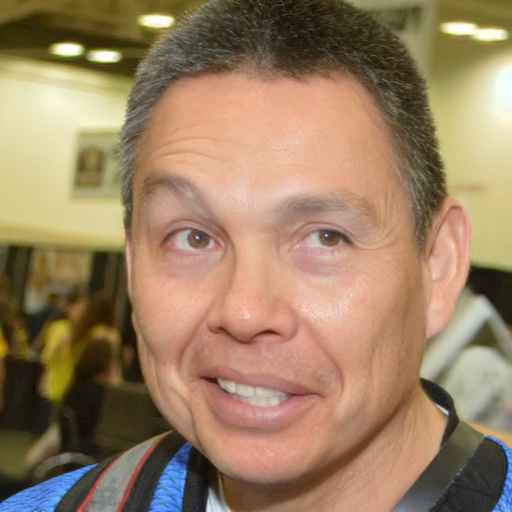} & 
\includegraphics[width=0.085\linewidth]{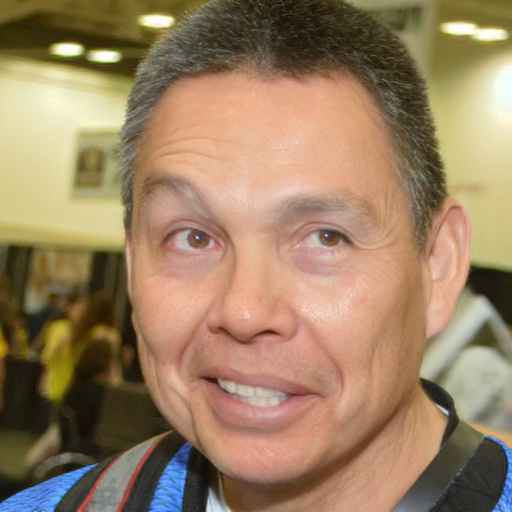} & 
\includegraphics[width=0.085\linewidth]{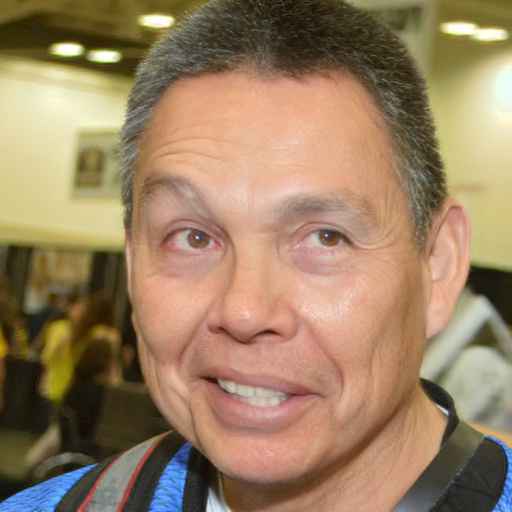} & 
\includegraphics[width=0.085\linewidth]{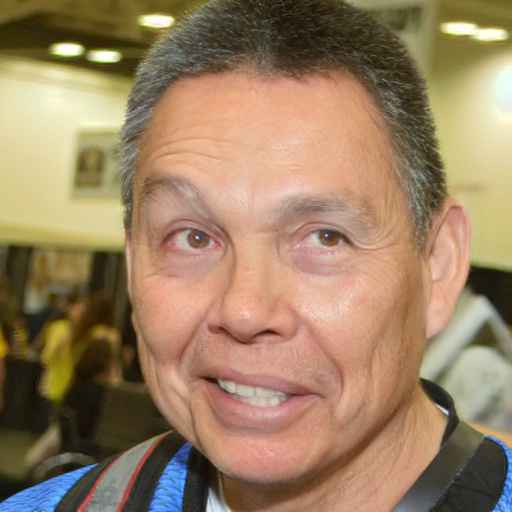} & 
\includegraphics[width=0.085\linewidth]{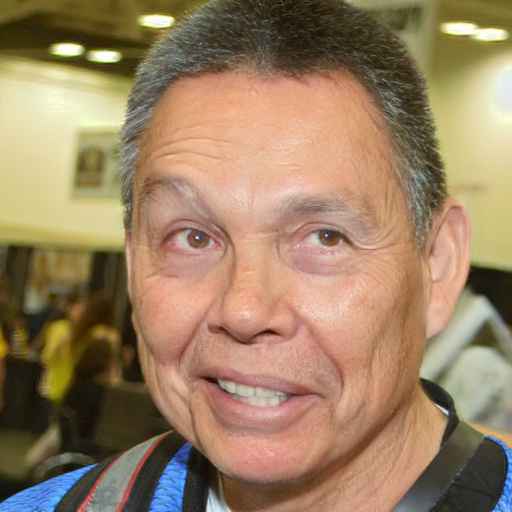} 
\end{tabular}
\end{center}
\caption{\textbf{Continuous face age editing results on FFHQ \cite{karras2019style}}. %
As can be observed, the difference between two adjacent results is nearly invisible, which demonstrates the smoothness of the aging process.}
\label{range}
\end{figure*}

Figure \ref{1024_1} presents age editing results on $1024 \times 1024$ input images in different age groups. Our approach yields visually satisfying results with sharp details (best viewed when zooming on the results) and without introducing significant artifacts. Only the age relevant facial features are modified, while the identity, haircut, emotion and background are well preserved. This is all the more satisfying that no mask has been used to isolate the face from the rest of the image. 
Figure \ref{range} presents age editing results with a smooth evolution of the target age. The difference between two adjacent results is nearly invisible, which illustrates the smoothness of the aging process. 

\begin{figure*}[t]
\begin{center}
\scriptsize
\setlength{\tabcolsep}{1pt}
\begin{tabular}{cc}
\begin{tabular}{ccccc}
&Input&31-40&41-50&51+
\\
\rotatebox{90}{IPCGAN}&
\includegraphics[width=0.115\linewidth]{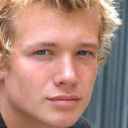} &
\includegraphics[width=0.115\linewidth]{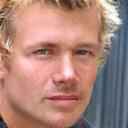} & 
\includegraphics[width=0.115\linewidth]{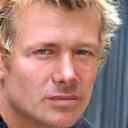} & 
\includegraphics[width=0.115\linewidth]{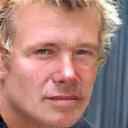}
\\
\rotatebox{90}{Ours}&
\includegraphics[width=0.115\linewidth]{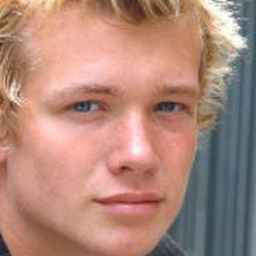} &
\includegraphics[width=0.115\linewidth]{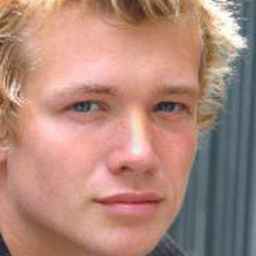} &
\includegraphics[width=0.115\linewidth]{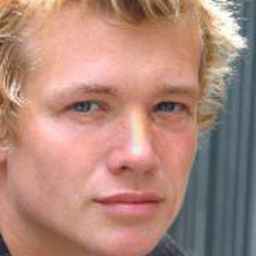} &
\includegraphics[width=0.115\linewidth]{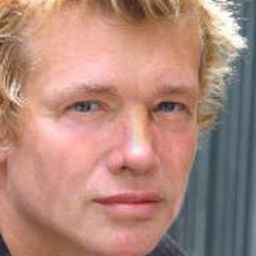} \\
\end{tabular}
&
\begin{tabular}{cccc}
Input&31-40&41-50&51+
\\
\includegraphics[width=0.115\linewidth]{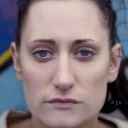} &
\includegraphics[width=0.115\linewidth]{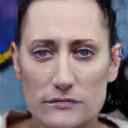} & 
\includegraphics[width=0.115\linewidth]{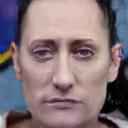} & 
\includegraphics[width=0.115\linewidth]{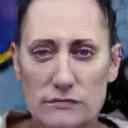}
\\
\includegraphics[width=0.115\linewidth]{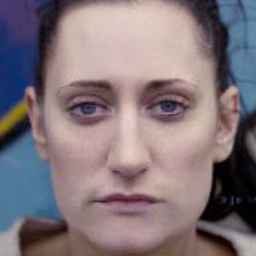} &
\includegraphics[width=0.115\linewidth]{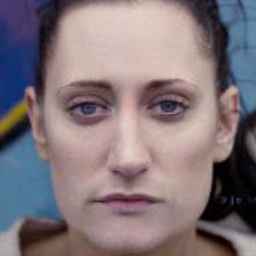} &
\includegraphics[width=0.115\linewidth]{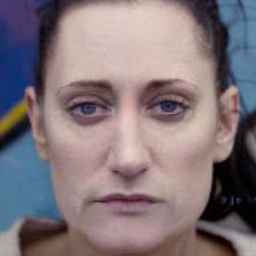} &
\includegraphics[width=0.115\linewidth]{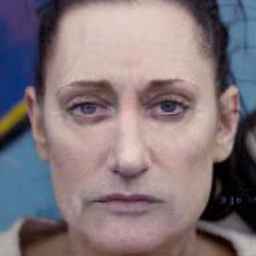} \\
\end{tabular}
\\
\multicolumn{2}{c}{(a) Comparison with IPCGAN.}
\\
\end{tabular}
\begin{tabular}{ccc|cc|cc|cc}
\\
&Input&51+&Input&51+&Input&51+&Input&51+
\\
\rotatebox{90}{PAGGAN}&
\includegraphics[width=0.115\linewidth]{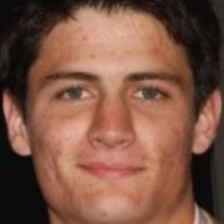} &
\includegraphics[width=0.115\linewidth]{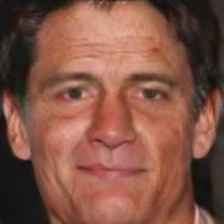} &
\includegraphics[width=0.115\linewidth]{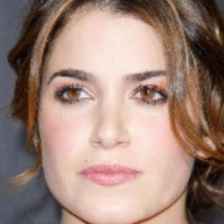}  &
\includegraphics[width=0.115\linewidth]{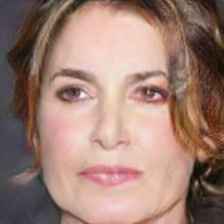}  &
\includegraphics[width=0.115\linewidth]{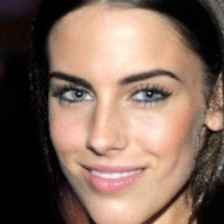} &
\includegraphics[width=0.115\linewidth]{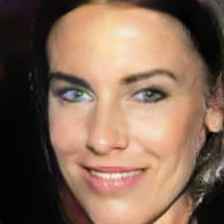} &
\includegraphics[width=0.115\linewidth]{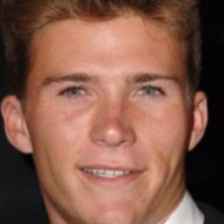}  &
\includegraphics[width=0.115\linewidth]{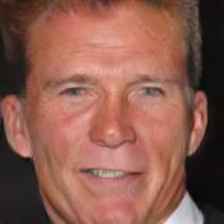} 
\\
\rotatebox{90}{Ours}&
\includegraphics[width=0.115\linewidth]{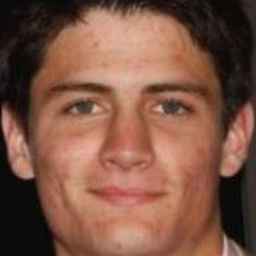}  &
\includegraphics[width=0.115\linewidth]{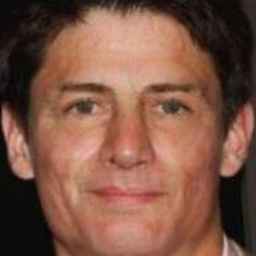}  &
\includegraphics[width=0.115\linewidth]{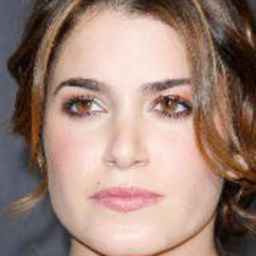}  &
\includegraphics[width=0.115\linewidth]{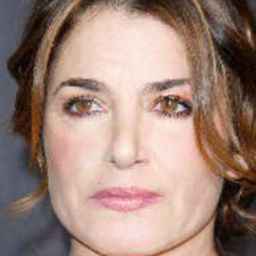}  &
\includegraphics[width=0.115\linewidth]{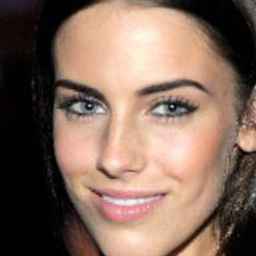}  &
\includegraphics[width=0.115\linewidth]{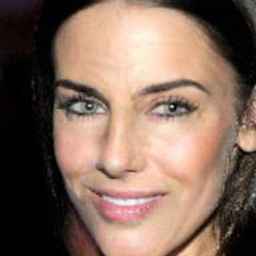}  &
\includegraphics[width=0.115\linewidth]{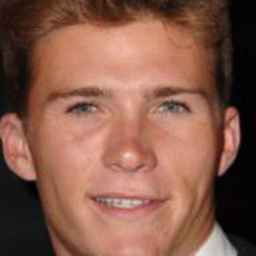}  &
\includegraphics[width=0.115\linewidth]{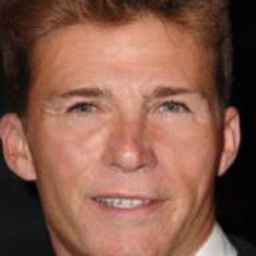}  
\\
\multicolumn{9}{c}{(b) Comparison with PAGGAN.}
\\
\end{tabular}
\end{center}
\caption{\textbf{Comparison with IPCGAN~\cite{wang2018face} and PAGGAN~\cite{yang2018learning} on CACD~\cite{chen2014cross}}. For each subfigure in (a), the top row corresponds to the aging results of IPCGAN. The second row shows the images generated by our method. For each subfigure in (b), the top row corresponds to the aging results of PAGGAN. The second row shows the images generated by our method.}
\label{compare_ipcgan}
\end{figure*}
We compare our method to the two most recent state-of-the-art methods on face aging for which the official codes are released - IPCGAN~\cite{wang2018face} and PAGGAN~\cite{yang2018learning}. We also compare our results to those obtained with FaderNet~\cite{lample2017fader}, which allows one to manipulate several facial attributes including the age.  

Figure \ref{compare_ipcgan} present the face aging results of IPCGAN, PAGGAN and our method on CACD~\cite{chen2014cross}. The output size of each method is: $128 \times 128$ for IPCGAN, $224 \times 224$ for PAGGAN, $256 \times 256$ for our method. IPCGAN generates satisfying aging results and preserves well the identity of input images. However, as can be seen e.g. in Figure \ref{compare_ipcgan}(a) row 1 column 4, the generated image presents noticeable artifacts. PAGGAN generates impressive aging effects but also introduce colored artifacts as shown in Figure \ref{compare_ipcgan}(b) row 1 column 2. IPCGAN and PAGGAN both degrade the quality of input images. Our method is able to generate consistent aging effects, and preserve well the fine details of the input images.

\begin{figure*}[t]
\centering
\scriptsize
\setlength{\tabcolsep}{2pt}
\begin{tabular}{ccccc}
Input ($1024^2$)&Fader ($256^2$)&PAGGAN ($224^2$)&IPCGAN ($128^2$)&Ours ($1024^2$)\\
\includegraphics[width=0.18\linewidth]{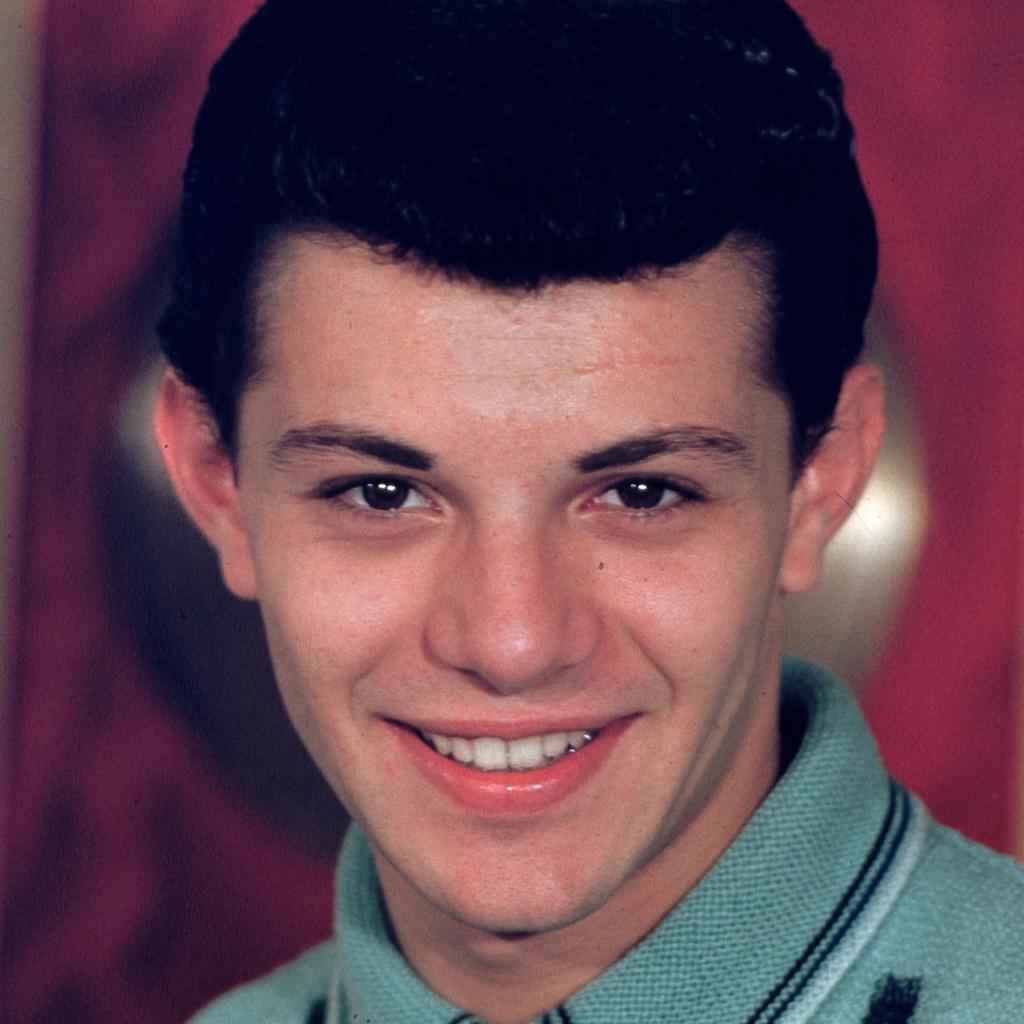} &
\includegraphics[width=0.18\linewidth]{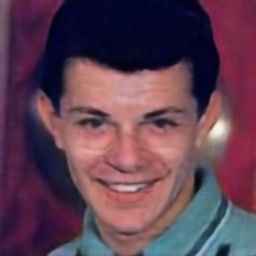} &
\includegraphics[width=0.18\linewidth]{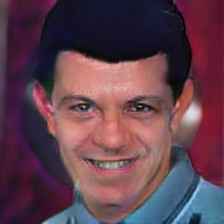} &
\includegraphics[width=0.18\linewidth]{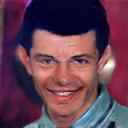} &
\includegraphics[width=0.18\linewidth]{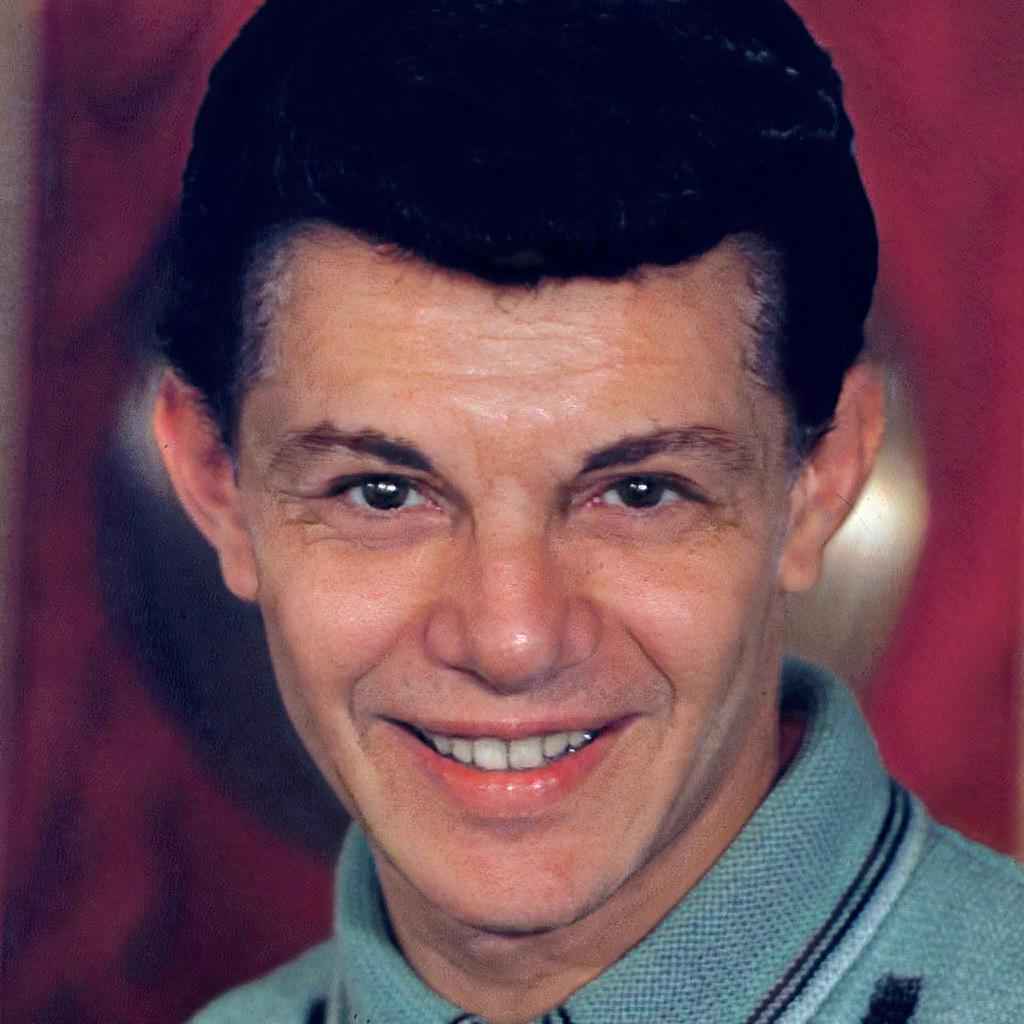} \\
\includegraphics[width=0.18\linewidth]{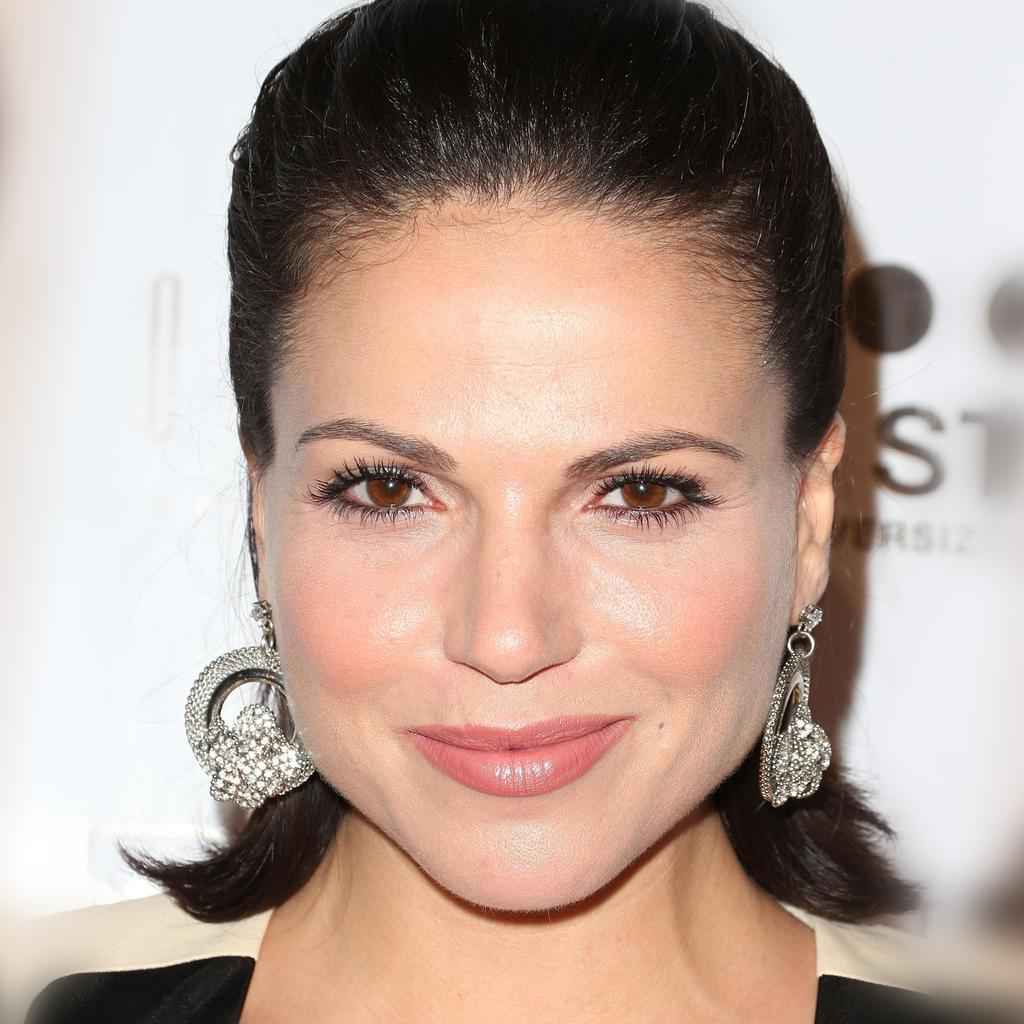} &
\includegraphics[width=0.18\linewidth]{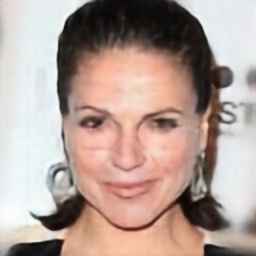} &
\includegraphics[width=0.18\linewidth]{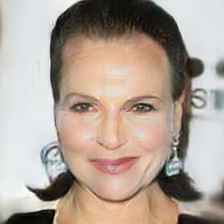} &
\includegraphics[width=0.18\linewidth]{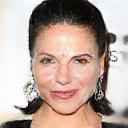} &
\includegraphics[width=0.18\linewidth]{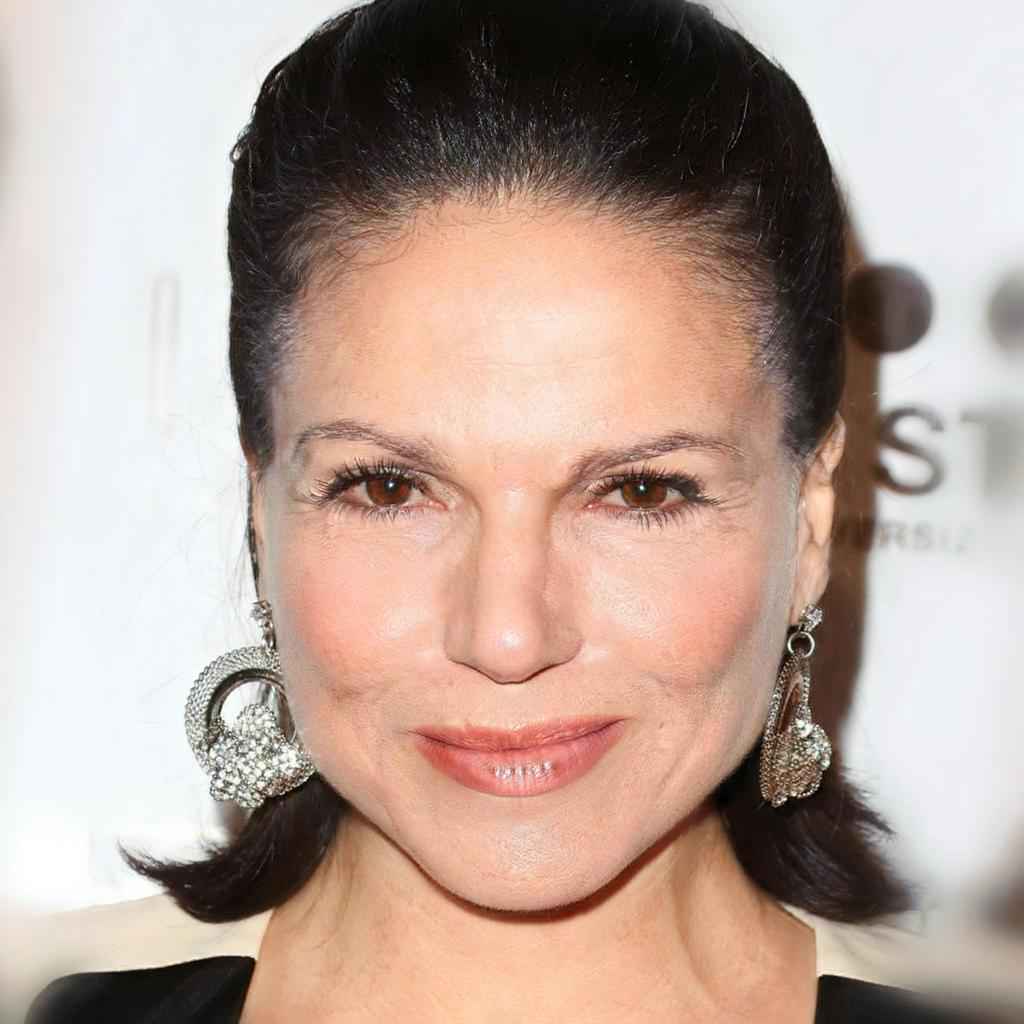} \\
\includegraphics[width=0.18\linewidth]{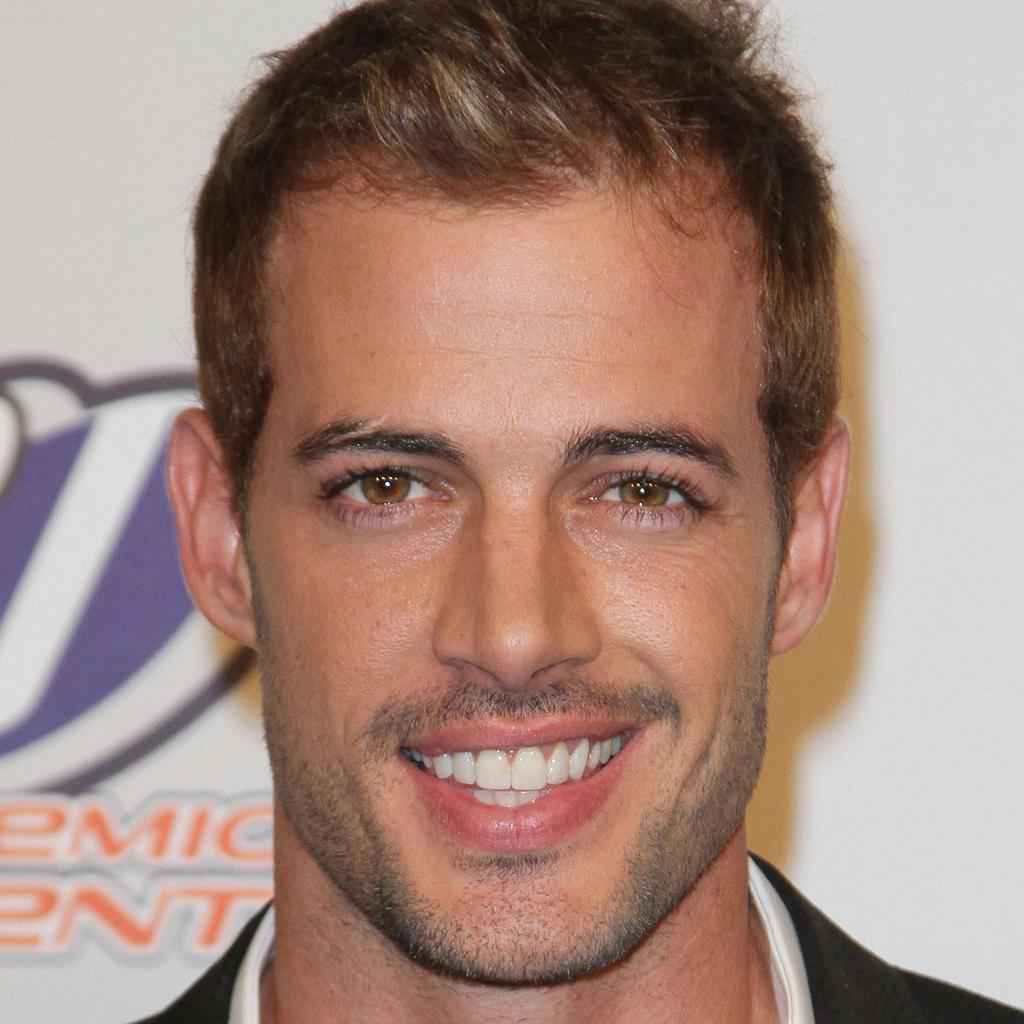} &
\includegraphics[width=0.18\linewidth]{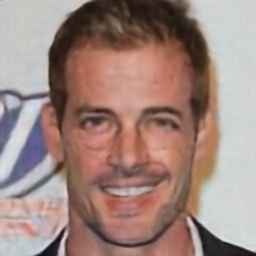} &
\includegraphics[width=0.18\linewidth]{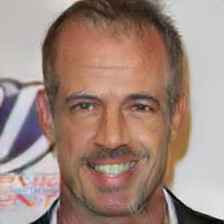} &
\includegraphics[width=0.18\linewidth]{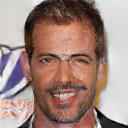} &
\includegraphics[width=0.18\linewidth]{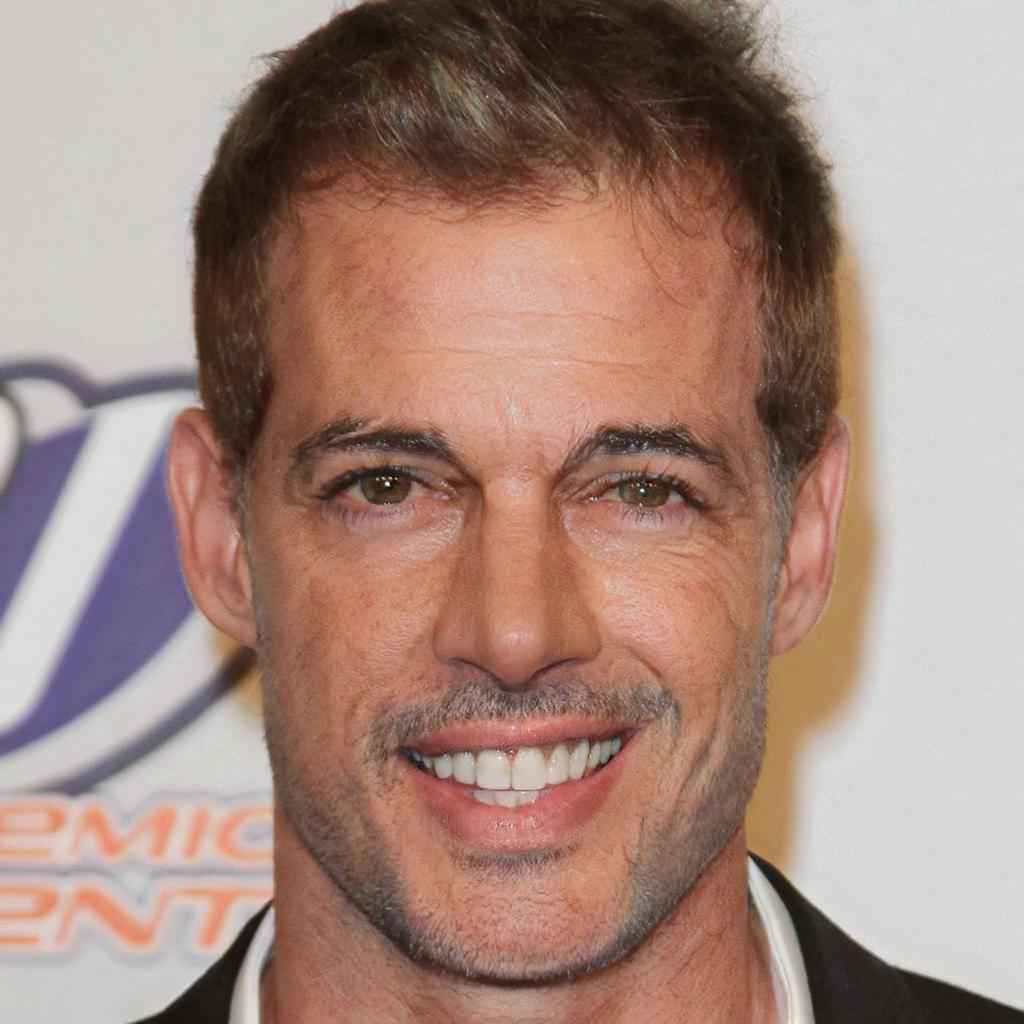} \\
\includegraphics[width=0.18\linewidth]{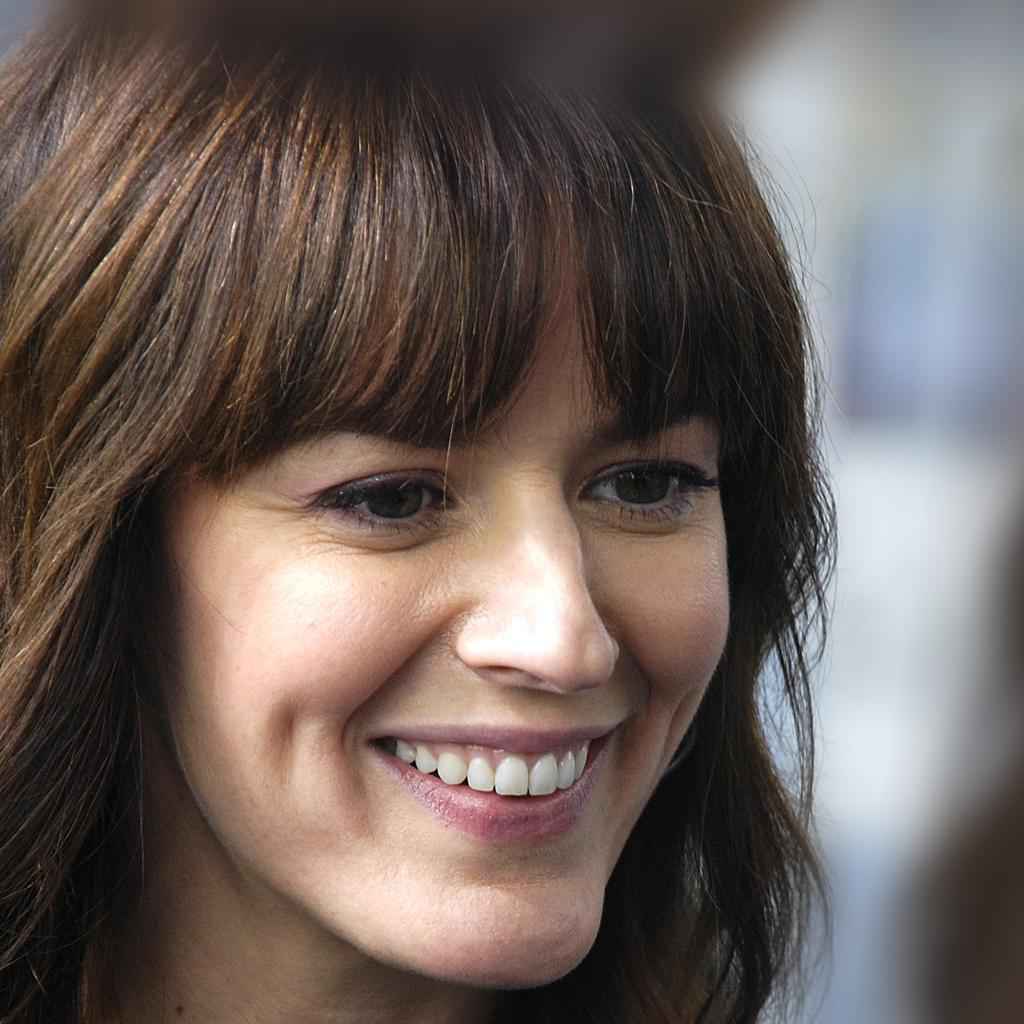} &
\includegraphics[width=0.18\linewidth]{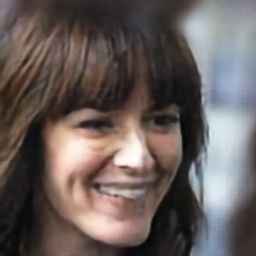} &
\includegraphics[width=0.18\linewidth]{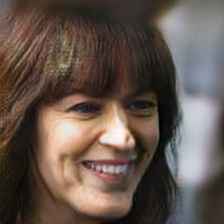} &
\includegraphics[width=0.18\linewidth]{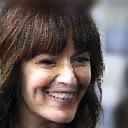} &
\includegraphics[width=0.18\linewidth]{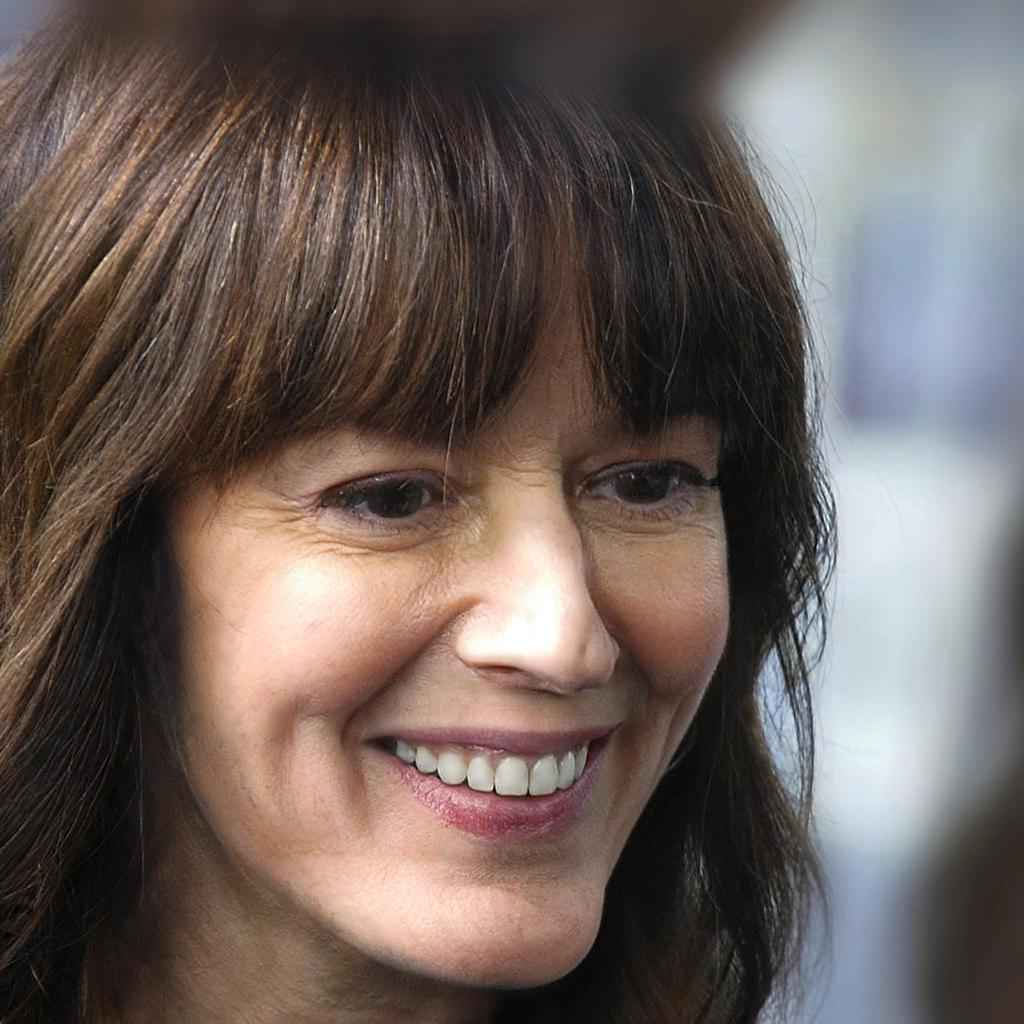} 
\end{tabular}
\caption{\textbf{Comparison of face aging results on CelebA-HQ \cite{karras2018progressive}}. The first column are the input images. The second to fifth column are outputs from Fader Network \cite{lample2017fader}, PAG-GAN \cite{yang2018learning}, IPC-GAN \cite{wang2018face} and our method. Our results reach the highest resolution without introducing significant artifacts. Our method preserves the background better compared to other techniques, see for instance the letters on the third row. In addition, compared to other techniques, our method leads to a result without artefacts nor blur. %
}
\label{compare}
\end{figure*}

\textbf{Generalisation capacity for images in unseen dataset} \quad For fair comparison and also to reduce the possible effect of overfitting on the training data, we evaluate all methods on a dataset not viewed at training time by any of the methods. We chose CelebA-HQ~\cite{karras2018progressive}, a high resolution version of the CelebA dataset. The input images are at $1024 \times 1024$ resolution, and are further downsampled at the resolution at which each method was trained using their official codes. The output size of each method is: $224 \times 224$ for PAGGAN, $128 \times 128$ for IPCGAN, $256 \times 256$ for FaderNet, and $1024 \times 1024$ for our method. We compare only the face aging results from young age group to old age group, since PAGGAN and IPCGAN are trained only for aging. Figure \ref{compare} shows the results obtained with the different methods. FaderNet~\cite{lample2017fader} introduces little modifications. PAGGAN~\cite{yang2018learning} generates satisfying age progression effects. However, noticeable artifacts are present on the face edges and hairs. IPCGAN~\cite{wang2018face} is limited to low resolution and thus introduces a strong degradation on the quality of the image. In comparison to these results, our approach introduces much less artifacts and preserves the fine details of the face and the background better. 

\subsection{Quantitative evaluation}

\begin{table}[t]
\scriptsize
\centering
\caption{\textbf{Quantitative evaluation using online face recognition API~\cite{megvii2013face++}}. We compare our method against three methods: Fader Network~\cite{lample2017fader}, PAGGAN~\cite{yang2018learning} and IPCGAN~\cite{wang2018face}. Images are transferred to the oldest age group ($50+$) for all the methods. The second column presents the average predicted age. The third column indicates the blurriness of the results (lower value means less blurry). The fourth column is the gender preservation rate, meaning to which percentage the original gender is preserved. The fifth column refers to expression preservation - smiling preservation rate. The last two columns indicate the emotion preservation rate}
\label{average_age}
\begin{center}
\begin{tabular}{cccccccc}
\toprule
&&&Gender&Smiling&&\multicolumn{2}{c}{Emotion Preservation(\%)}\\
Method & Predicted Age & Blur & Preservation(\%) &Preservation(\%)&&Neutral &Happiness\\
\cmidrule[0.4pt]{1-5}%
\cmidrule[0.4pt]{7-8}%
FaderNet~\cite{lample2017fader} & 44.34 $\pm$ 11.40 & 9.15 & \textbf{97.60} & 95.20 && 90.60 & 92.40 \\
PAGGAN~\cite{yang2018learning} & 49.07 $\pm$ 11.22 & 3.68 & 95.10 & 93.10 && 90.20 & 91.70\\
IPCGAN~\cite{wang2018face} & 49.72 $\pm$ 10.95 & 9.73 & 96.70 & 93.60 && 89.50 & 91.10 \\
Ours & 54.77 $\pm$ 8.40 & \textbf{2.15} & 97.10 & \textbf{96.30} && \textbf{91.30} & \textbf{92.70}\\
\bottomrule
\end{tabular}
\end{center}
\end{table}

Quantitative evaluation of image-to-image translation tasks is still an open question and there is no universal metric to measure photorealism or quantify artifacts in an image. The recent works~\cite{he2019s2gan,Liu_2019_CVPR,yang2018learning} on face aging use an online face recognition API to estimate the age and the identity preservation accuracy of the modified images. We thus employ a similar evaluation process. 

In our evaluation, the first $1,000$ images with true ``Young'' label of the CelebA-HQ dataset are extracted as test images. Using this test set, we make a quantitative comparison with FaderNet~\cite{lample2017fader}, IPCGAN~\cite{wang2018face} and PAGGAN~\cite{yang2018learning}. Each image is transferred to the oldest age group using their official released models. For IPCGAN and PAGGAN, the oldest age group refer to $50+$ and $[51, 60]$ respectively. For FaderNet, the old attribute is set to be the default largest value for aging in their official code. 
To have a fair comparison with groupwise methods, and since $50+$ is considered as the oldest age group, we choose a target age of $60$ (the mean of the age range $\{51, \ldots, 69\} \subset \mathcal{Q}$) for our age transformer.

Thus we get 1000 modified images for each method. We further evaluate these output images using the online face recognition API of Face++~\cite{megvii2013face++}. From the detect API, we obtain the following interesting metrics: age, gender, blurriness (whether the face is blurry or not, larger values means blurrier), smiling and emotion estimation. The emotion estimation contains a series of emotions: sadness, neutral, disgust, anger, surprise, fear and happiness. With a preliminary analysis on the results, $94.20\%$ of the input images are classified as neutral or happiness. Thus we just keep these two terms for emotion preservation comparison. We have also compared the identity preservation rate using the API to compare the modified images with the original inputs. However, since all methods achieve a nearly $100\%$ accuracy, this metric is not reported here.

Table \ref{average_age} shows the quantitative evaluation results. All the methods are given the oldest age group as aging target, and we notice that our method has the highest average predicted age.
The gender preservation rate is calculated by comparing the estimated gender with the original CelebA annotations. Using this metric, FaderNet achieves the best performance, followed by our method. For expression preservation (smiling) and emotion preservation (neutral, happiness), our approach yields the best results. It is to be noted however that all methods have similar results. For the blur evaluation, results are much more contrasted. Our method performs much better in generating sharper images, which is in agreement with the visual comparisons.

\subsection{Discussion}

\begin{figure*}[t]
\centering
\scriptsize
\begin{tabular}{ccccccc}
&Input&25&65&Input&25&65
\\
\raisebox{4\height}{(a)}&
{\color{yellow}%
\setlength{\fboxsep}{0pt}%
\setlength{\fboxrule}{2pt}%
\fbox{\includegraphics[width=0.14\linewidth]{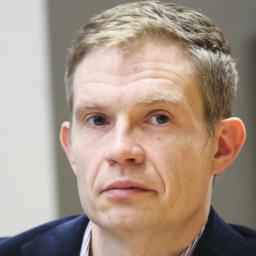}}}&
\includegraphics[width=0.14\linewidth]{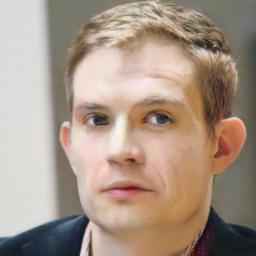} &
\includegraphics[width=0.14\linewidth]{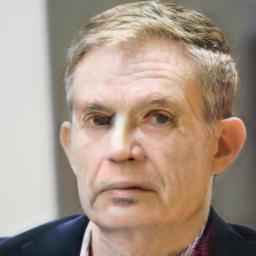} &
{\color{yellow}%
\setlength{\fboxsep}{0pt}%
\setlength{\fboxrule}{2pt}%
\fbox{\includegraphics[width=0.14\linewidth]{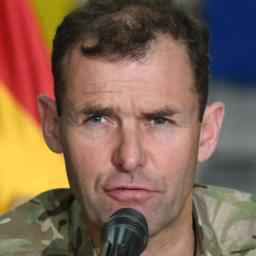}}}&
\includegraphics[width=0.14\linewidth]{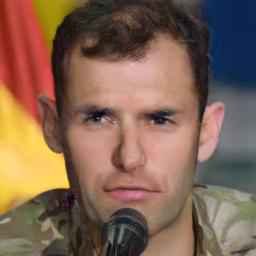} &
\includegraphics[width=0.14\linewidth]{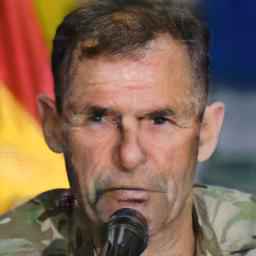} 
\\
\raisebox{4\height}{(b)}&
{\color{yellow}%
\setlength{\fboxsep}{0pt}%
\setlength{\fboxrule}{2pt}%
\fbox{\includegraphics[width=0.14\linewidth]{fig/ablation/c_dis/iter00099_content.jpg}}}&
\includegraphics[width=0.14\linewidth]{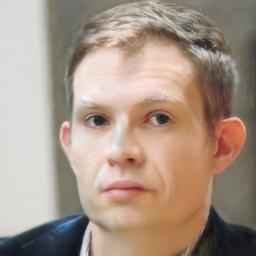} &
\includegraphics[width=0.14\linewidth]{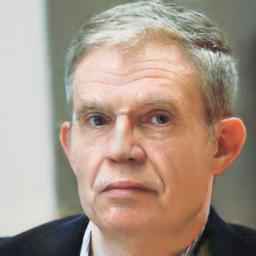} & 
{\color{yellow}%
\setlength{\fboxsep}{0pt}%
\setlength{\fboxrule}{2pt}%
\fbox{\includegraphics[width=0.14\linewidth]{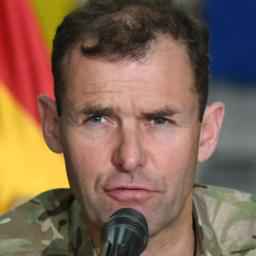}}}&
\includegraphics[width=0.14\linewidth]{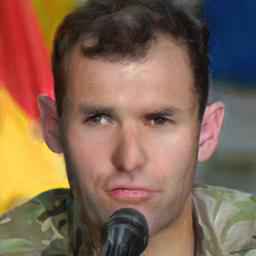} &
\includegraphics[width=0.14\linewidth]{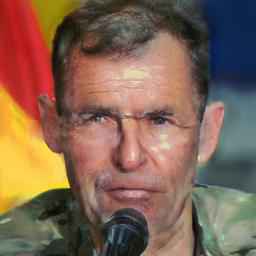} 
\\
\raisebox{4\height}{(c)}&
{\color{yellow}%
\setlength{\fboxsep}{0pt}%
\setlength{\fboxrule}{2pt}%
\fbox{\includegraphics[width=0.14\linewidth]{fig/ablation/c_dis/iter00099_content.jpg}}}&
\includegraphics[width=0.14\linewidth]{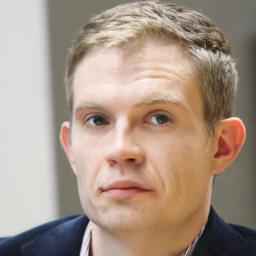} &
\includegraphics[width=0.14\linewidth]{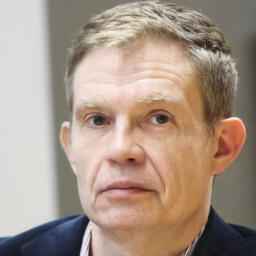} & 
{\color{yellow}%
\setlength{\fboxsep}{0pt}%
\setlength{\fboxrule}{2pt}%
\fbox{\includegraphics[width=0.14\linewidth]{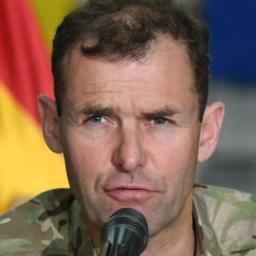}}}&
\includegraphics[width=0.14\linewidth]{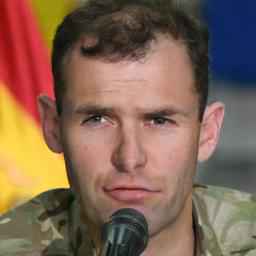} &
\includegraphics[width=0.14\linewidth]{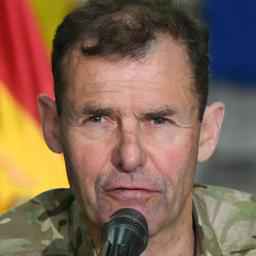} 
\\
\end{tabular}
\caption{
\textbf{Face age editing results with different types of discriminator}. (a) Conditional discriminator. (b) Two separate discriminators. One receives images only from old age groups, the other receives images from young age groups. (c) Our proposed method - using one single discriminator. Comparing to the results in (a) and (b), the proposed method (c), which uses a single discriminator, generates reliable face aging/de-aging effets with the least artifacts. 
}
\label{ablation}
\end{figure*}

\textbf{Ablation study on discriminator} \quad We have explored three different types of discriminators to train the age transformer. Figure \ref{ablation} presents the face age editing results corresponding to the different settings.

\begin{itemize}
\item \textbf{Conditional discriminator}. We adopt a patch discriminator~\cite{pix2pix2016} with a label projection applied on the features before the last convolutional layer, similar to the settings in~\cite{miyato2018cgans}. The discriminator is conditioned on four age groups: $20$-$35$, $35$-$45$, $45$-$55$, $55$-$70$. At the training stage we find it essential to give the same number of real and fake images from each class to the discriminator to make the training successful. If we sample a target age $\alpha_1$ from the set $\set{Q}_{\alpha_0} = \{\alpha \in \set{Q}\ : \left|\alpha - \alpha_0\right| \geq \alpha_*\}$ at training time, the discriminator will receive more manipulated images in the youngest and oldest group. Thus it tends to classify all the images in these two groups as fake. The conditional discriminator is very sensitive to the original data distribution and needs much more hyper-parameter fine-tuning to converge. Figure \ref{ablation}(a) presents the age editing results with conditional discriminator. Strong artifacts can be observed in the aging results.

\item \textbf{Two separate discriminators}. One discriminator receives manipulated and real images with a desired age lies in the old age group ($45$-$70$), while the other one takes manipulated and real images in the young age group ($20$-$44$). With this setting, the task of generating aging/de-aging effects is shared among the classifier and the discriminators. Although the results in \ref{ablation}(b) are better than those in \ref{ablation}(a), over-smoothing artifacts are perceived in the de-aging results and colored artifacts appear in the aging results. 

\item \textbf{One single discriminator}. This is our proposed method. The discriminator can be considered as a regularizer which imposes photorealism, as it takes all the manipulated and real images as input. The generation of aging/de-aging effects is solely dictated by the age classifier. We are able to achieve high resolution results only with this last setting.
\end{itemize}

\begin{figure*}[t]
\begin{center}
\setlength{\tabcolsep}{2pt}
\scriptsize
\begin{tabular}{cc|cc|cc}
Input&Reconstructed&Input&Reconstructed&Input&Reconstructed
\\
\includegraphics[width=0.14\linewidth]{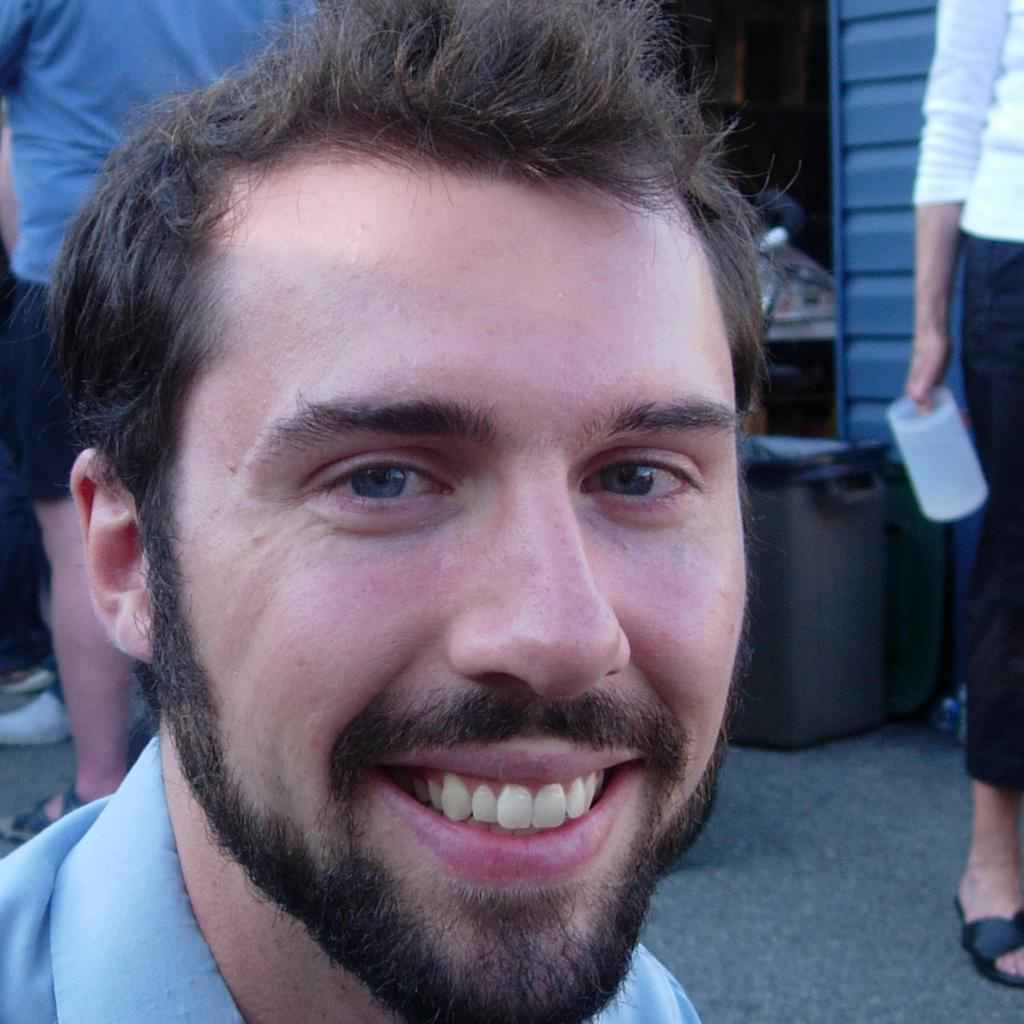}  &
\includegraphics[width=0.14\linewidth]{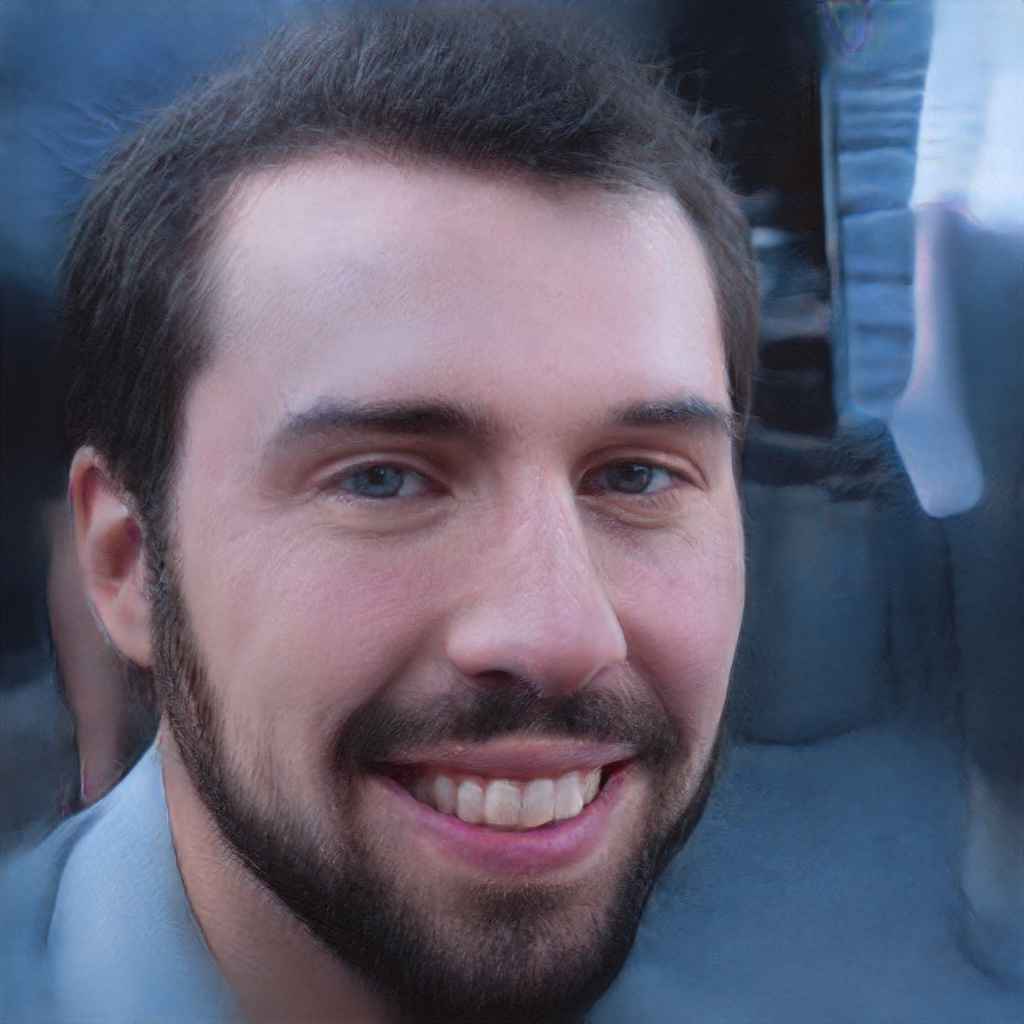}  &
\includegraphics[width=0.14\linewidth]{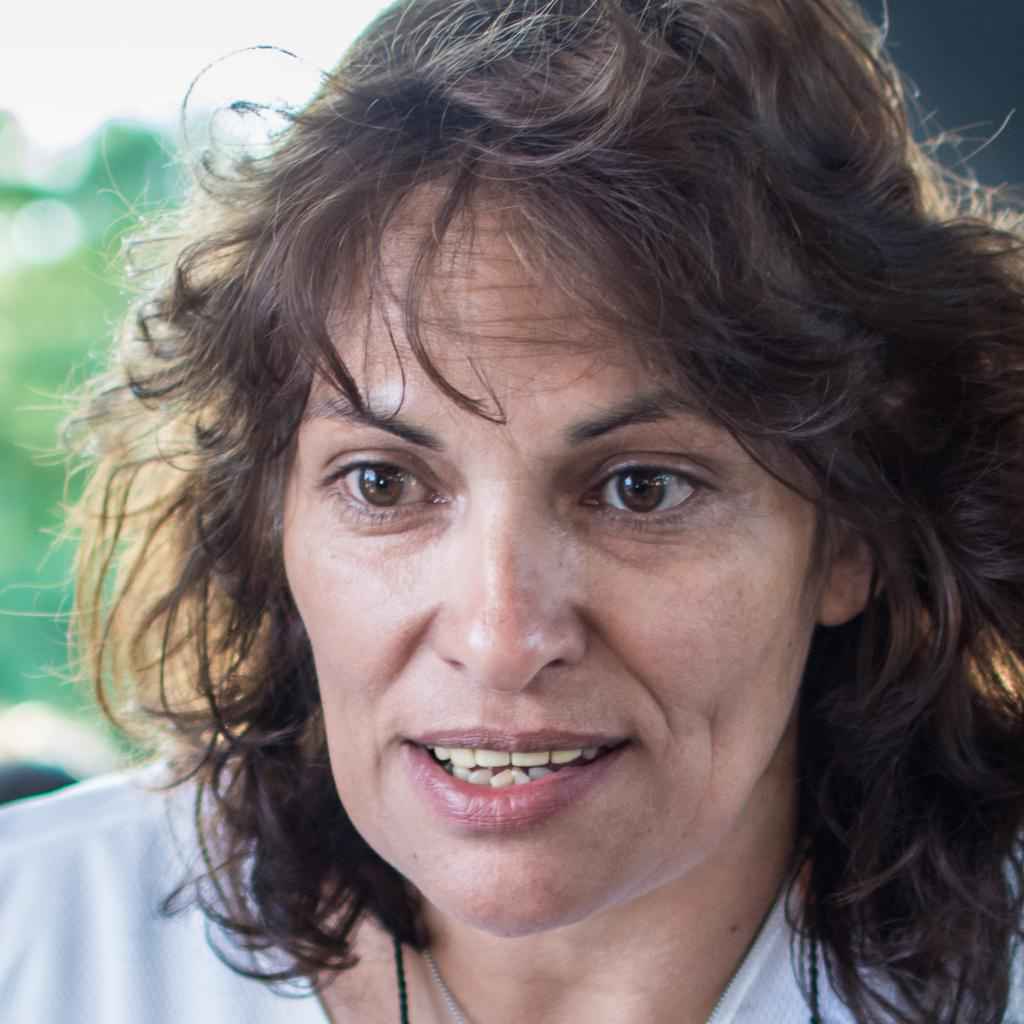}  &
\includegraphics[width=0.14\linewidth]{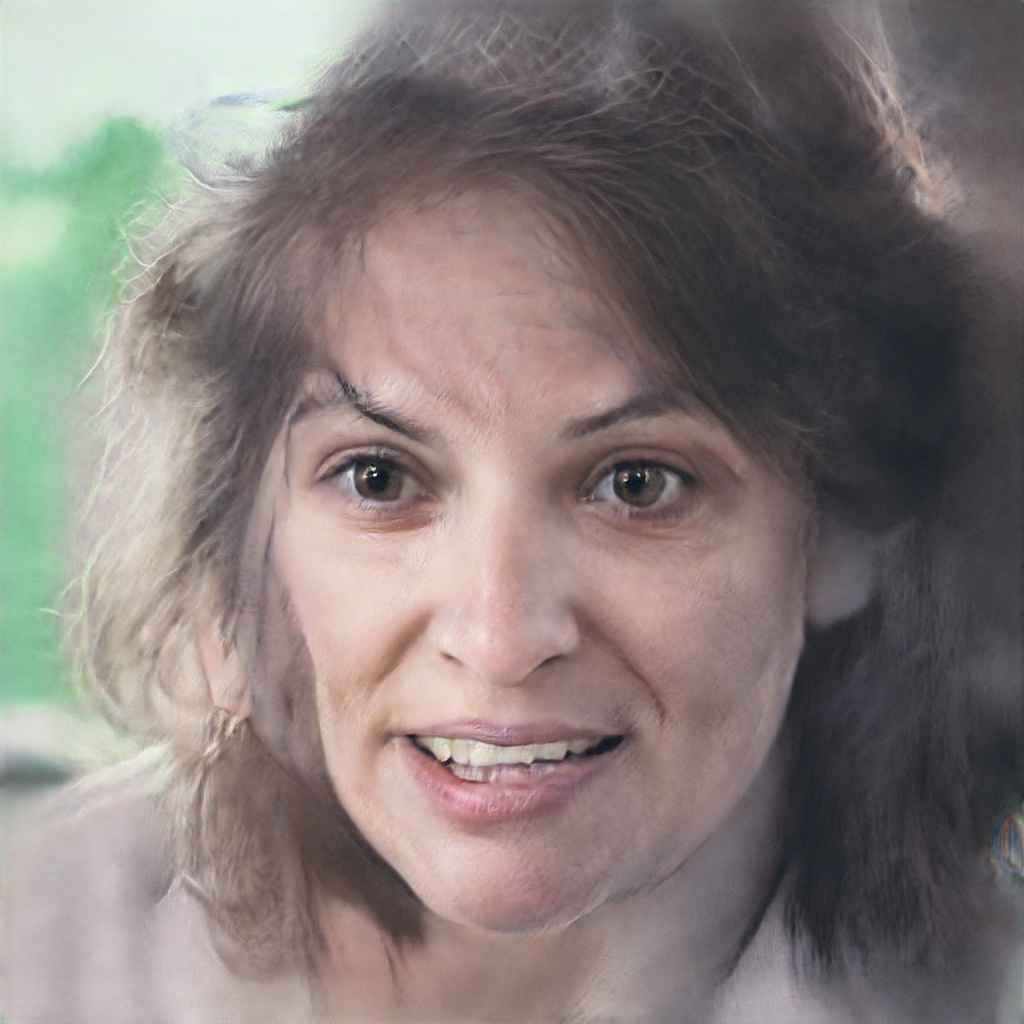}  &
\includegraphics[width=0.14\linewidth]{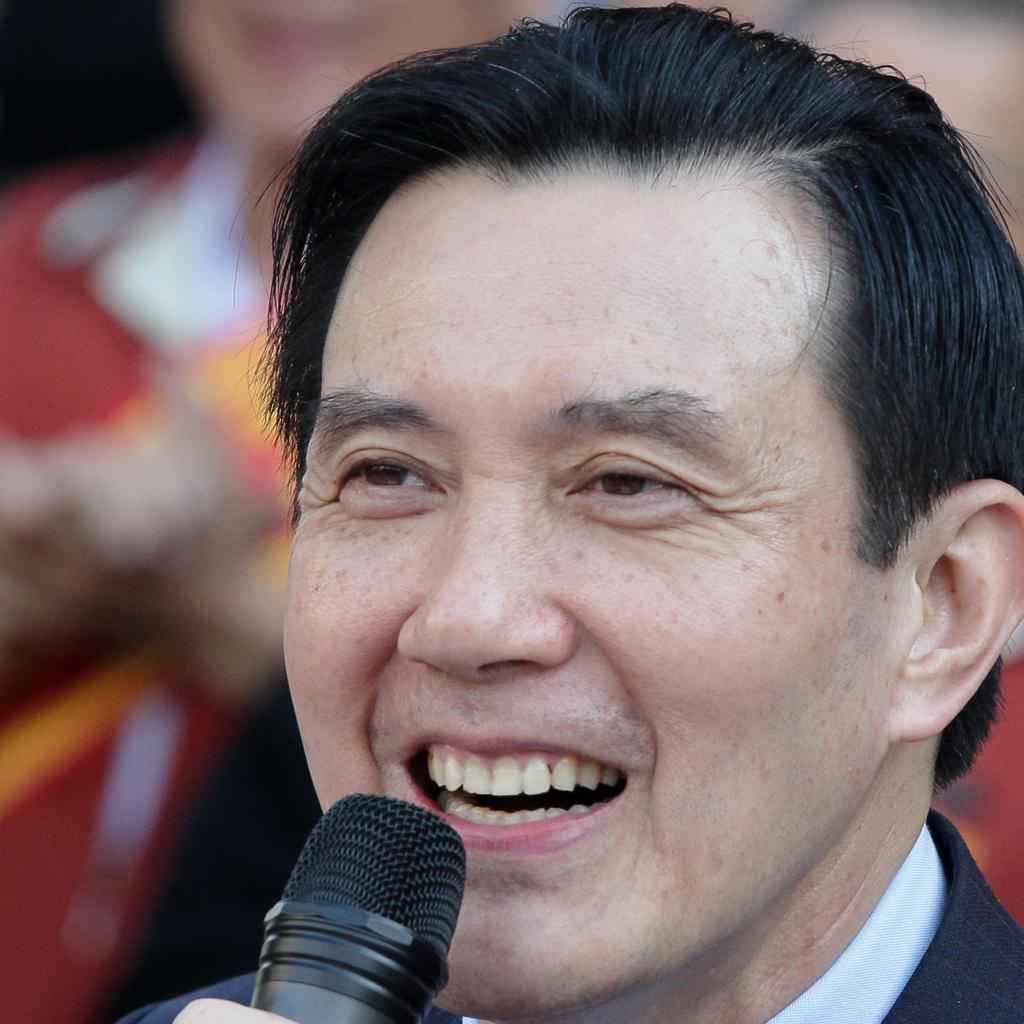}  &
\includegraphics[width=0.14\linewidth]{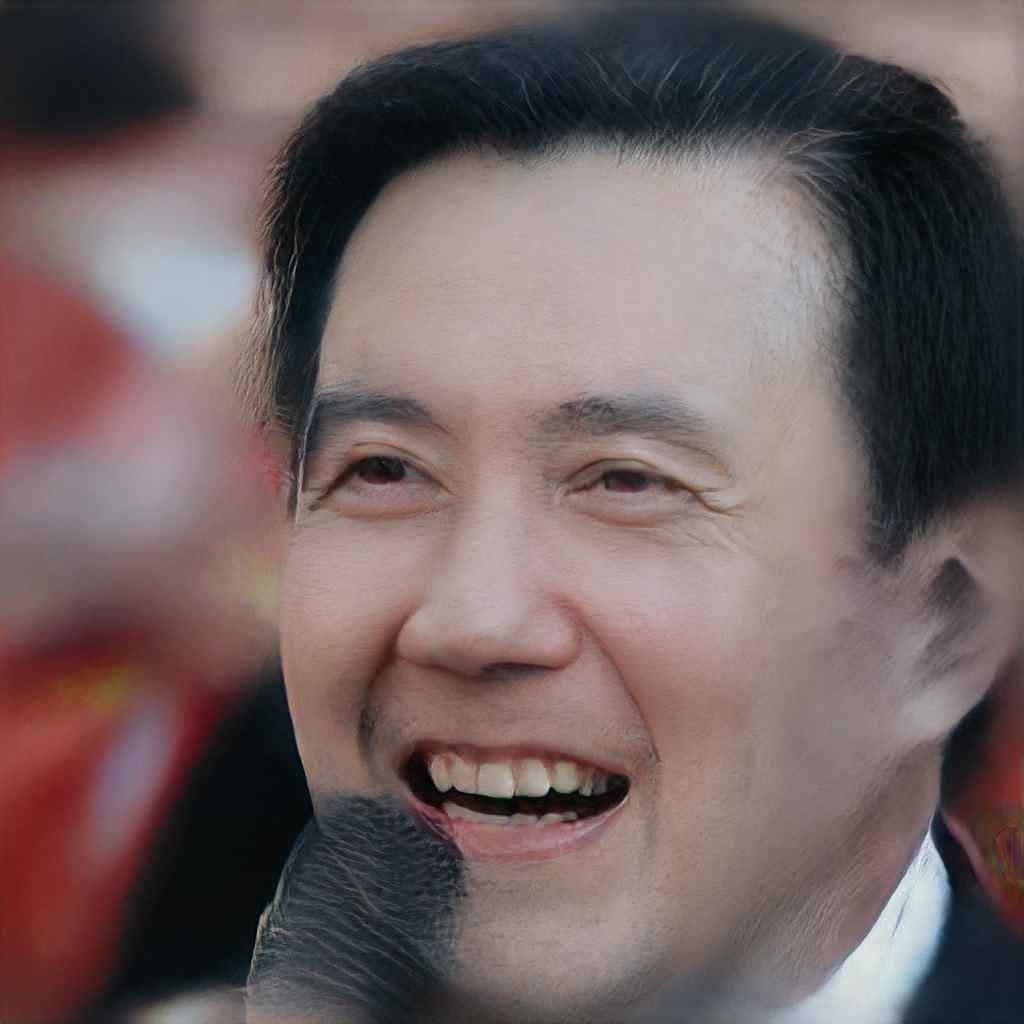} 
\end{tabular}
\end{center}
\caption{\textbf{Images reconstructed from a latent code optimization.} We analyze the possibility of encoding natural images to the latent space of StyleGAN \cite{karras2019style}, through optimization in the latent space minimizing the distance between the generated image and the input image. Each image is then reconstructed from this optimized latent code. %
The relatively low quality of the reconstruction strongly suggest that editing performed in the latent space cannot lead to a sharp and artifact-free result.}
\label{interface}
\end{figure*}

\textbf{Image reconstructed from a latent code optimization} \quad As mentioned in Section \ref{sec:related}, the recent work of Shen \etal \cite{shen2019interpreting} proposes an effective way to manipulate the latent code of an image generator to achieve high visual quality manipulation of synthesized images. It is therefore tempting to manipulate the latent code directly to produce face manipulation (and thus age editing) on natural images with this approach. However, finding such a latent code for an arbitrary face image %
is still a challenging problem. According to our experiments using StyleGAN \cite{karras2019style}, only a fraction of natural face images can be accurately reconstructed from the latent code \footnote{The latent code is obtained through optimization in the latent space by finding a latent code that minimizes the distance between the generated image and the input image.} by \cite{stylegan-encoder}. Consequently, this type of method is impractical until a better StyleGAN encoder is made available. Figure~\ref{interface} is meant to support this claim, where reconstruction results of natural face images can be assessed. We notice that the reconstructed images have painting-like artifacts, blurry backgrounds, and sometimes fail to preserve the identity of the person in the input image. Indeed, StyleGAN is much more efficient at sampling random faces from the latent space than at approximating a given face image. This is due to the fact that a GAN is not necessarily invertible. Hence, an editing method based on this latent code reconstruction will struggle to handle correctly natural images and to achieve the high visual quality of our method.

\textbf{Weakly supervised training} \quad To the best of our knowledge, our work is the first to use unlabeled data for training among recent face aging studies~\cite{he2019s2gan,Liu_2019_CVPR,song2018dual,wang2018face,zhang2017age}. A classifier pretrained on IMDB-WIKI~\cite{rothe2015dex}, a low resolution face dataset, is used to provide age information. Moreover, the discriminator in our method is used only to distinguish real and manipulated images. Relying solely on the classifier, we successfully extract the age specific features and further realize age transform on high resolution images. This reveals the capacity of the classifier, even trained on low quality images. 
Our method could be potentially generalized to other face attributes manipulation tasks, by using a separate pair of modulating network and classifier for each attribute. %

\section{Conclusion}
In this paper, we have proposed an age transformer architecture, enabling continuous face age editing with a single network, which we have endeavoured to keep as simple as possible. We believe that this approach, combined with an encoder-decoder architecture, rather than relying on a complex GAN, is the best path towards high quality, high resolution face editing results. 
We have demonstrated the capacity of our model to produce photorealistic and sharp results, without introducing significant artifacts, on images of resolution $1024 \times 1024$. 
The proposed feature modulation block appears to  achieve efficient separation of age and identity information. 
Given the performance achieved, this design can be potentially useful for other face attribute manipulation tasks.

\clearpage
\bibliographystyle{splncs04}
\bibliography{egbib}

\begin{thebibliography}{10}
\providecommand{\url}[1]{\texttt{#1}}
\providecommand{\urlprefix}{URL }
\providecommand{\doi}[1]{https://doi.org/#1}

\bibitem{antipov2017face}
Antipov, G., Baccouche, M., Dugelay, J.L.: Face aging with conditional
  generative adversarial networks. In: 2017 IEEE International Conference on
  Image Processing (ICIP). pp. 2089--2093. IEEE (2017)

\bibitem{chen2014cross}
Chen, B.C., Chen, C.S., Hsu, W.H.: Cross-age reference coding for age-invariant
  face recognition and retrieval. In: European conference on computer vision.
  pp. 768--783. Springer (2014)

\bibitem{chen2019semantic}
Chen, Y.C., Shen, X., Lin, Z., Lu, X., Pao, I., Jia, J., et~al.: Semantic
  component decomposition for face attribute manipulation. In: Proceedings of
  the IEEE Conference on Computer Vision and Pattern Recognition. pp.
  9859--9867 (2019)

\bibitem{choi2018stargan}
Choi, Y., Choi, M., Kim, M., Ha, J.W., Kim, S., Choo, J.: Stargan: Unified
  generative adversarial networks for multi-domain image-to-image translation.
  In: Proceedings of the IEEE Conference on Computer Vision and Pattern
  Recognition. pp. 8789--8797 (2018)

\bibitem{dumoulin17}
Dumoulin, V., Shlens, J., Kudlur, M.: A learned representation for artistic
  style. Proc. of ICLR  (2017)

\bibitem{fu2010age}
Fu, Y., Guo, G., Huang, T.S.: Age synthesis and estimation via faces: A survey.
  IEEE transactions on pattern analysis and machine intelligence
  \textbf{32}(11),  1955--1976 (2010)

\bibitem{Goodfellow2014}
Goodfellow, I., Pouget-Abadie, J., Mirza, M., Xu, B., Warde-Farley, D., Ozair,
  S., Courville, A., Bengio, Y.: Generative adversarial nets. In: Advances in
  neural information processing systems. pp. 2672--2680 (2014)

\bibitem{he_resnet16}
He, K., Zhang, X., Ren, S., Sun, J.: Deep residual learning for image
  recognition. In: CVPR (2016)

\bibitem{he2019s2gan}
He, Z., Kan, M., Shan, S., Chen, X.: S2gan: Share aging factors across ages and
  share aging trends among individuals. In: Proceedings of the IEEE
  International Conference on Computer Vision. pp. 9440--9449 (2019)

\bibitem{huang17}
Huang, X., Belongie, S.J.: Arbitrary style transfer in real-time with adaptive
  instance normalization. In: ICCV (2017)

\bibitem{huang2018multimodal}
Huang, X., Liu, M.Y., Belongie, S., Kautz, J.: Multimodal unsupervised
  image-to-image translation. In: Proceedings of the European Conference on
  Computer Vision (ECCV). pp. 172--189 (2018)

\bibitem{megvii2013face++}
Inc, M.: Face++ research toolkit. \url{http://www. faceplusplus.com.} (2013)

\bibitem{pix2pix2016}
Isola, P., Zhu, J.Y., Zhou, T., Efros, A.A.: Image-to-image translation with
  conditional adversarial networks. In: CVPR (2017)

\bibitem{johnson2016perceptual}
Johnson, J., Alahi, A., Li, F.F.: Perceptual losses for real-time style
  transfer and super-resolution. In: ECCV. Springer (2016)

\bibitem{karras2018progressive}
Karras, T., Aila, T., Laine, S., Lehtinen, J.: Progressive growing of {GAN}s
  for improved quality, stability, and variation. In: International Conference
  on Learning Representations (2018),
  \url{https://openreview.net/forum?id=Hk99zCeAb}

\bibitem{karras2019style}
Karras, T., Laine, S., Aila, T.: A style-based generator architecture for
  generative adversarial networks. In: Proceedings of the IEEE Conference on
  Computer Vision and Pattern Recognition. pp. 4401--4410 (2019)

\bibitem{karras2019analyzing}
Karras, T., Laine, S., Aittala, M., Hellsten, J., Lehtinen, J., Aila, T.:
  Analyzing and improving the image quality of stylegan. arXiv preprint
  arXiv:1912.04958  (2019)

\bibitem{lample2017fader}
Lample, G., Zeghidour, N., Usunier, N., Bordes, A., DENOYER, L., et~al.: Fader
  networks: Manipulating images by sliding attributes. In: Advances in Neural
  Information Processing Systems (2017)

\bibitem{li2019global}
Li, P., Hu, Y., He, R., Sun, Z.: Global and local consistent wavelet-domain age
  synthesis. IEEE Transactions on Information Forensics and Security  (2019)

\bibitem{Liu_2019_CVPR}
Liu, Y., Li, Q., Sun, Z.: Attribute-aware face aging with wavelet-based
  generative adversarial networks. In: The IEEE Conference on Computer Vision
  and Pattern Recognition (CVPR) (June 2019)

\bibitem{maas2013rectifier}
Maas, A.L., Hannun, A.Y., Ng, A.Y.: Rectifier nonlinearities improve neural
  network acoustic models. In: in ICML Workshop on Deep Learning for Audio,
  Speech and Language Processing. Citeseer (2013)

\bibitem{mao2017least}
Mao, X., Li, Q., Xie, H., Lau, R.Y., Wang, Z., Paul~Smolley, S.: Least squares
  generative adversarial networks. In: Proceedings of the IEEE International
  Conference on Computer Vision. pp. 2794--2802 (2017)

\bibitem{Mescheder2018ICML}
Mescheder, L., Nowozin, S., Geiger, A.: Which training methods for gans do
  actually converge? In: International Conference on Machine Learning (ICML)
  (2018)

\bibitem{mirza2014conditional}
Mirza, M., Osindero, S.: Conditional generative adversarial nets. arXiv
  preprint arXiv:1411.1784  (2014)

\bibitem{miyato2018spectral}
Miyato, T., Kataoka, T., Koyama, M., Yoshida, Y.: Spectral normalization for
  generative adversarial networks. In: International Conference on Learning
  Representations (2018), \url{https://openreview.net/forum?id=B1QRgziT-}

\bibitem{miyato2018cgans}
Miyato, T., Koyama, M.: cgans with projection discriminator. arXiv preprint
  arXiv:1802.05637  (2018)

\bibitem{stylegan-encoder}
Nikitko, D.: Stylegan encoder for official tensorflow implementation.
  \url{https://github.com/Puzer/stylegan-encoder} (2019)

\bibitem{paszke2017automatic}
Paszke, A., Gross, S., Chintala, S., Chanan, G., Yang, E., DeVito, Z., Lin, Z.,
  Desmaison, A., Antiga, L., Lerer, A.: Automatic differentiation in {PyTorch}.
  In: NIPS Autodiff Workshop (2017)

\bibitem{pumarola2018ganimation}
Pumarola, A., Agudo, A., Martinez, A.M., Sanfeliu, A., Moreno-Noguer, F.:
  Ganimation: Anatomically-aware facial animation from a single image. In:
  Proceedings of the European Conference on Computer Vision (ECCV). pp.
  818--833 (2018)

\bibitem{qian2019make}
Qian, S., Lin, K.Y., Wu, W., Liu, Y., Wang, Q., Shen, F., Qian, C., He, R.:
  Make a face: Towards arbitrary high fidelity face manipulation. In:
  Proceedings of the IEEE International Conference on Computer Vision. pp.
  10033--10042 (2019)

\bibitem{rothe2015dex}
Rothe, R., Timofte, R., Van~Gool, L.: Dex: Deep expectation of apparent age
  from a single image. In: Proceedings of the IEEE International Conference on
  Computer Vision Workshops. pp. 10--15 (2015)

\bibitem{shen2019interpreting}
Shen, Y., Gu, J., Tang, X., Zhou, B.: Interpreting the latent space of gans for
  semantic face editing. arXiv preprint arXiv:1907.10786  (2019)

\bibitem{song2018dual}
Song, J., Zhang, J., Gao, L., Liu, X., Shen, H.T.: Dual conditional gans for
  face aging and rejuvenation. In: IJCAI. pp. 899--905 (2018)

\bibitem{upchurch2017deep}
Upchurch, P., Gardner, J., Pleiss, G., Pless, R., Snavely, N., Bala, K.,
  Weinberger, K.: Deep feature interpolation for image content changes. In:
  Proceedings of the IEEE conference on computer vision and pattern
  recognition. pp. 7064--7073 (2017)

\bibitem{wang2018pix2pixHD}
Wang, T.C., Liu, M.Y., Zhu, J.Y., Tao, A., Kautz, J., Catanzaro, B.:
  High-resolution image synthesis and semantic manipulation with conditional
  gans. In: CVPR (2018)

\bibitem{wang2018face}
Wang, Z., Tang, X., Luo, W., Gao, S.: Face aging with identity-preserved
  conditional generative adversarial networks. In: Proceedings of the IEEE
  Conference on Computer Vision and Pattern Recognition. pp. 7939--7947 (2018)

\bibitem{xiao2018elegant}
Xiao, T., Hong, J., Ma, J.: Elegant: Exchanging latent encodings with gan for
  transferring multiple face attributes. In: Proceedings of the European
  Conference on Computer Vision (ECCV). pp. 168--184 (2018)

\bibitem{yang2018learning}
Yang, H., Huang, D., Wang, Y., Jain, A.K.: Learning face age progression: A
  pyramid architecture of gans. In: Proceedings of the IEEE Conference on
  Computer Vision and Pattern Recognition. pp. 31--39 (2018)

\bibitem{zhang2017age}
Zhang, Z., Song, Y., Qi, H.: Age progression/regression by conditional
  adversarial autoencoder. In: IEEE Conference on Computer Vision and Pattern
  Recognition (CVPR) (2017)

\bibitem{zhu2017unpaired}
Zhu, J.Y., Park, T., Isola, P., Efros, A.A.: Unpaired image-to-image
  translation using cycle-consistent adversarial networks. In: Proceedings of
  the IEEE international conference on computer vision. pp. 2223--2232 (2017)

\end{thebibliography}

\clearpage
\appendix

\section{Network architecture}

Table \ref{arch_table} presents the hyperparameters of the proposed network architecture. The discriminator is a $142 \times 142$ patch discriminator. Each element of the output feature map corresponds to a receptive field of $142 \times 142$ on the original input image.

\section{Age classifier}

To obtain the age information of FFHQ dataset \cite{karras2019style}, we use the age classifier \cite{rothe2015dex}, which has been pretrained on IMDB-WIKI. This dataset contains $523,051$ face images of $20,284$ celebrities collected from the IMDB and Wikipedia websites. The dataset mostly covers the $[20, 65]$ age interval, and has only very few samples for the younger and older age intervals. Consequently, the age classifier might yield less accurate age estimation for faces of people younger than $20$ years old or much older than $65$ years old. We therefore choose to use images in the age range $\mathcal{Q} = \{20, \ldots, 69\}$ for training. We pass the images of FFHQ dataset into the age classifier and observe that FFHQ contains much more samples of young faces than of old ones. We then augment the dataset with synthetic images generated by StyleGAN \cite{karras2019style} to achieve a quasi-uniform age distribution over the age range $\mathcal{Q}$, as described in section $4.1$ of the paper.

\section{Additional results}

In this section, we present supplementary results on $1024 \times 1024$ images. 

\subsection{Results on FFHQ dataset}

\begin{figure*}[ht]
\begin{center}
\setlength{\tabcolsep}{1pt}
\begin{tabular}{ccccc}
25&35&45&55&65 \\
{\color{yellow}%
\setlength{\fboxsep}{0pt}%
\setlength{\fboxrule}{2pt}%
\fbox{\includegraphics[width=0.19\linewidth]{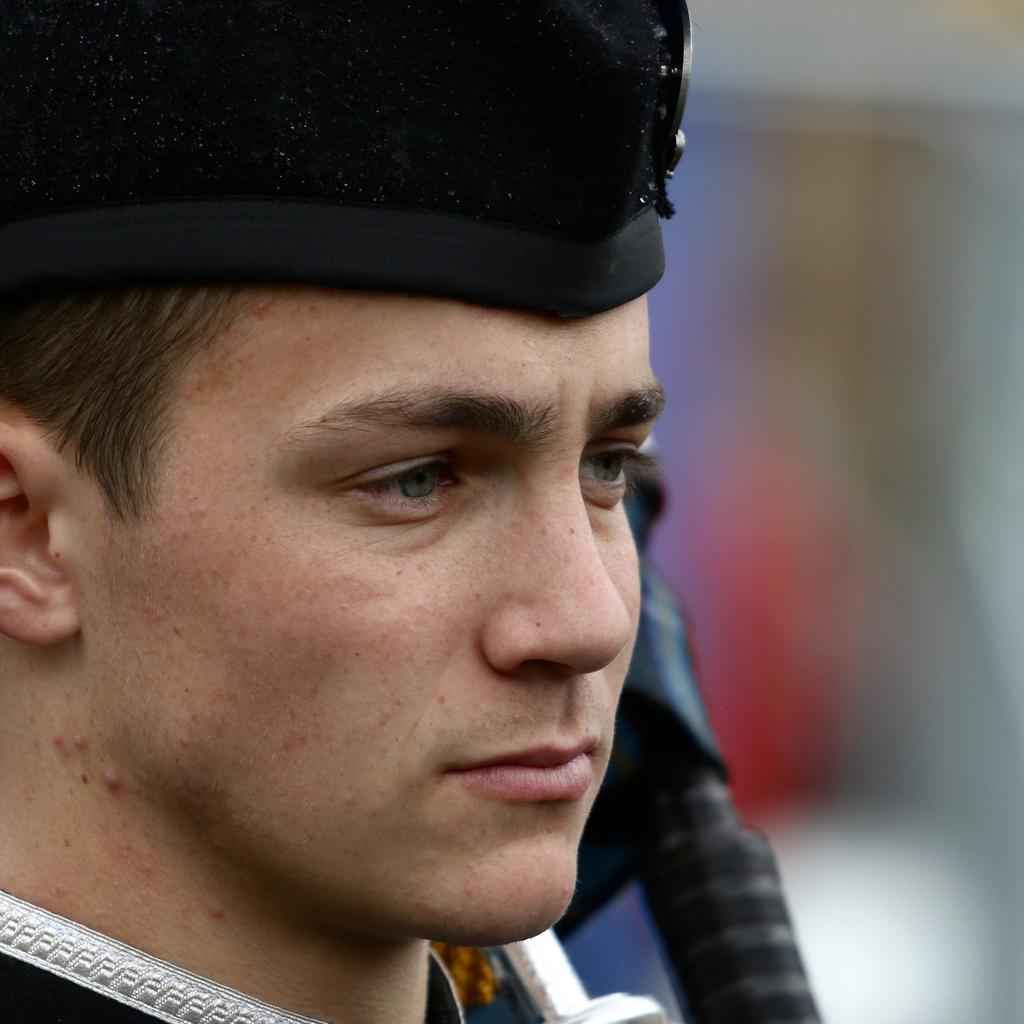}}} &
\includegraphics[width=0.19\linewidth]{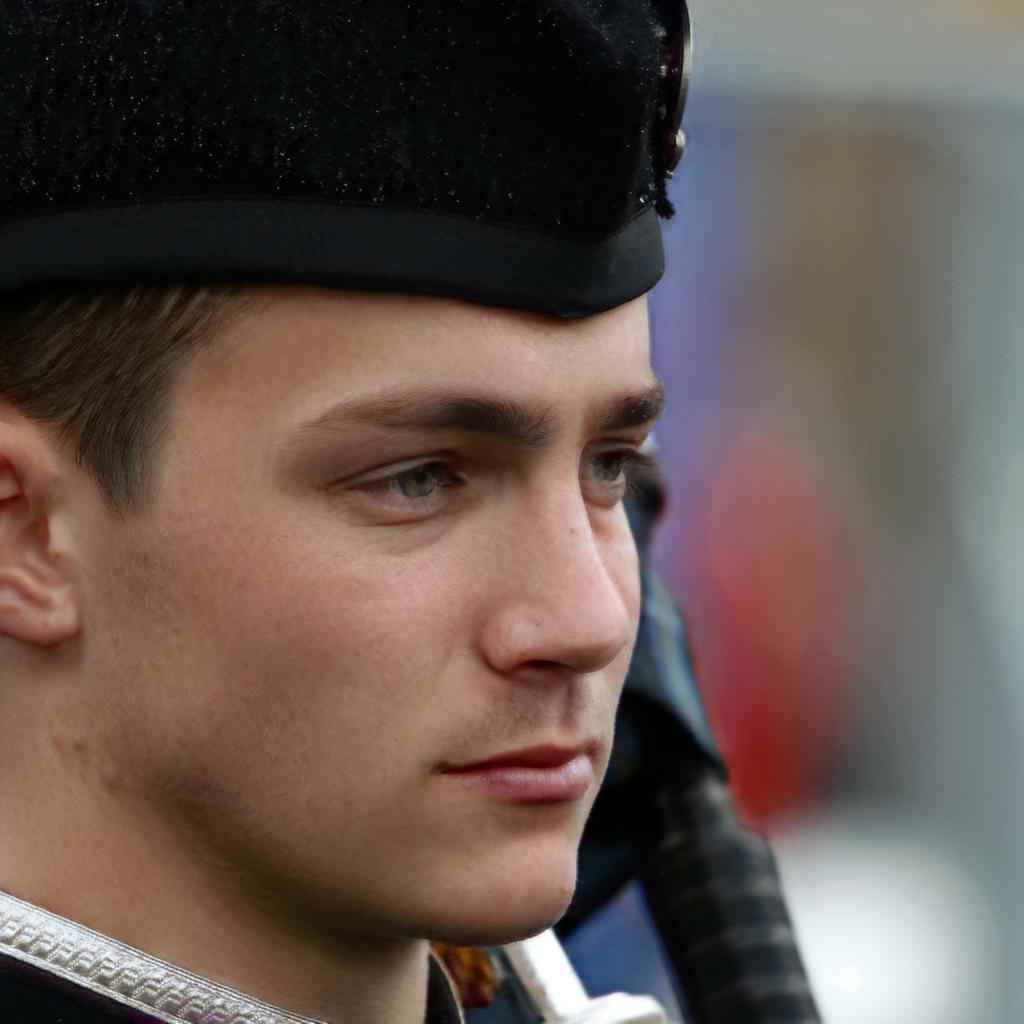} & 
\includegraphics[width=0.19\linewidth]{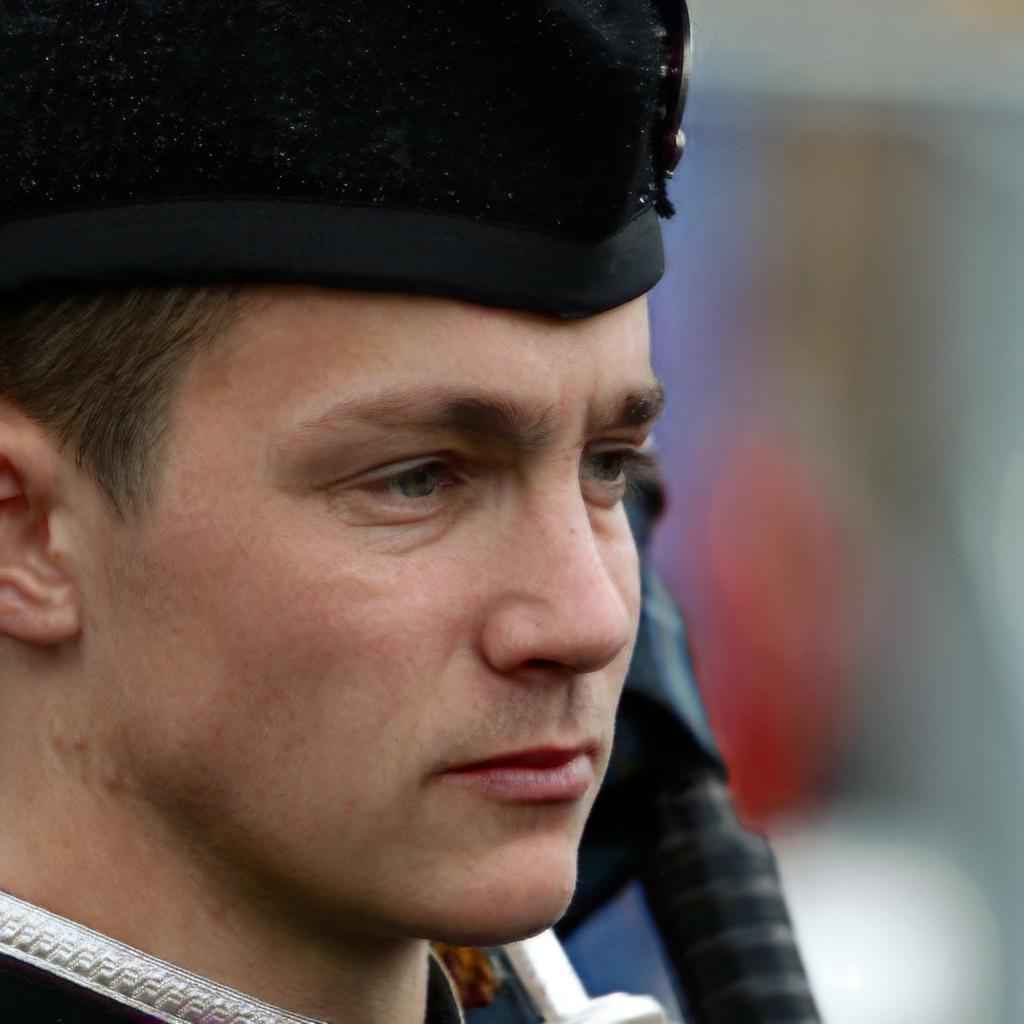} &
\includegraphics[width=0.19\linewidth]{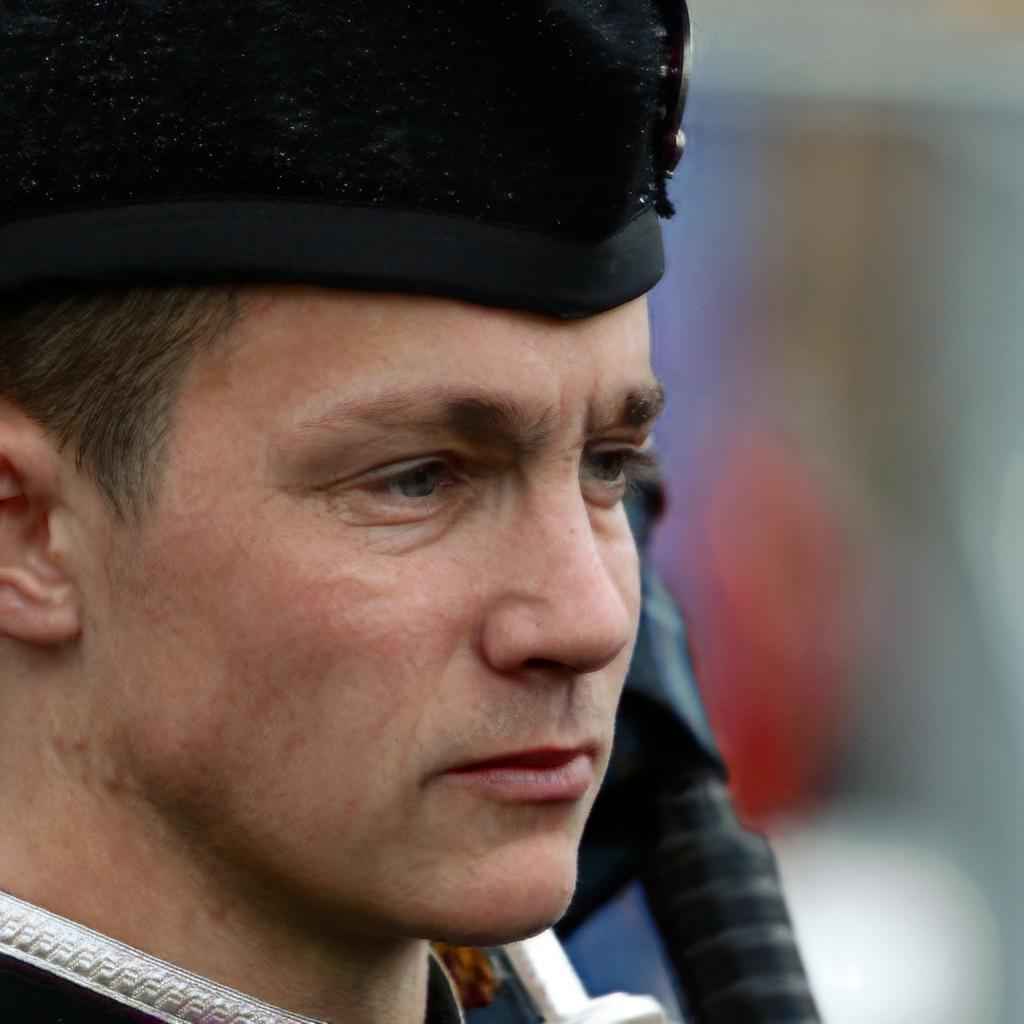} & 
\includegraphics[width=0.19\linewidth]{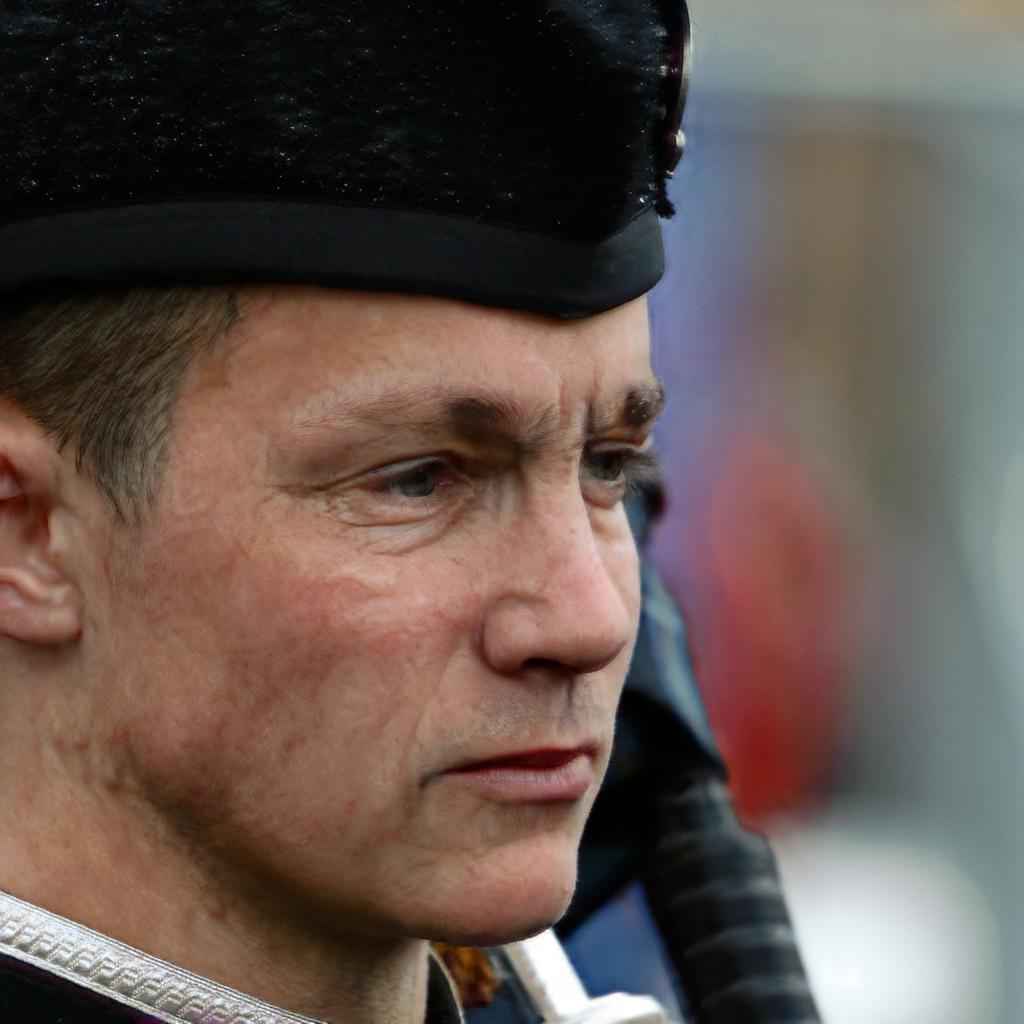} 
\\
\includegraphics[width=0.19\linewidth]{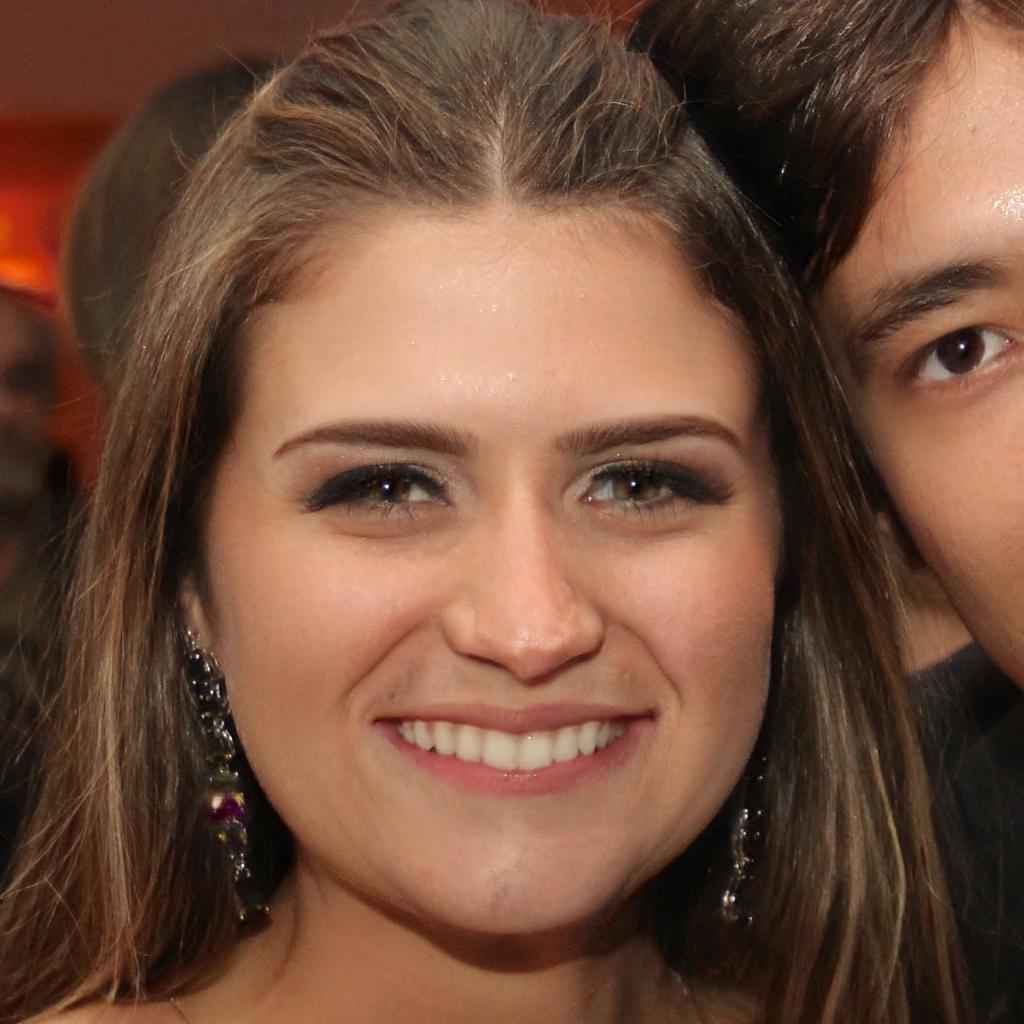} & 
{\color{yellow}%
\setlength{\fboxsep}{0pt}%
\setlength{\fboxrule}{2pt}%
\fbox{\includegraphics[width=0.19\linewidth]{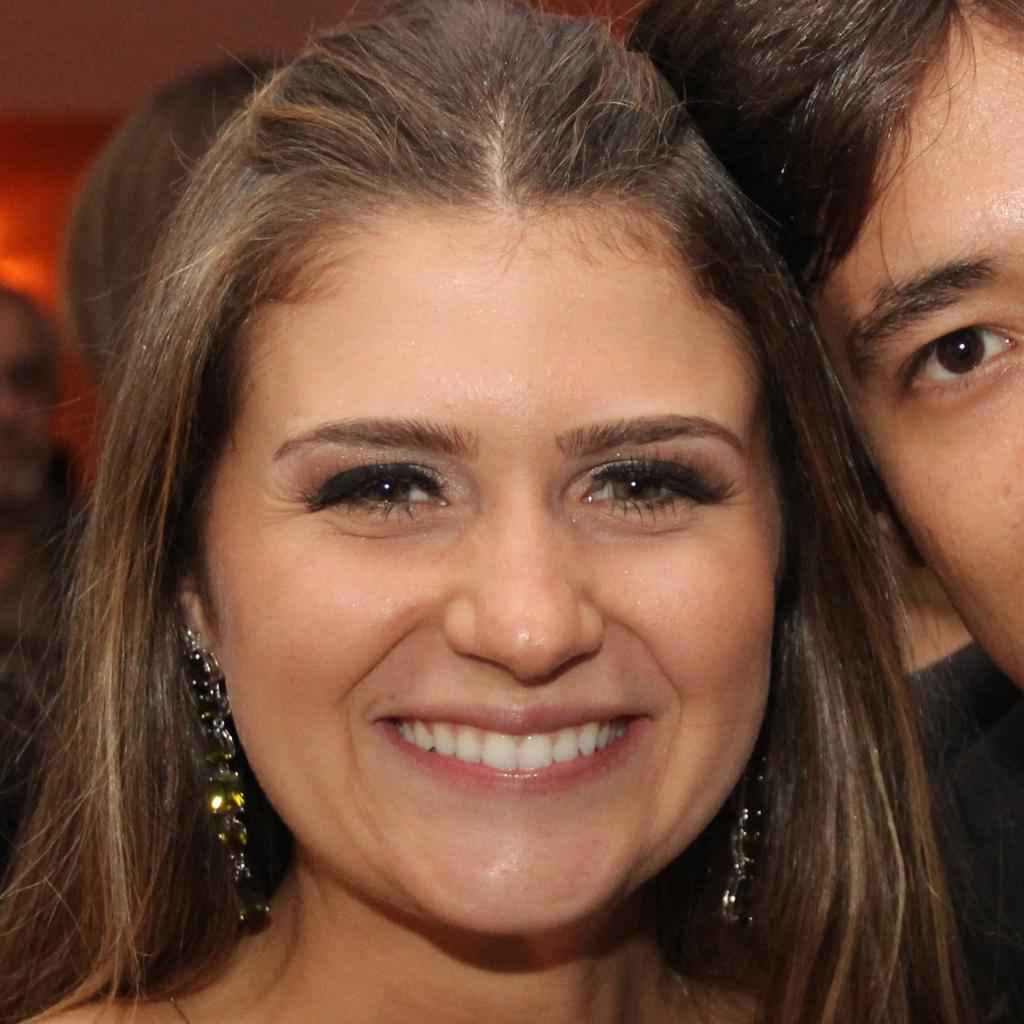}}} & 
\includegraphics[width=0.19\linewidth]{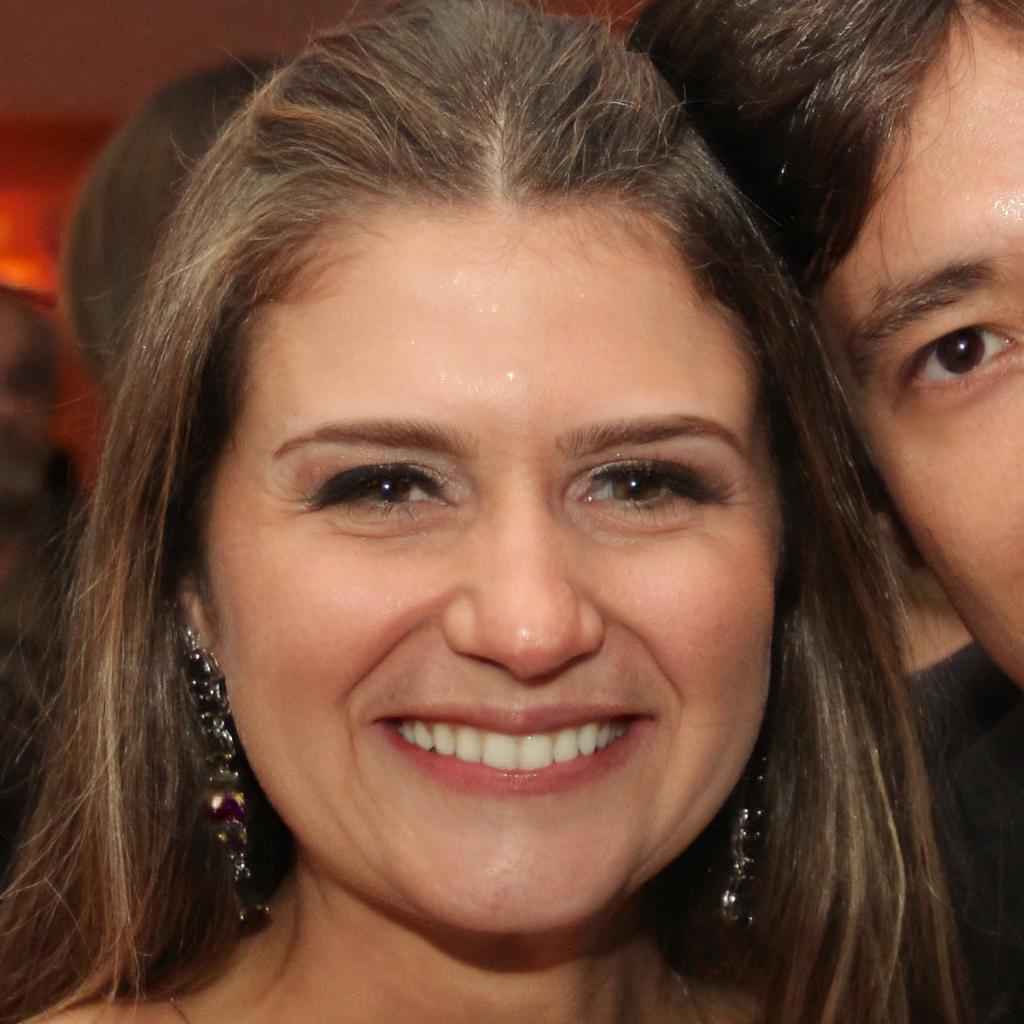} &
\includegraphics[width=0.19\linewidth]{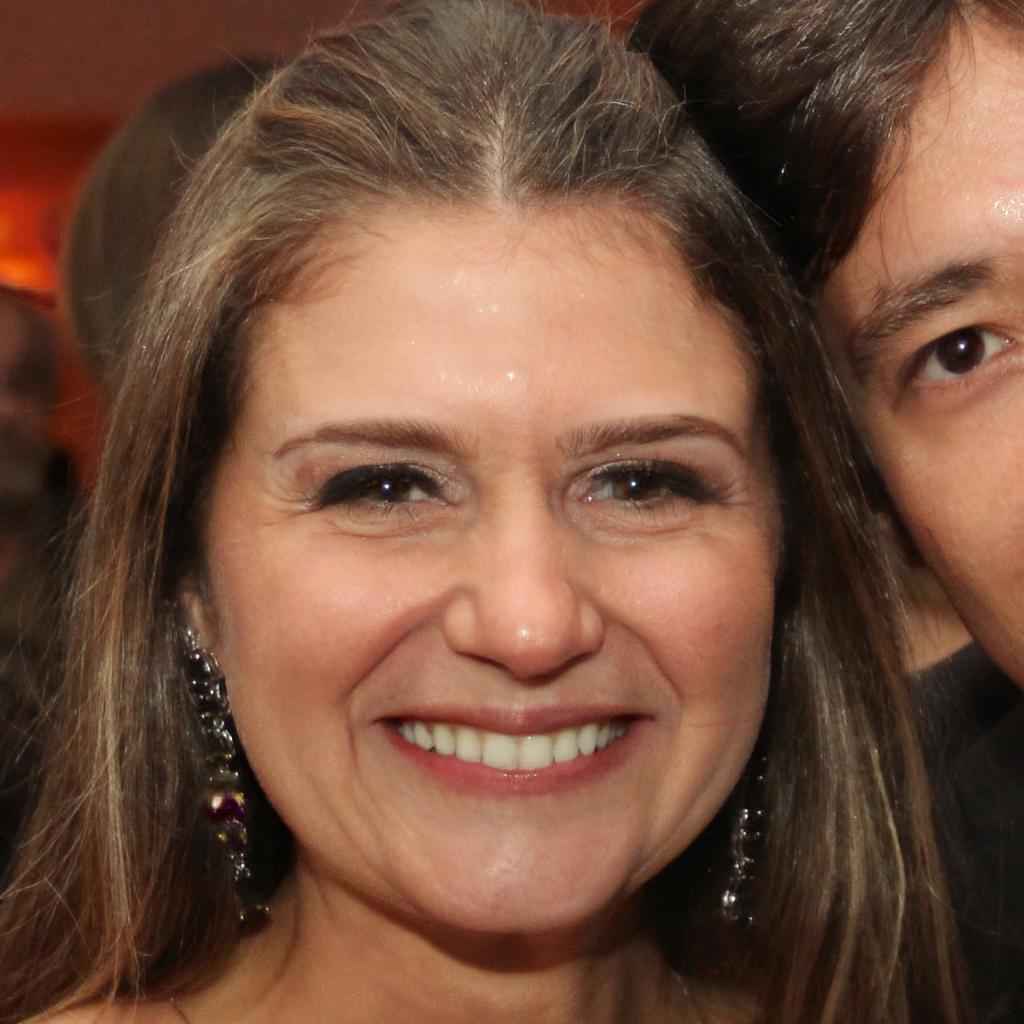} & 
\includegraphics[width=0.19\linewidth]{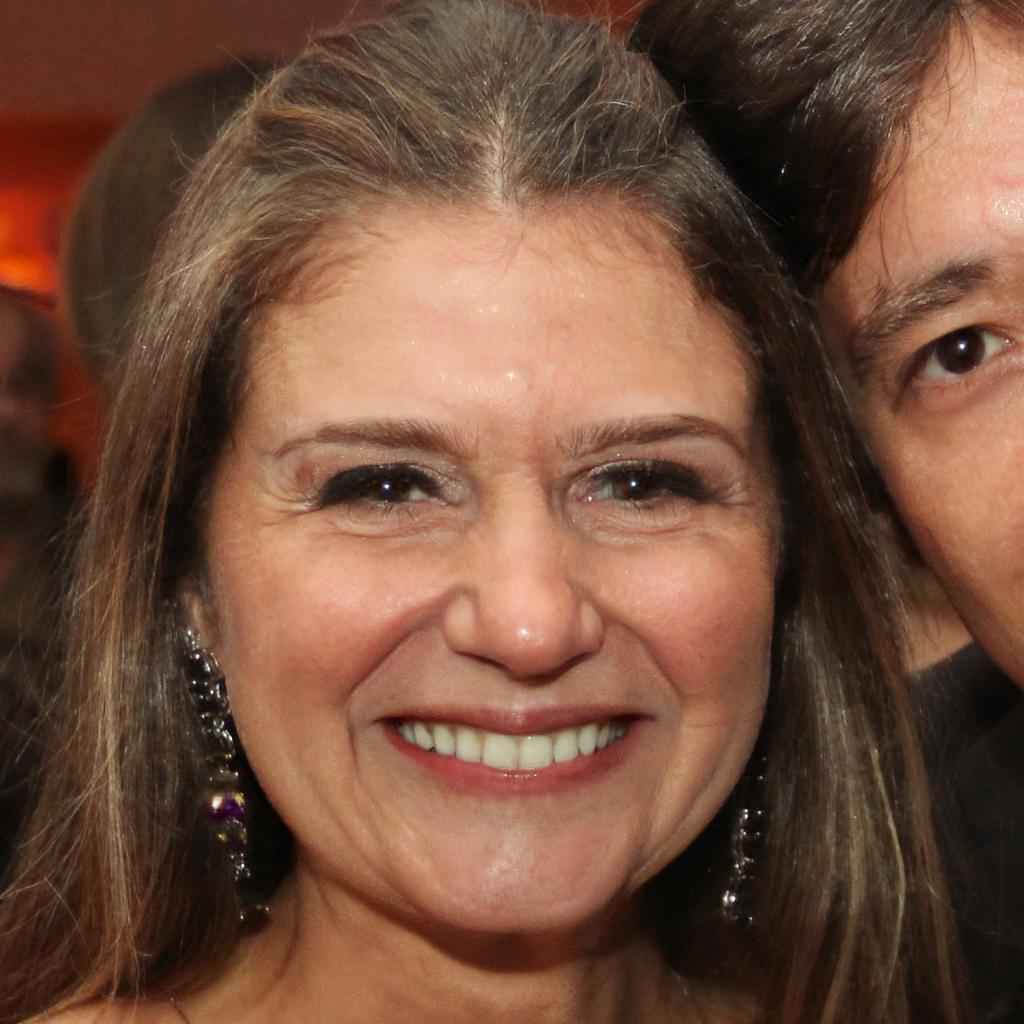} 
\\
\includegraphics[width=0.19\linewidth]{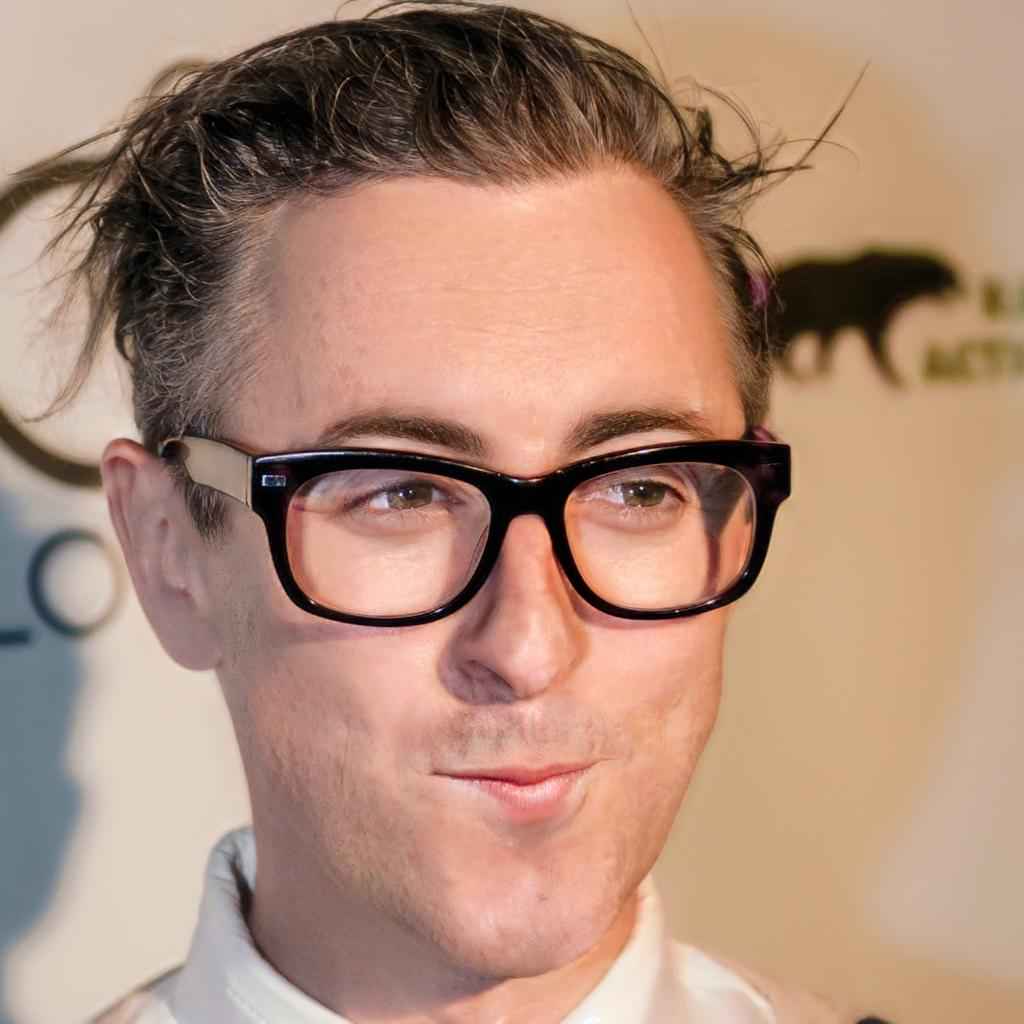} & 
\includegraphics[width=0.19\linewidth]{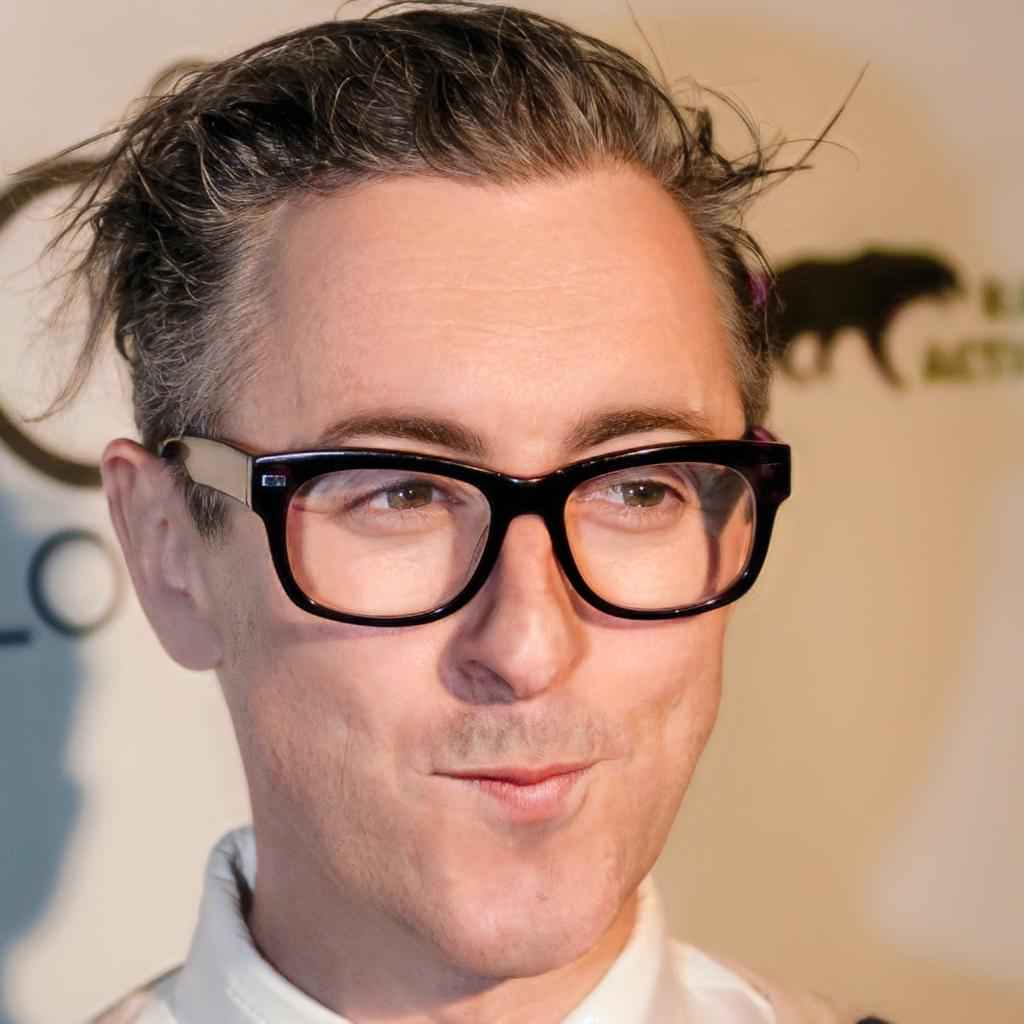} &
{\color{yellow}%
\setlength{\fboxsep}{0pt}%
\setlength{\fboxrule}{2pt}%
\fbox{\includegraphics[width=0.19\linewidth]{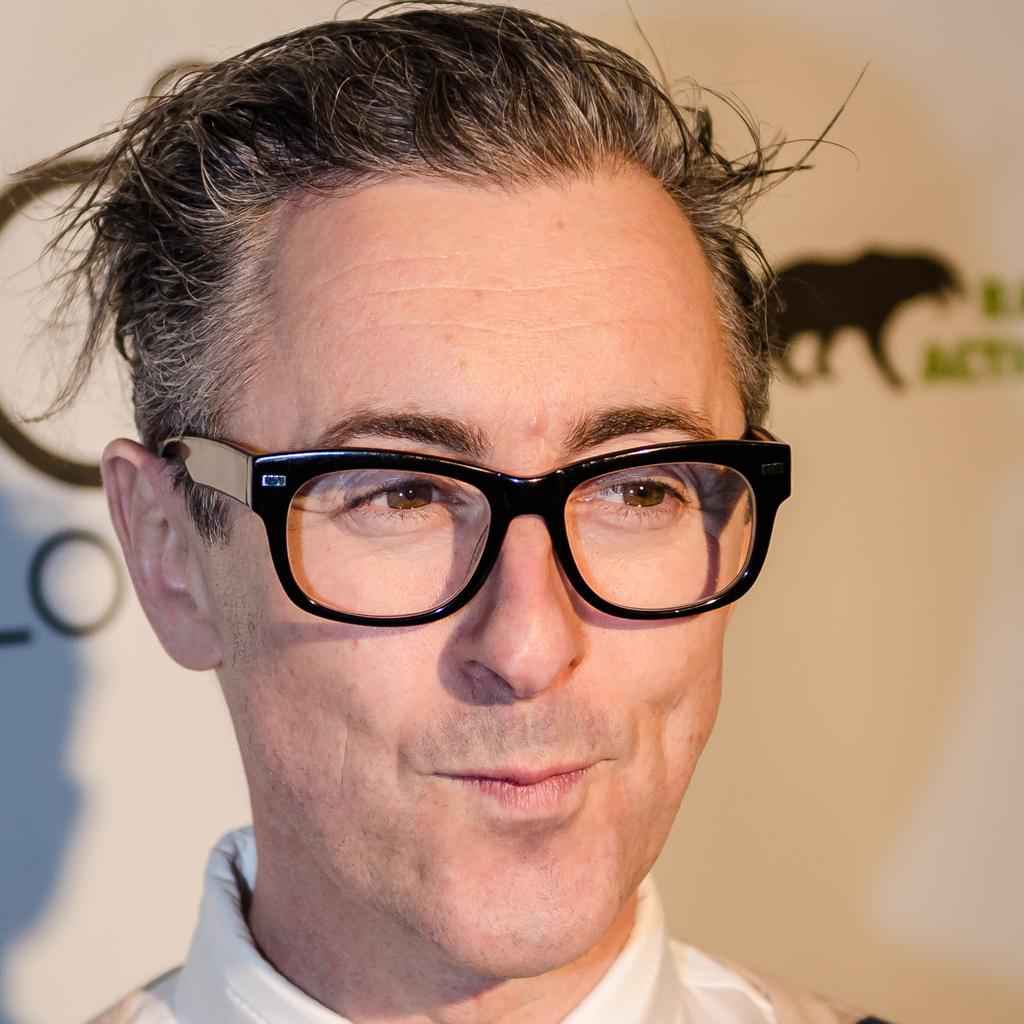}}} & 
\includegraphics[width=0.19\linewidth]{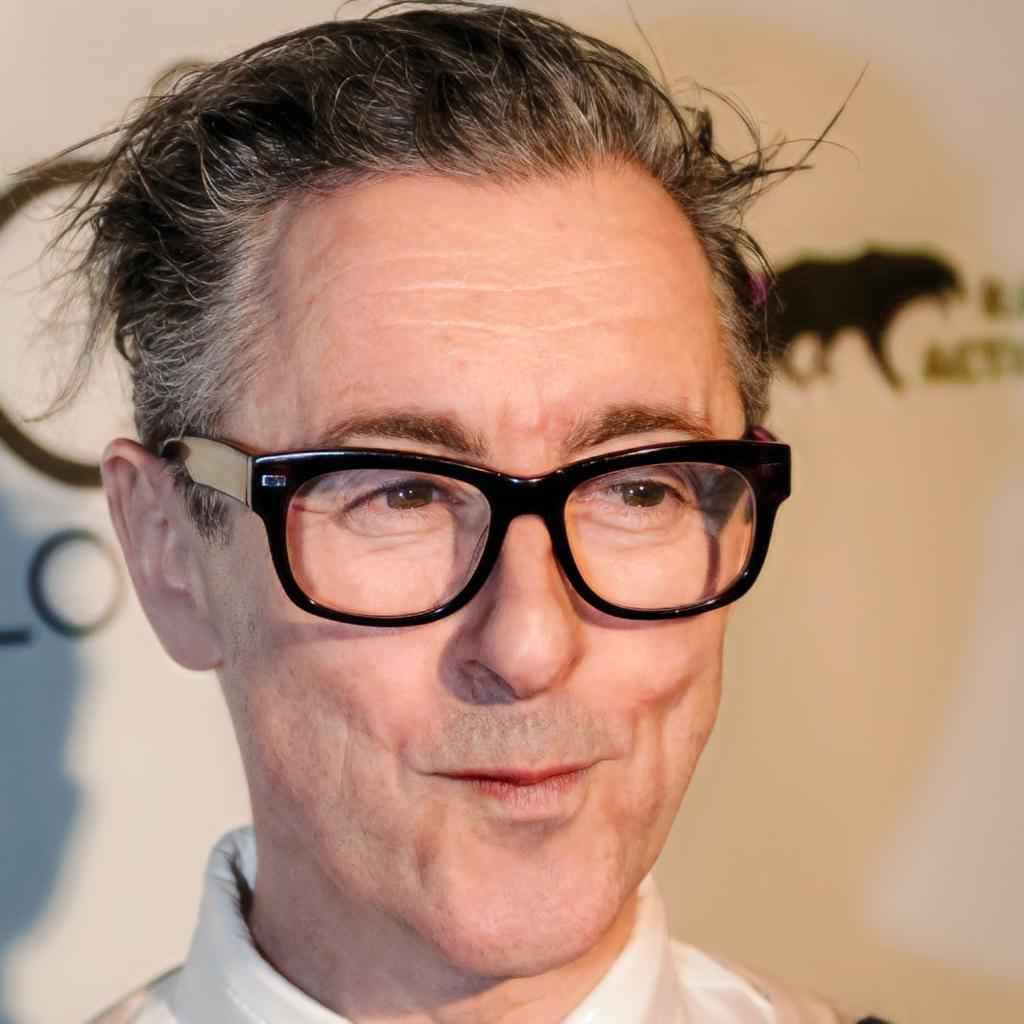} & 
\includegraphics[width=0.19\linewidth]{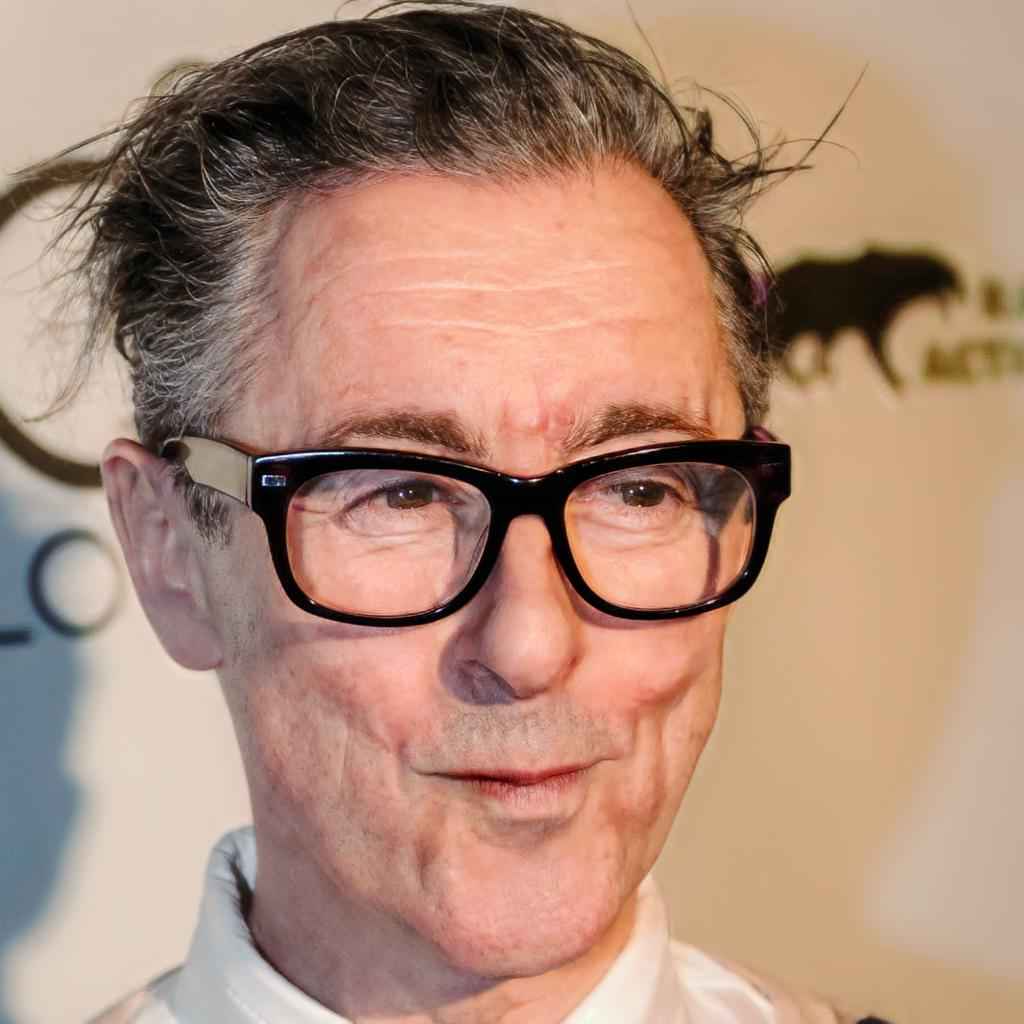} 
\\
\includegraphics[width=0.19\linewidth]{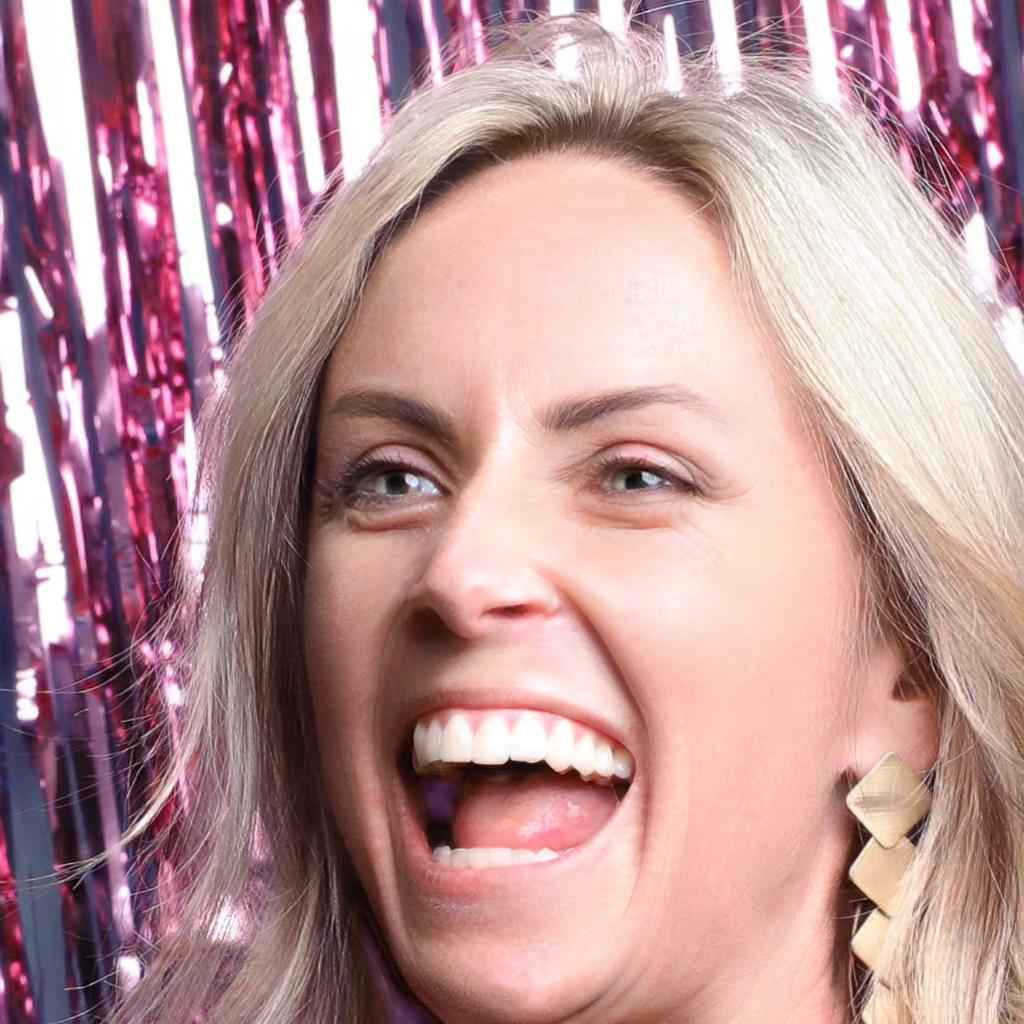} & 
\includegraphics[width=0.19\linewidth]{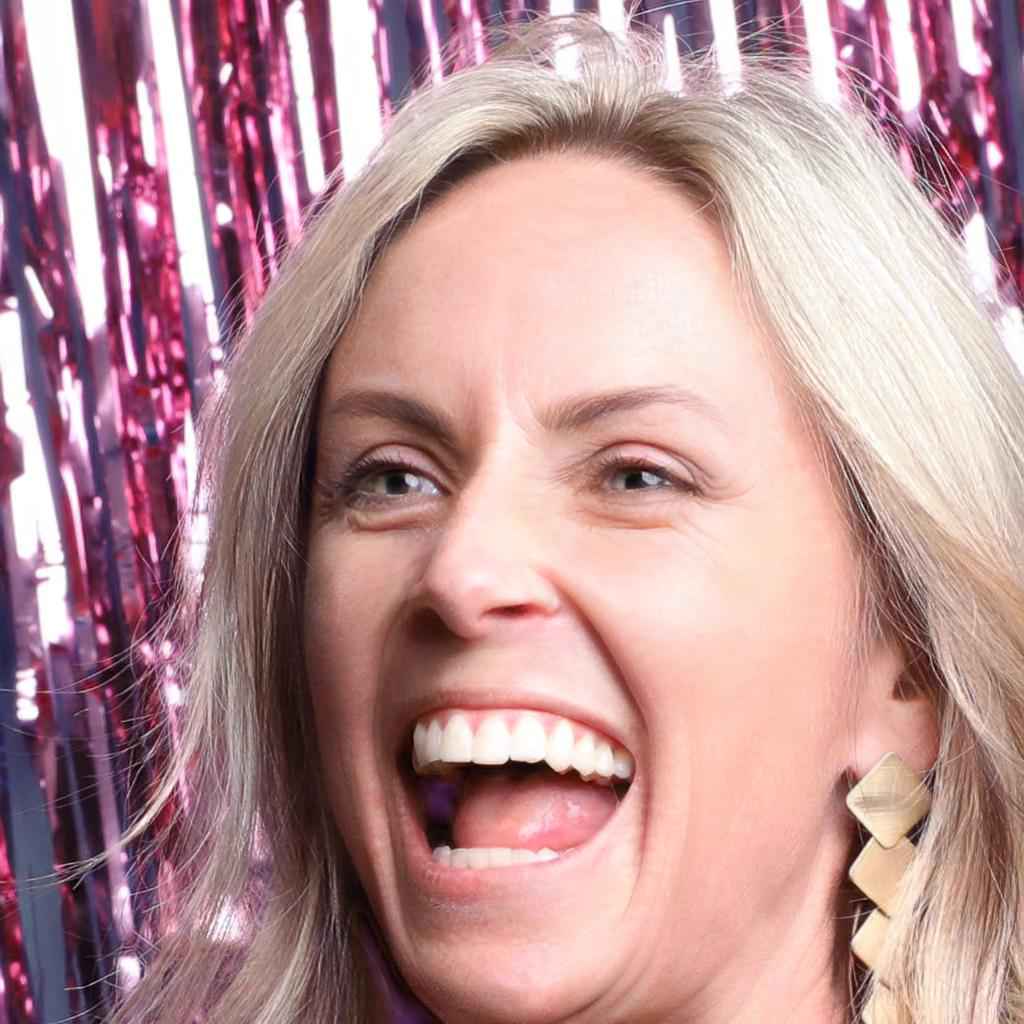} &
\includegraphics[width=0.19\linewidth]{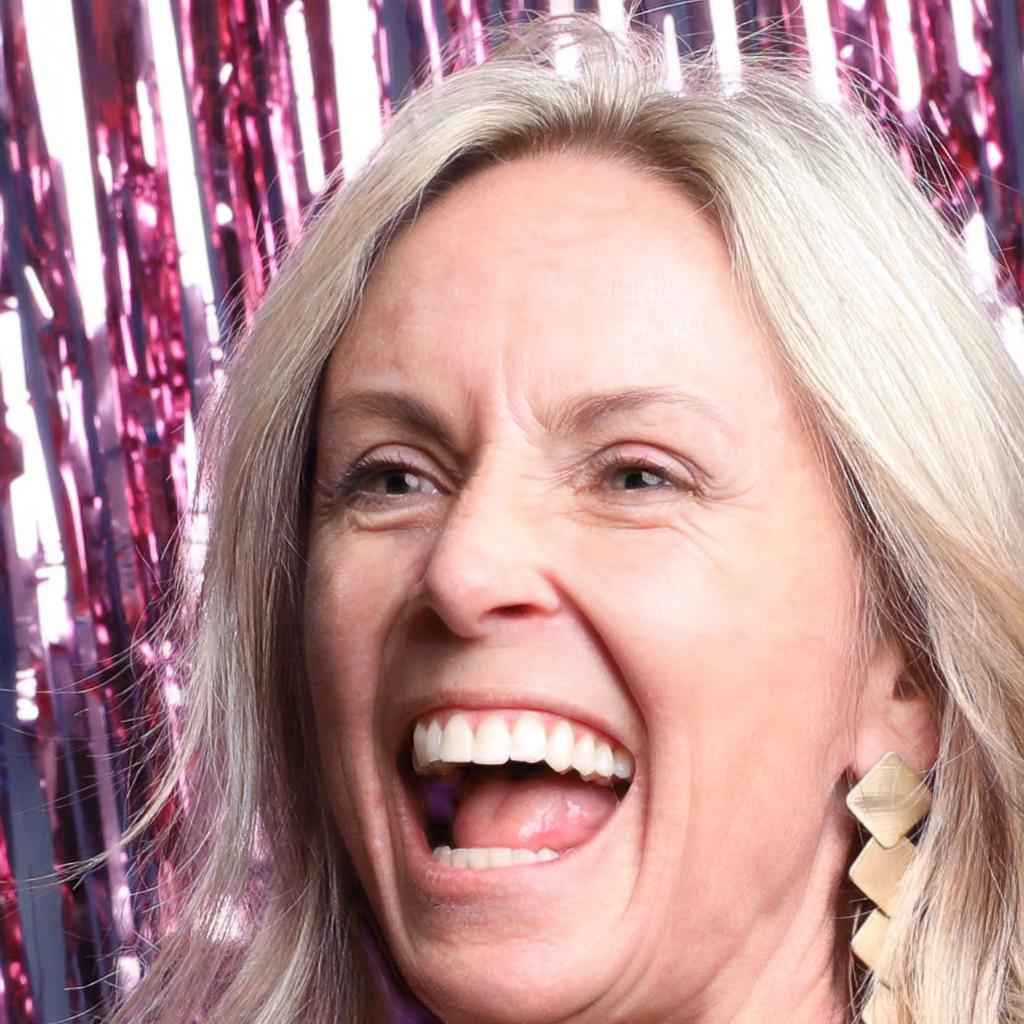} &
{\color{yellow}%
\setlength{\fboxsep}{0pt}%
\setlength{\fboxrule}{2pt}%
\fbox{\includegraphics[width=0.19\linewidth]{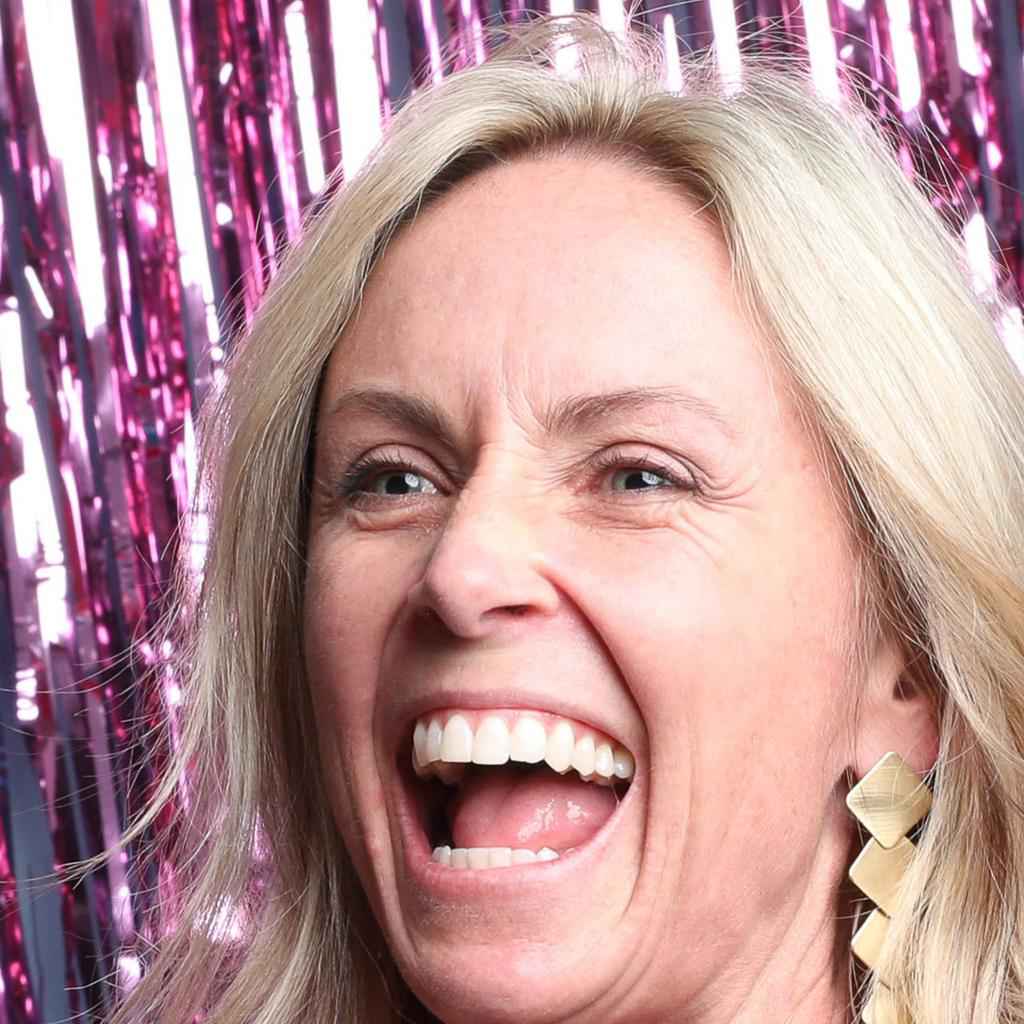}}} &  
\includegraphics[width=0.19\linewidth]{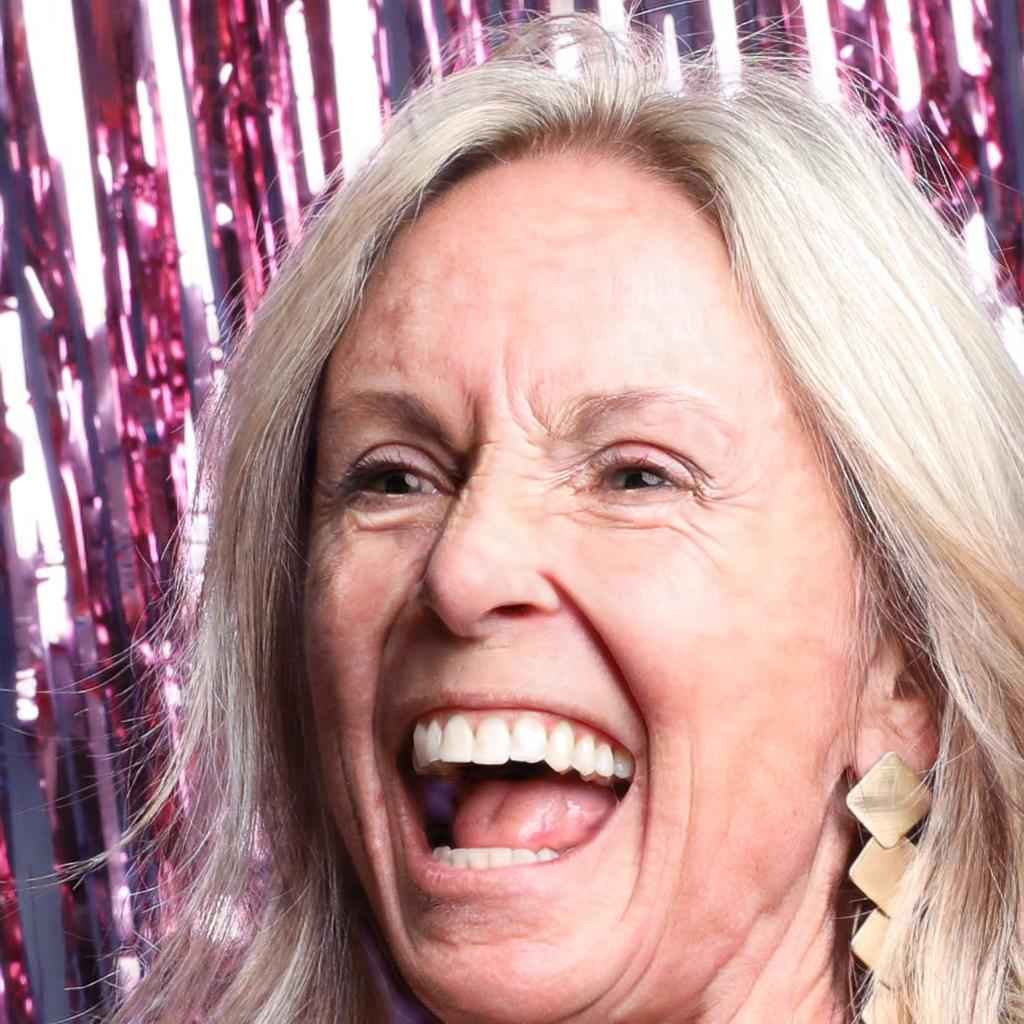} 
\\
\includegraphics[width=0.19\linewidth]{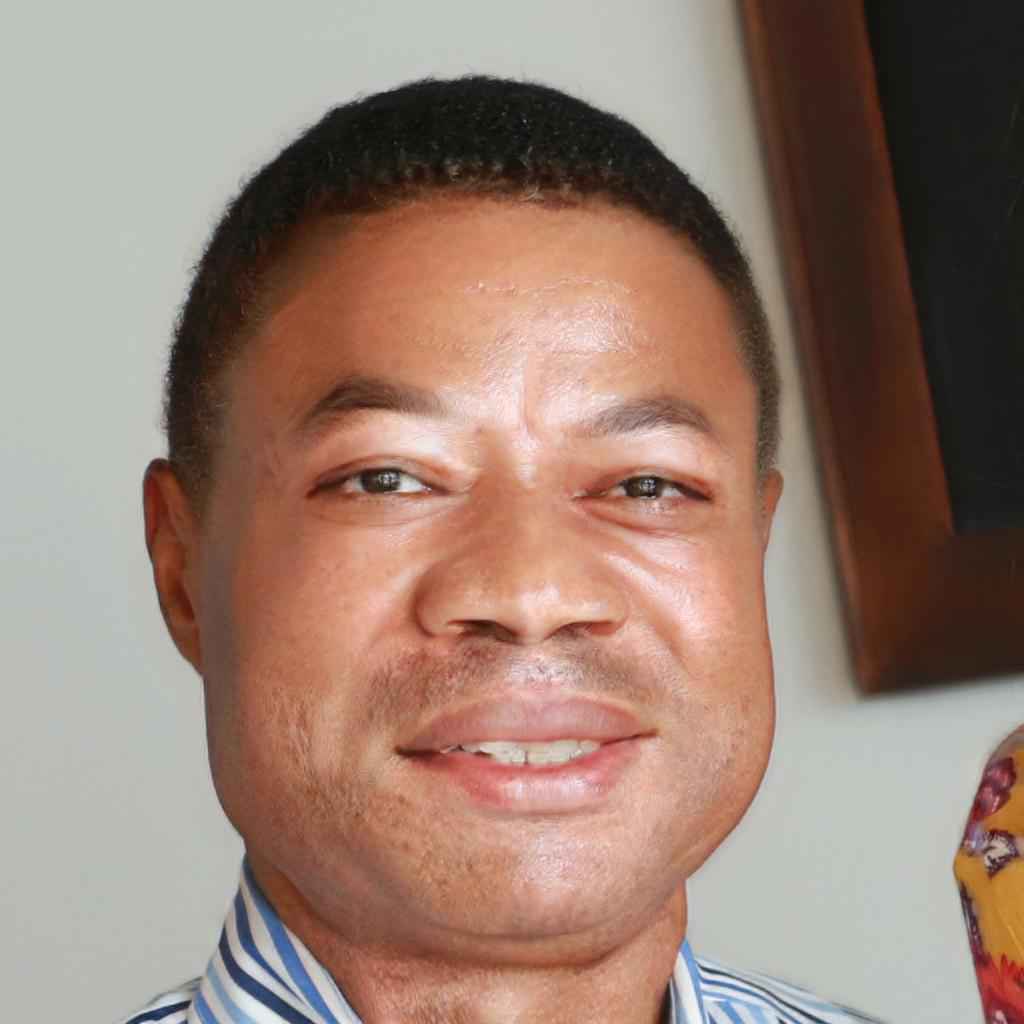} & 
\includegraphics[width=0.19\linewidth]{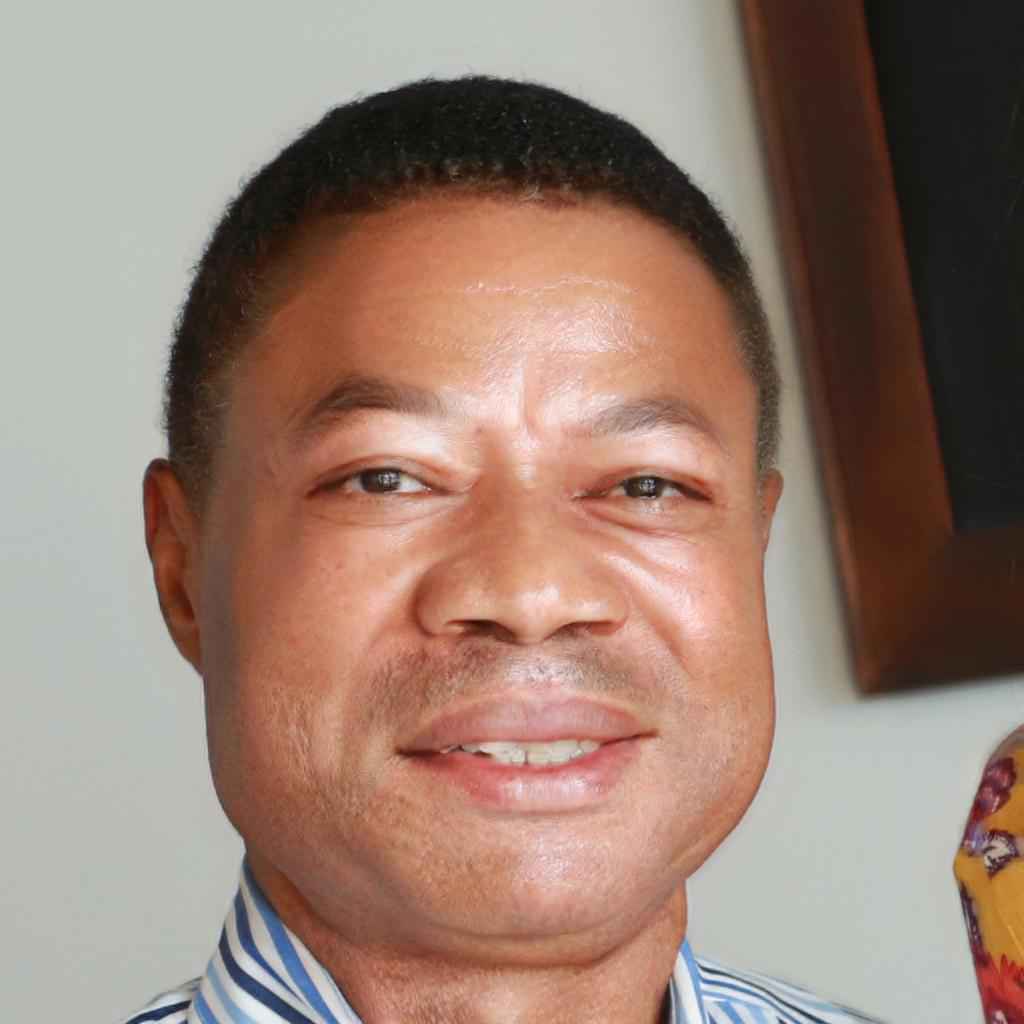} &
\includegraphics[width=0.19\linewidth]{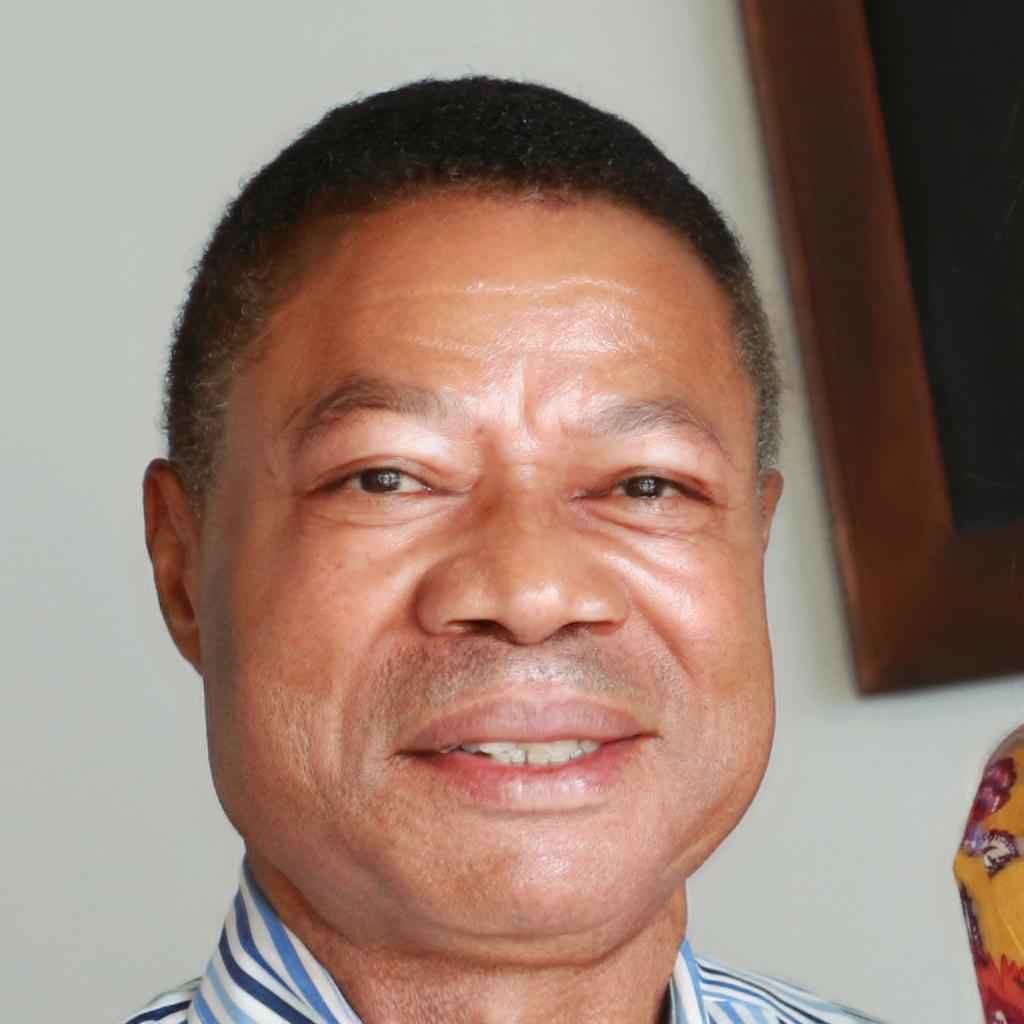} &
\includegraphics[width=0.19\linewidth]{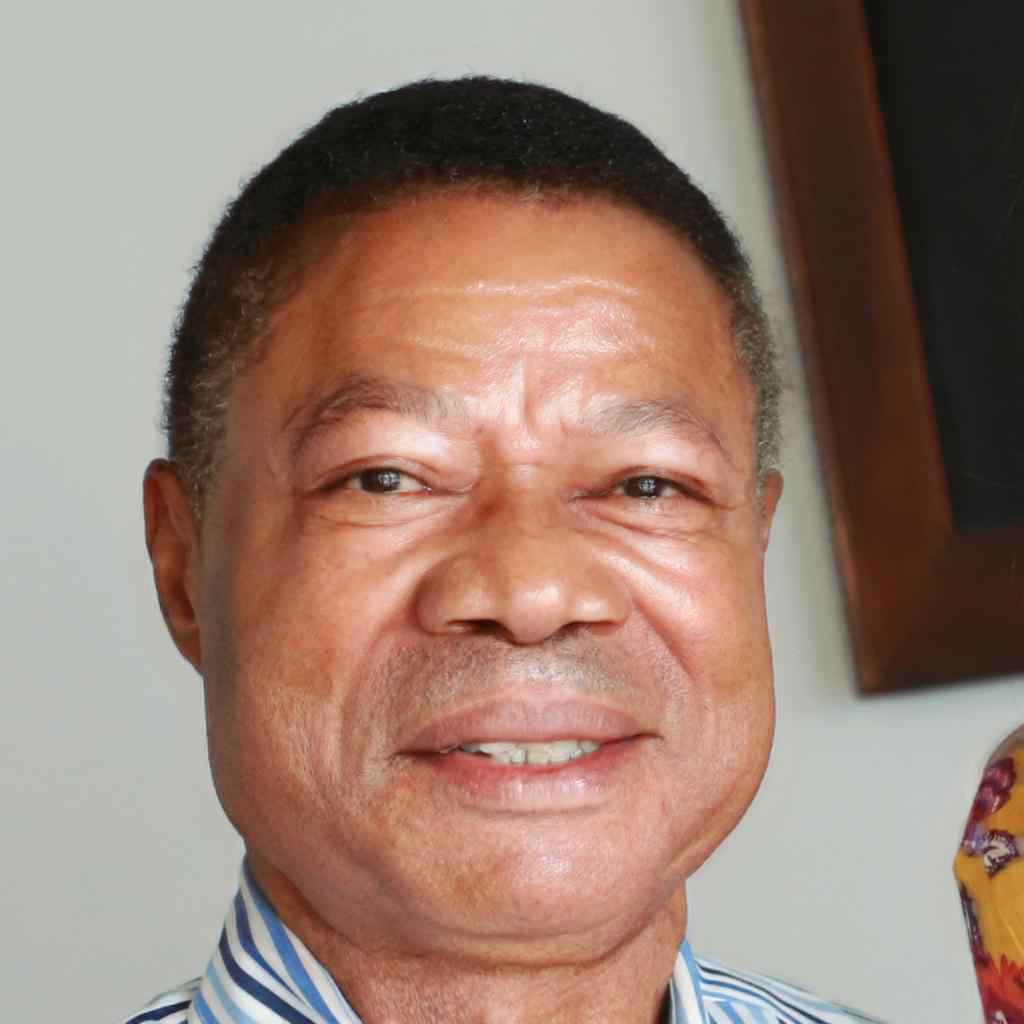} & 
{\color{yellow}%
\setlength{\fboxsep}{0pt}%
\setlength{\fboxrule}{2pt}%
\fbox{\includegraphics[width=0.19\linewidth]{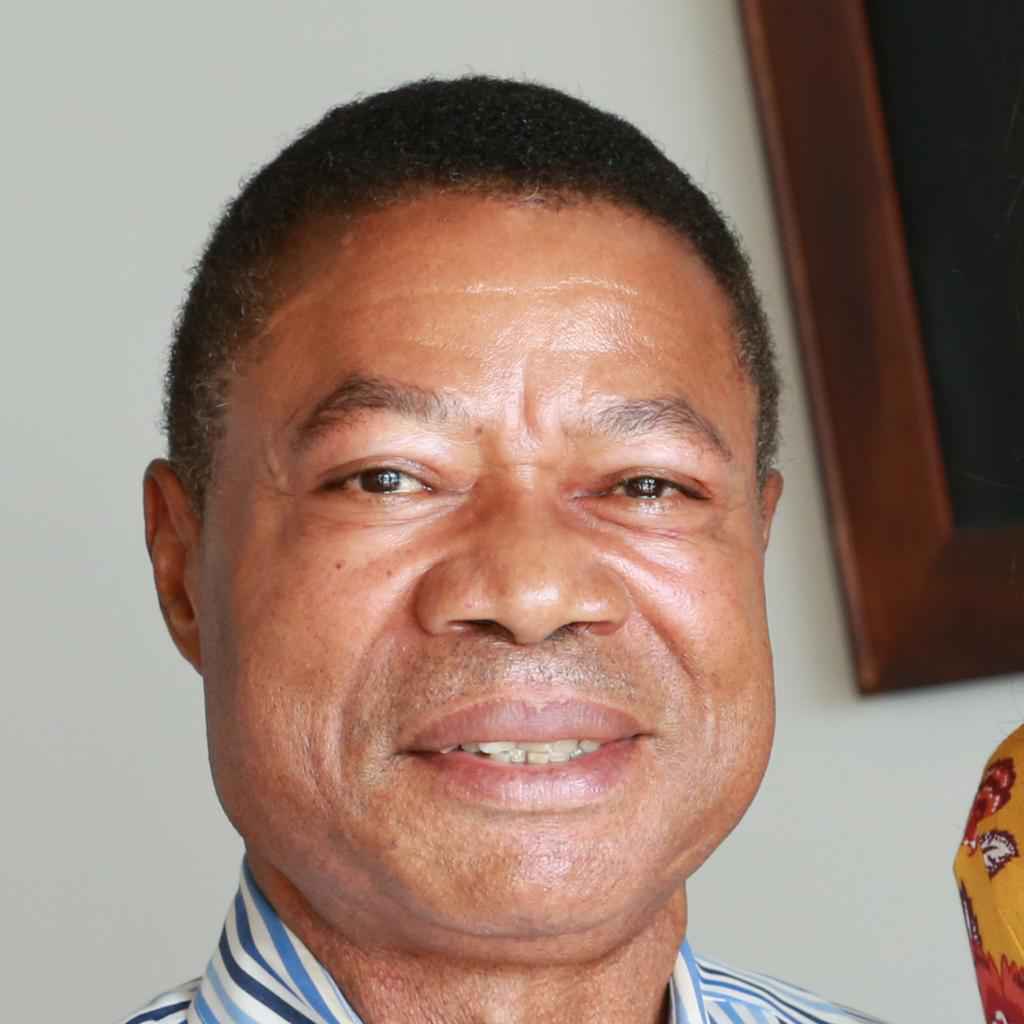}}} 
\end{tabular}
\end{center}
\caption{\textbf{Age transformation on $\bf 1024 \times 1024$ images}. On each row, the yellow frame indicates the original image. Each column corresponds to a target age of: $25$, $35$, $45$, $55$, $65$. Our approach yields visually satisfying results without introducing significant artifacts. Only age relevant features are modified, while the identity, haircut, emotion and background are perfectly preserved. 
}
\label{1024_2}
\end{figure*}

More age transform results on $1024 \times 1024$ images of FFHQ dataset are presented in Figure \ref{1024_1} and \ref{1024_2}. 

\subsection{Comparison with other methods}

\begin{figure*}
\centering
\setlength{\tabcolsep}{2pt}
\begin{tabular}{ccccc}
Input ($1024^2$)&Fader ($256^2$)&PAGGAN ($224^2$)&IPCGAN ($128^2$)&Ours ($1024^2$)\\
\includegraphics[width=0.19\linewidth]{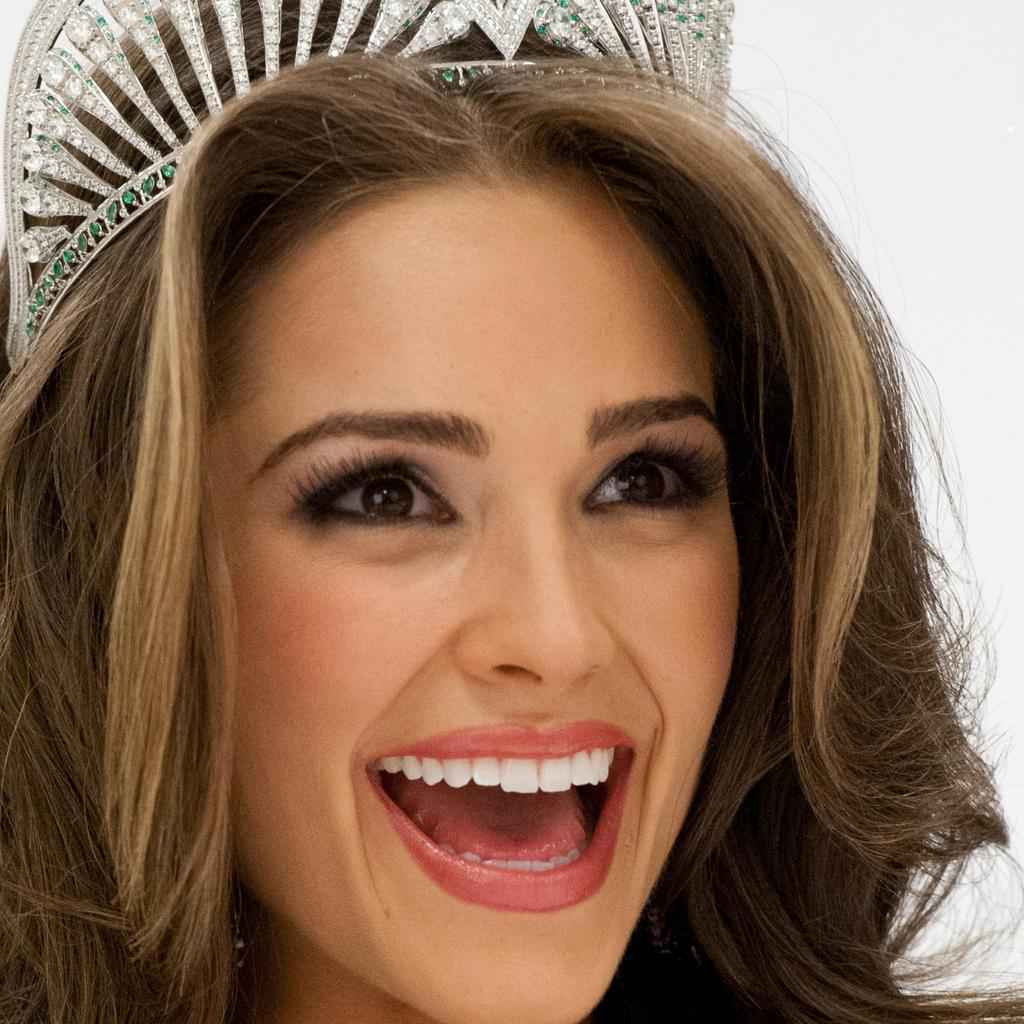} &
\includegraphics[width=0.19\linewidth]{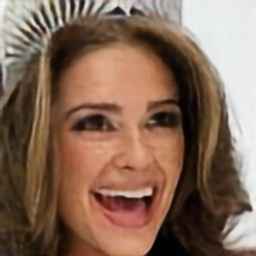} &
\includegraphics[width=0.19\linewidth]{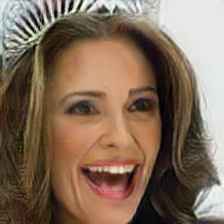} &
\includegraphics[width=0.19\linewidth]{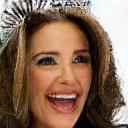} &
\includegraphics[width=0.19\linewidth]{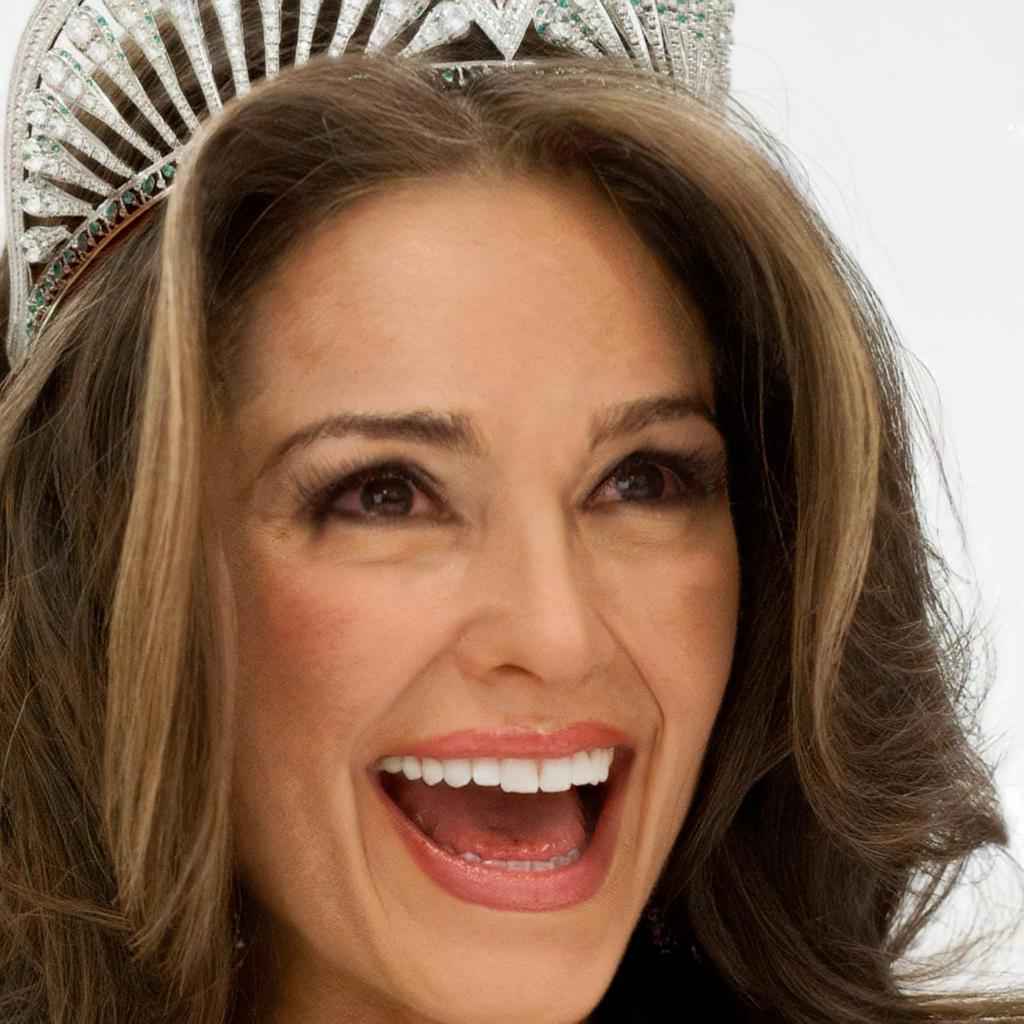} \\
\includegraphics[width=0.19\linewidth]{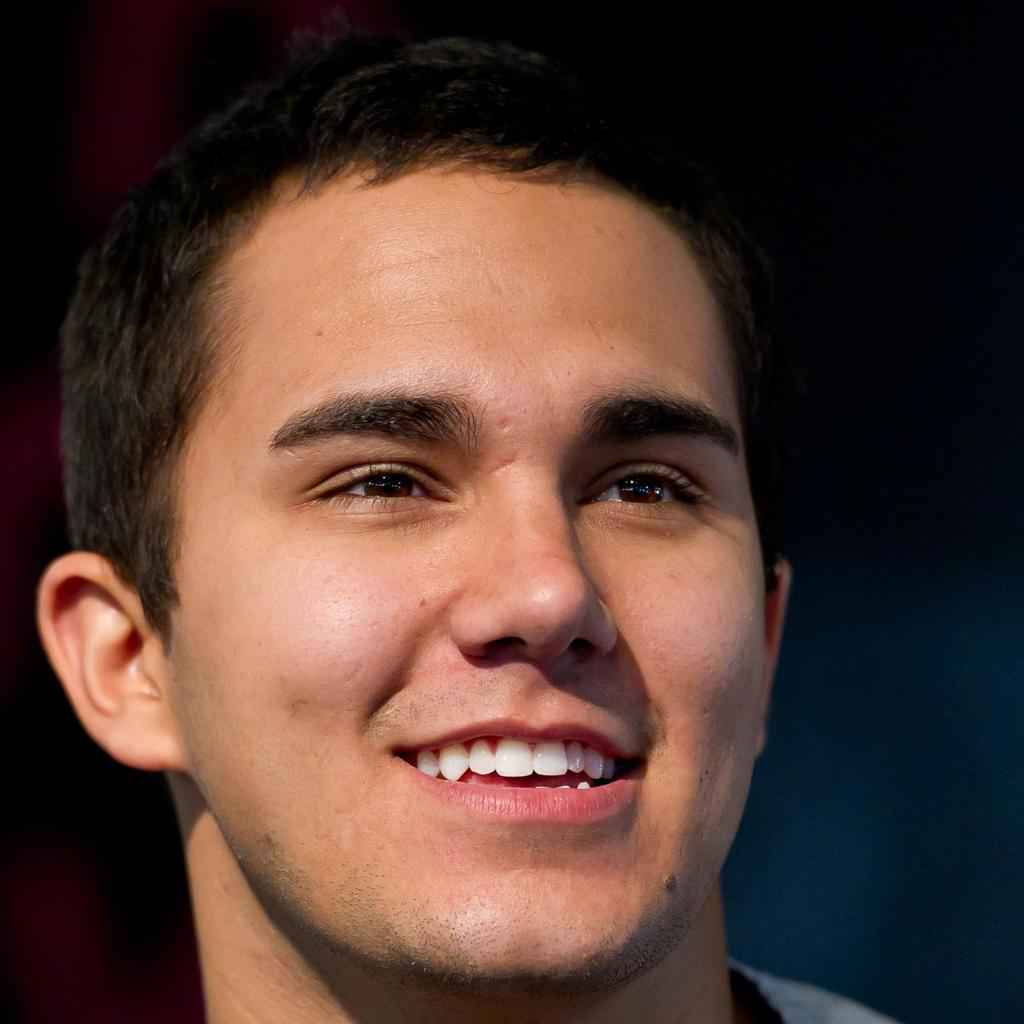} &
\includegraphics[width=0.19\linewidth]{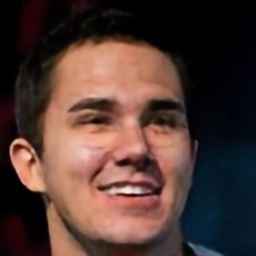} &
\includegraphics[width=0.19\linewidth]{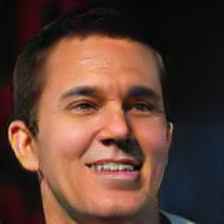} &
\includegraphics[width=0.19\linewidth]{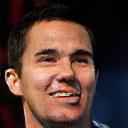} &
\includegraphics[width=0.19\linewidth]{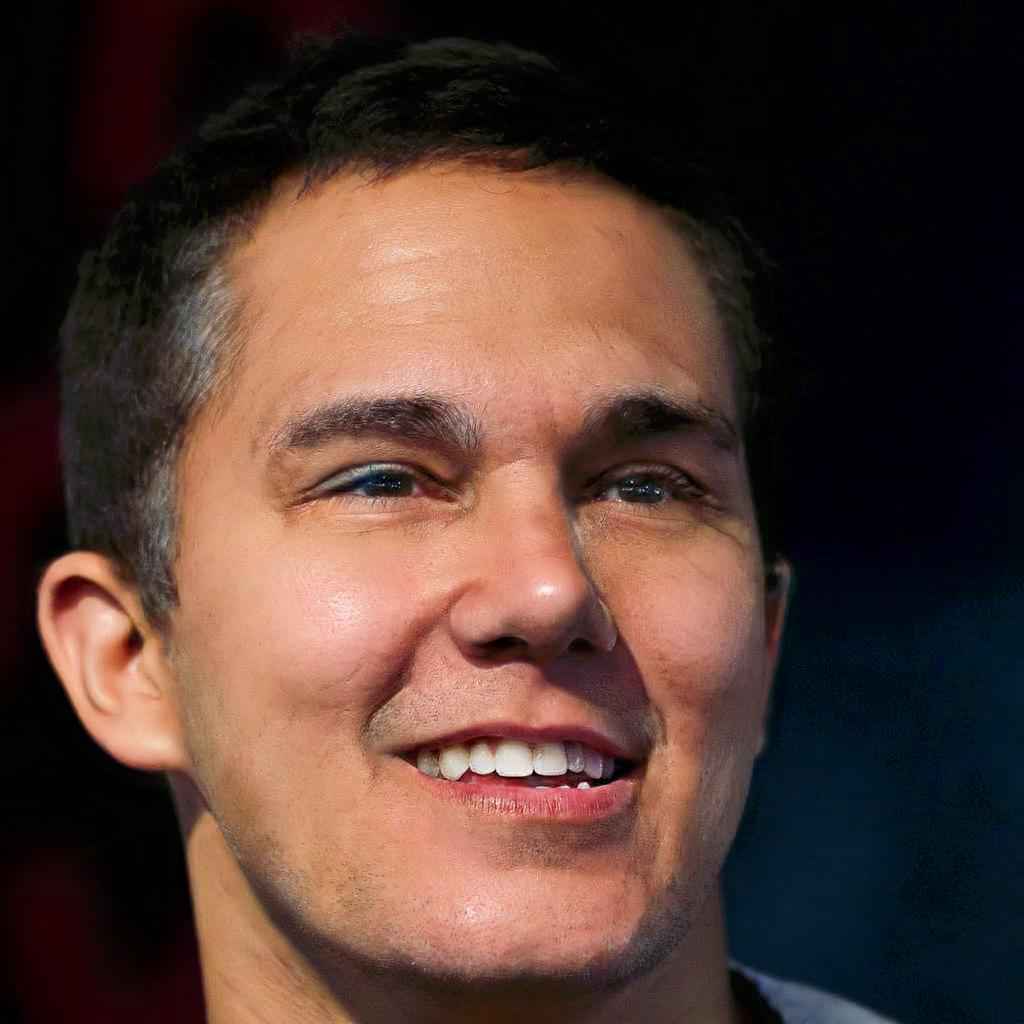} \\
\includegraphics[width=0.19\linewidth]{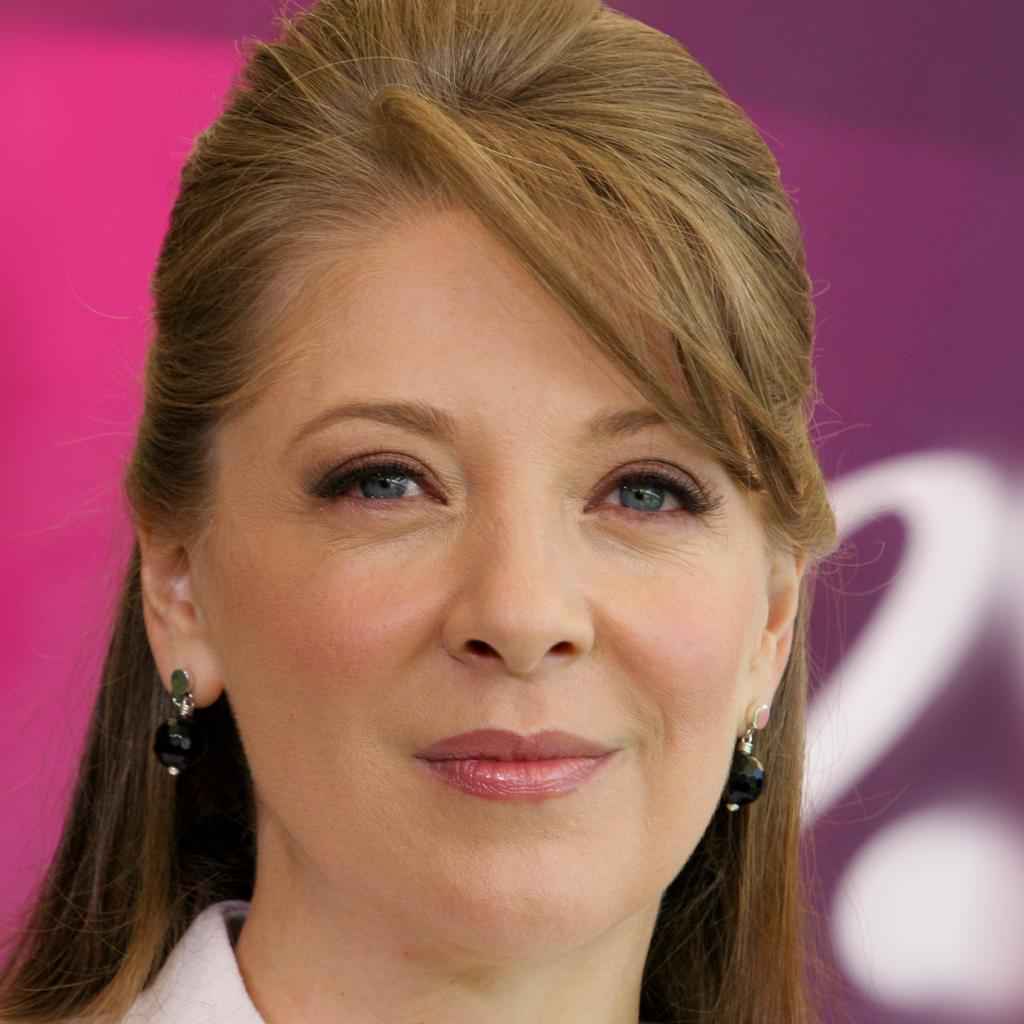} &
\includegraphics[width=0.19\linewidth]{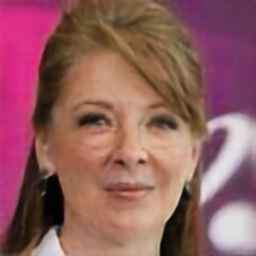} &
\includegraphics[width=0.19\linewidth]{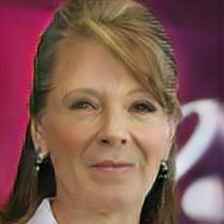} &
\includegraphics[width=0.19\linewidth]{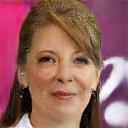} &
\includegraphics[width=0.19\linewidth]{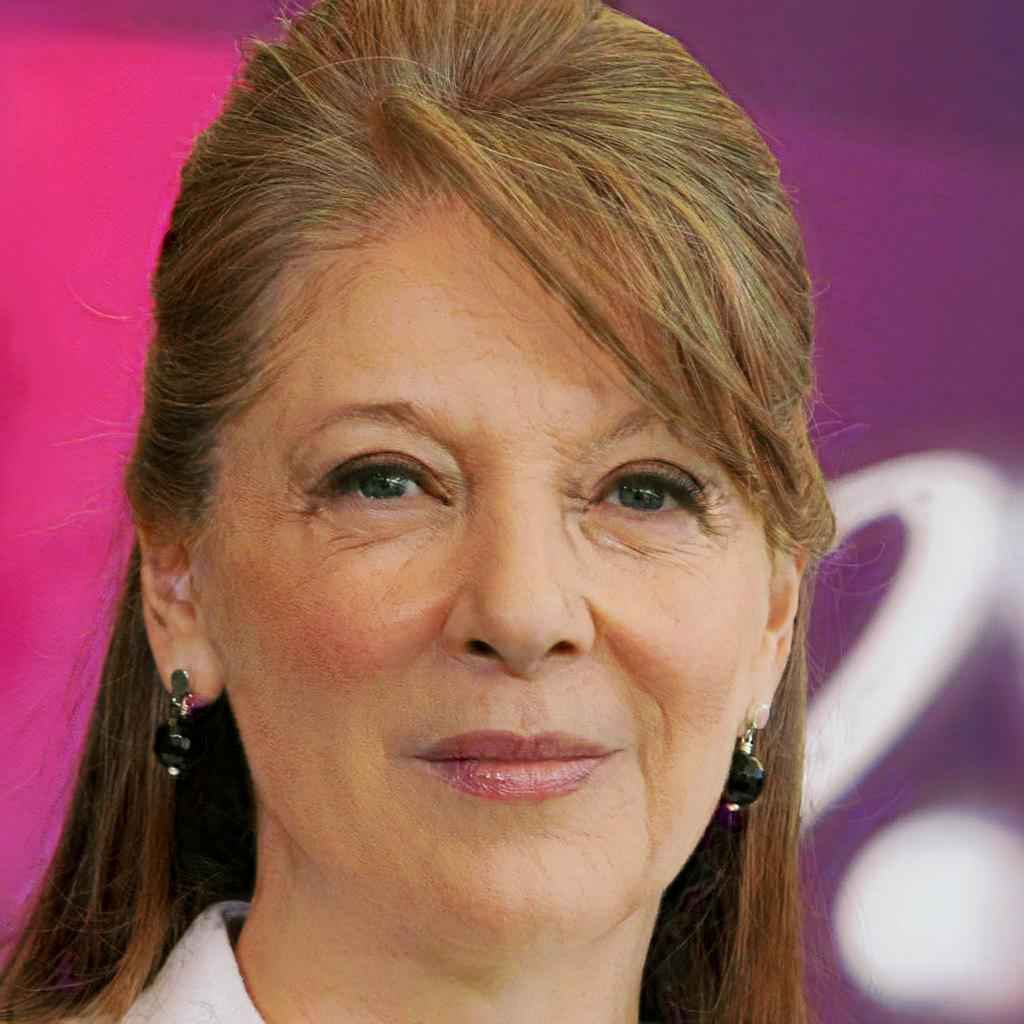} \\
\includegraphics[width=0.19\linewidth]{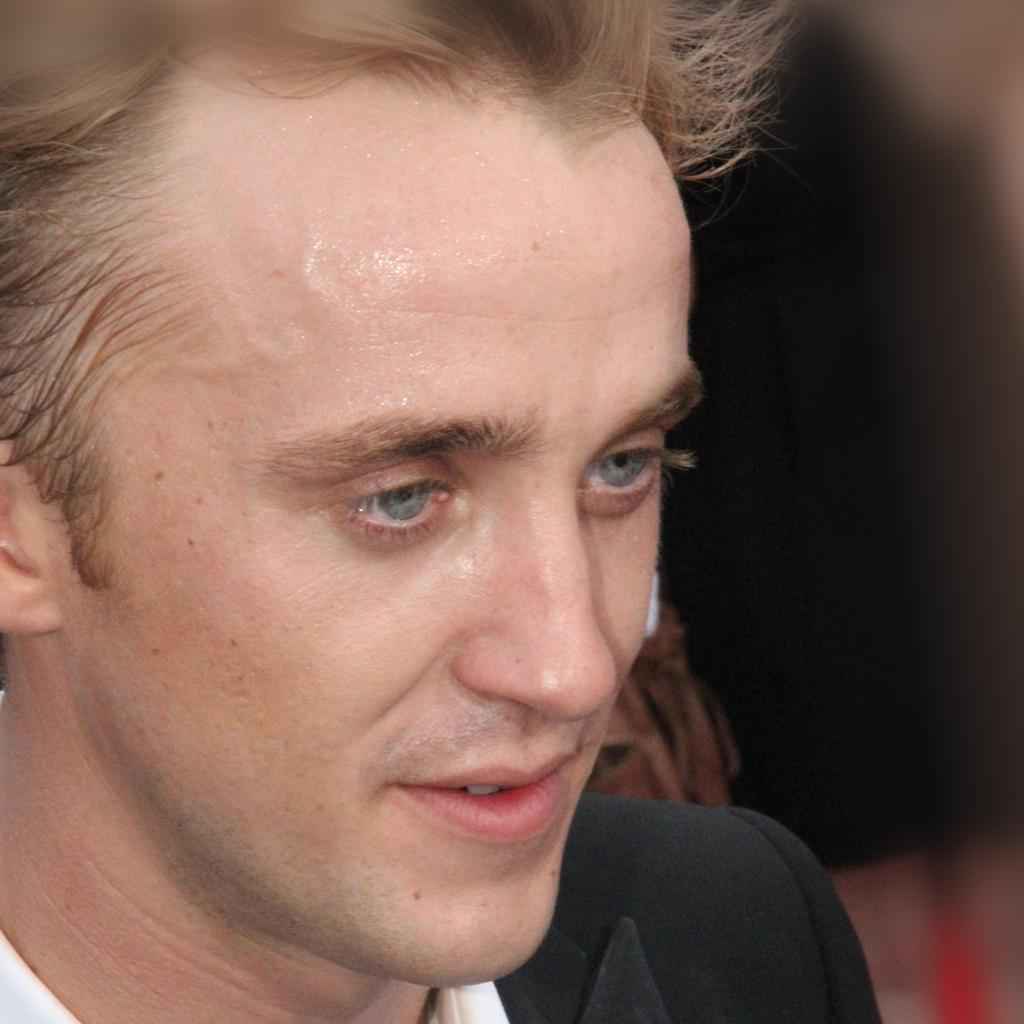} &
\includegraphics[width=0.19\linewidth]{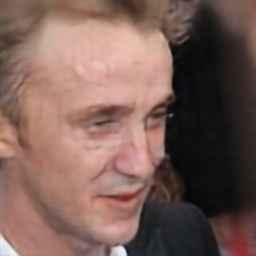} &
\includegraphics[width=0.19\linewidth]{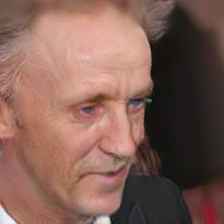} &
\includegraphics[width=0.19\linewidth]{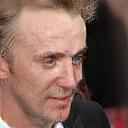} &
\includegraphics[width=0.19\linewidth]{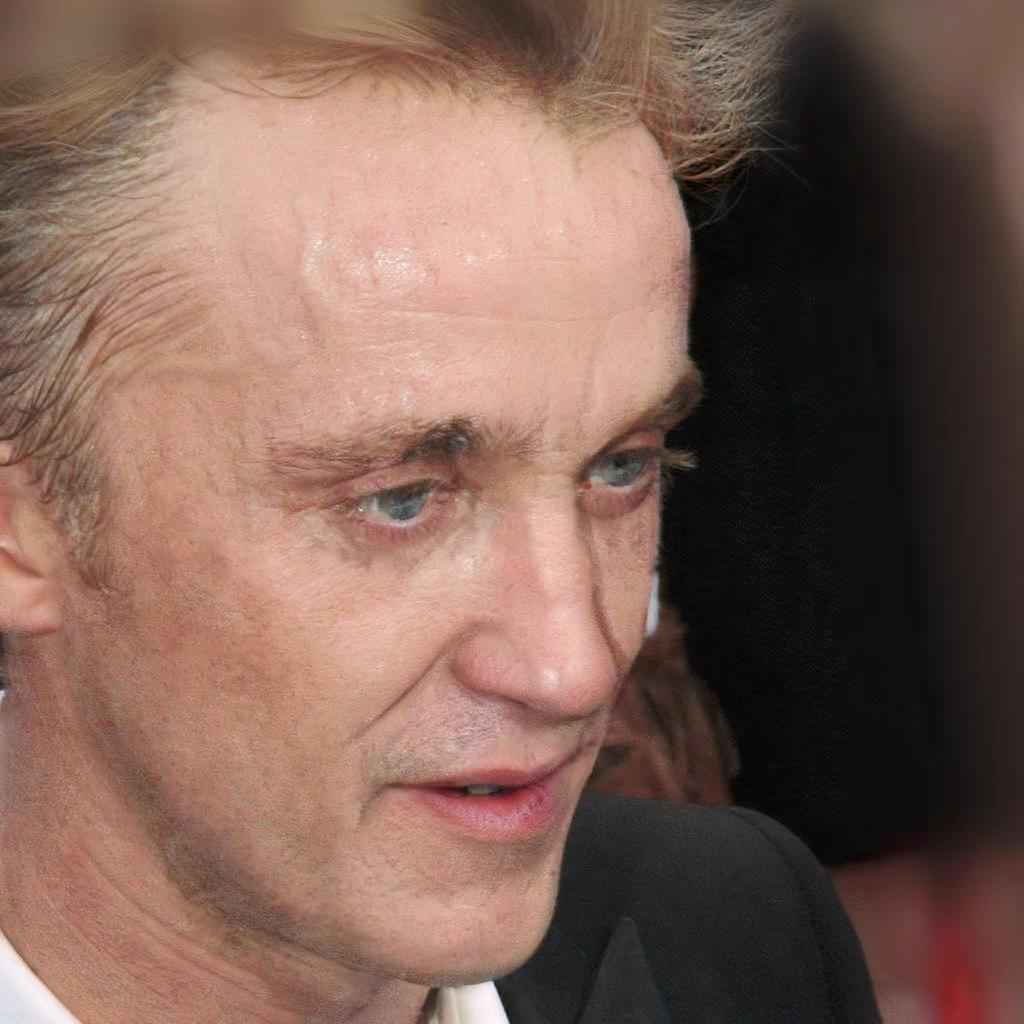} \\
\includegraphics[width=0.19\linewidth]{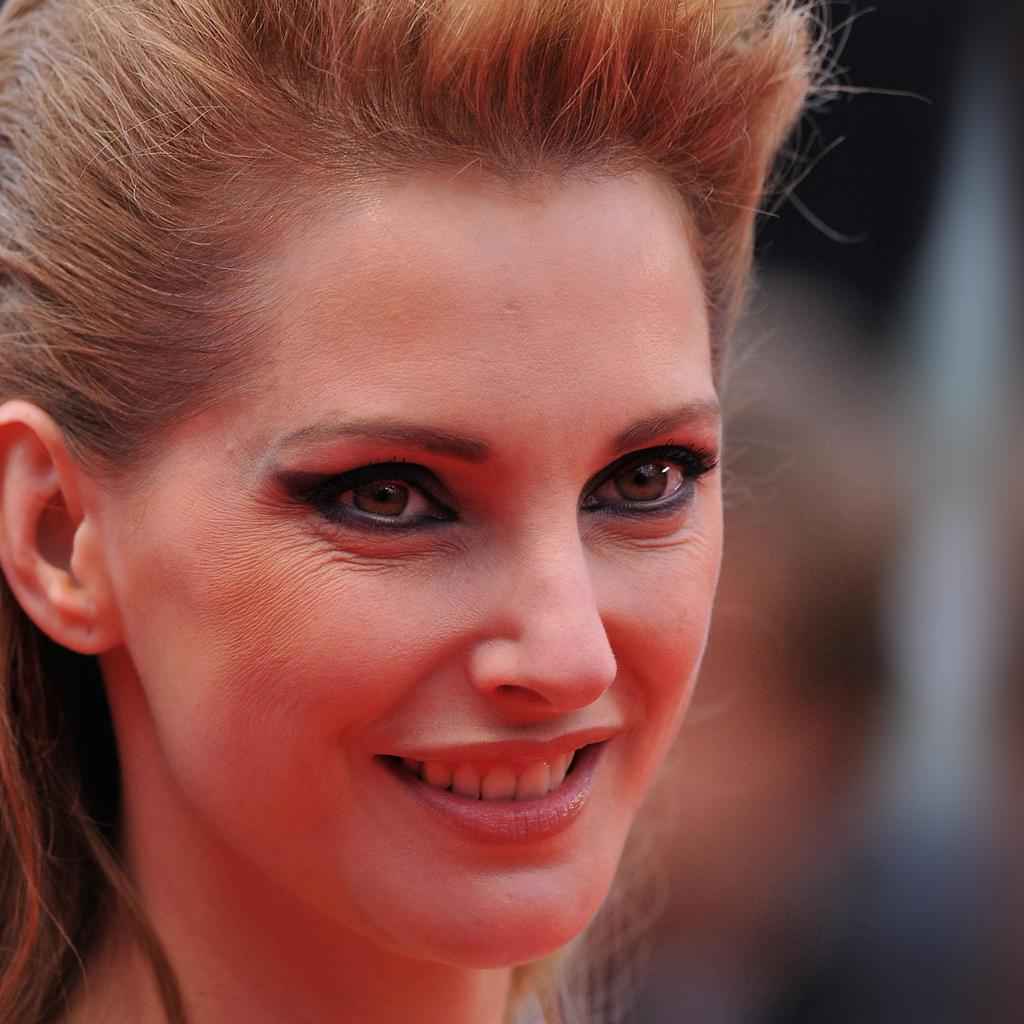} &
\includegraphics[width=0.19\linewidth]{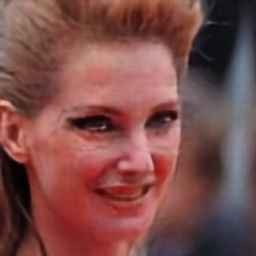} &
\includegraphics[width=0.19\linewidth]{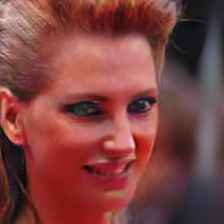} &
\includegraphics[width=0.19\linewidth]{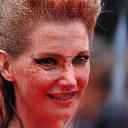} &
\includegraphics[width=0.19\linewidth]{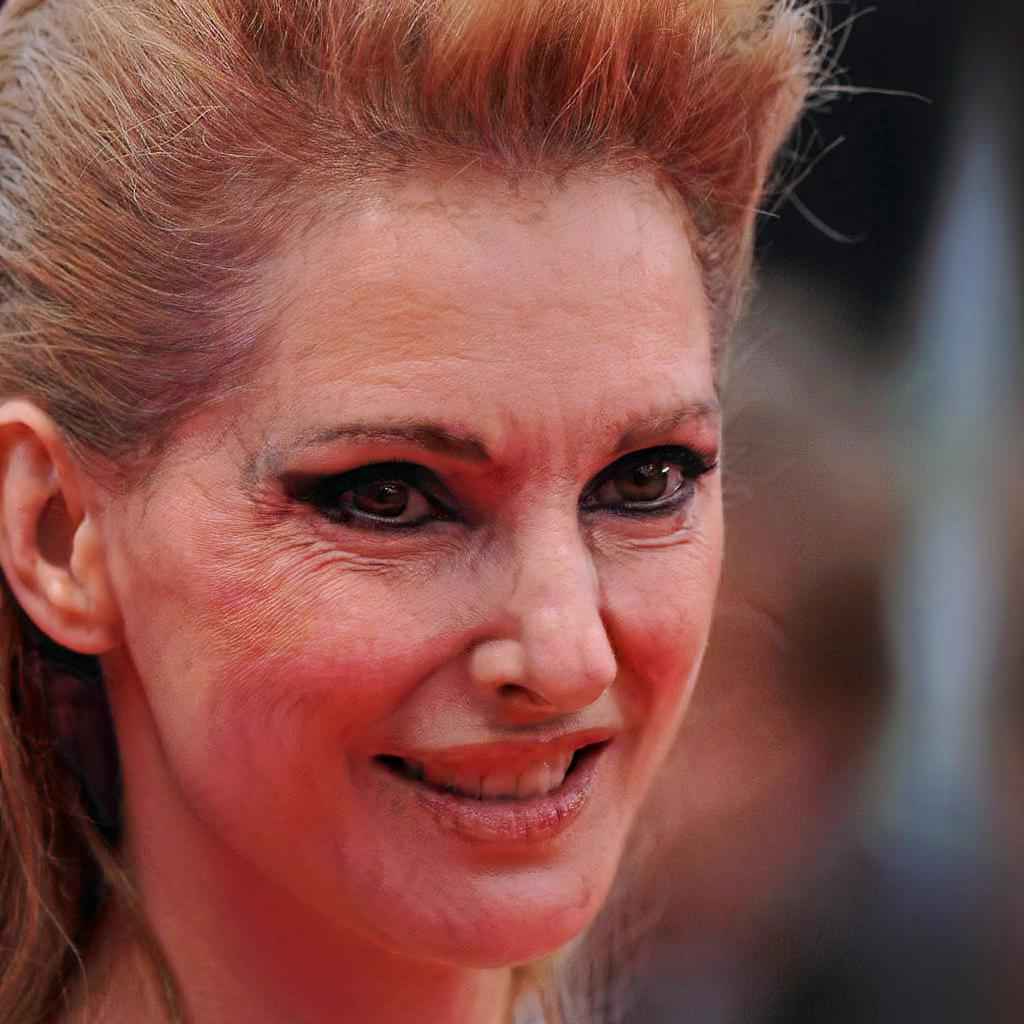} \\
\includegraphics[width=0.19\linewidth]{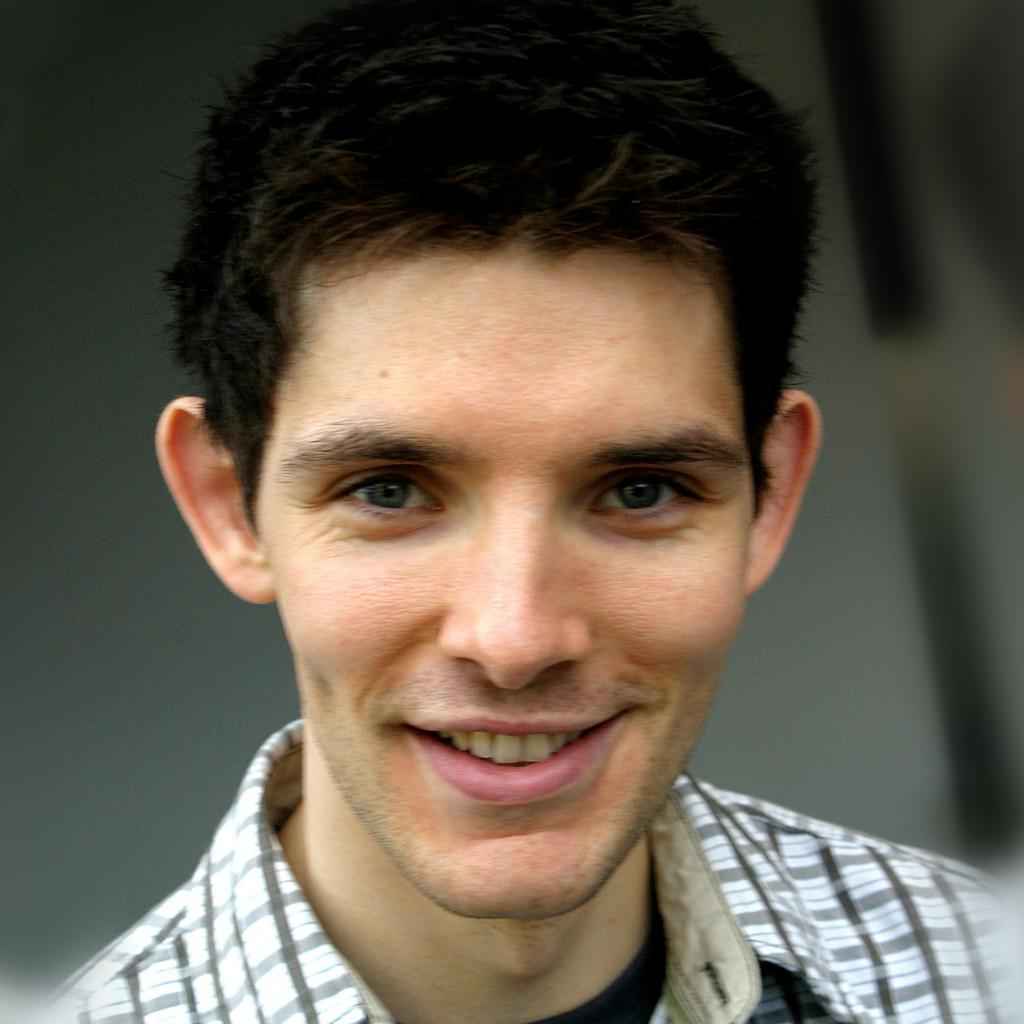} &
\includegraphics[width=0.19\linewidth]{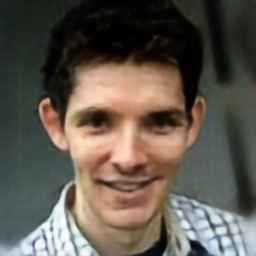} &
\includegraphics[width=0.19\linewidth]{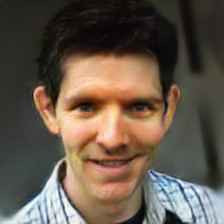} &
\includegraphics[width=0.19\linewidth]{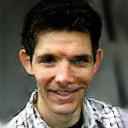} &
\includegraphics[width=0.19\linewidth]{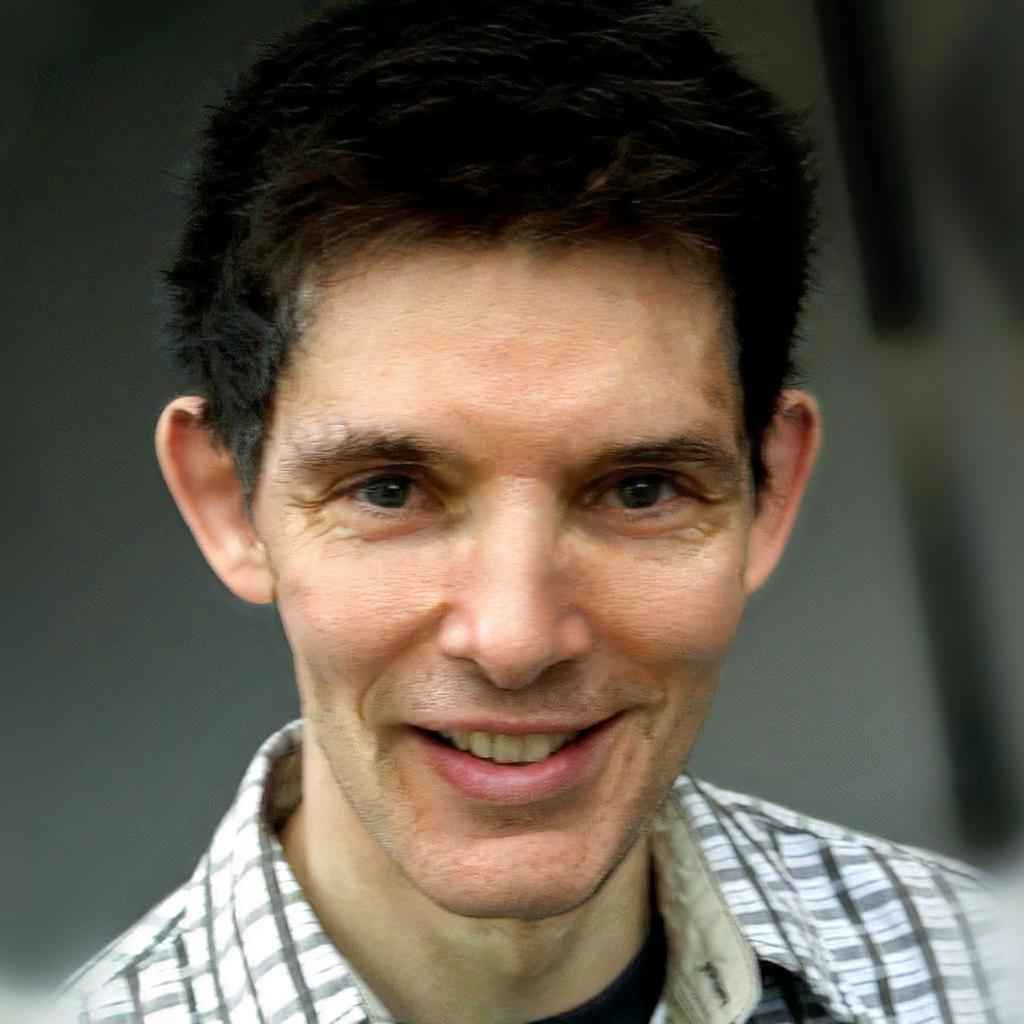} \\
\end{tabular}
\caption{
\textbf{Comparison of face aging results on CelebA HQ \cite{karras2018progressive}}. The first column are the input images. The second to fifth column are outputs from Fader Network \cite{lample2017fader}, PAG-GAN \cite{yang2018learning}, IPC-GAN \cite{wang2018face} and our method. Our results reach the highest resolution without introducing significant artifacts. Our method preserves the background better compared to other techniques, see for instance the letters on the third row. In addition, compared to other techniques, our method leads to  results that are free of artefacts and blur.
}
\label{celeba}
\end{figure*}

In Figure \ref{celeba}, we show additional comparison of face aging results on Celeba-HQ \cite{karras2018progressive}. As mentioned in the paper, we compare our method against the two most recent state-of-the-art methods on face aging for which the official codes are released - PAGGAN \cite{yang2018learning} and IPCGAN \cite{wang2018face}. We also compare our results to those obtained with Fader Network \cite{lample2017fader}, which allows one to manipulate several facial attributes including the age. Each input image is transformed to the oldest age group using their official released models. For IPCGAN and PAGGAN, the oldest age group refer to $50+$ and $[51, 60]$ respectively. For Fader Network, the age attribute is set to be the default largest value for aging in their official code. To have a fair comparison with groupwise methods, and since $50+$ is considered as the oldest age group, we choose a target age of $60$ (the mean of the age range $\{51, \ldots, 69\} \subset \mathcal{Q}$) for our age transformer.

\begin{table}[t]
\small
\centering
\caption{\textbf{Hyperparameters of the proposed network architecture}. The input size is $1024 \times 1024 \times3$. For the age transformer, except the last one, each convolution is followed by an instance normalization and a LeakyReLU activation. For the discriminator, except the first and the last one, each convolution is followed by a batch normalization and a LeakyReLU activation. }
\label{arch_table}
\begin{tabular}{>{\raggedleft\arraybackslash}p{3cm}ccc}
\toprule
Operation & Kernel size & Stride & Channel
\\
\midrule
\multicolumn{4}{l}{\textbf{Age transformer}}
\\
\multicolumn{4}{l}{\textbf{Encoder}}
\\
Convolution & $9 \times 9$ & $1$ & $32$ 
\\
Convolution & $3 \times 3$ & $2$ & $64$ 
\\
\multicolumn{4}{c}{\textit{Skip connection 1}}
\\
Convolution & $3 \times 3$ & $2$ & $128$ 
\\
\multicolumn{4}{c}{\textit{Skip connection 2}}
\\
Residual block & $3 \times 3$ & $1$ & $128$ 
\\
Residual block & $3 \times 3$ & $1$ & $128$ 
\\
Residual block & $3 \times 3$ & $1$ & $128$ 
\\
Residual block & $3 \times 3$ & $1$ & $128$ 
\\
\multicolumn{4}{l}{\textbf{Modulation layer}}
\\
\multicolumn{4}{l}{\textbf{Decoder}}
\\
\multicolumn{4}{c}{\textit{Concatenation with skip connection 2}}
\\
Upsampling &&&
\\
Convolution & $3 \times 3$ & $1$ & $64$
\\
\multicolumn{4}{c}{\textit{Concatenation with skip connection 1}}
\\
Upsampling &&&
\\
Convolution & $3 \times 3$ & $1$ & $32$ 
\\
Convolution & $9 \times 9$ & $1$ & $3$ 
\\
\midrule
\multicolumn{4}{l}{\textbf{Discriminator}}
\\
Convolution & $4 \times 4$ & $2$ & $32$ 
\\
Convolution & $4 \times 4$ & $2$ & $64$ 
\\
Convolution & $4 \times 4$ & $2$ & $128$ 
\\
Convolution & $4 \times 4$ & $2$ & $256$ 
\\
Convolution & $4 \times 4$ & $1$ & $512$ 
\\
Convolution & $4 \times 4$ & $1$ & $1$ 
\\
\midrule
Upsampling mode & \multicolumn{3}{l}{Nearest (scale factor = $2$)}
\\
Padding mode & \multicolumn{3}{l}{Reflection}
\\
Normalization & \multicolumn{3}{l}{InstanceNorm for age transformer}
\\
 & \multicolumn{3}{l}{BatchNorm for discriminator}
\\
Activation & \multicolumn{3}{l}{LeakyReLU (negative slope = $0.2$)}
\\
\bottomrule
\end{tabular}
\end{table}
\begin{figure}[ht]
\begin{center}
\setlength{\tabcolsep}{1pt}
\begin{tabular}{ccccc}
25&35&45&55&65 \\
{\color{yellow}%
\setlength{\fboxsep}{0pt}%
\setlength{\fboxrule}{2pt}%
\fbox{\includegraphics[width=0.19\linewidth]{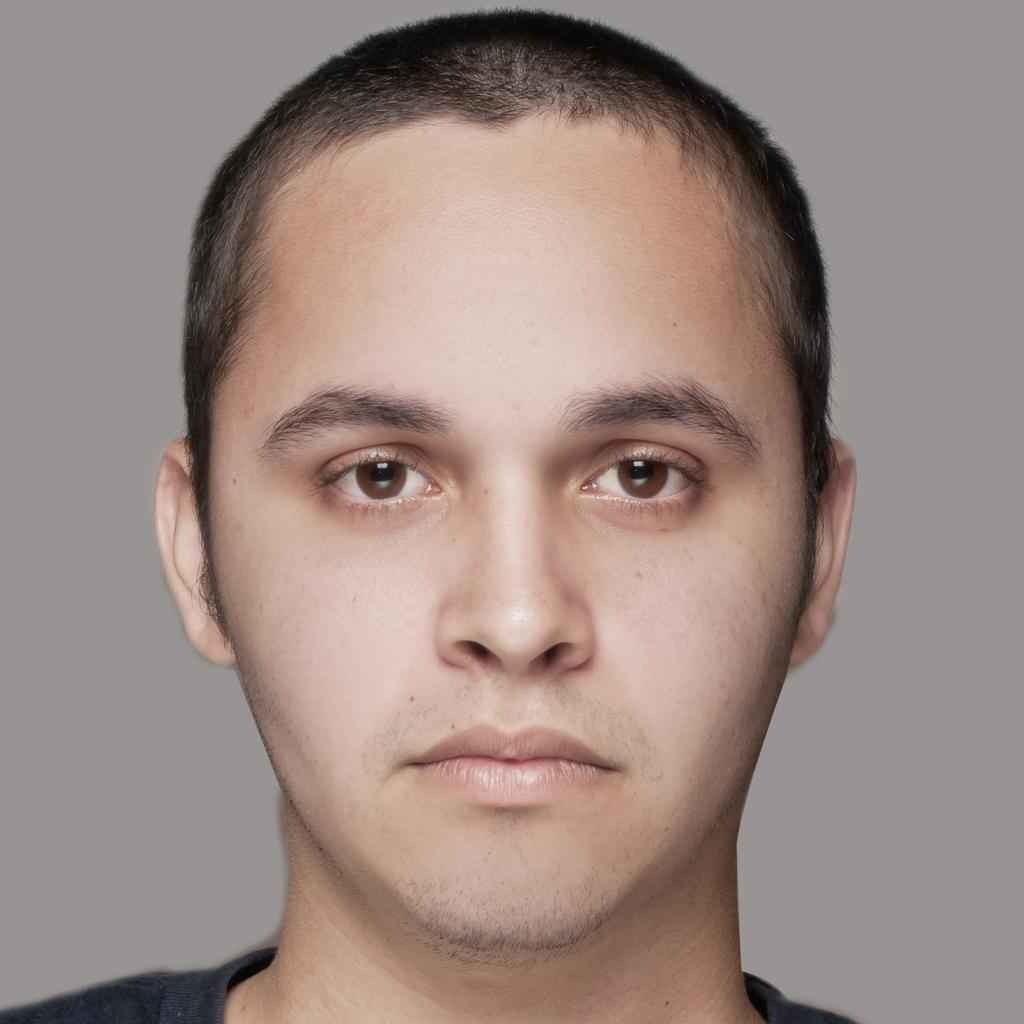}}} &
\includegraphics[width=0.19\linewidth]{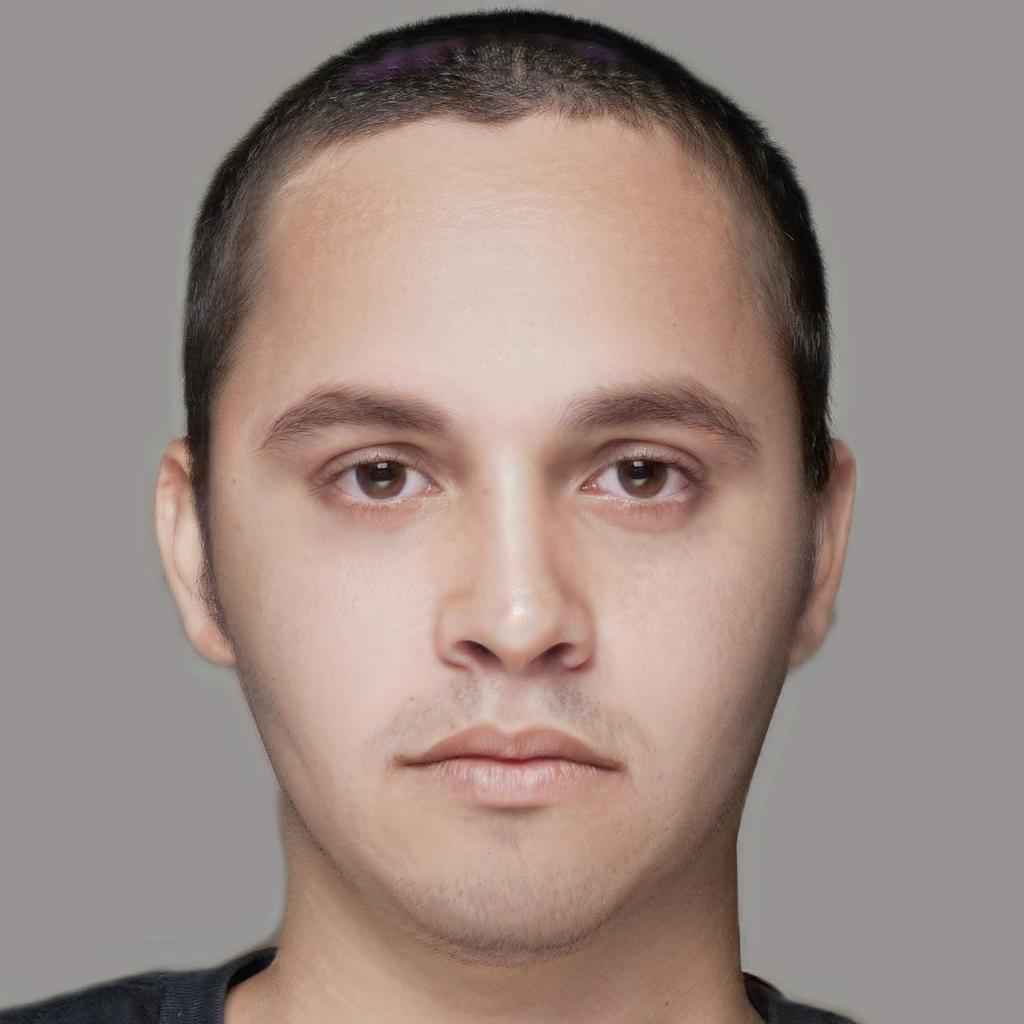} & 
\includegraphics[width=0.19\linewidth]{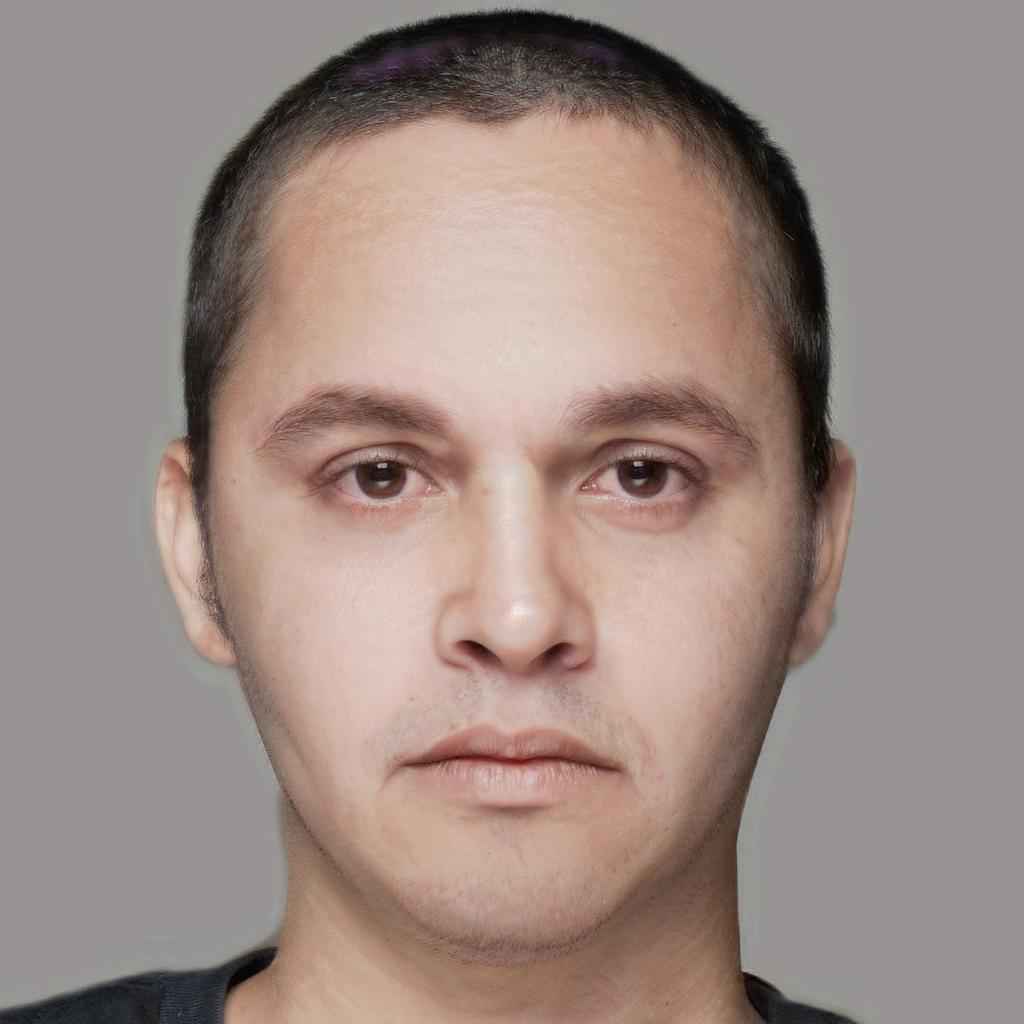} &
\includegraphics[width=0.19\linewidth]{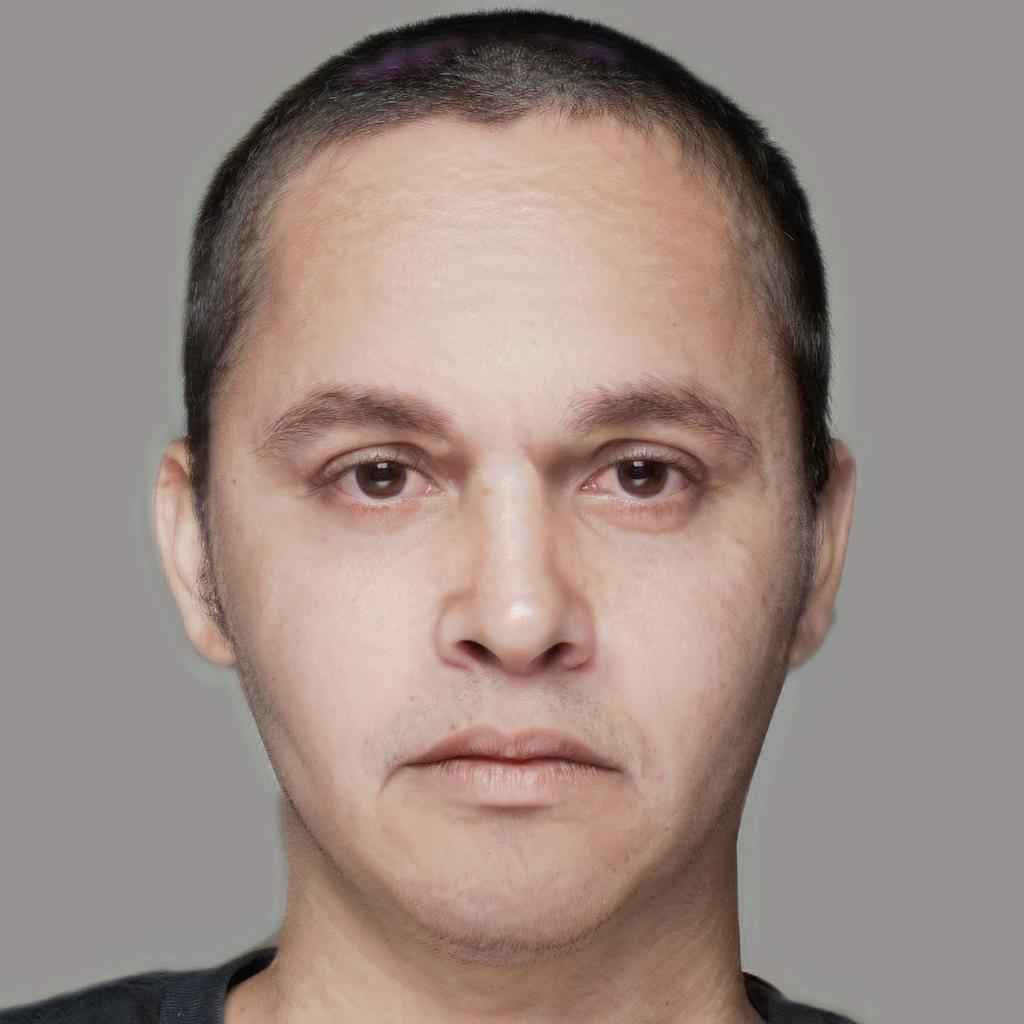} & 
\includegraphics[width=0.19\linewidth]{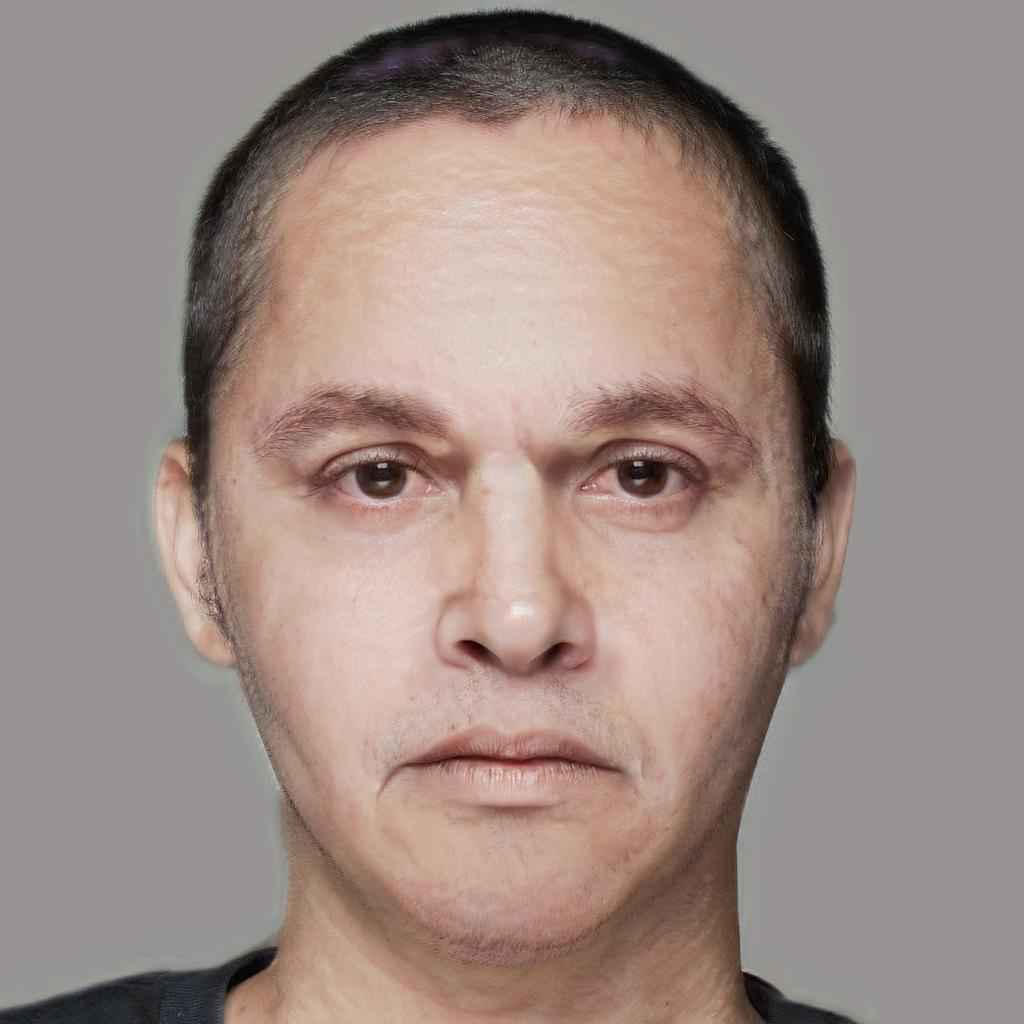} 
\\
\includegraphics[width=0.19\linewidth]{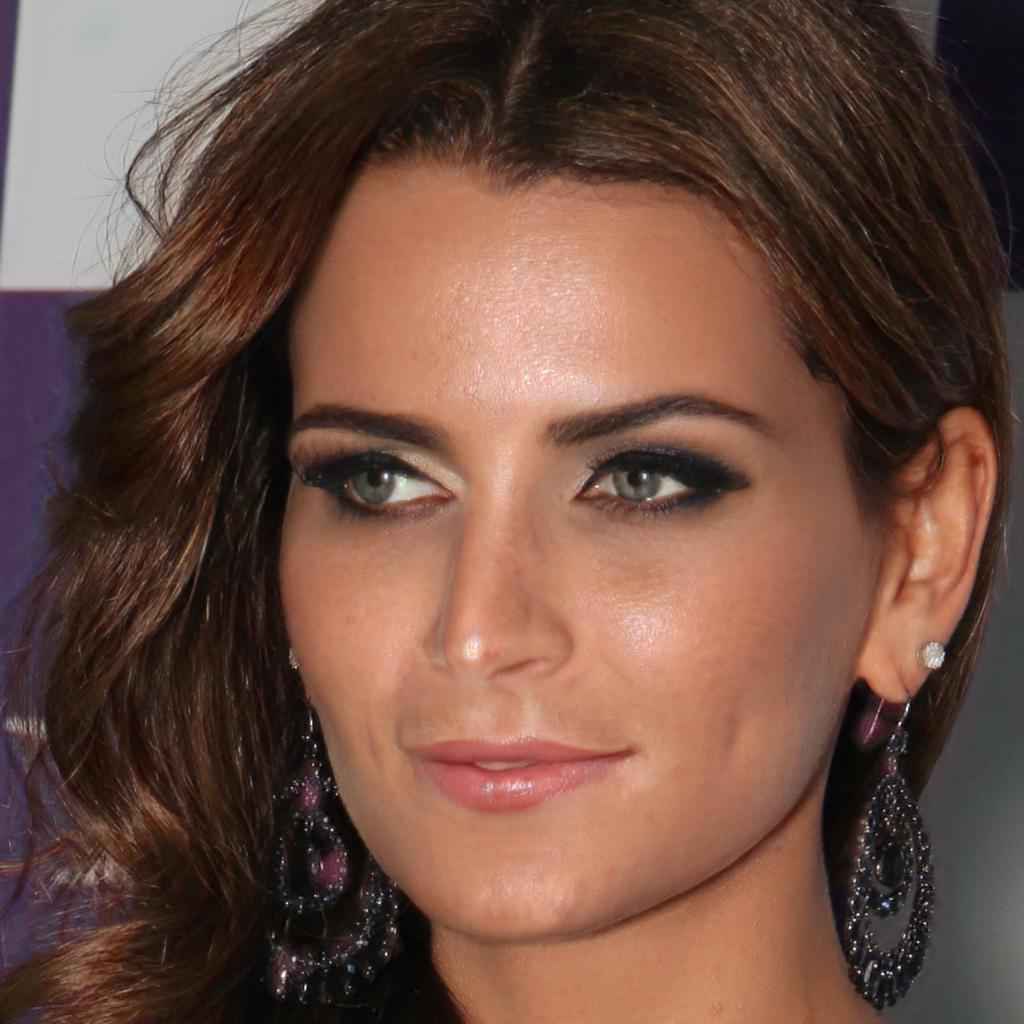} & 
{\color{yellow}%
\setlength{\fboxsep}{0pt}%
\setlength{\fboxrule}{2pt}%
\fbox{\includegraphics[width=0.19\linewidth]{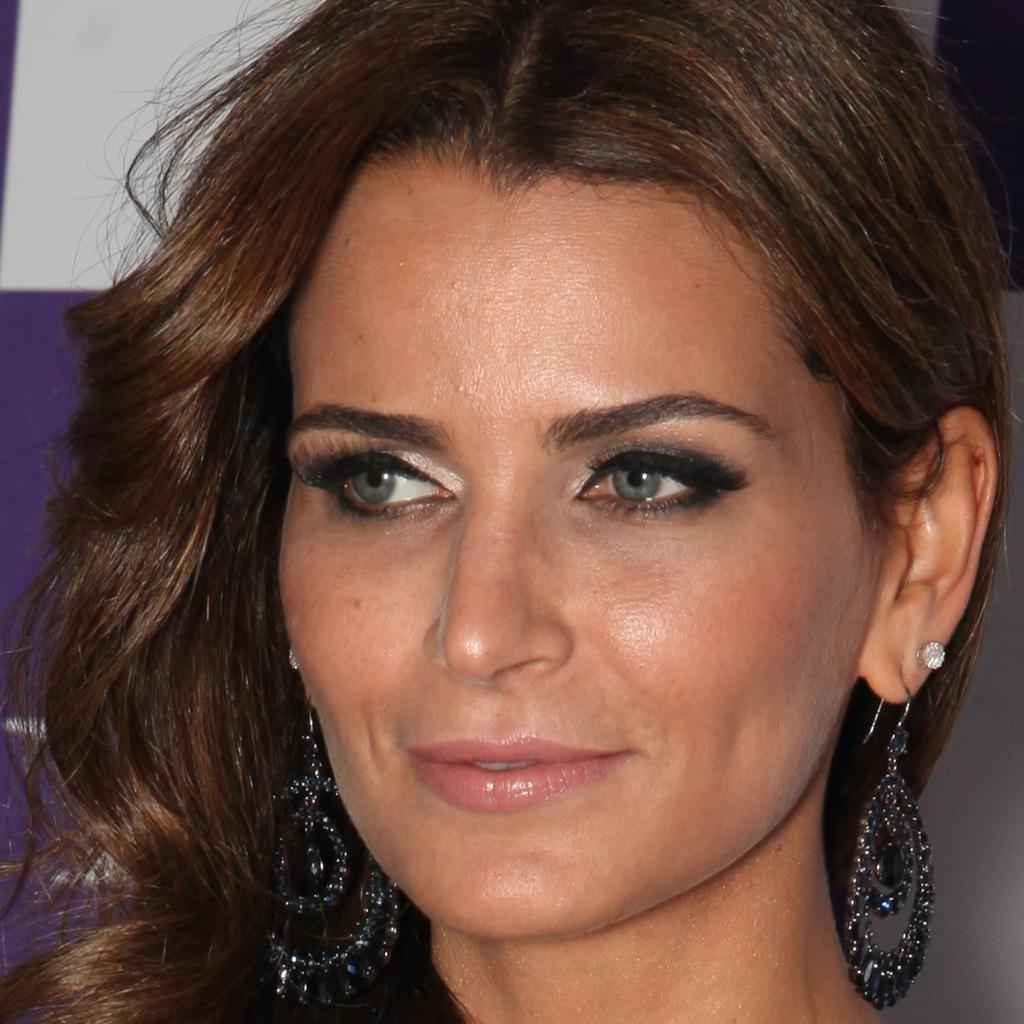}}} & 
\includegraphics[width=0.19\linewidth]{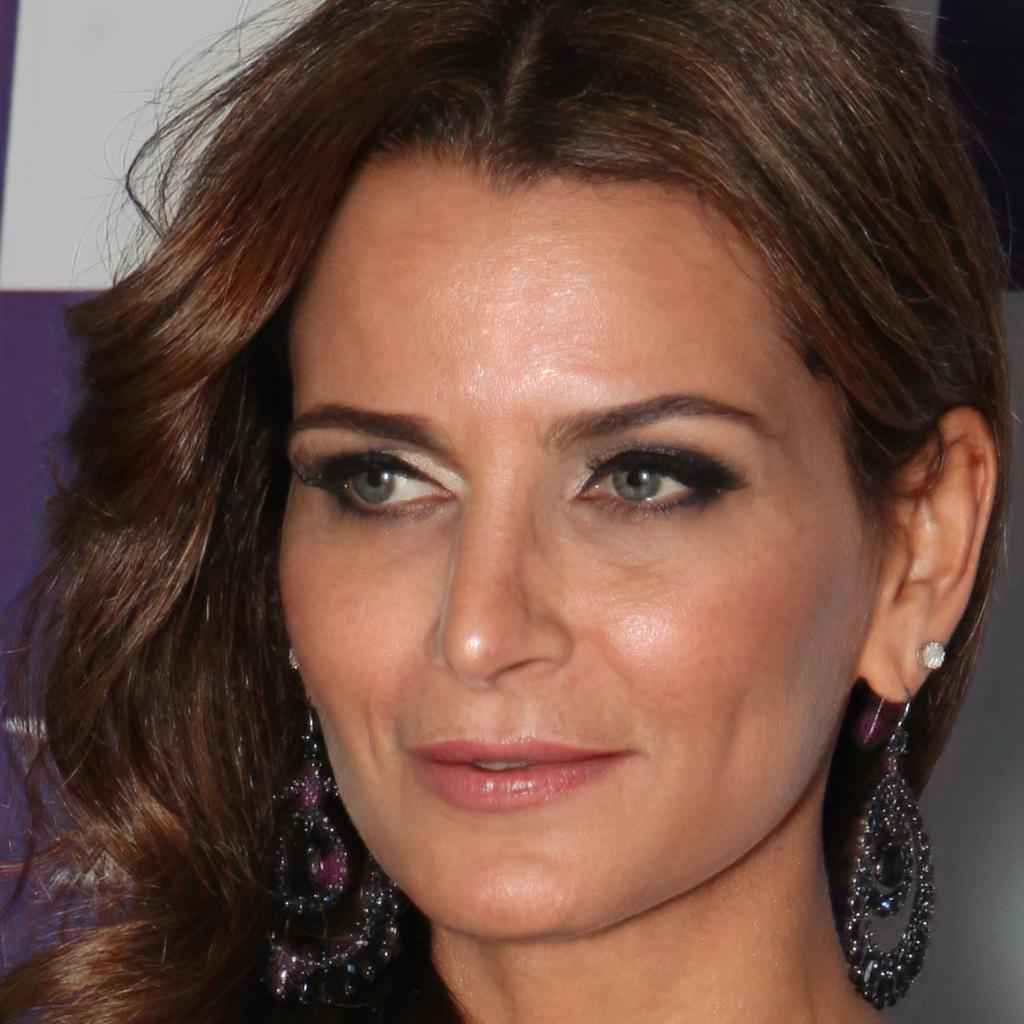} &
\includegraphics[width=0.19\linewidth]{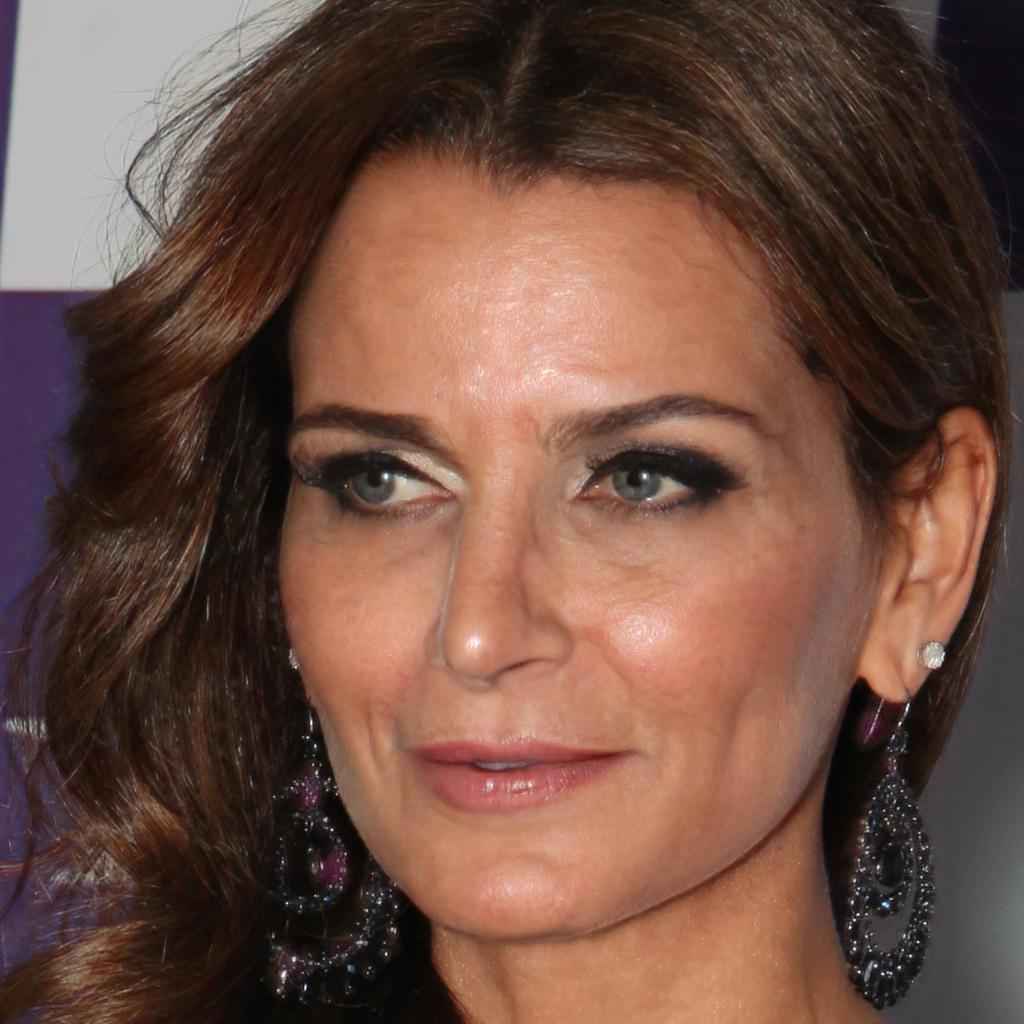} & 
\includegraphics[width=0.19\linewidth]{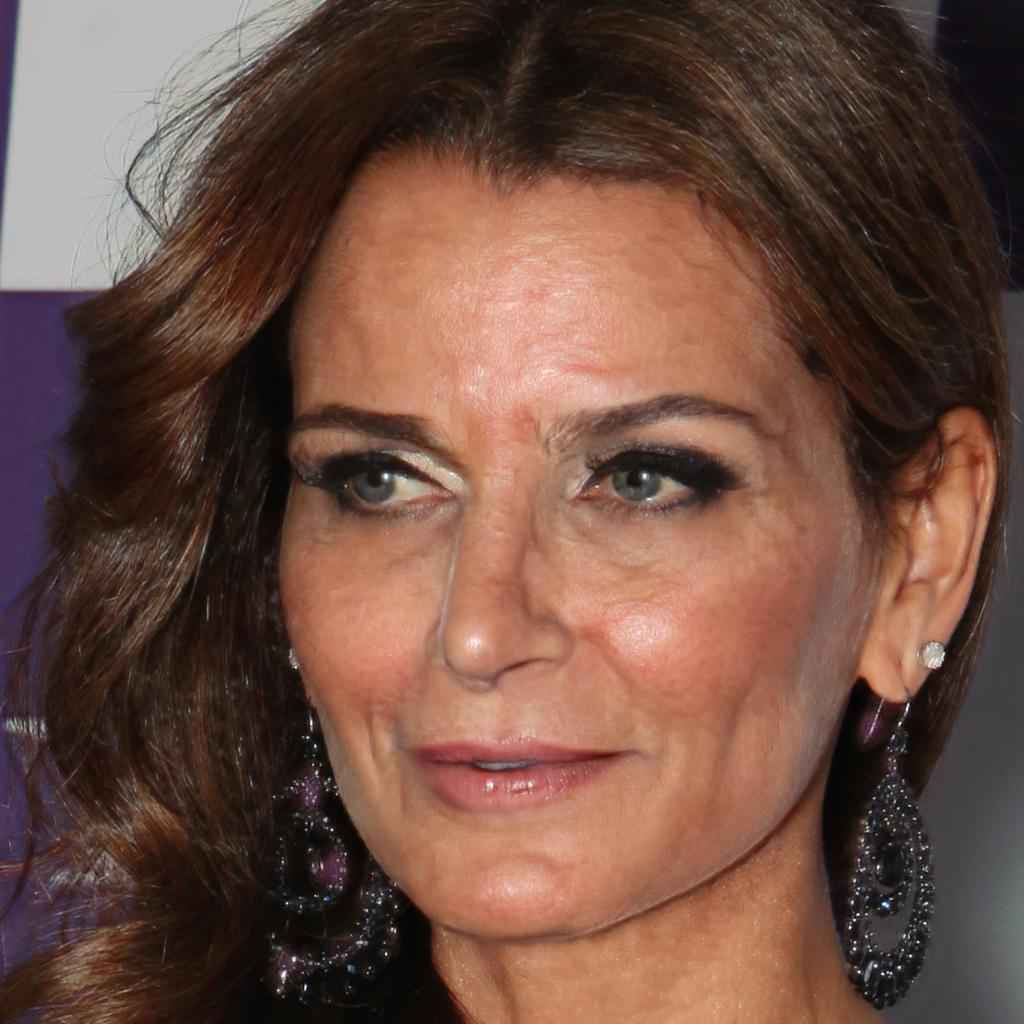} 
\\
\includegraphics[width=0.19\linewidth]{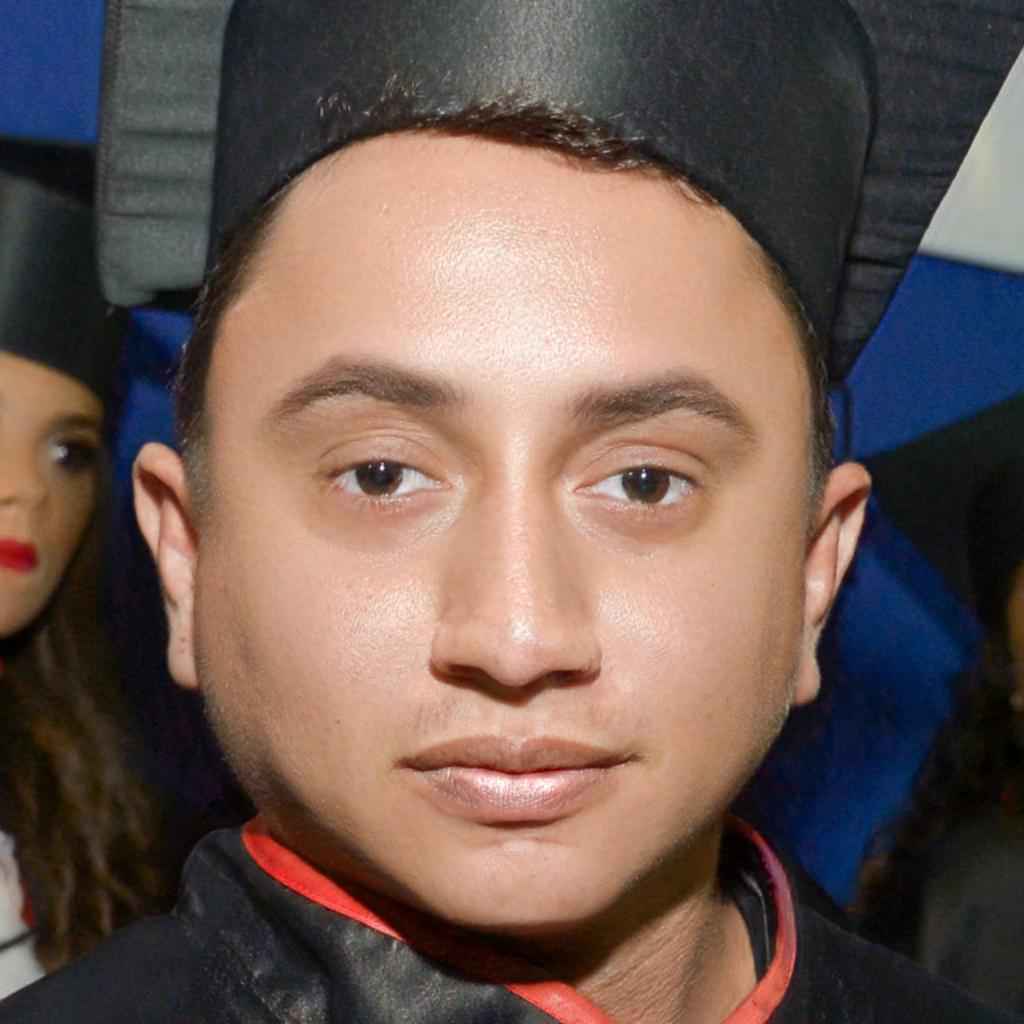} & 
\includegraphics[width=0.19\linewidth]{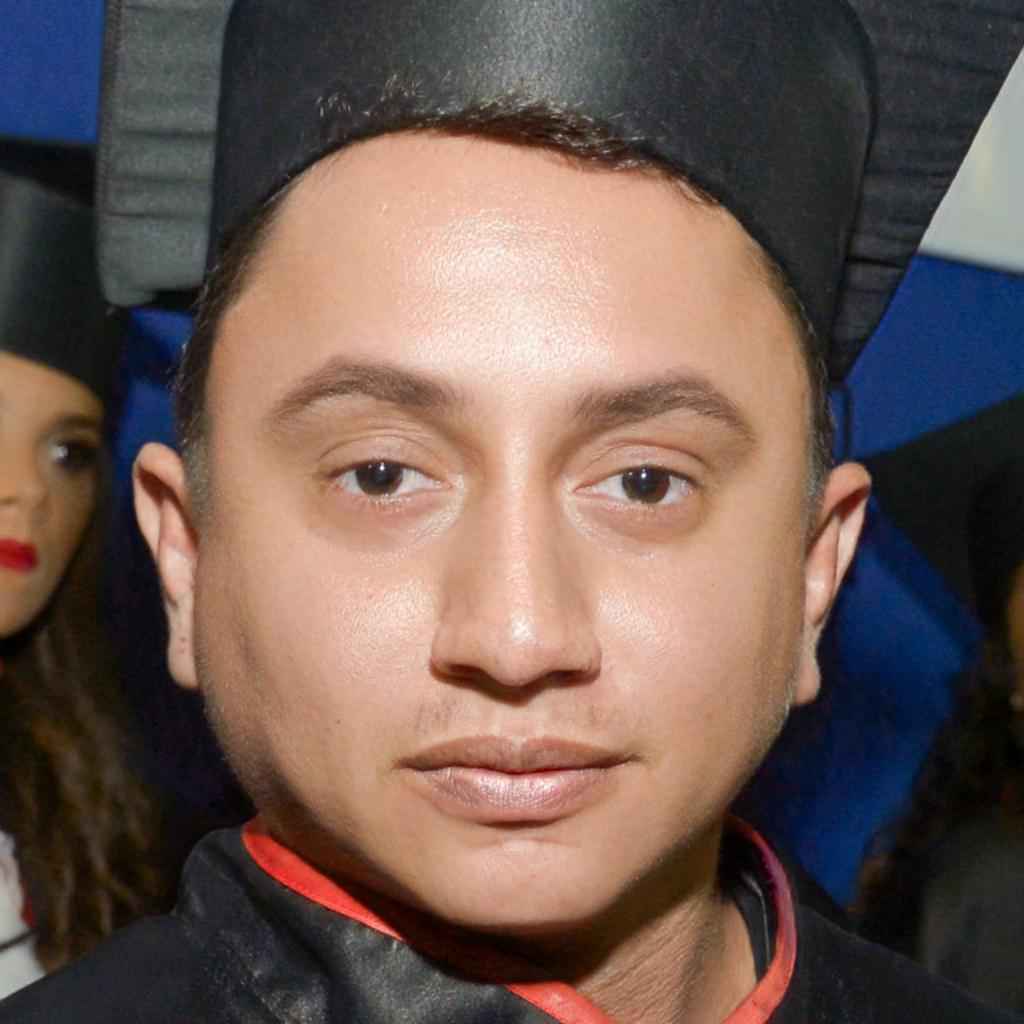} &
{\color{yellow}%
\setlength{\fboxsep}{0pt}%
\setlength{\fboxrule}{2pt}%
\fbox{\includegraphics[width=0.19\linewidth]{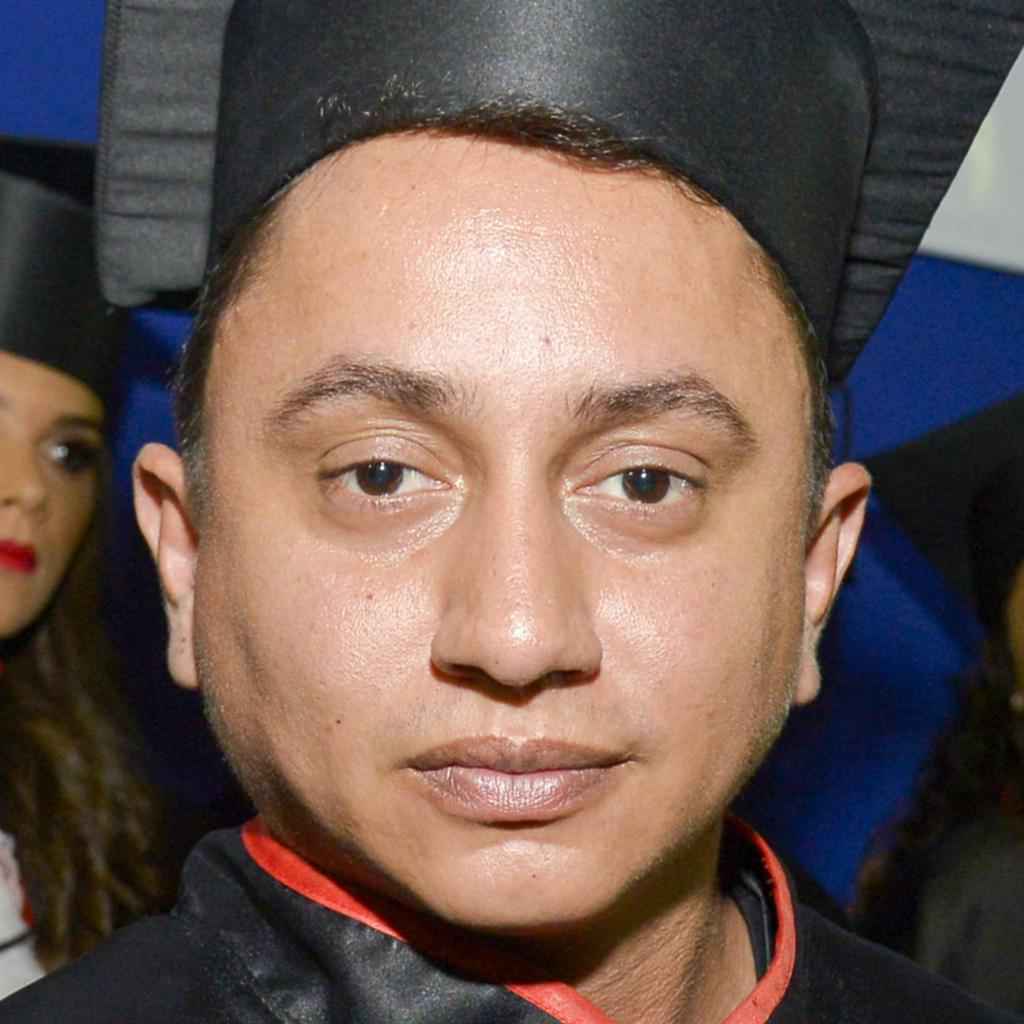}}} & 
\includegraphics[width=0.19\linewidth]{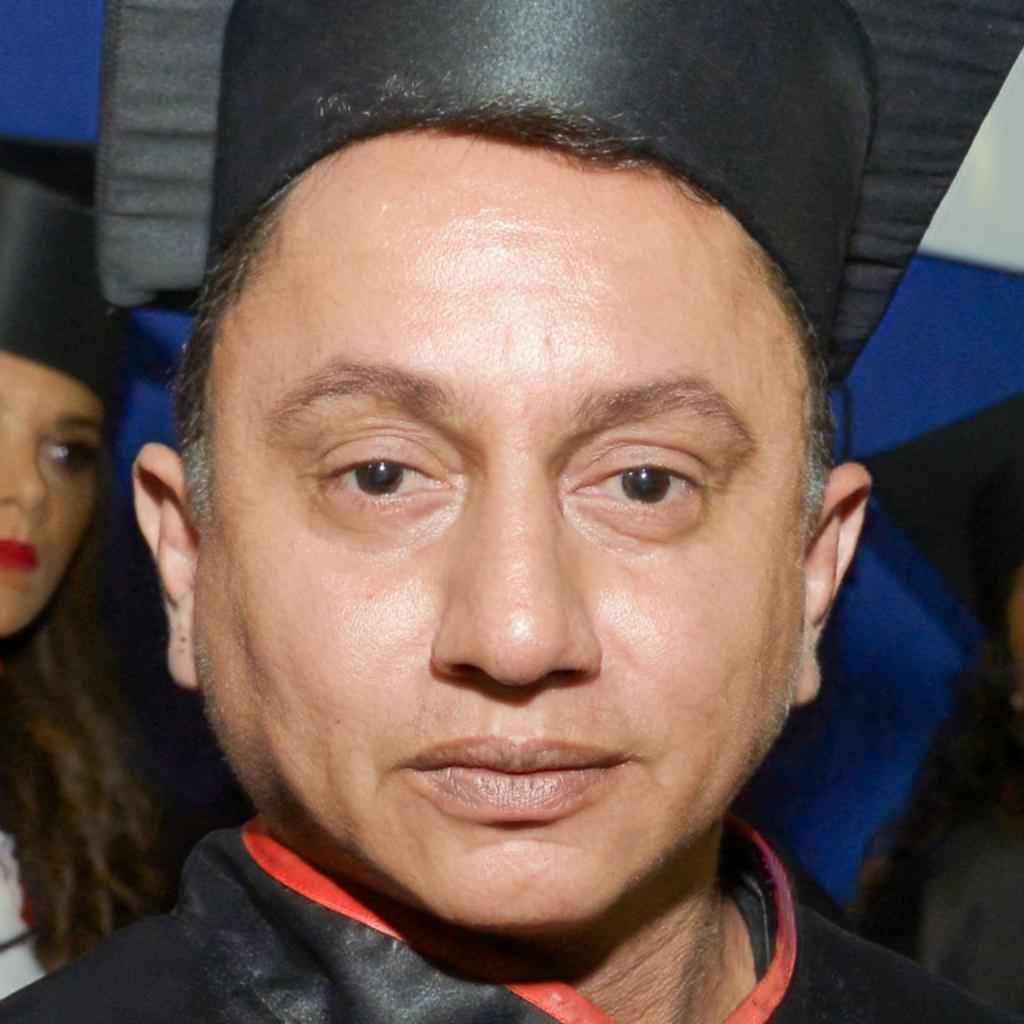} & 
\includegraphics[width=0.19\linewidth]{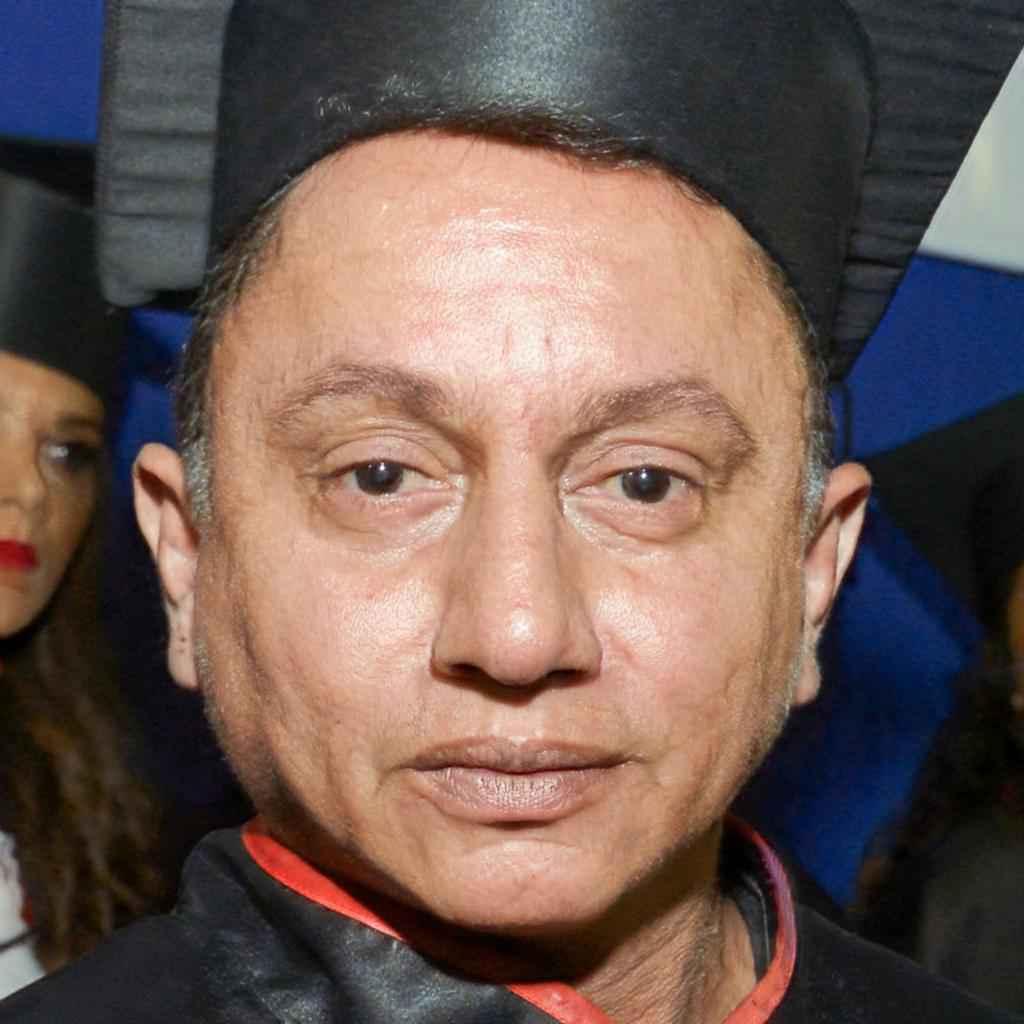} 
\\
\includegraphics[width=0.19\linewidth]{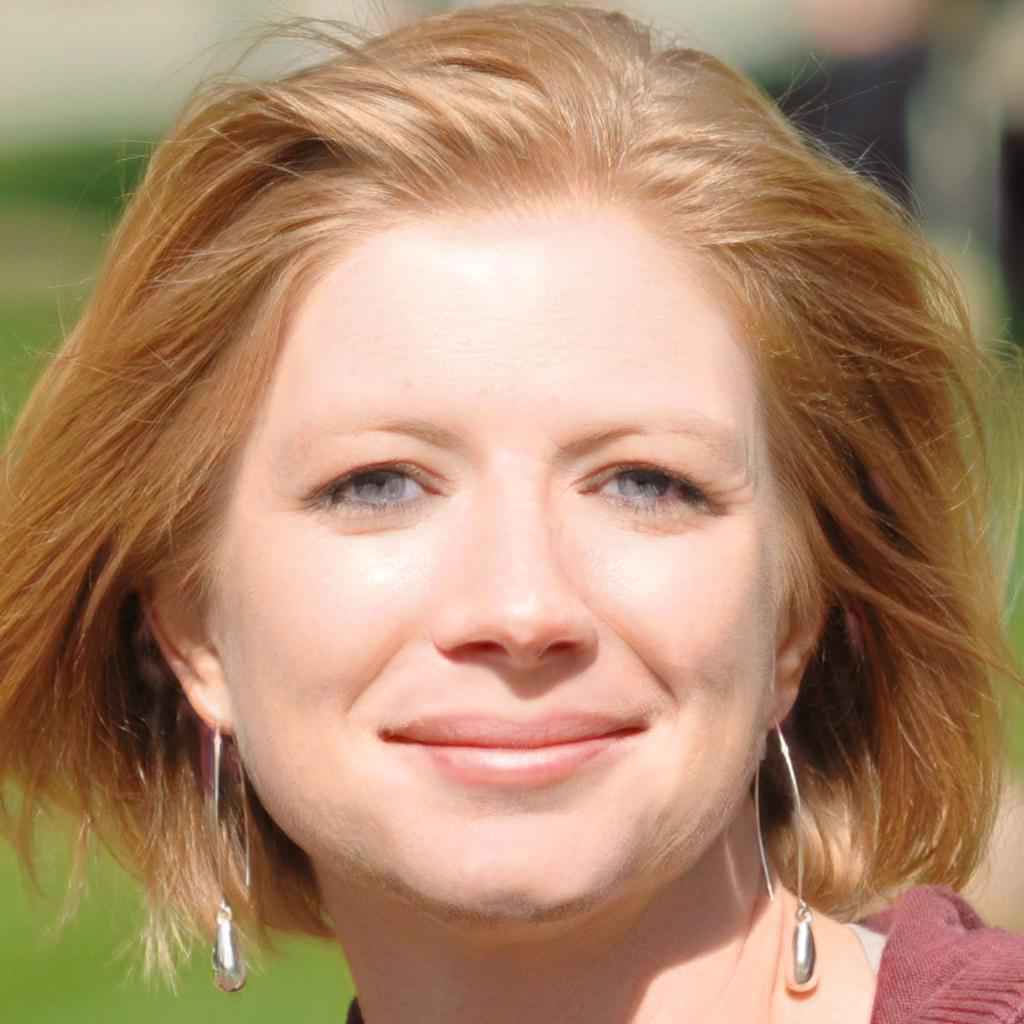} & 
\includegraphics[width=0.19\linewidth]{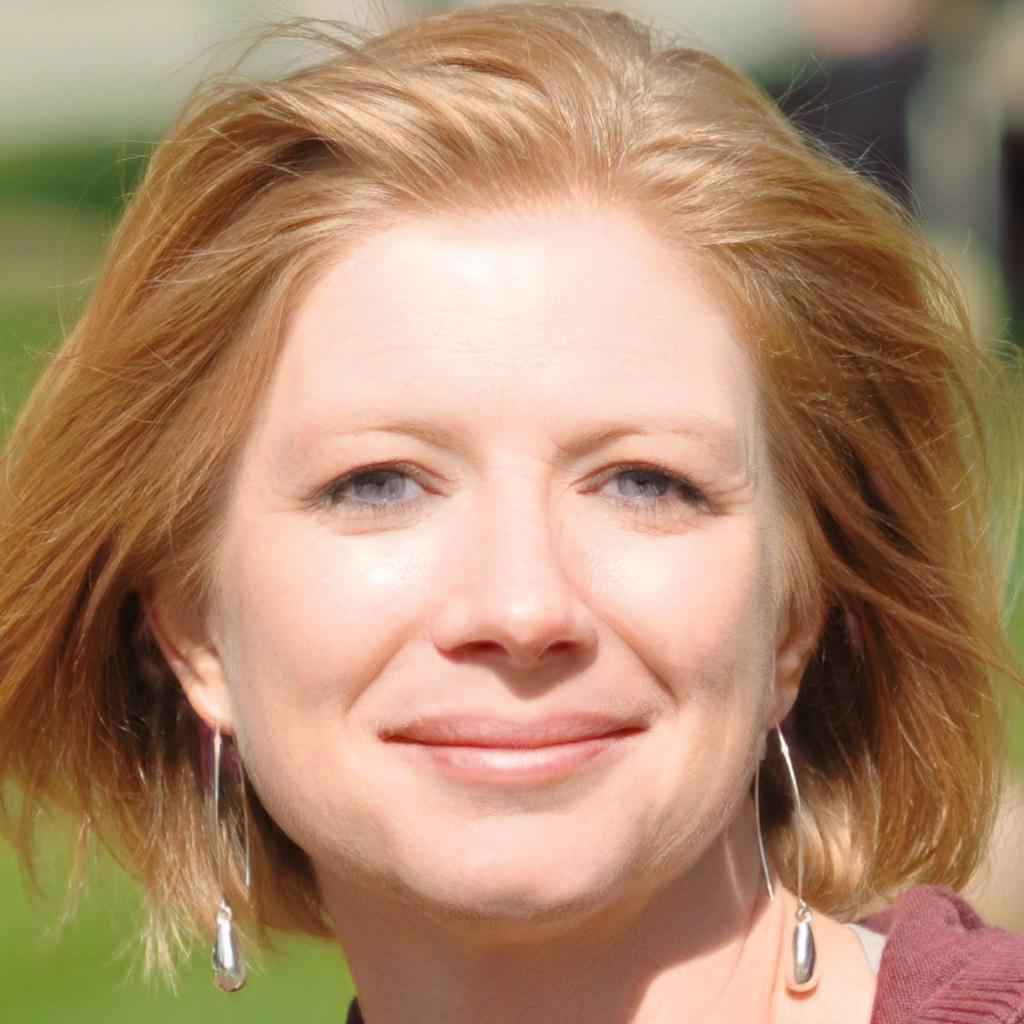} &
\includegraphics[width=0.19\linewidth]{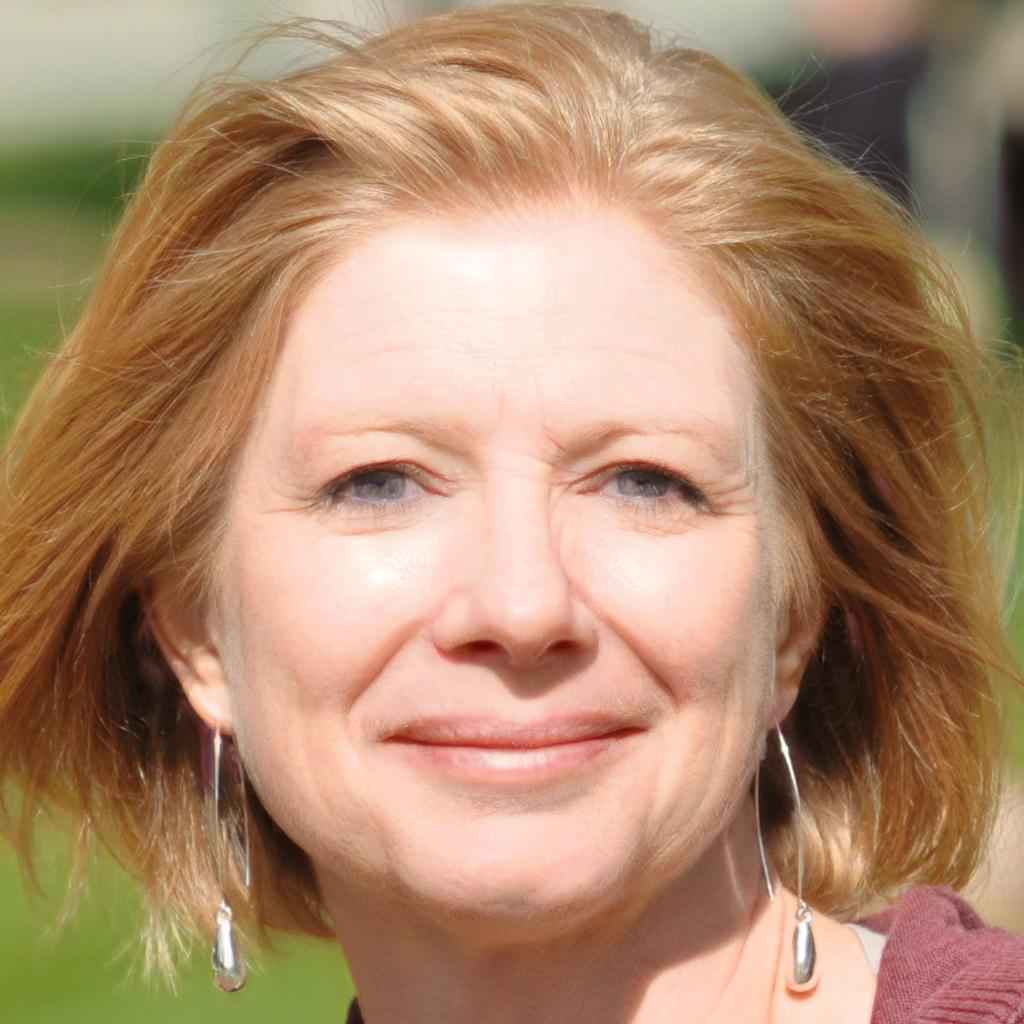} &
{\color{yellow}%
\setlength{\fboxsep}{0pt}%
\setlength{\fboxrule}{2pt}%
\fbox{\includegraphics[width=0.19\linewidth]{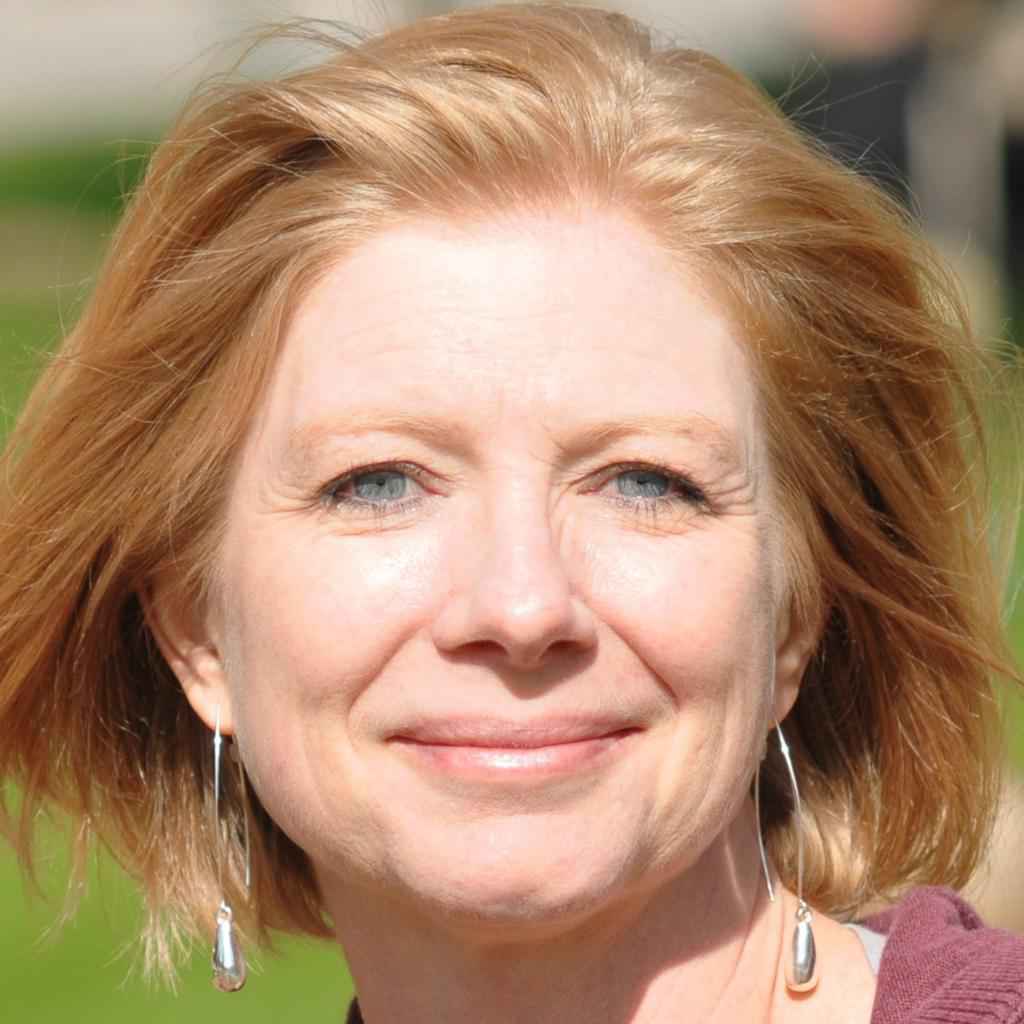}}} &  
\includegraphics[width=0.19\linewidth]{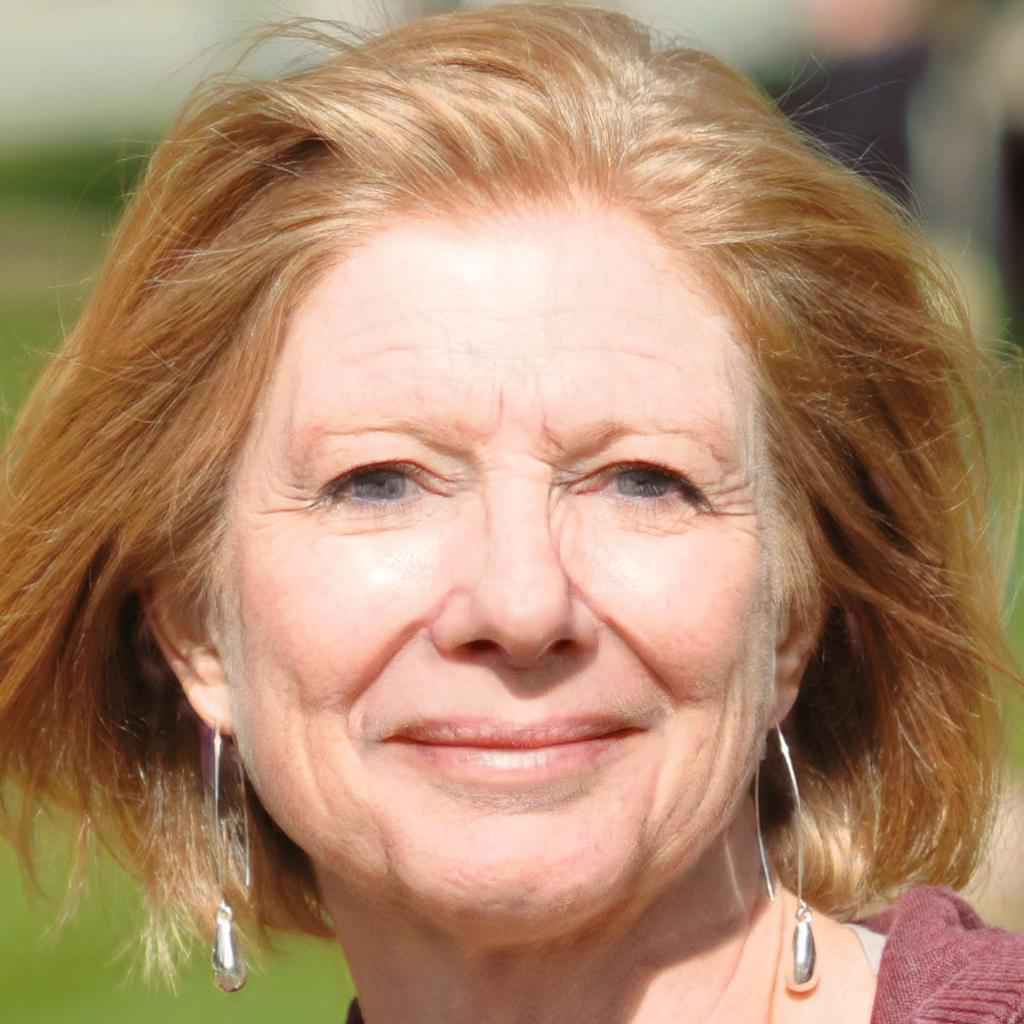} 
\\
\includegraphics[width=0.19\linewidth]{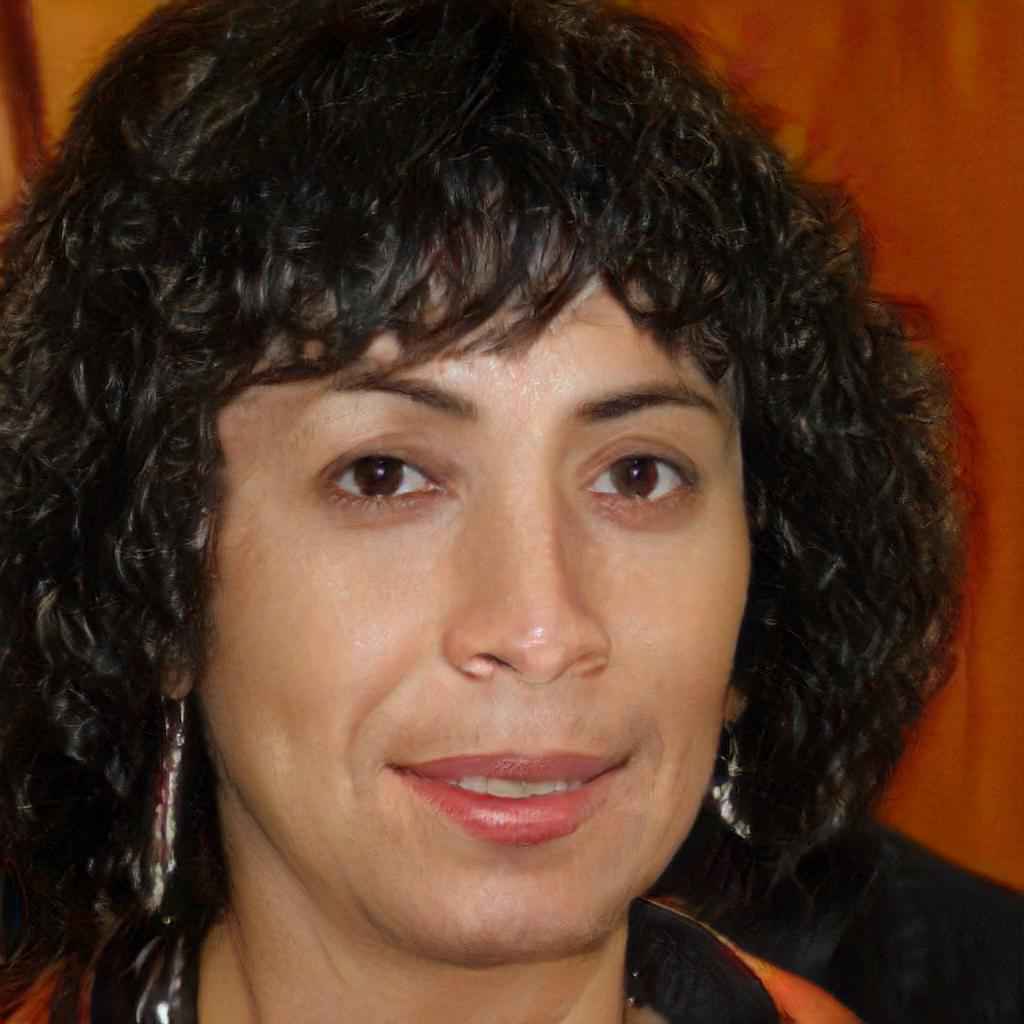} & 
\includegraphics[width=0.19\linewidth]{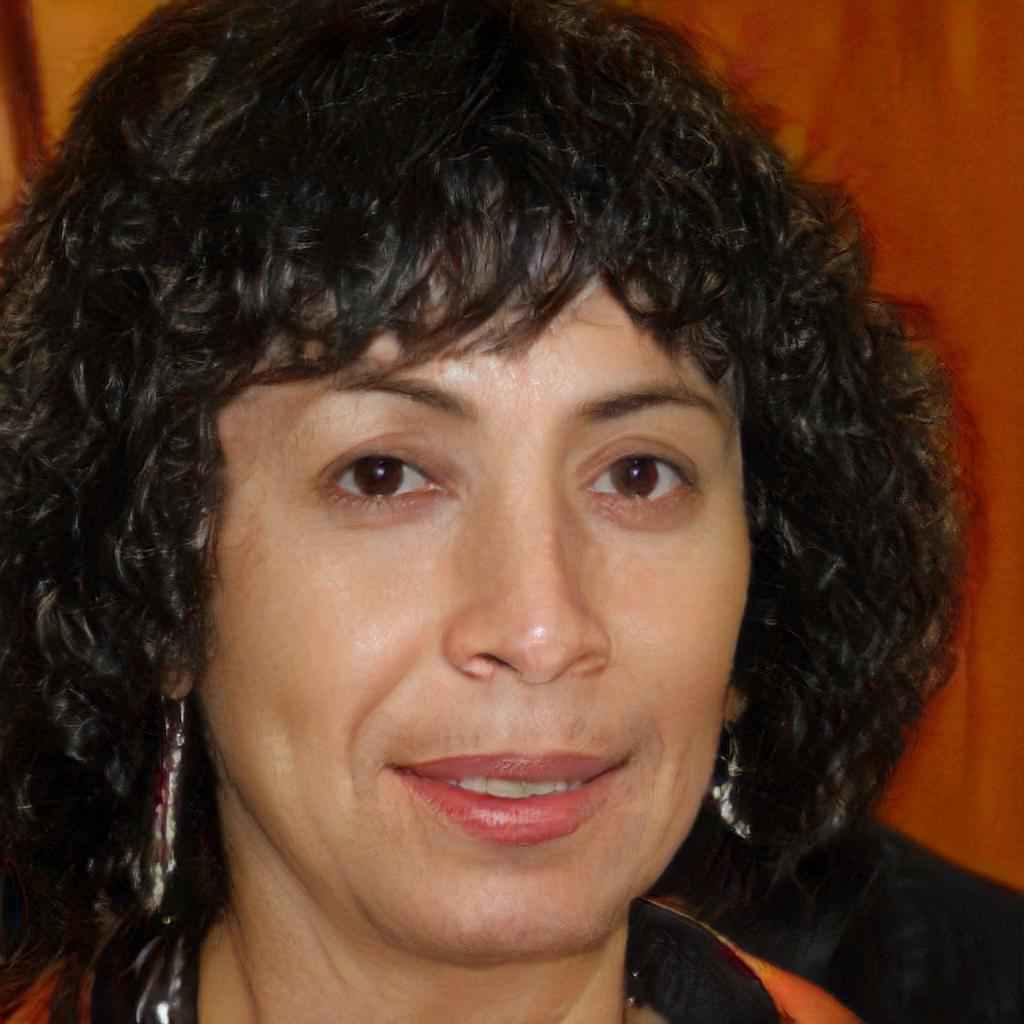} &
\includegraphics[width=0.19\linewidth]{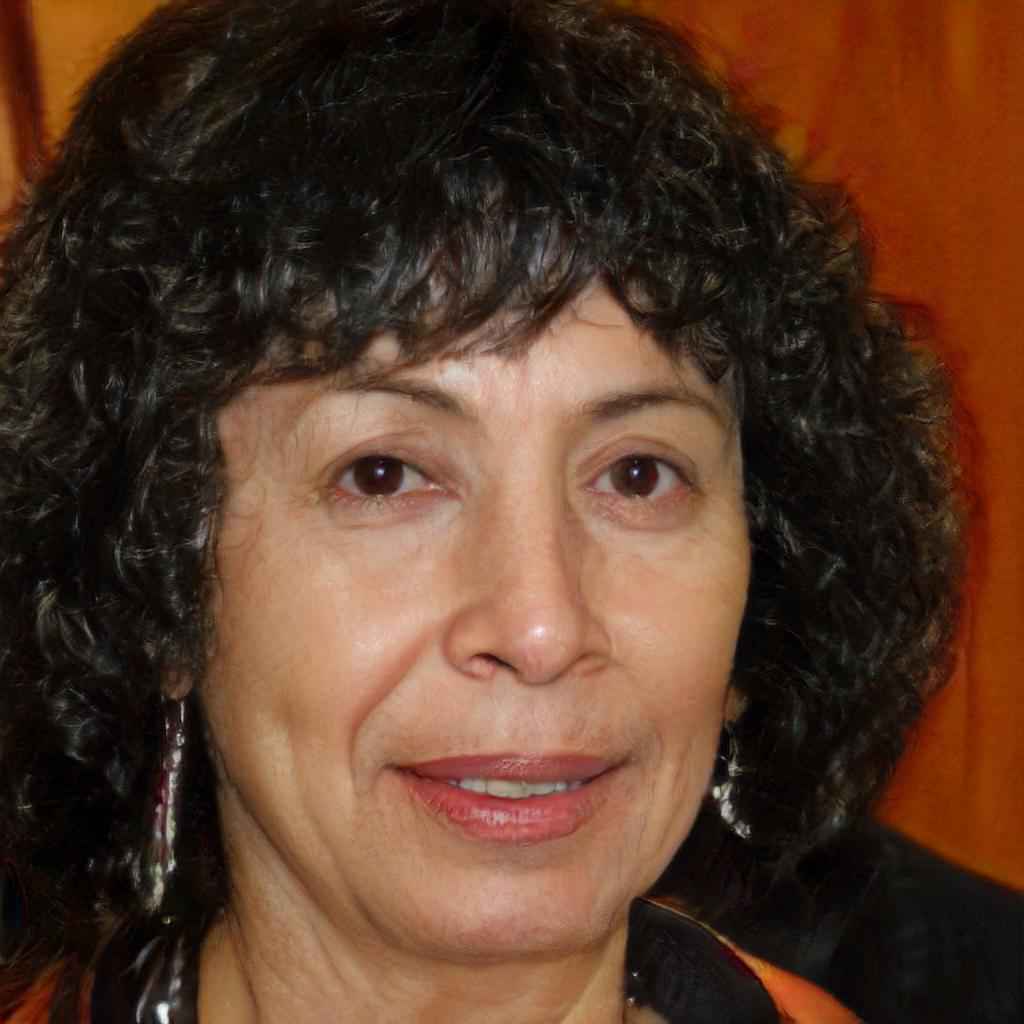} &
\includegraphics[width=0.19\linewidth]{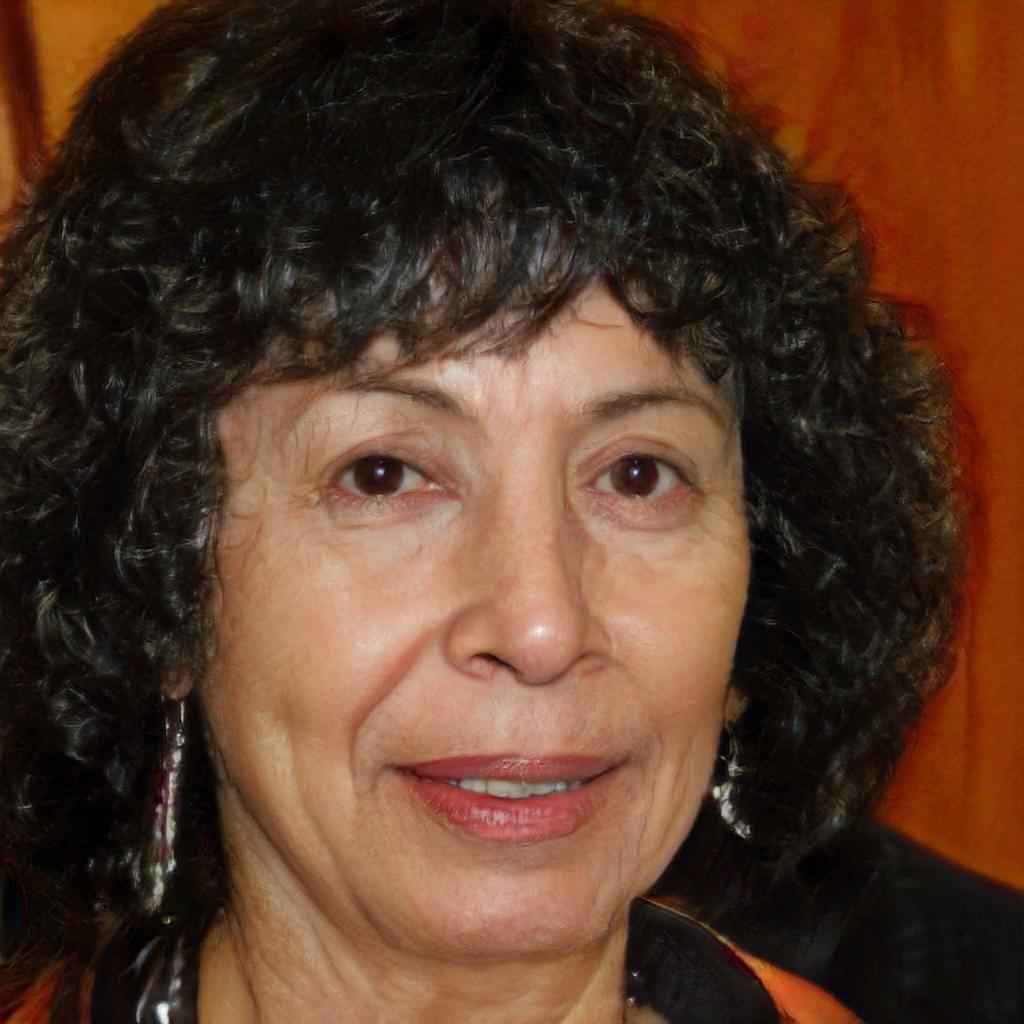} & 
{\color{yellow}%
\setlength{\fboxsep}{0pt}%
\setlength{\fboxrule}{2pt}%
\fbox{\includegraphics[width=0.19\linewidth]{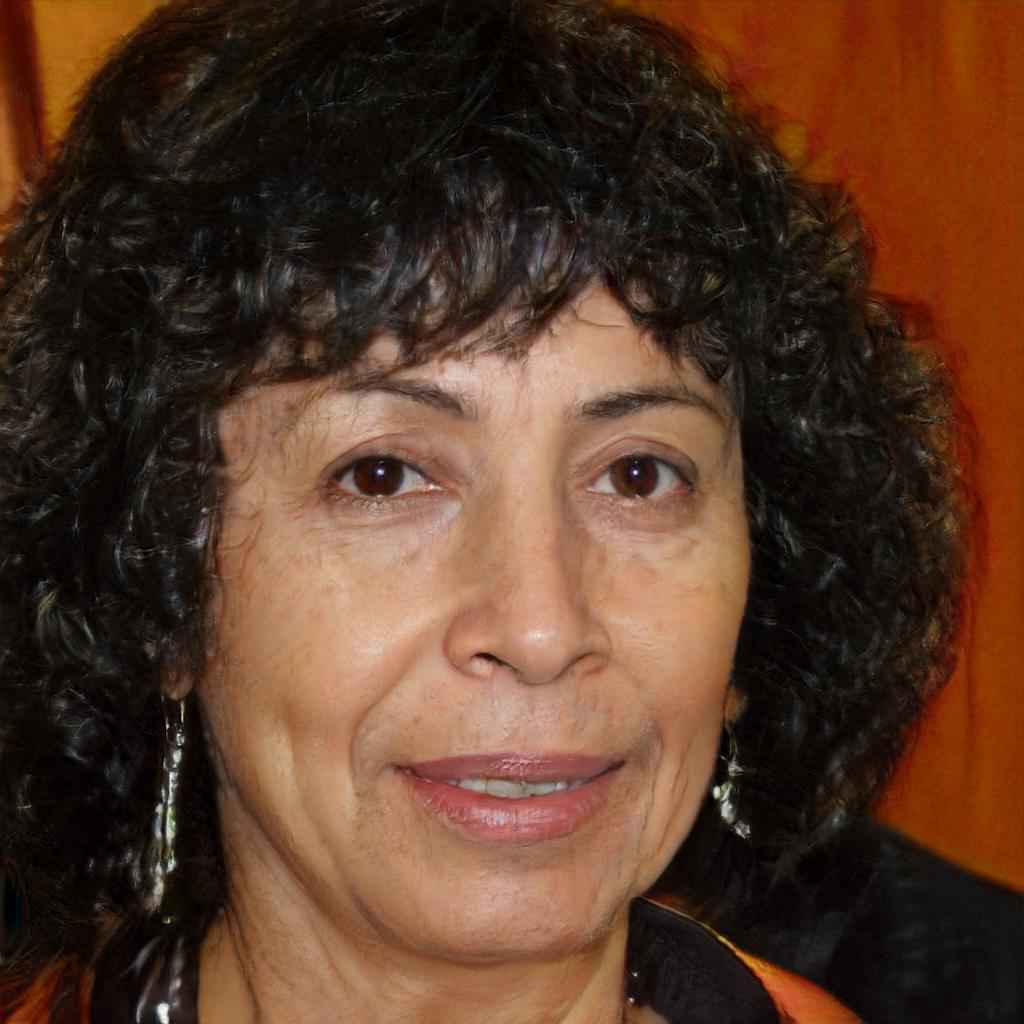}}} 
\end{tabular}
\end{center}
\caption{\textbf{Age transformation on $\bf 1024 \times 1024$ images}. On each row, the yellow frame indicates the original image. Each column corresponds to a target age of: $25$, $35$, $45$, $55$, $65$. Our approach yields visually satisfying results without introducing significant artifacts. Only age relevant features are modified, while the identity, haircut, emotion and background are perfectly preserved. 
}
\label{1024_1}
\end{figure}
\begin{figure}[ht]
\begin{center}
\setlength{\tabcolsep}{1pt}
\begin{tabular}{ccccc}
25&35&45&55&65 \\
{\color{yellow}%
\setlength{\fboxsep}{0pt}%
\setlength{\fboxrule}{2pt}%
\fbox{\includegraphics[width=0.19\linewidth]{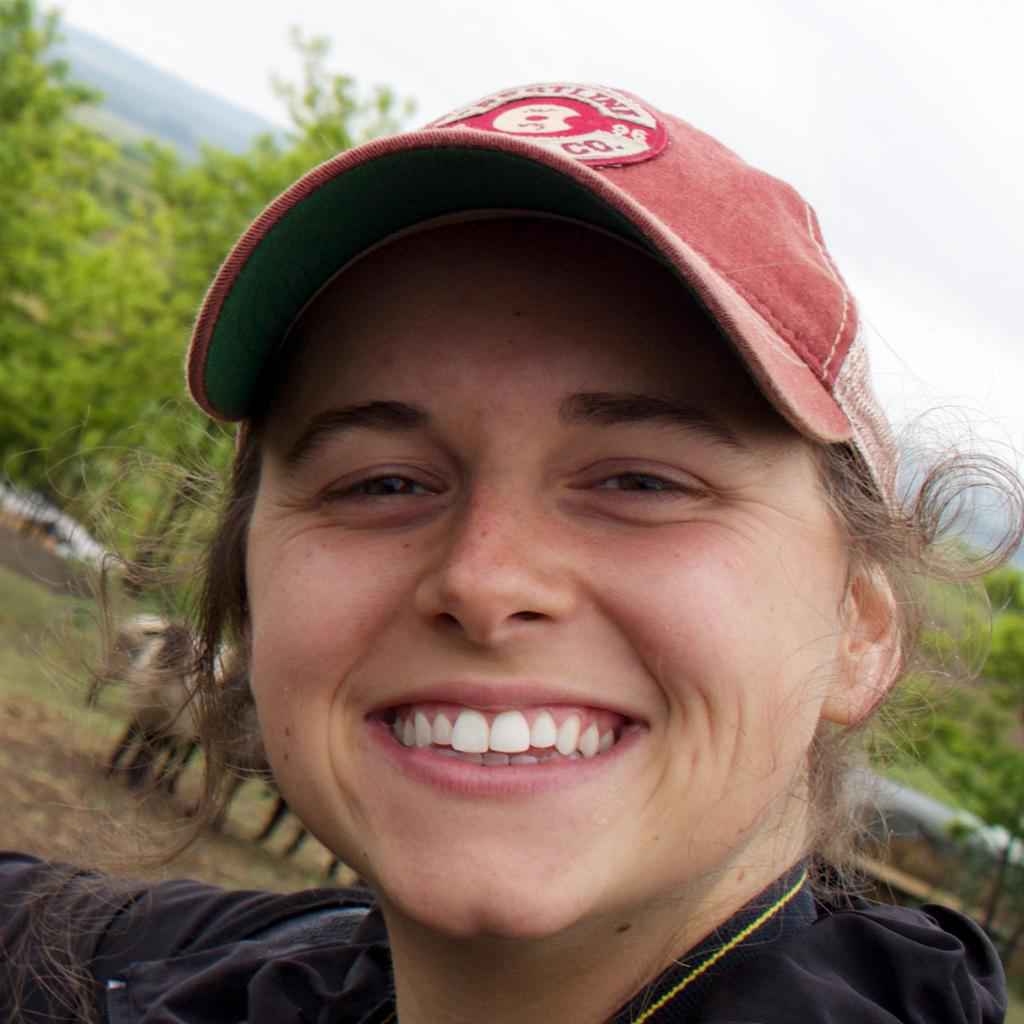}}} &
\includegraphics[width=0.19\linewidth]{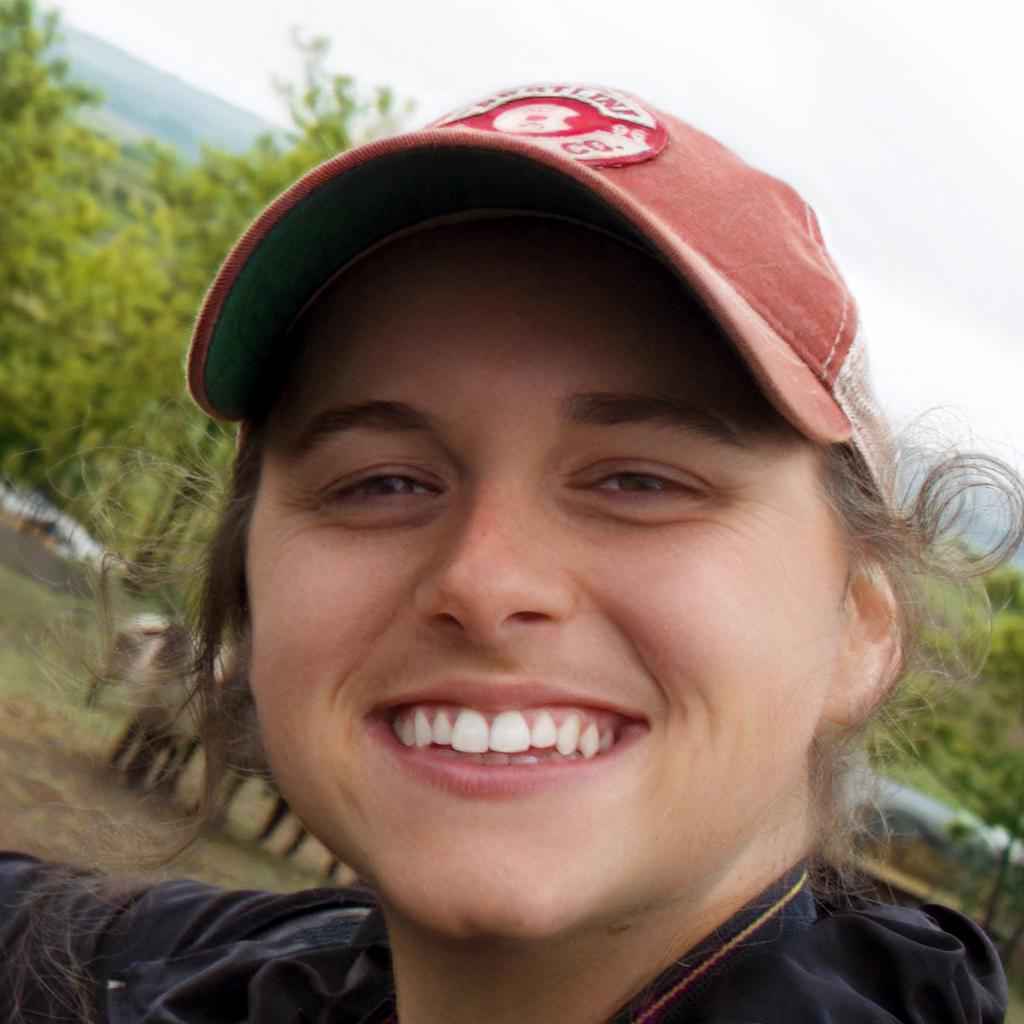} & 
\includegraphics[width=0.19\linewidth]{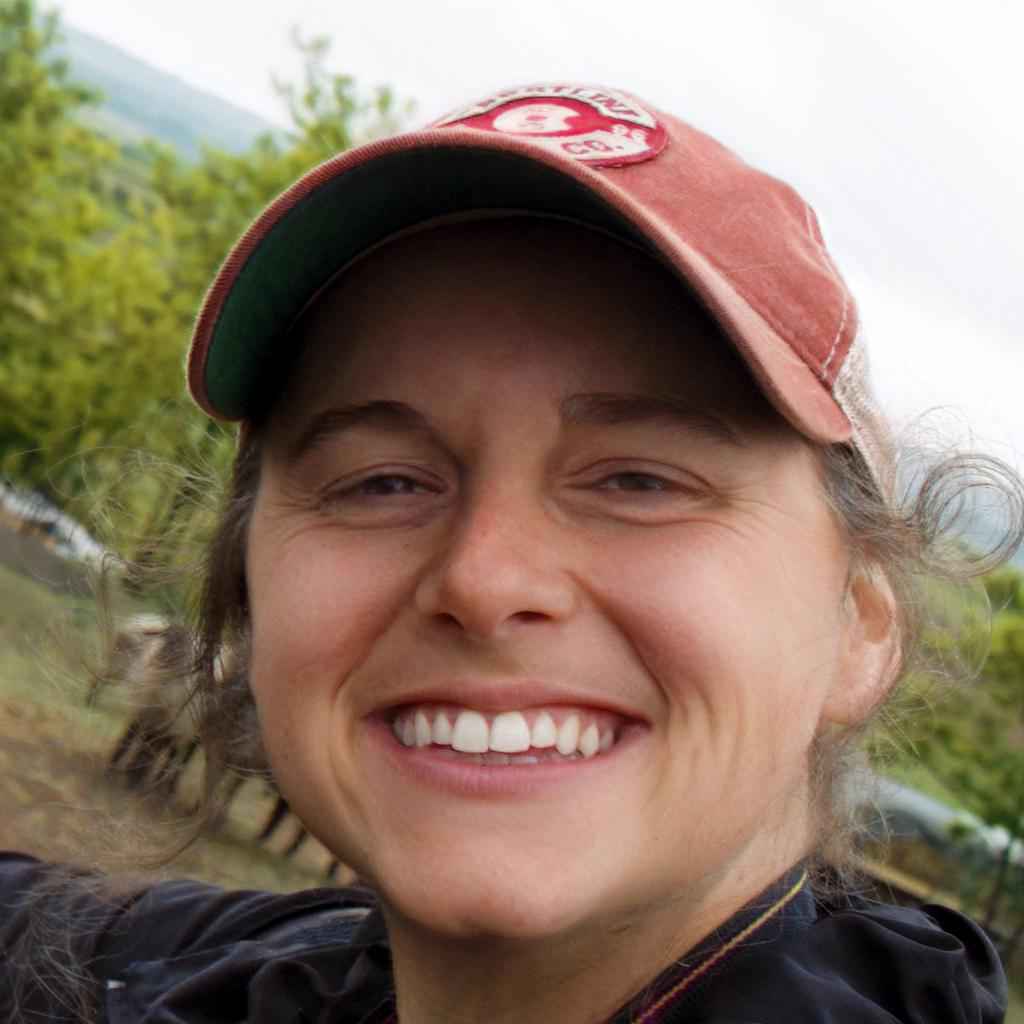} &
\includegraphics[width=0.19\linewidth]{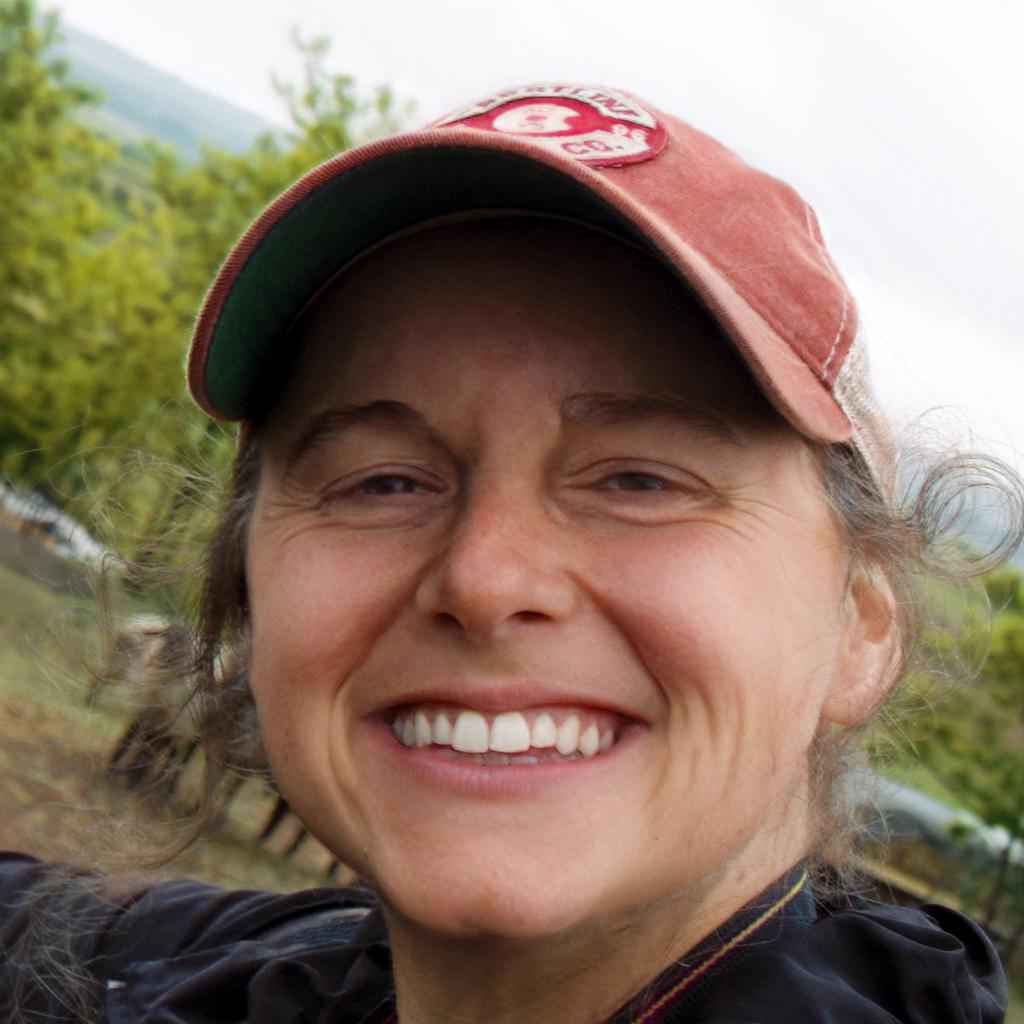} & 
\includegraphics[width=0.19\linewidth]{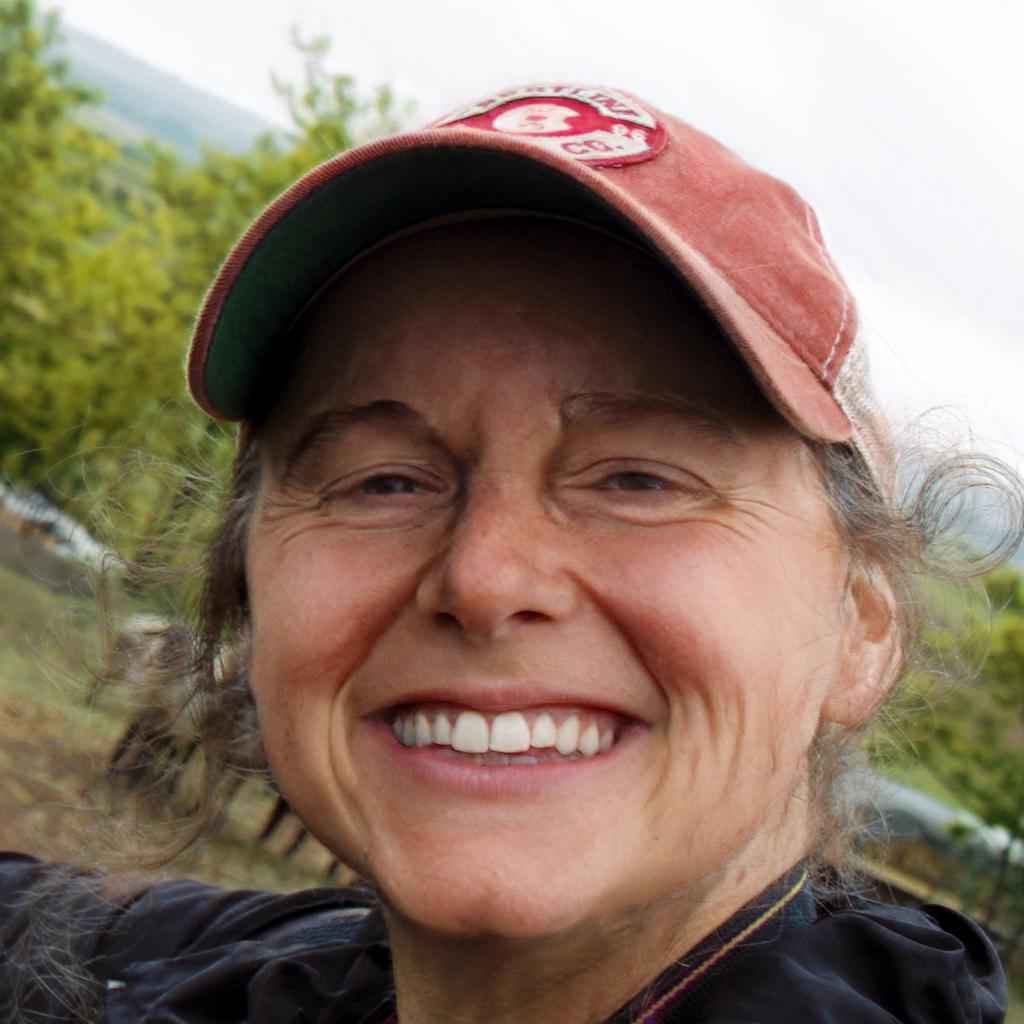} 
\\
\includegraphics[width=0.19\linewidth]{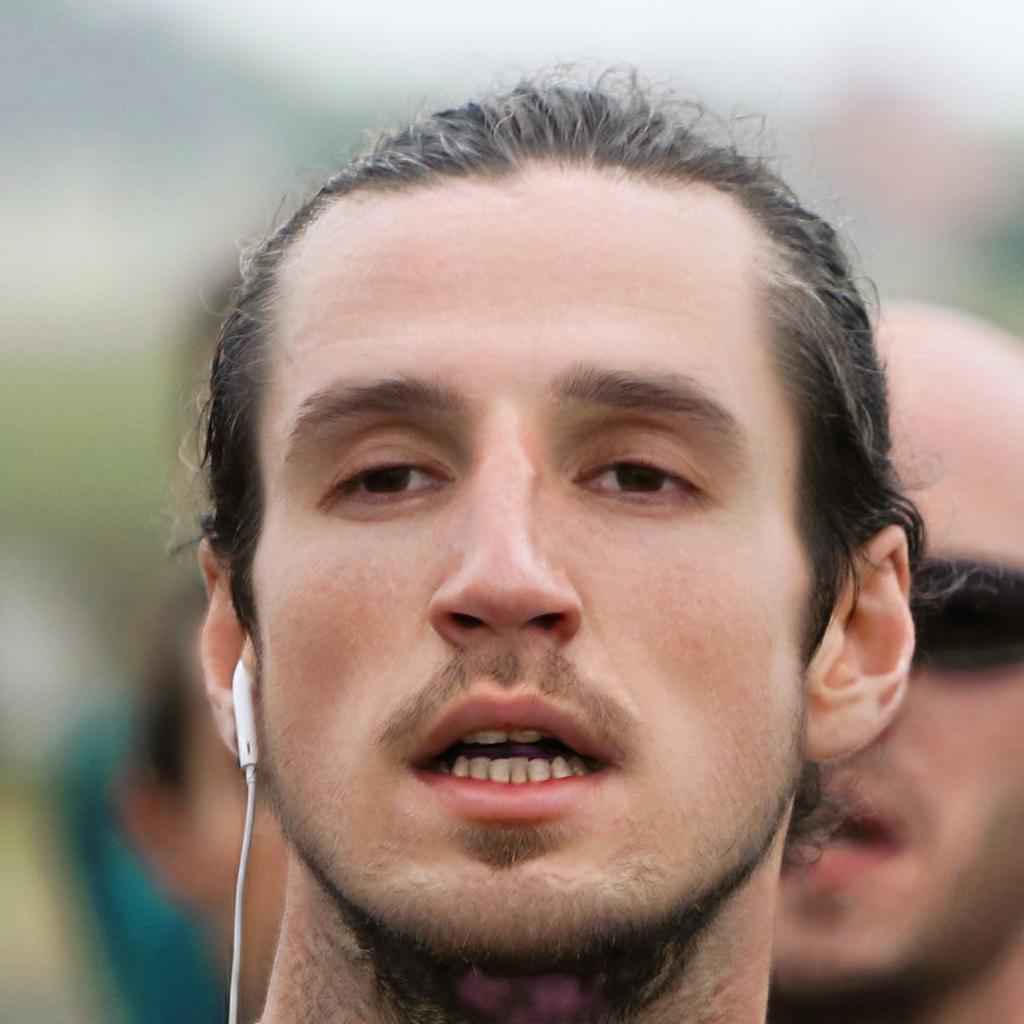} & 
{\color{yellow}%
\setlength{\fboxsep}{0pt}%
\setlength{\fboxrule}{2pt}%
\fbox{\includegraphics[width=0.19\linewidth]{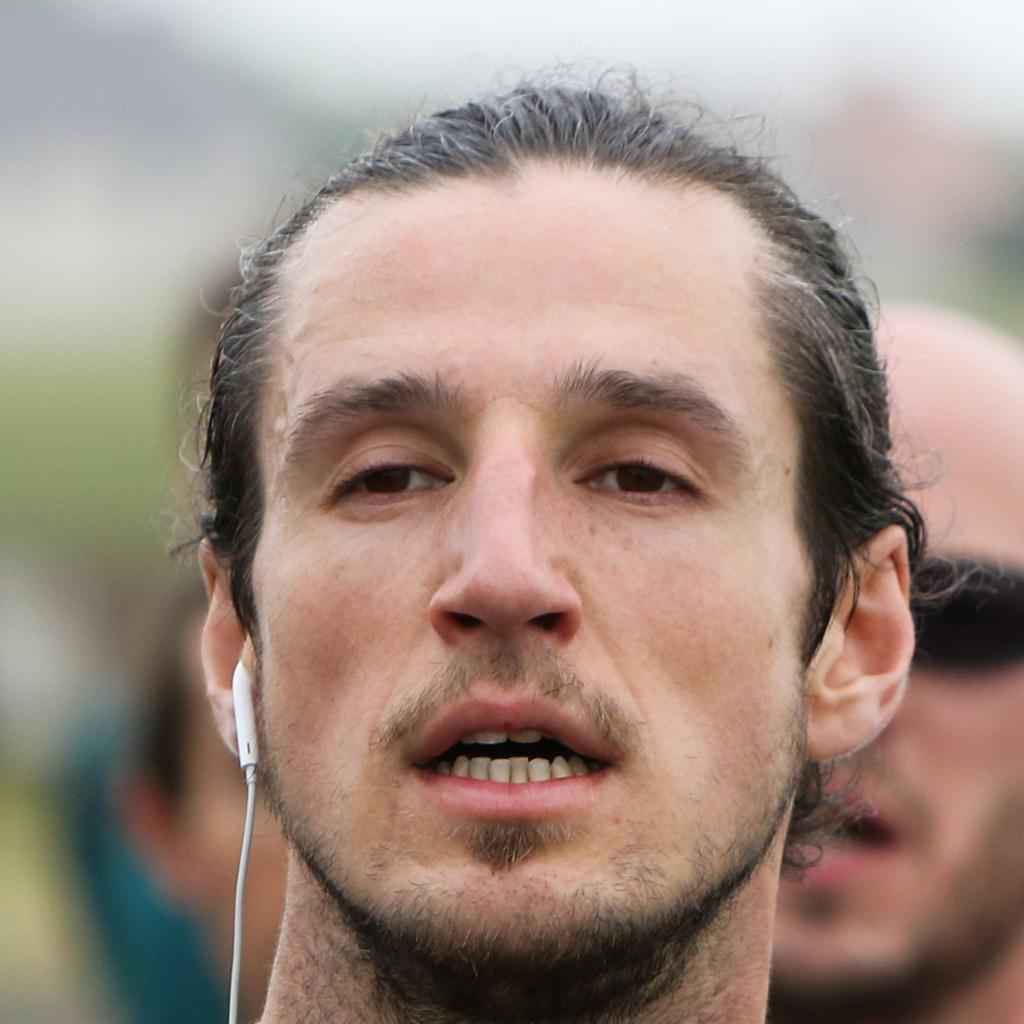}}} & 
\includegraphics[width=0.19\linewidth]{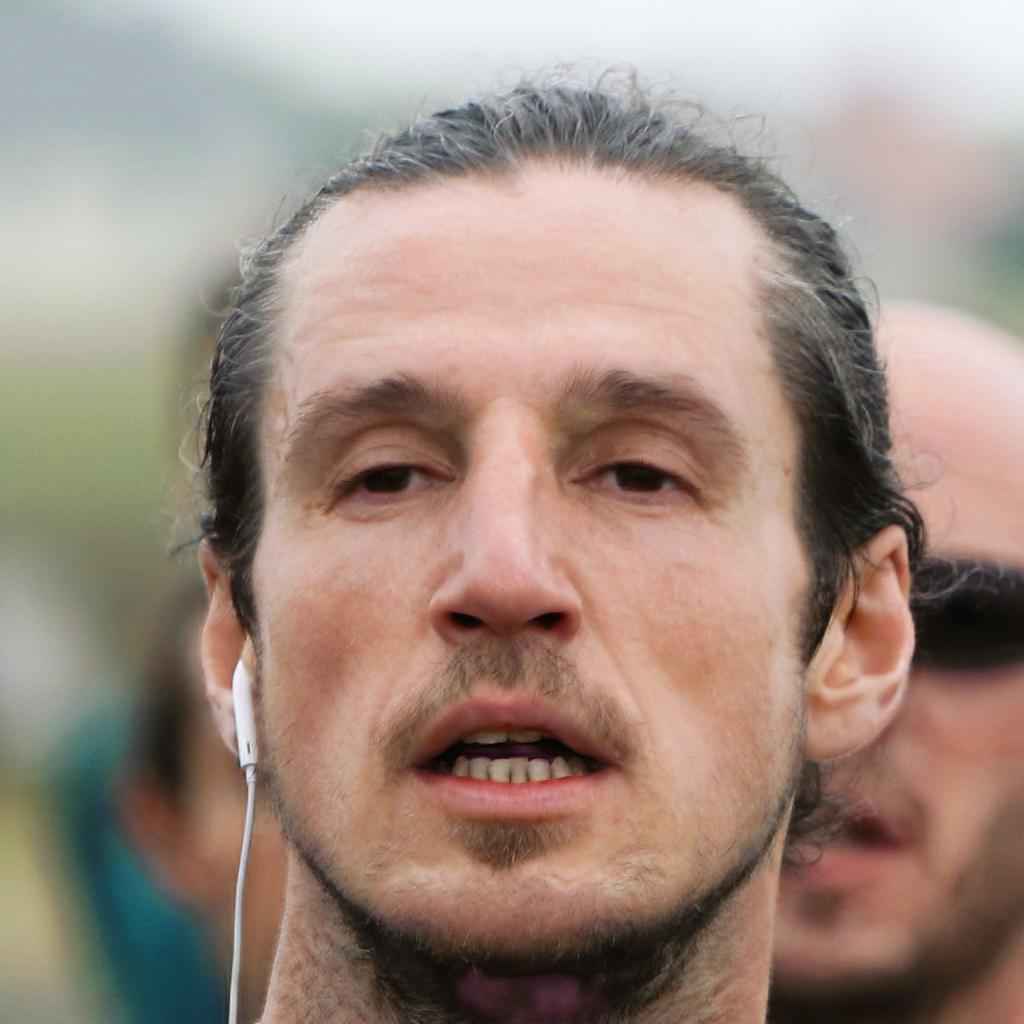} &
\includegraphics[width=0.19\linewidth]{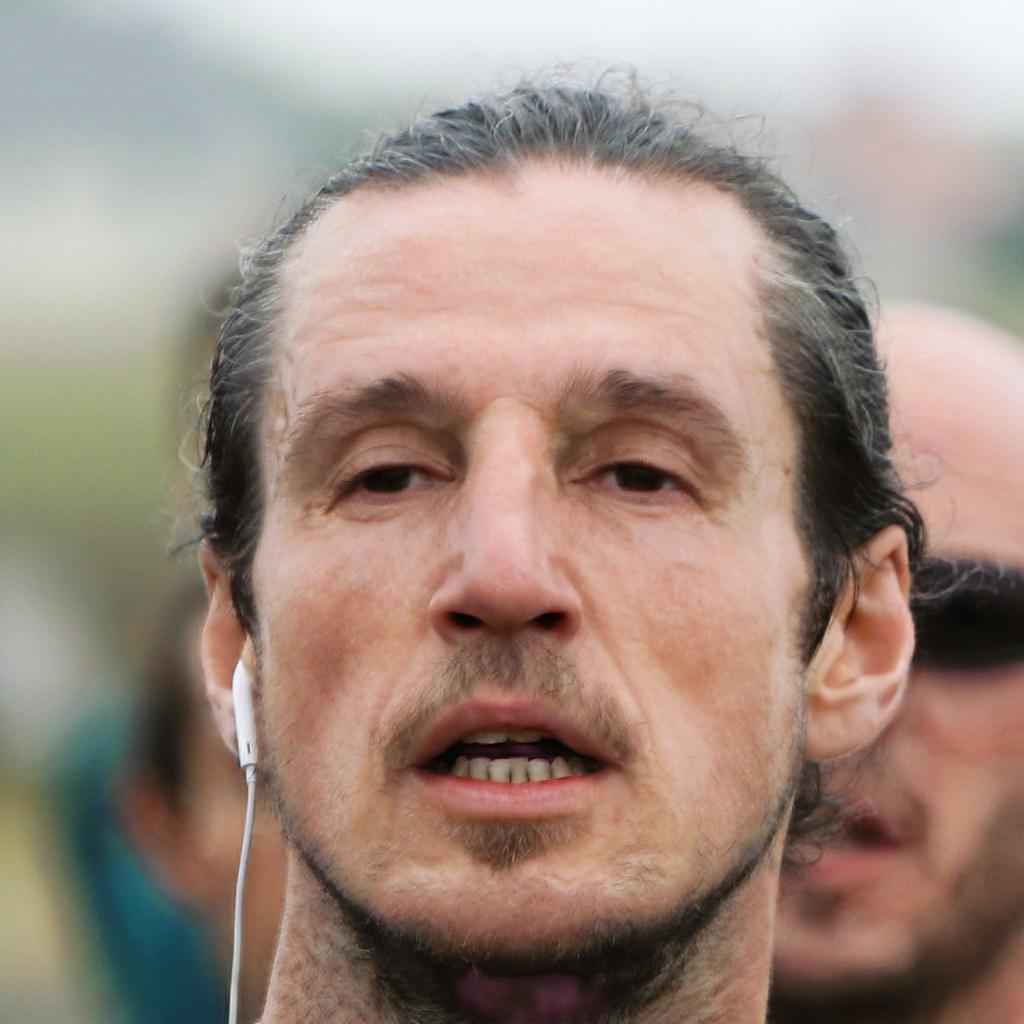} & 
\includegraphics[width=0.19\linewidth]{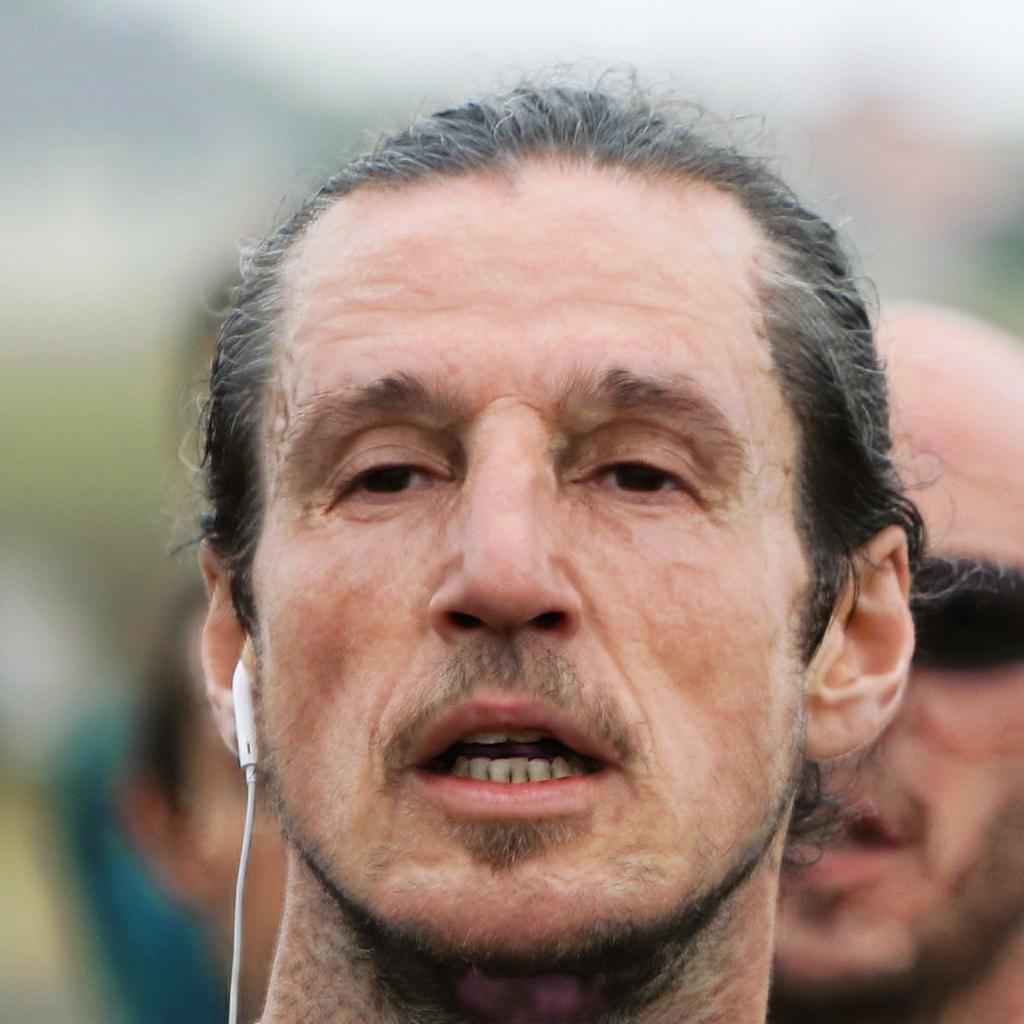} 
\\
\includegraphics[width=0.19\linewidth]{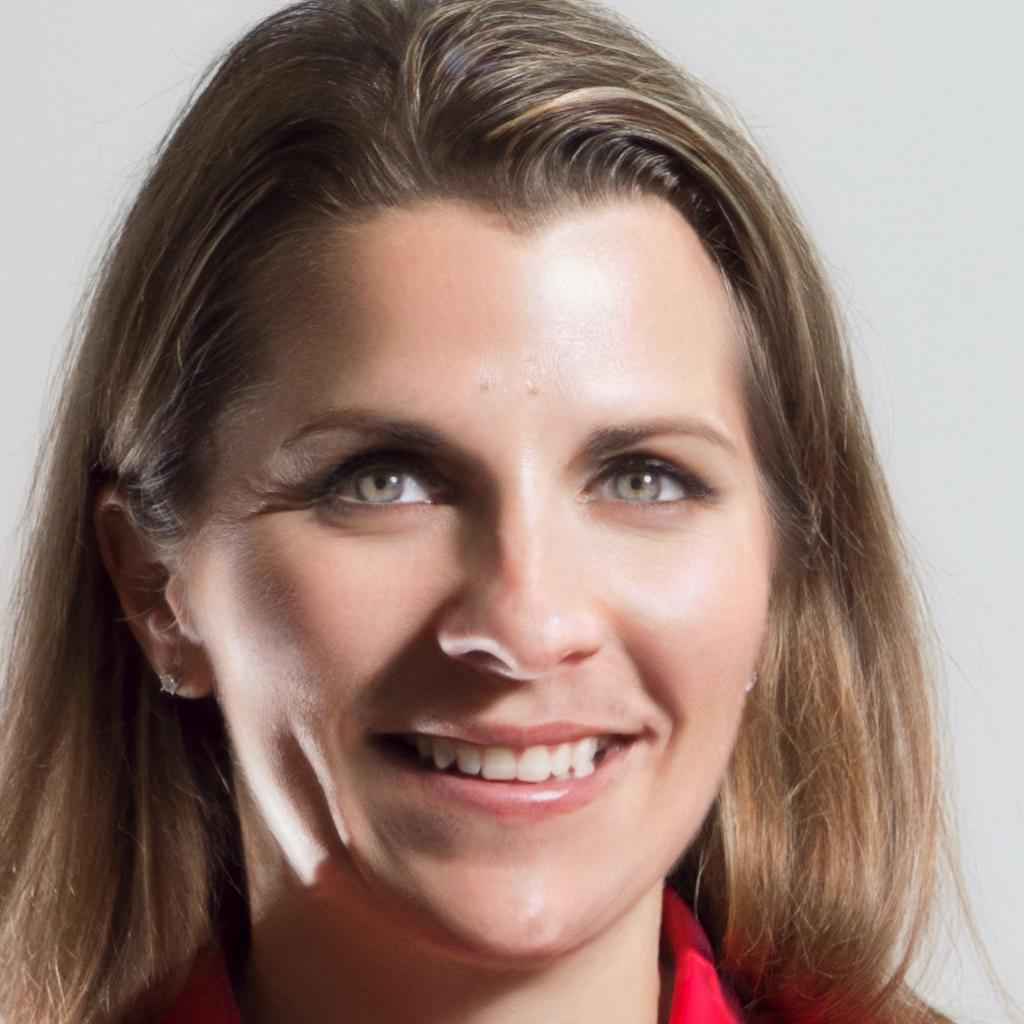} & 
\includegraphics[width=0.19\linewidth]{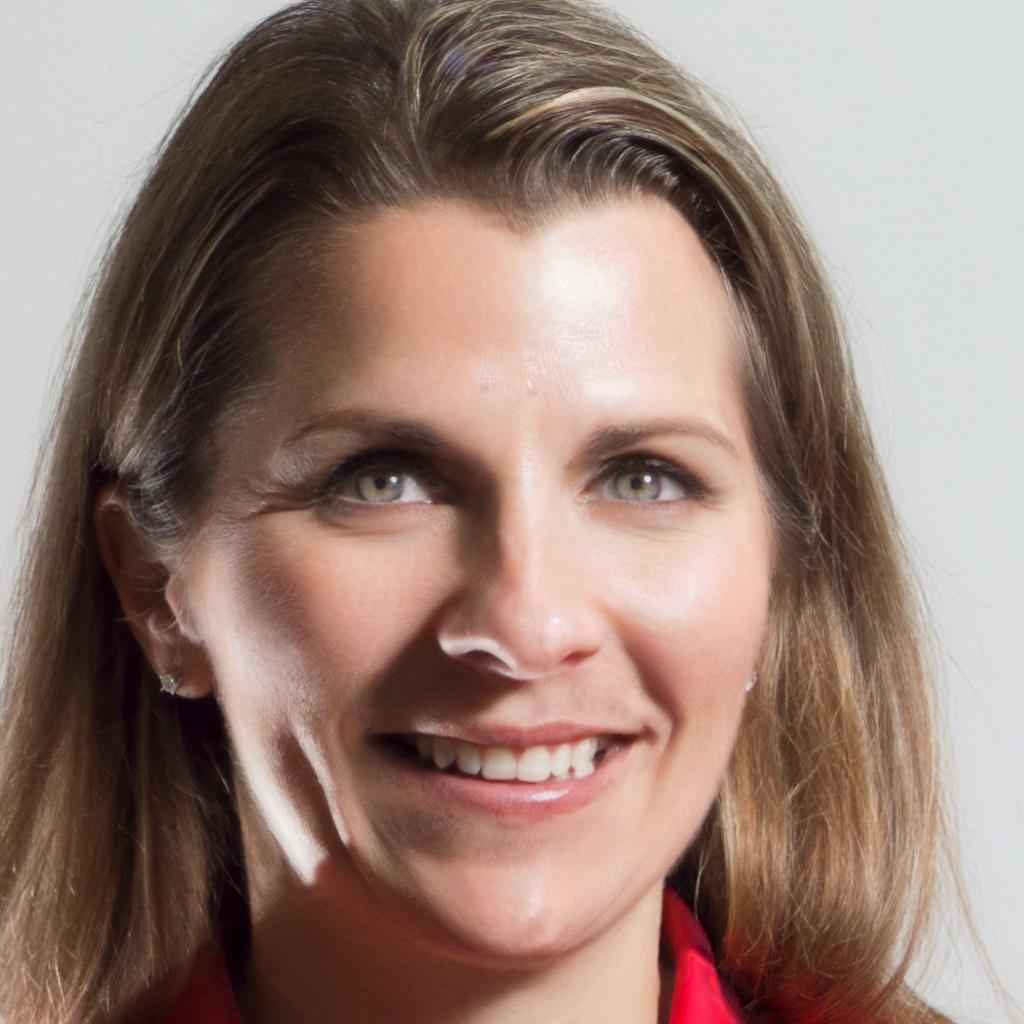} &
{\color{yellow}%
\setlength{\fboxsep}{0pt}%
\setlength{\fboxrule}{2pt}%
\fbox{\includegraphics[width=0.19\linewidth]{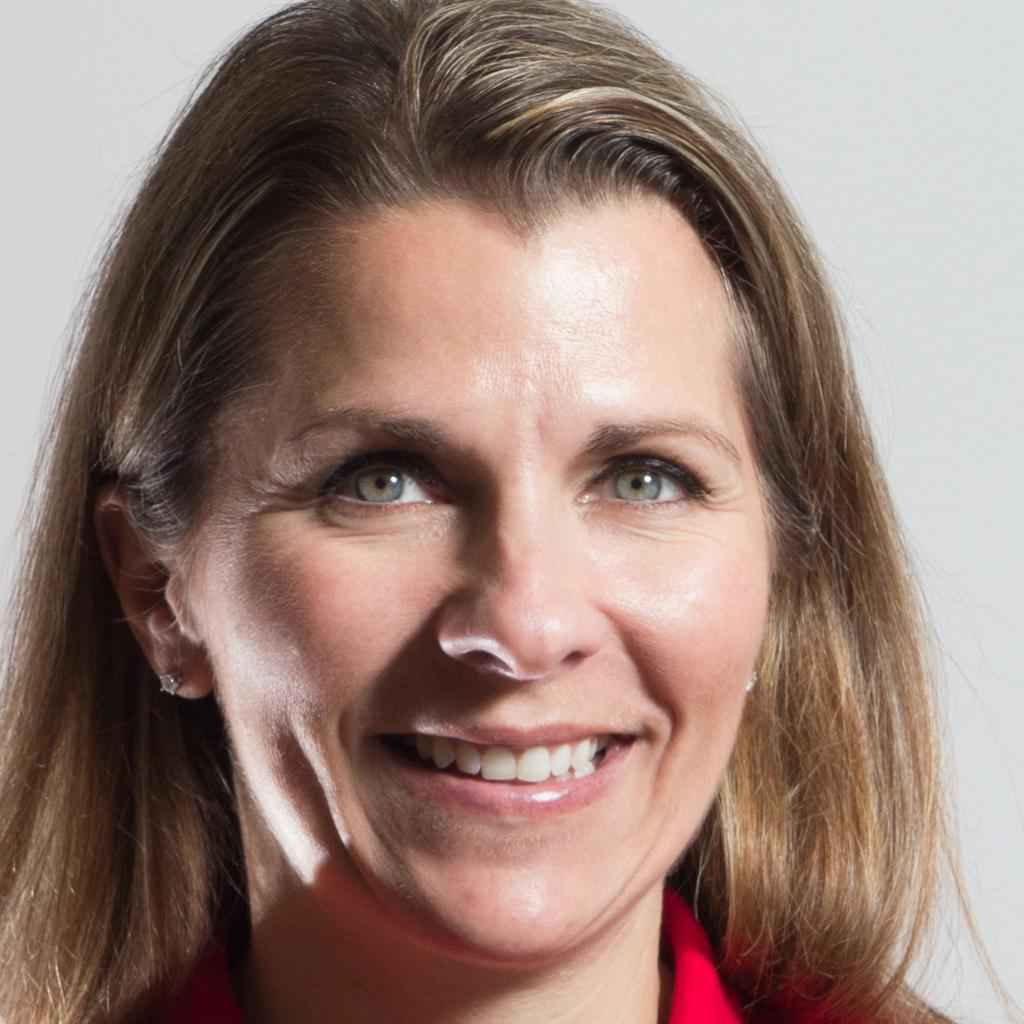}}} & 
\includegraphics[width=0.19\linewidth]{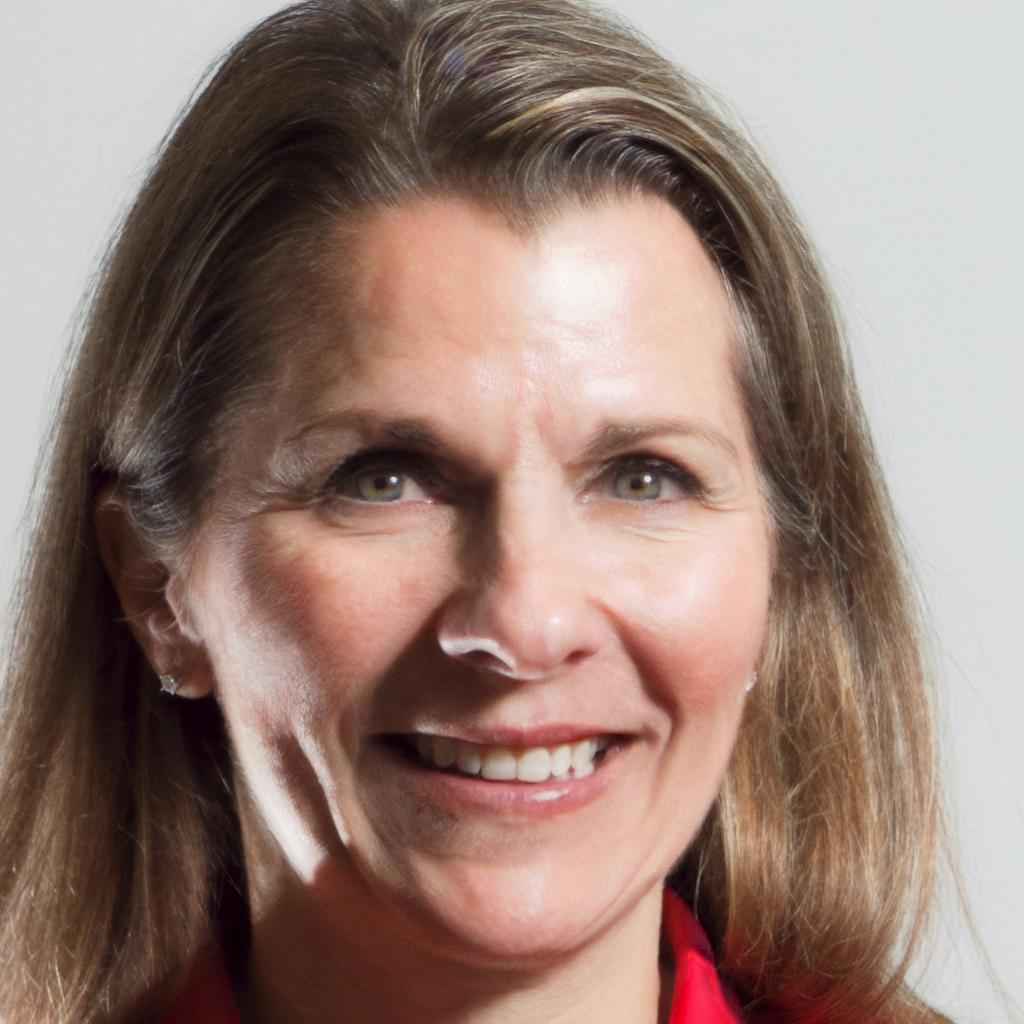} & 
\includegraphics[width=0.19\linewidth]{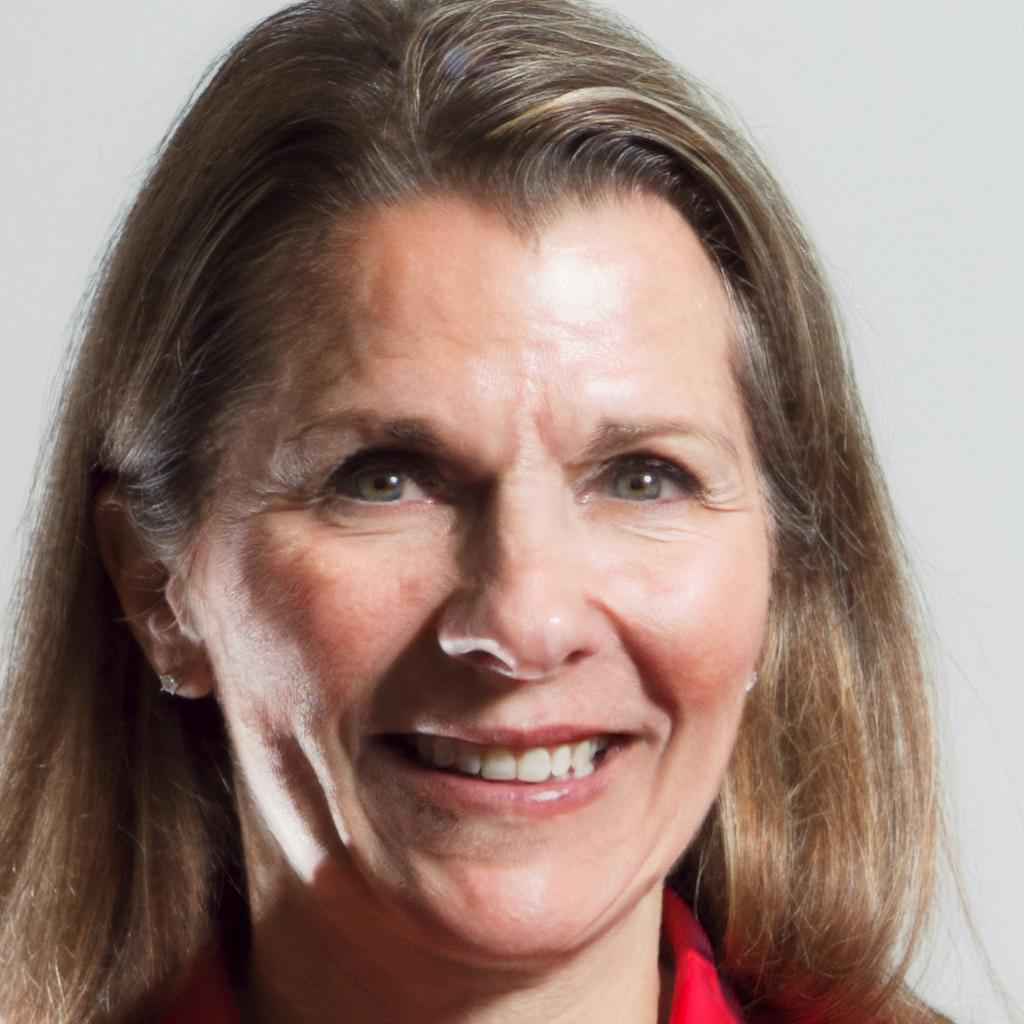} 
\\
\includegraphics[width=0.19\linewidth]{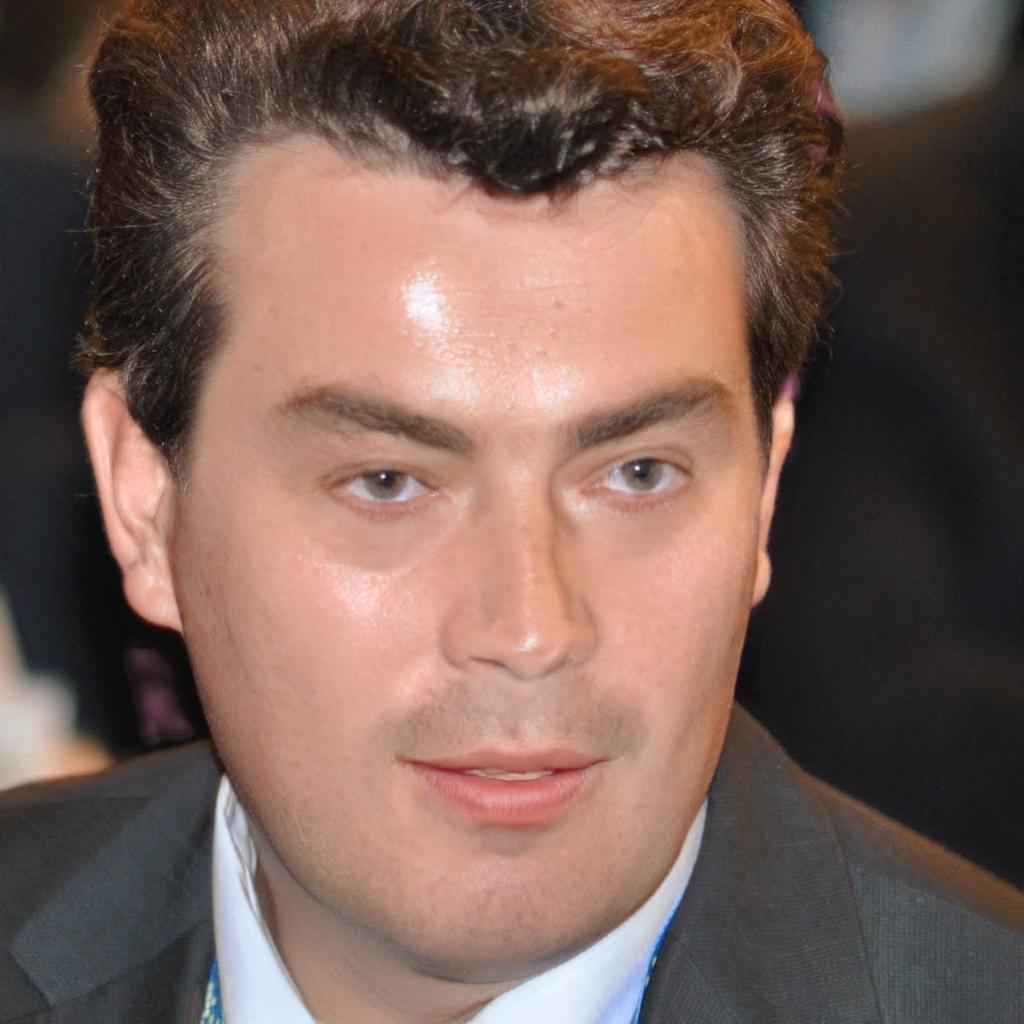} & 
\includegraphics[width=0.19\linewidth]{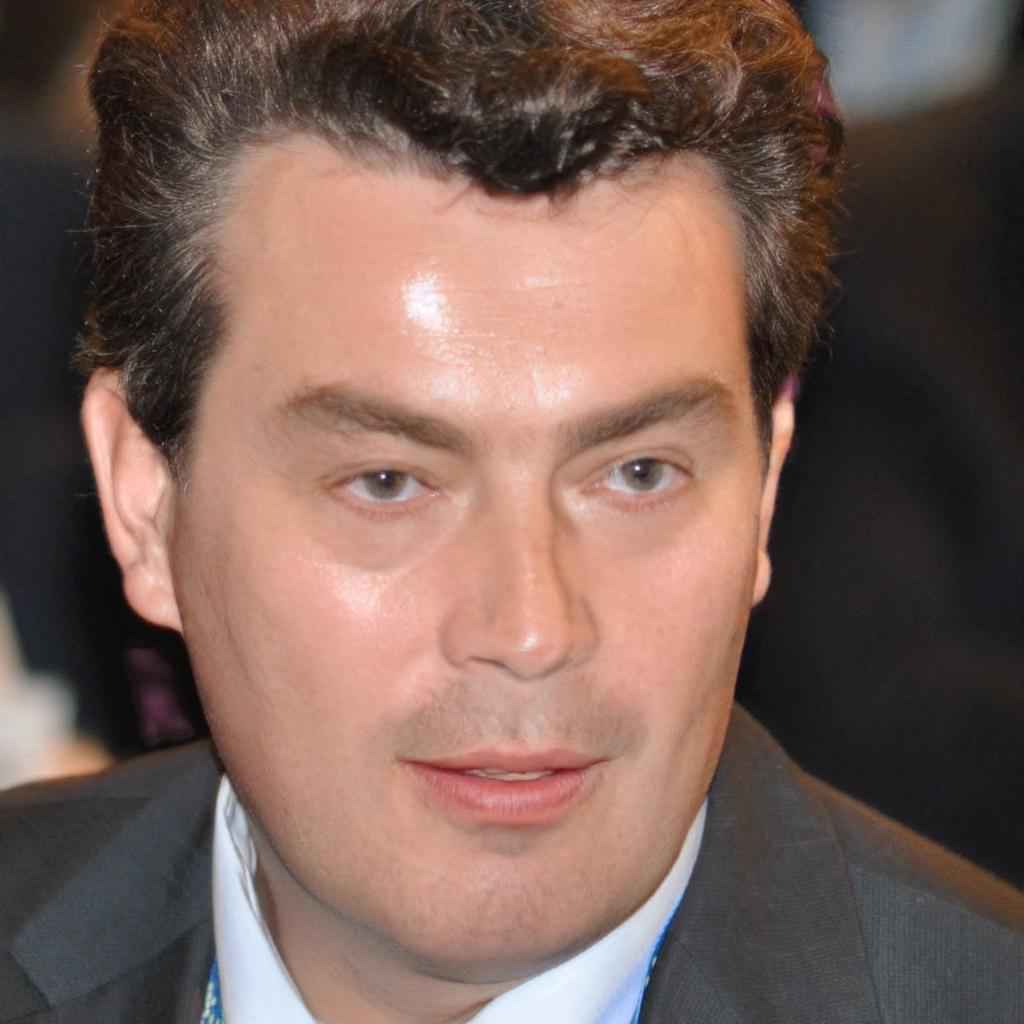} &
\includegraphics[width=0.19\linewidth]{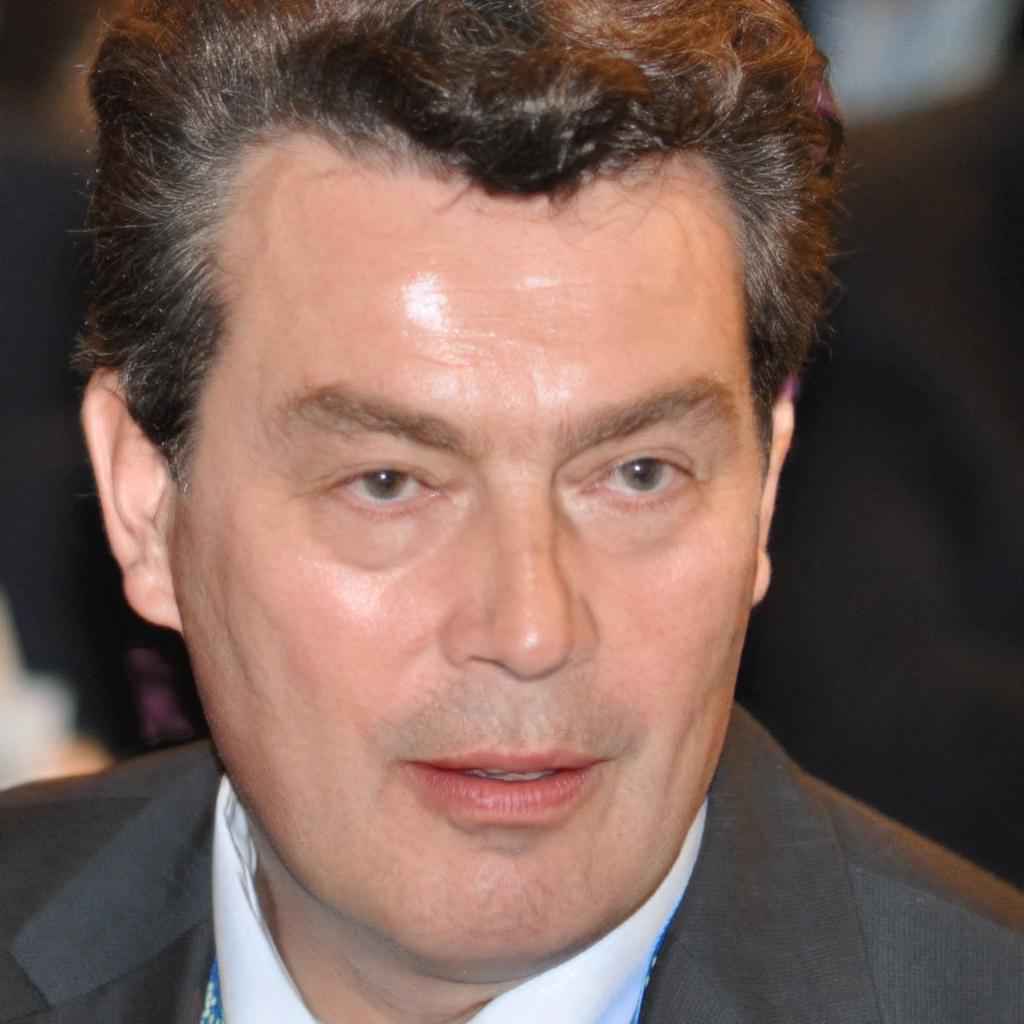} &
{\color{yellow}%
\setlength{\fboxsep}{0pt}%
\setlength{\fboxrule}{2pt}%
\fbox{\includegraphics[width=0.19\linewidth]{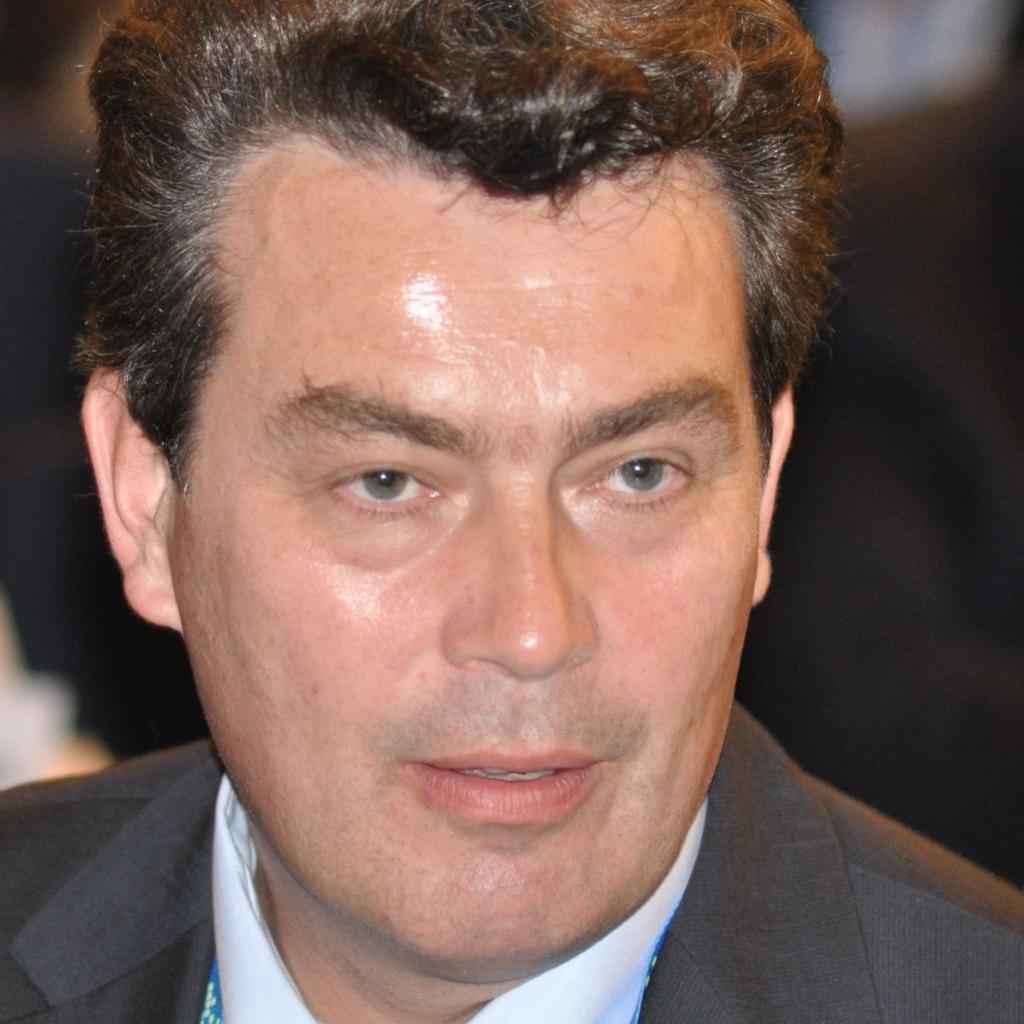}}} &  
\includegraphics[width=0.19\linewidth]{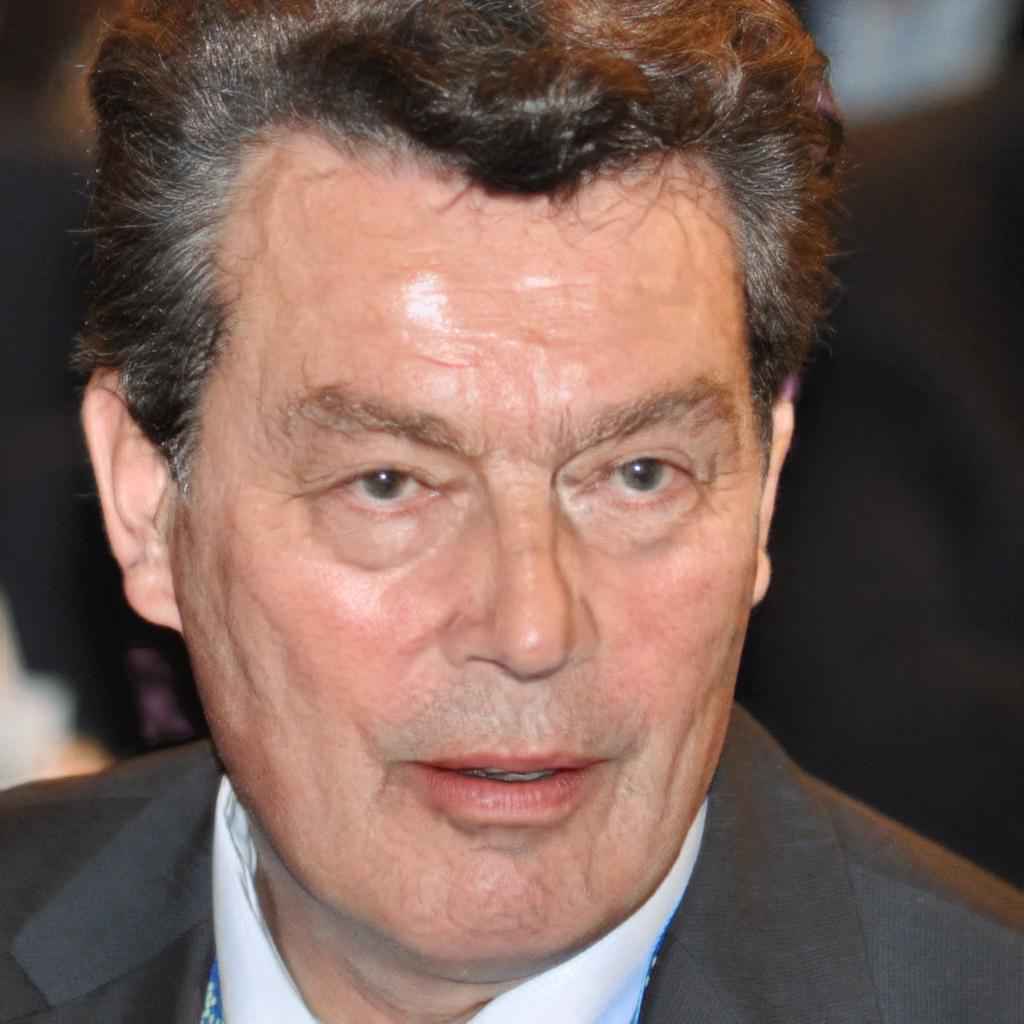} 
\\
\includegraphics[width=0.19\linewidth]{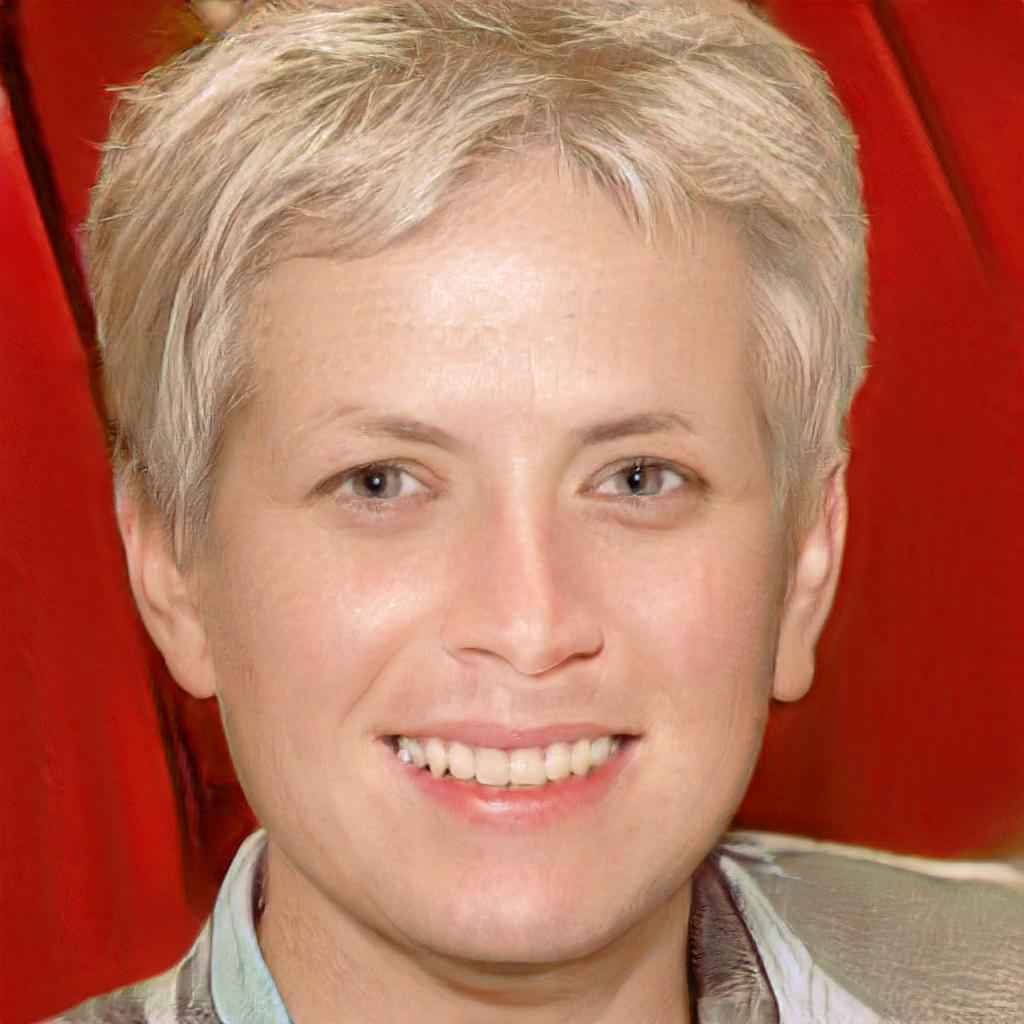} & 
\includegraphics[width=0.19\linewidth]{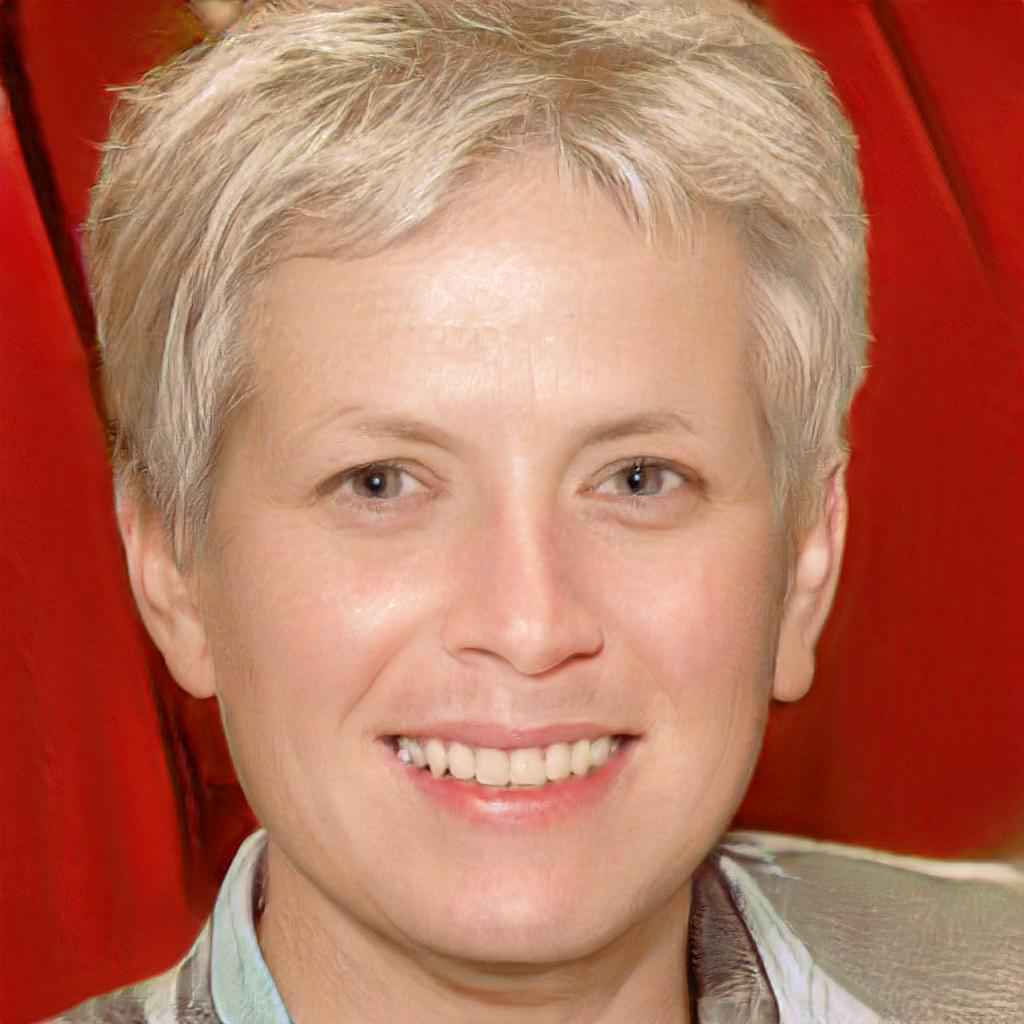} &
\includegraphics[width=0.19\linewidth]{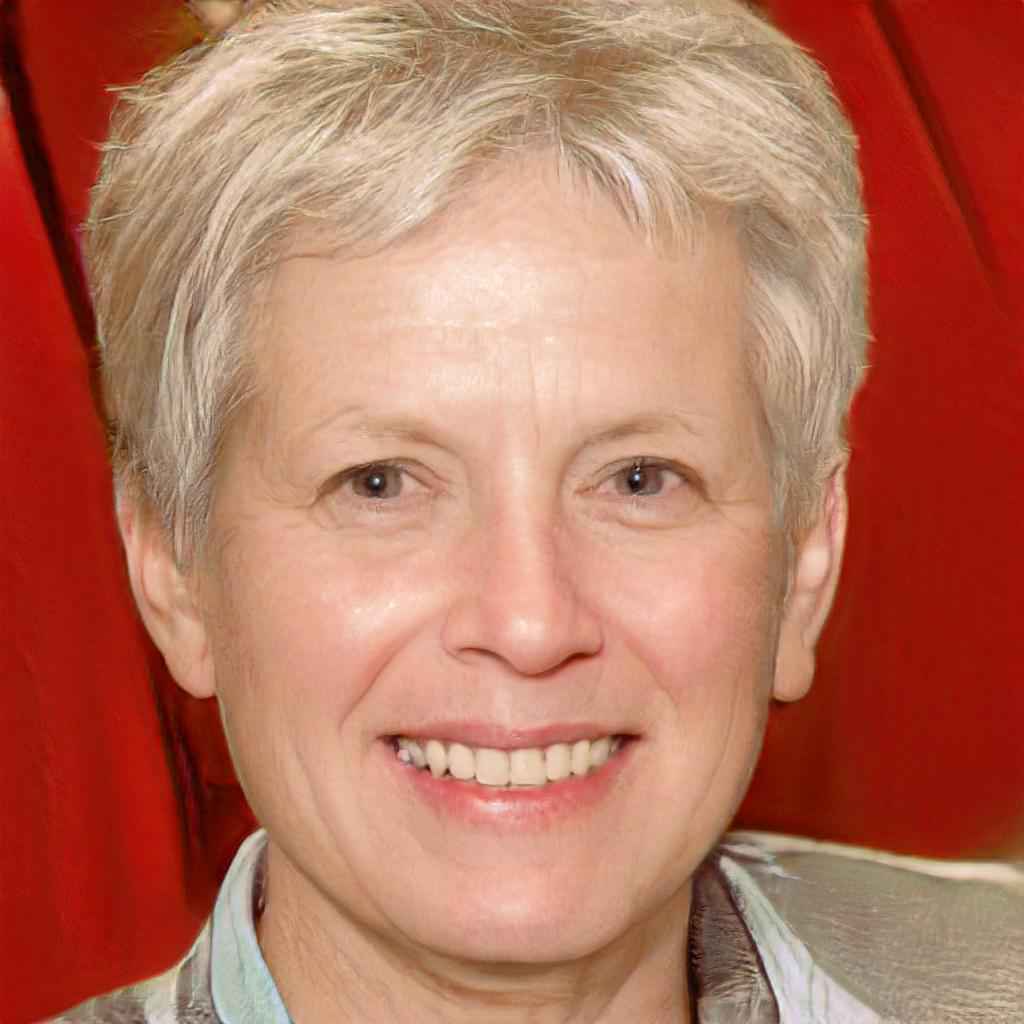} &
\includegraphics[width=0.19\linewidth]{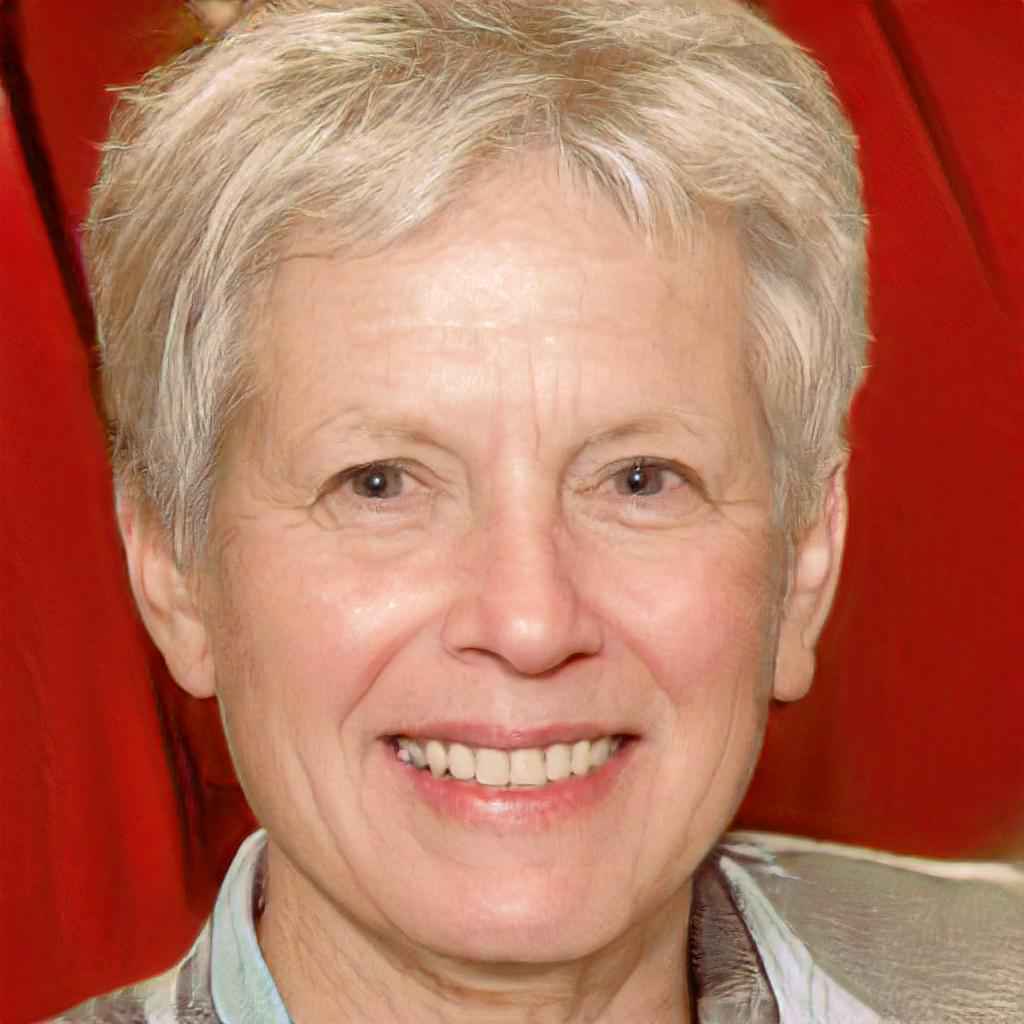} & 
{\color{yellow}%
\setlength{\fboxsep}{0pt}%
\setlength{\fboxrule}{2pt}%
\fbox{\includegraphics[width=0.19\linewidth]{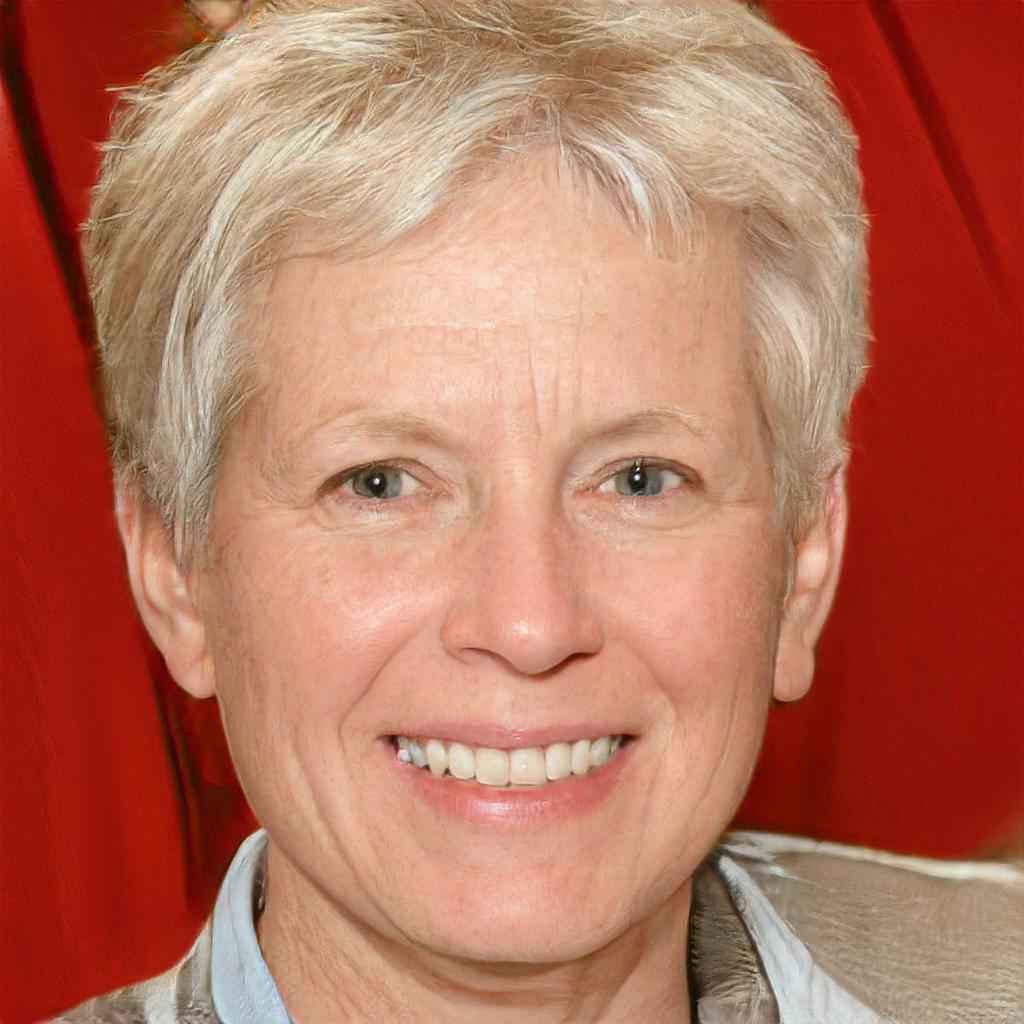}}} 
\end{tabular}
\end{center}
\caption{\textbf{Age transformation on $\bf 1024 \times 1024$ images}. On each row, the yellow frame indicates the original image. Each column corresponds to a target age of: $25$, $35$, $45$, $55$, $65$. Our approach yields visually satisfying results without introducing significant artifacts. Only age relevant features are modified, while the identity, haircut, emotion and background are perfectly preserved. 
}
\label{1024_3}
\end{figure}
\begin{figure}[ht]
\begin{center}
\setlength{\tabcolsep}{1pt}
\begin{tabular}{ccccc}
25&35&45&55&65 \\
{\color{yellow}%
\setlength{\fboxsep}{0pt}%
\setlength{\fboxrule}{2pt}%
\fbox{\includegraphics[width=0.19\linewidth]{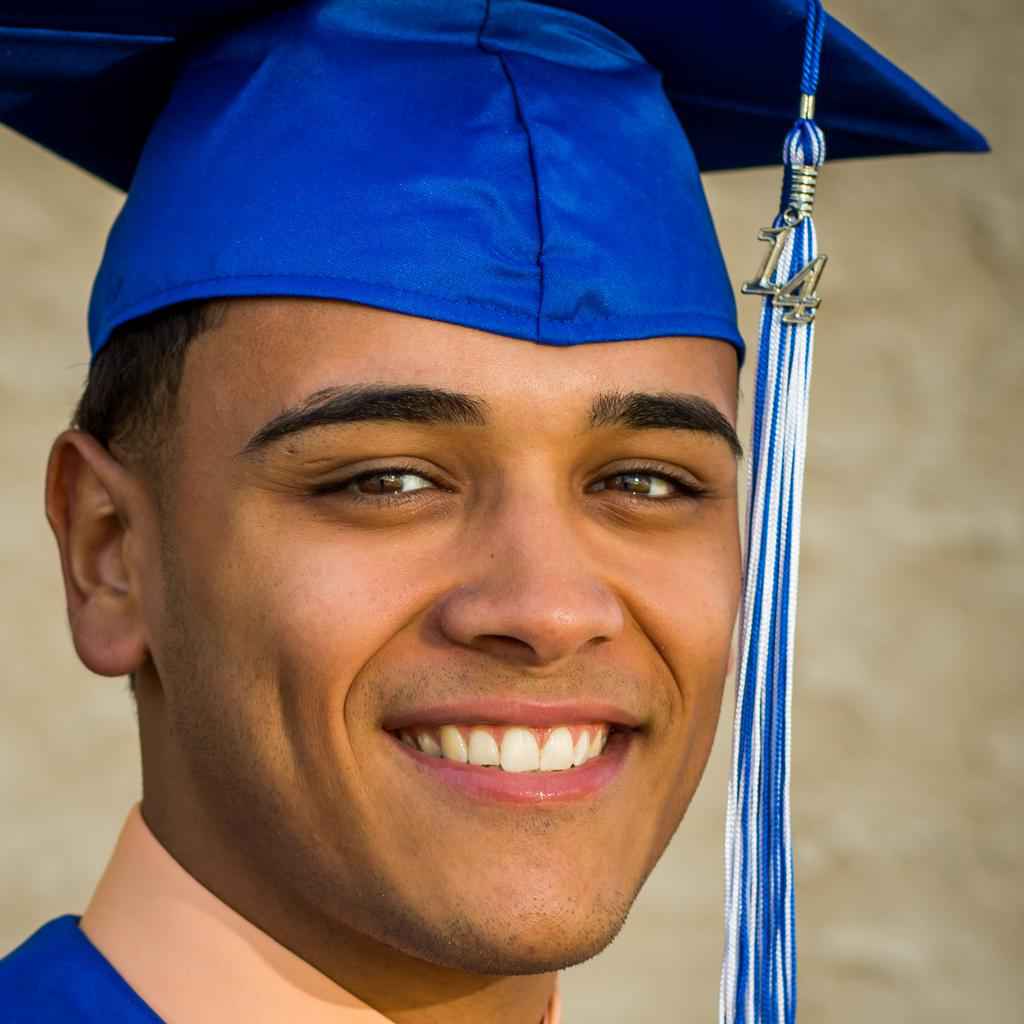}}} &
\includegraphics[width=0.19\linewidth]{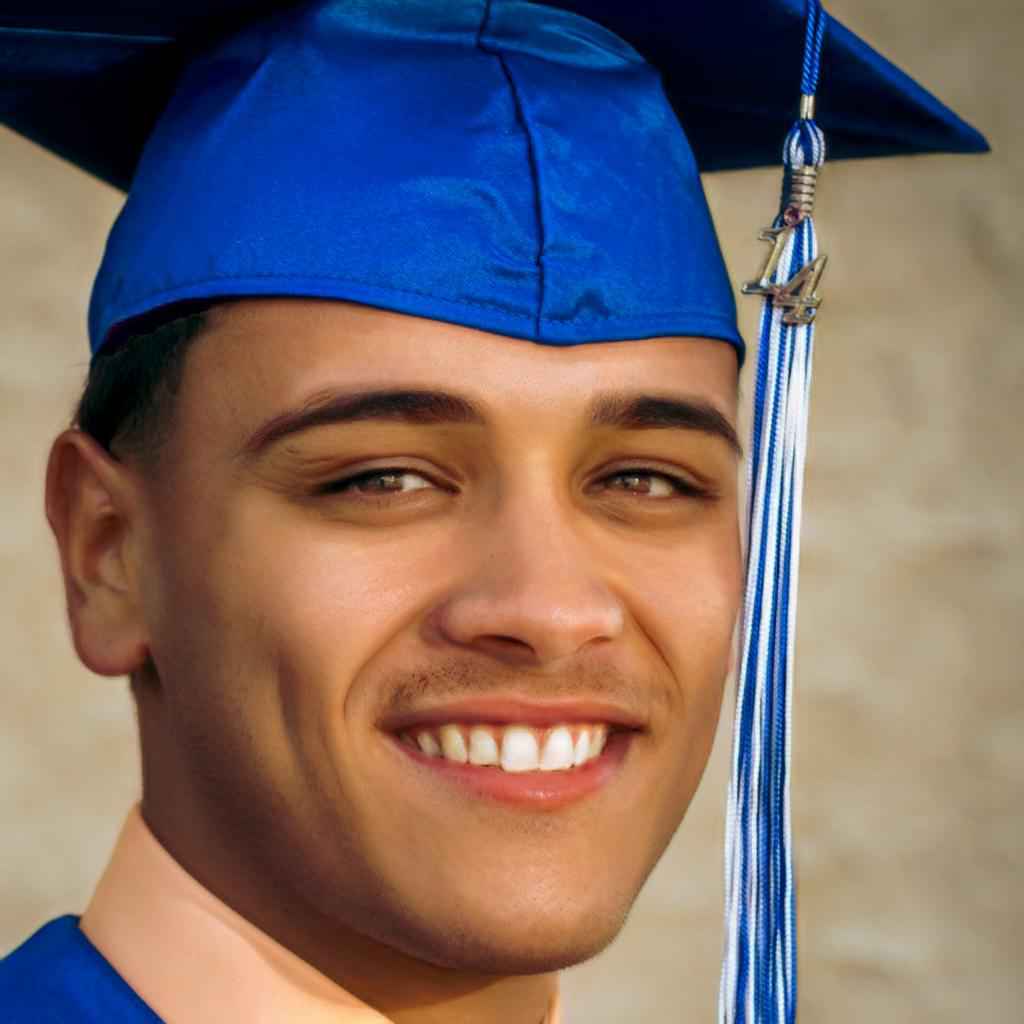} & 
\includegraphics[width=0.19\linewidth]{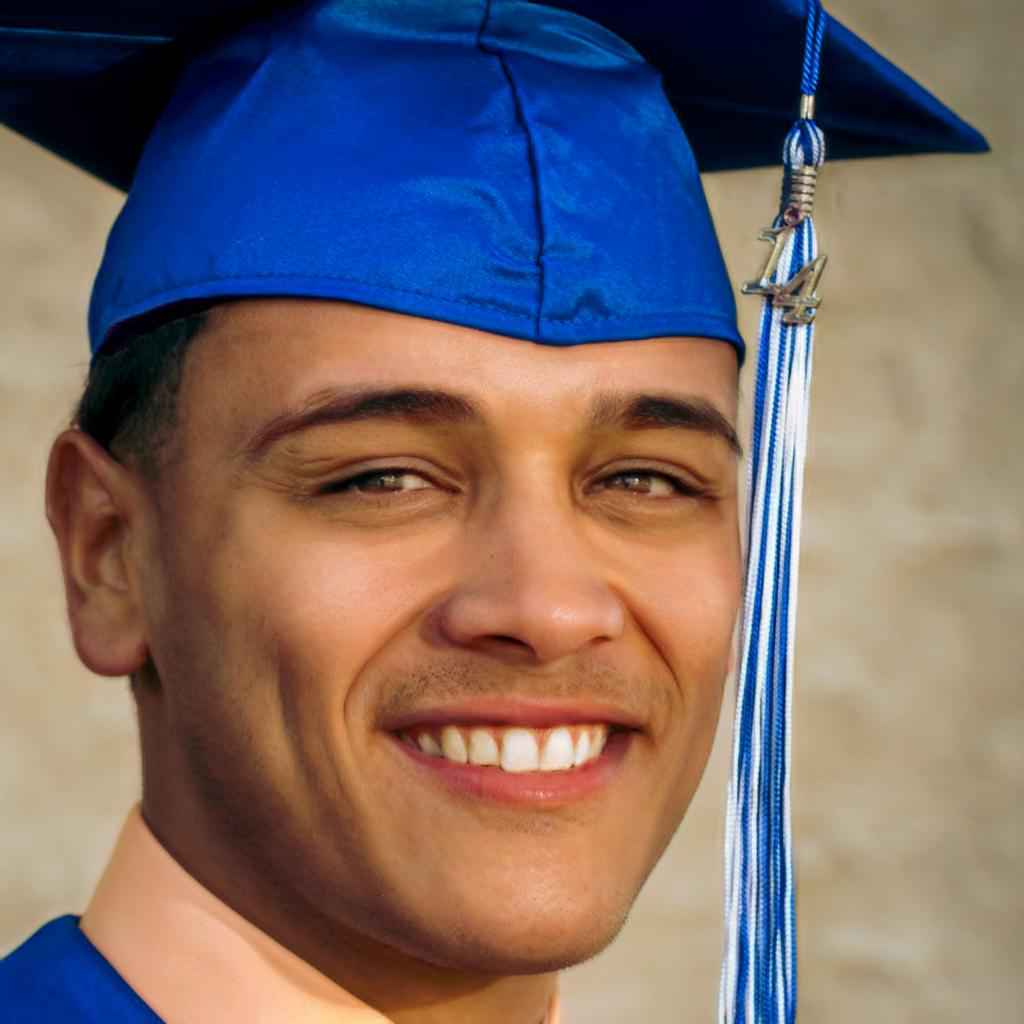} &
\includegraphics[width=0.19\linewidth]{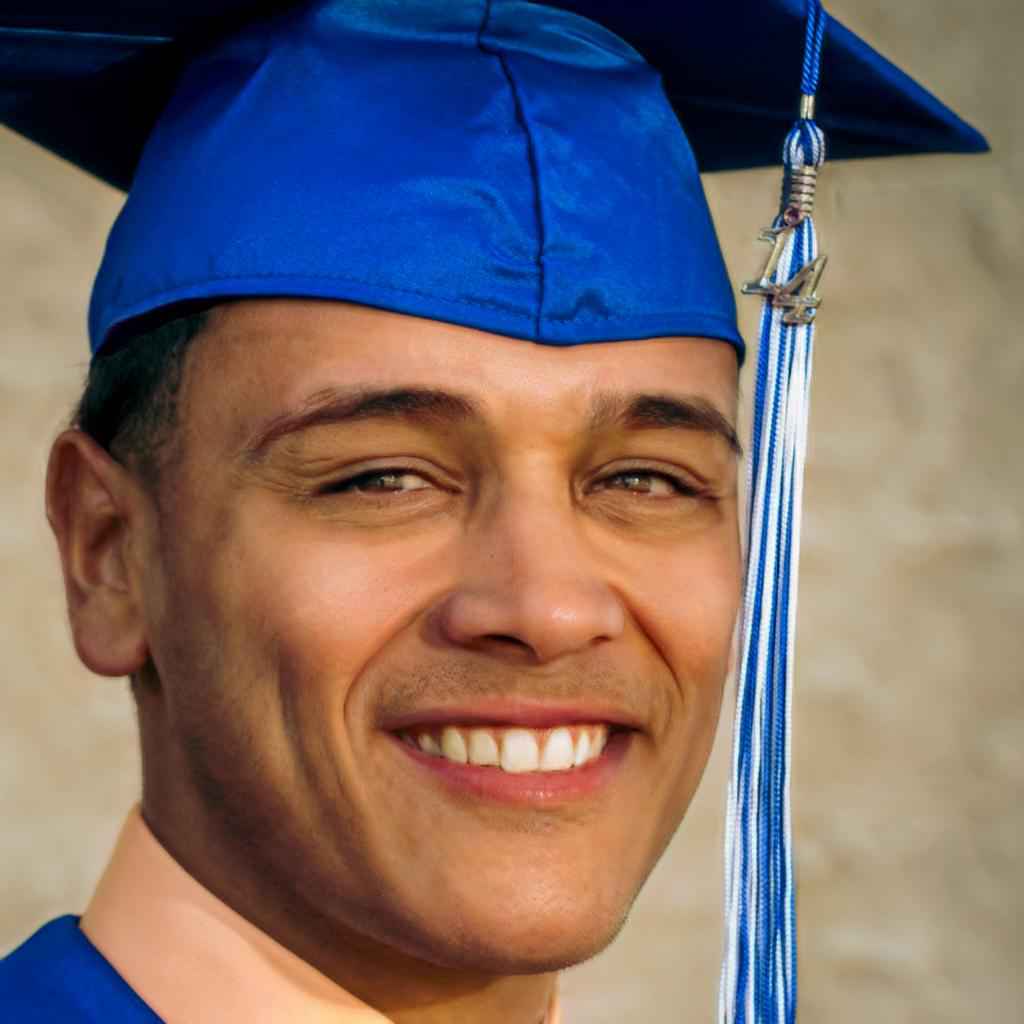} & 
\includegraphics[width=0.19\linewidth]{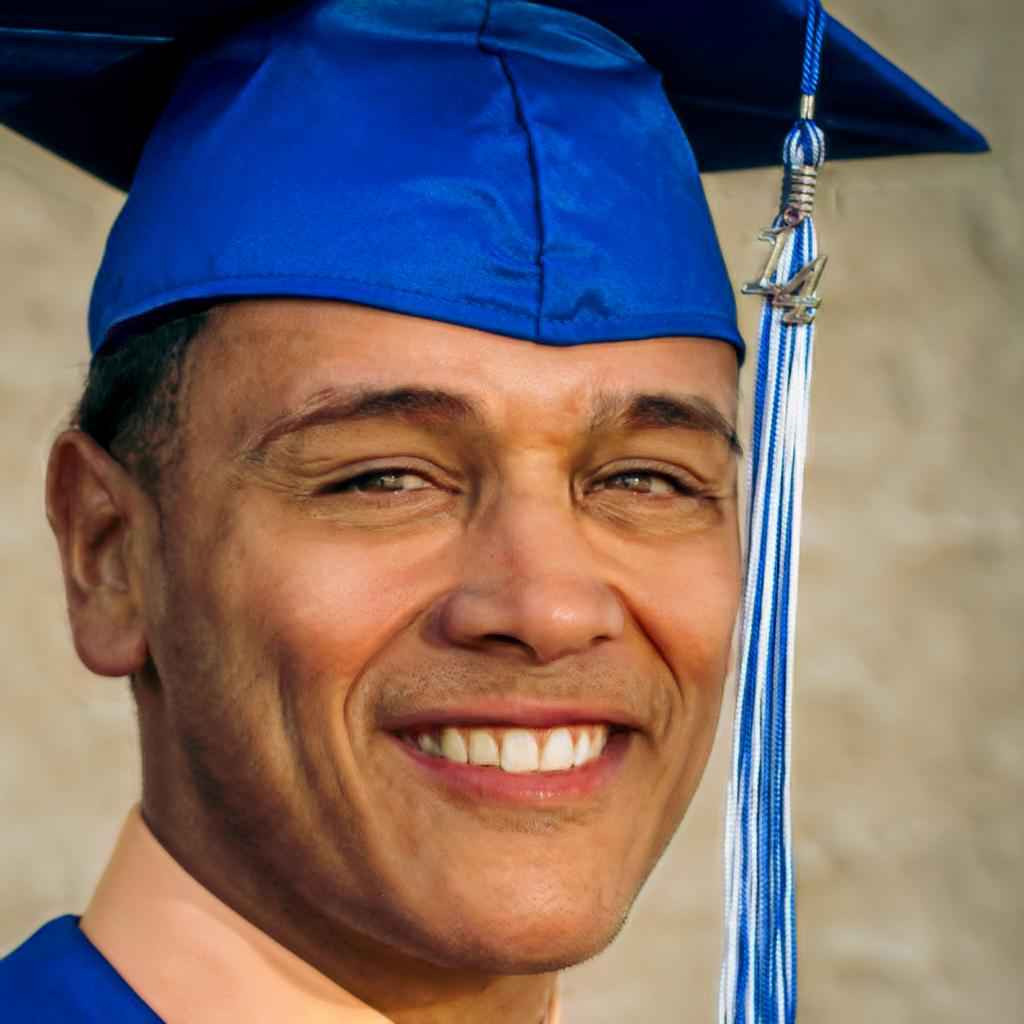} 
\\
\includegraphics[width=0.19\linewidth]{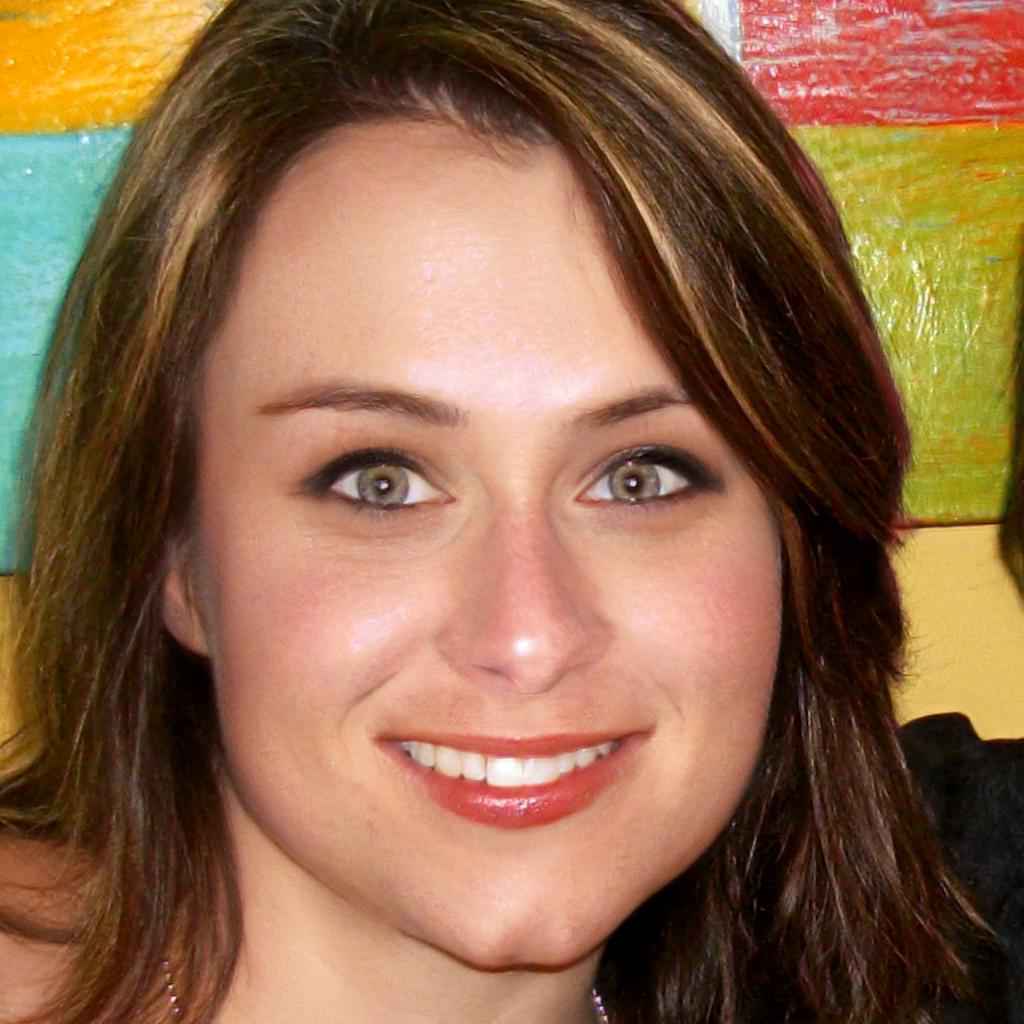} & 
{\color{yellow}%
\setlength{\fboxsep}{0pt}%
\setlength{\fboxrule}{2pt}%
\fbox{\includegraphics[width=0.19\linewidth]{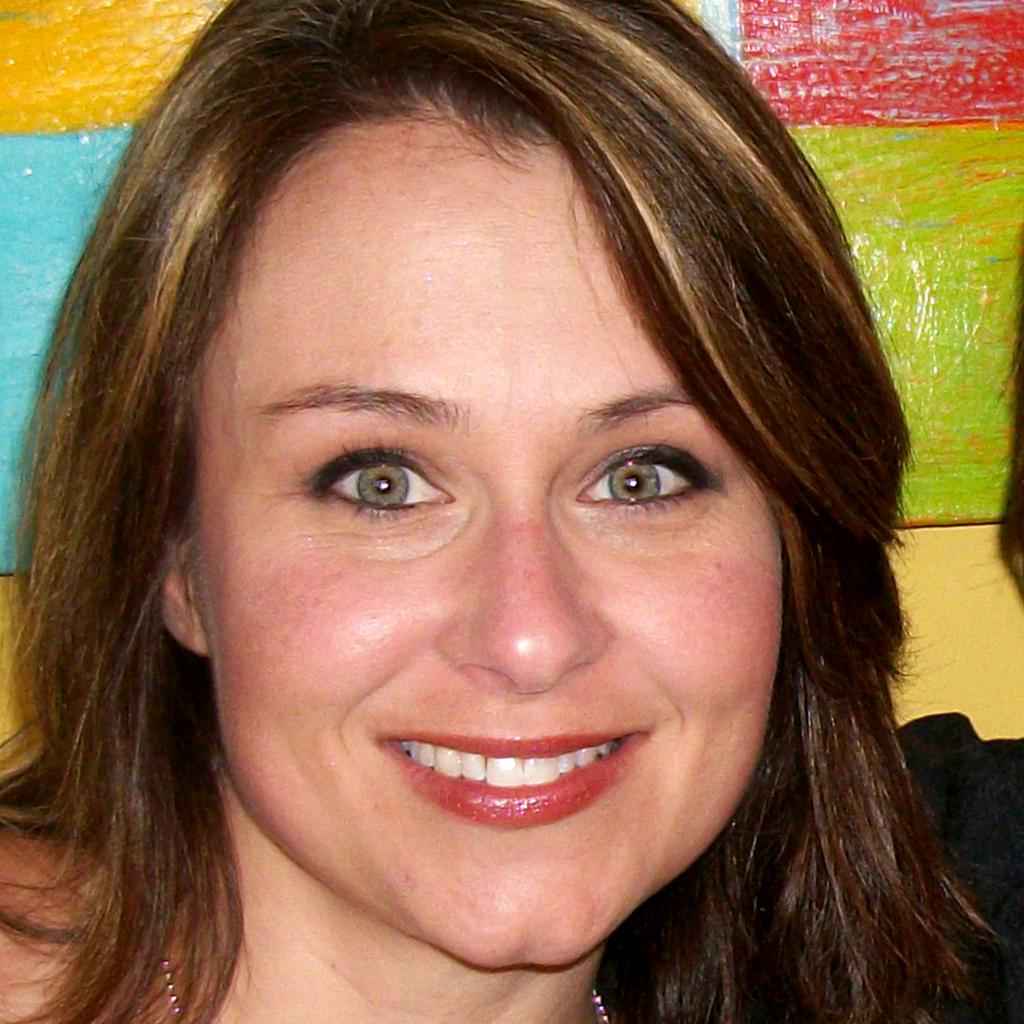}}} & 
\includegraphics[width=0.19\linewidth]{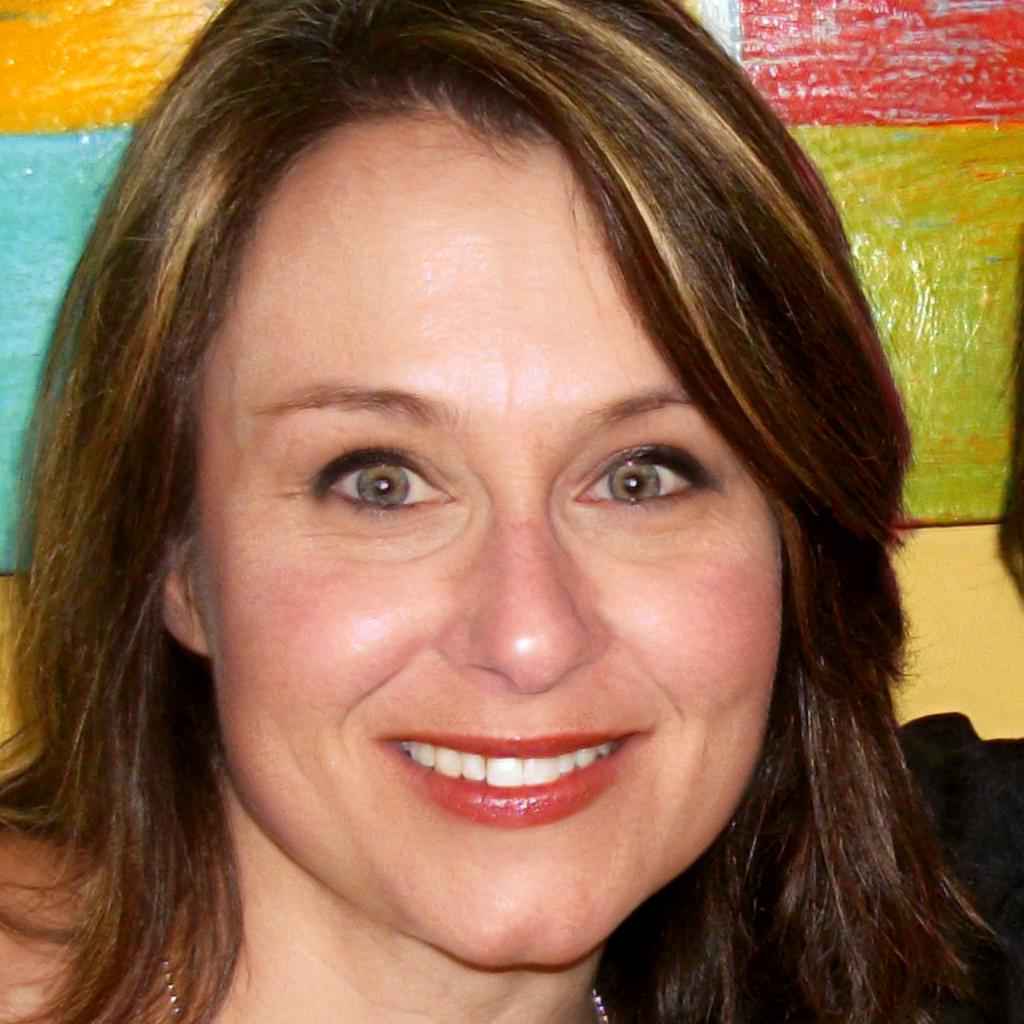} &
\includegraphics[width=0.19\linewidth]{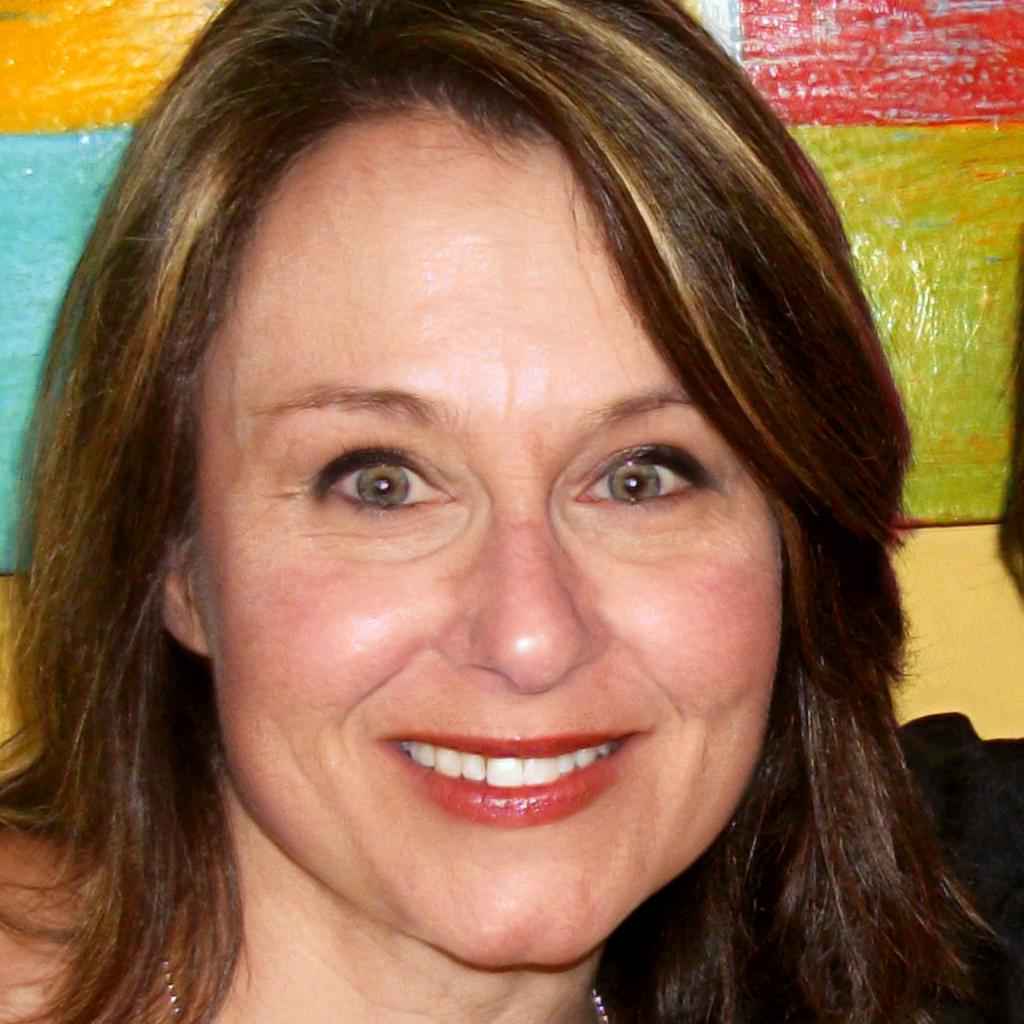} & 
\includegraphics[width=0.19\linewidth]{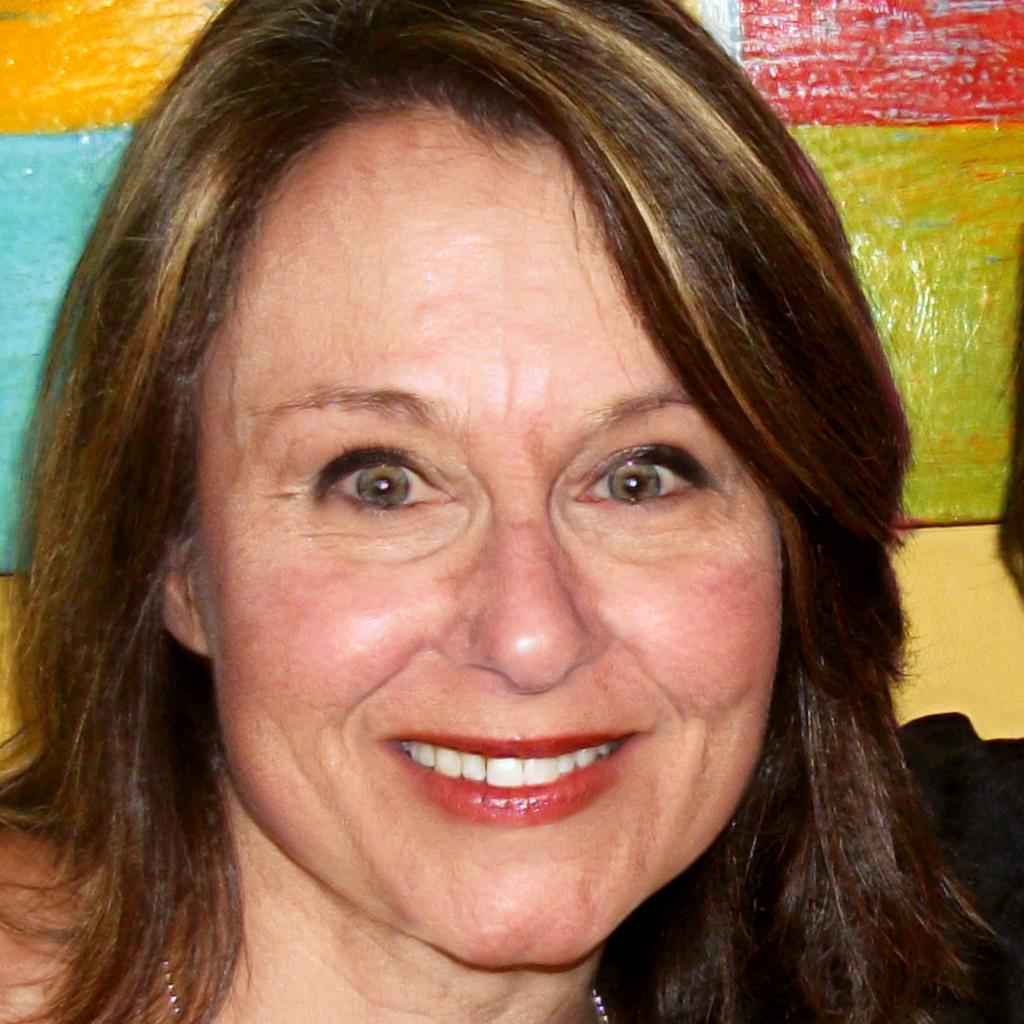} 
\\
\includegraphics[width=0.19\linewidth]{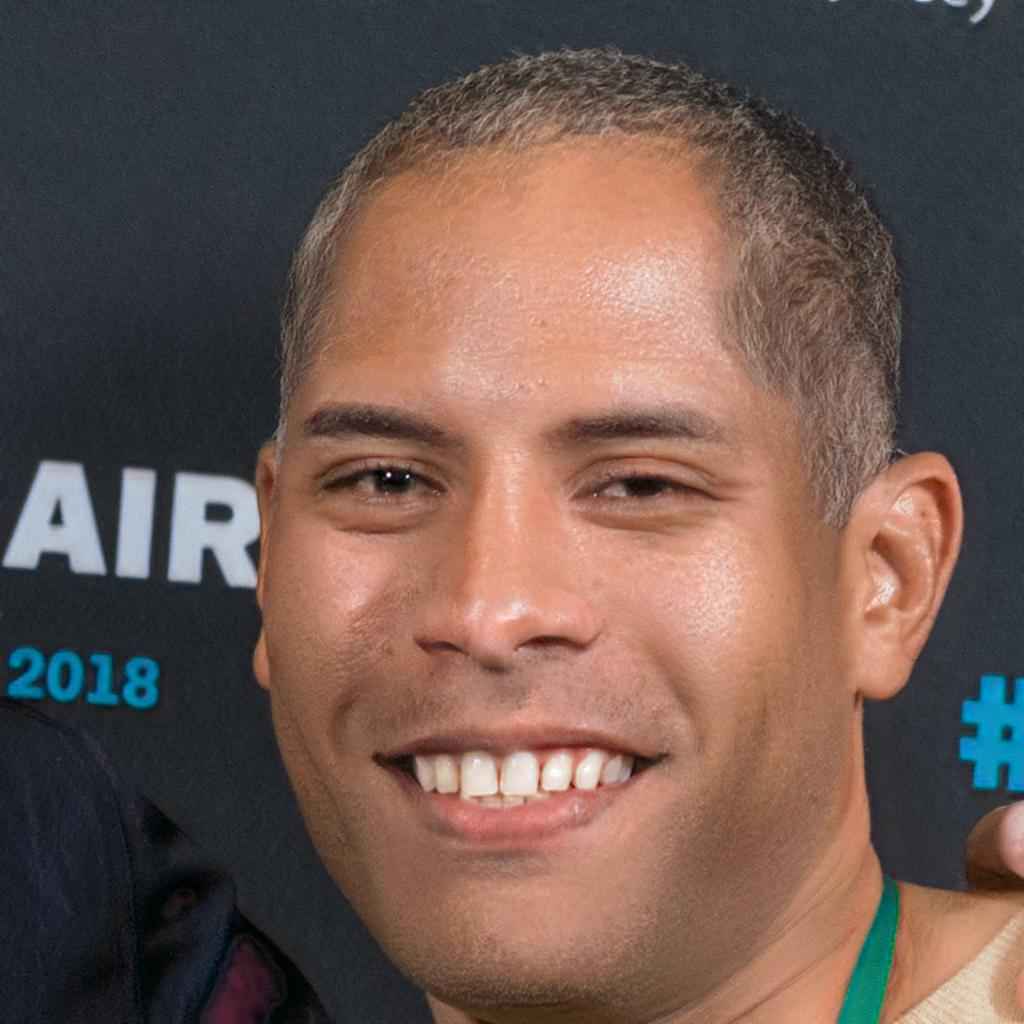} & 
\includegraphics[width=0.19\linewidth]{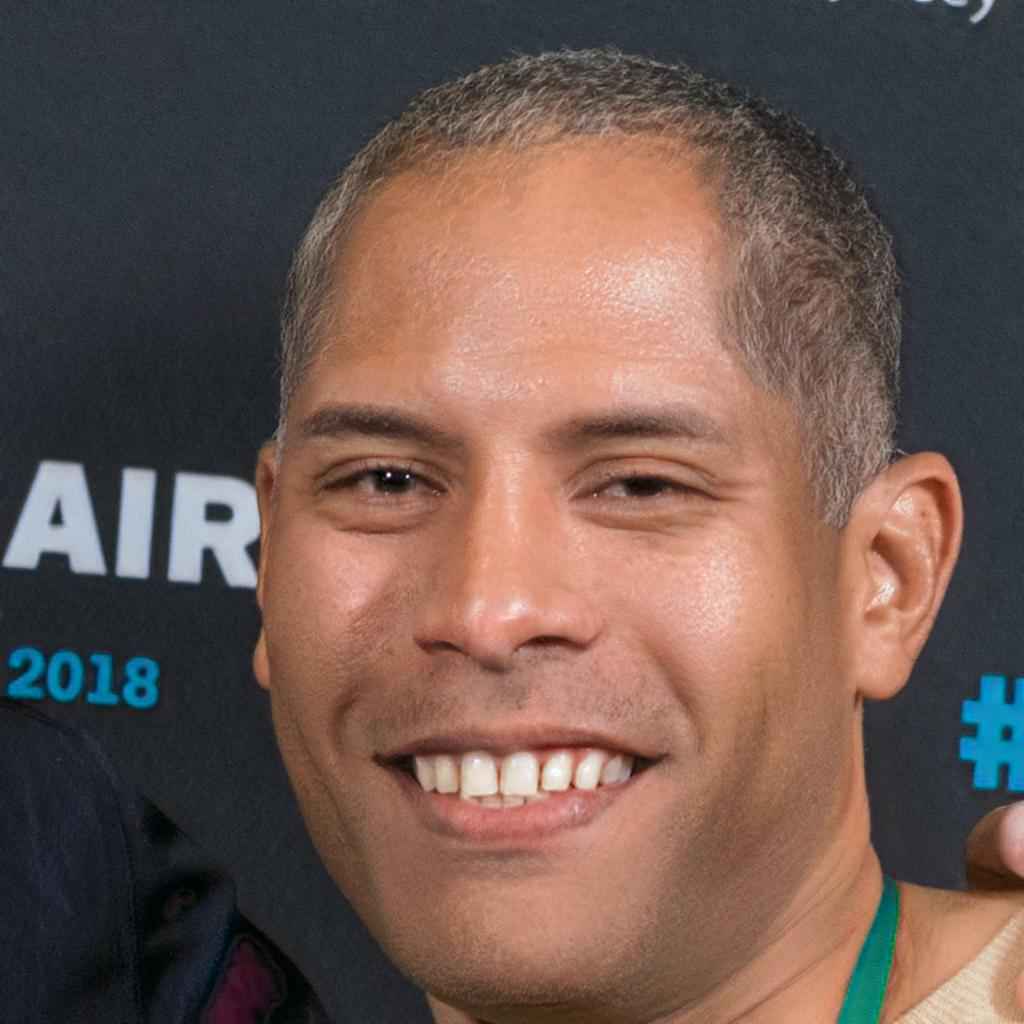} &
{\color{yellow}%
\setlength{\fboxsep}{0pt}%
\setlength{\fboxrule}{2pt}%
\fbox{\includegraphics[width=0.19\linewidth]{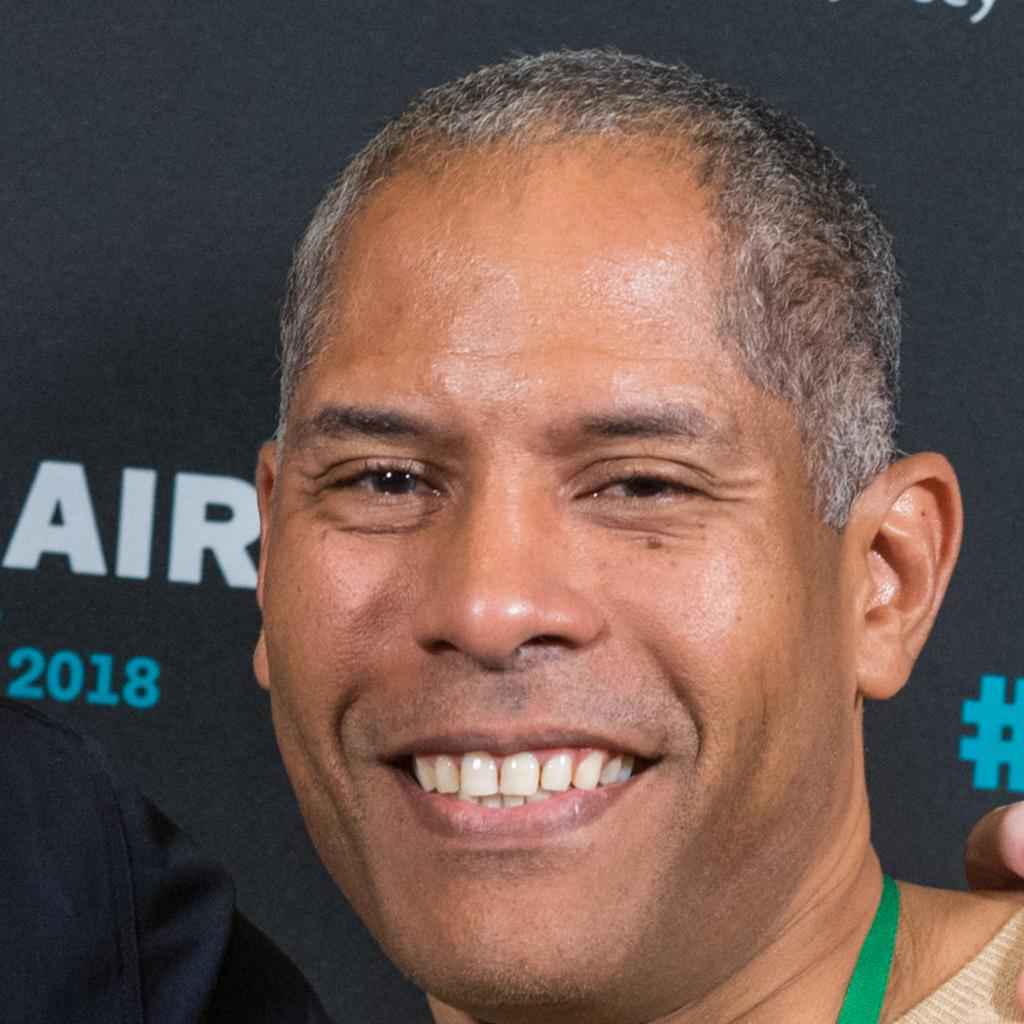}}} & 
\includegraphics[width=0.19\linewidth]{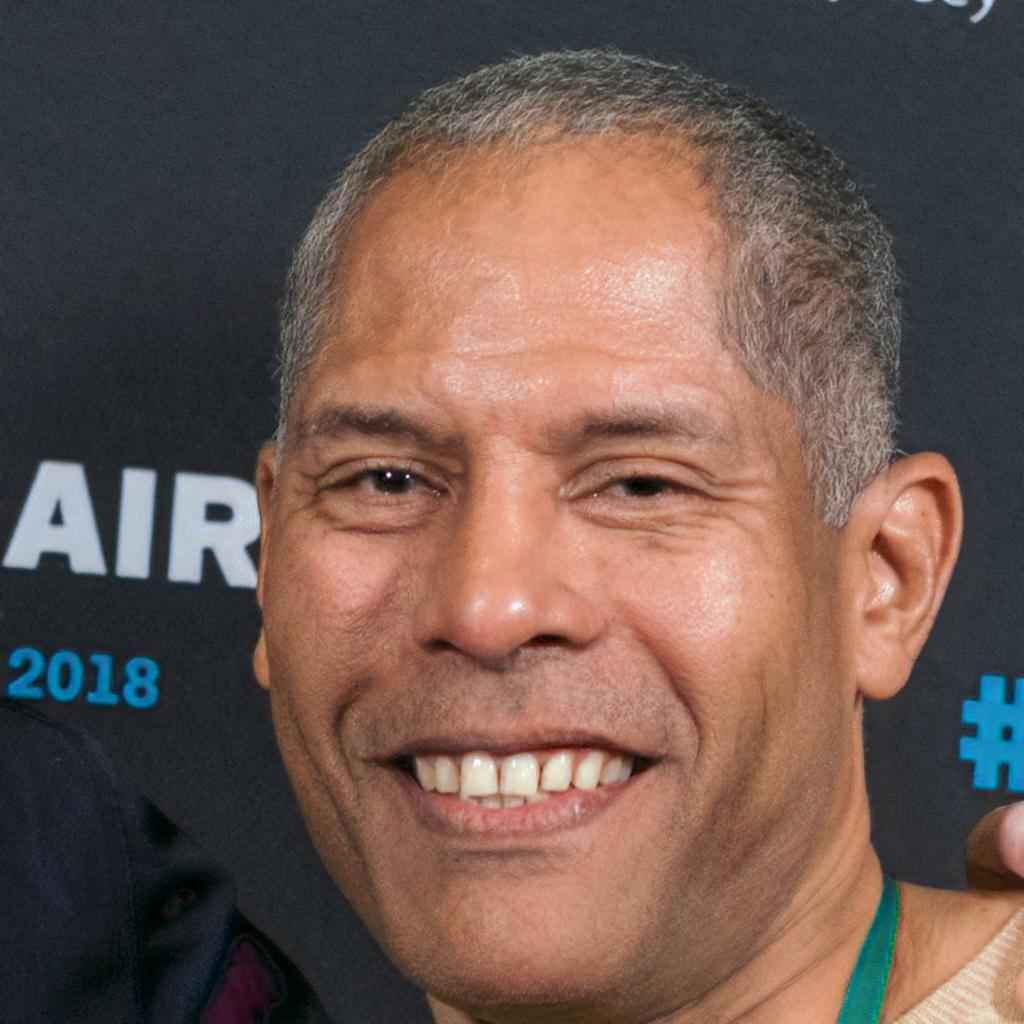} & 
\includegraphics[width=0.19\linewidth]{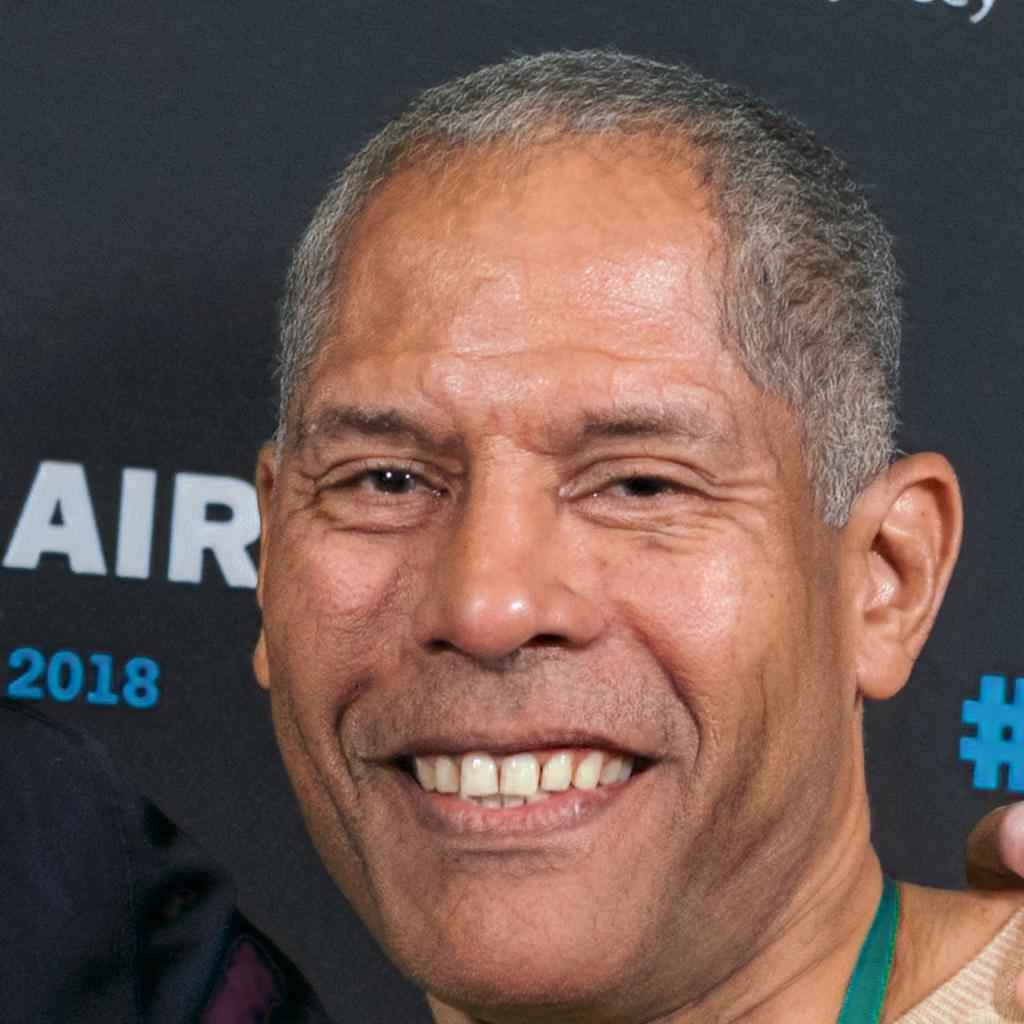} 
\\
\includegraphics[width=0.19\linewidth]{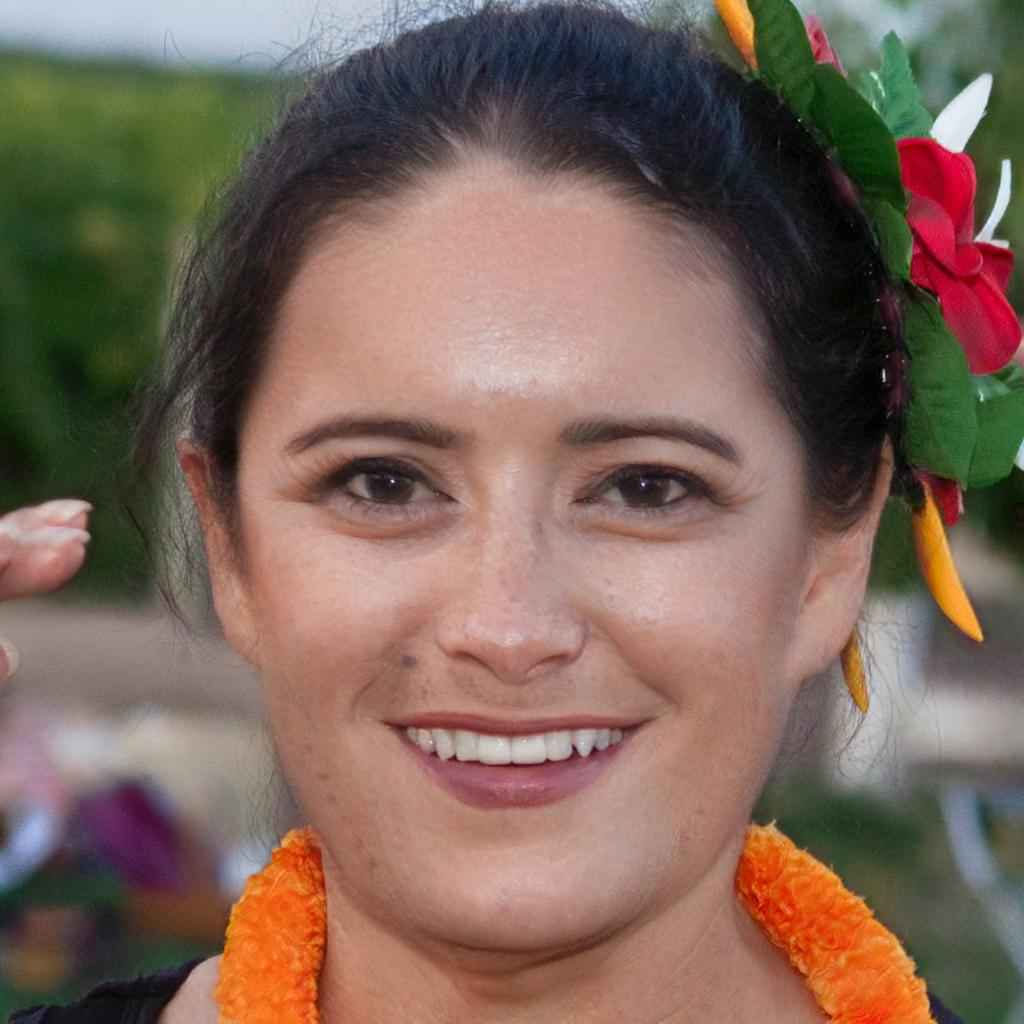} & 
\includegraphics[width=0.19\linewidth]{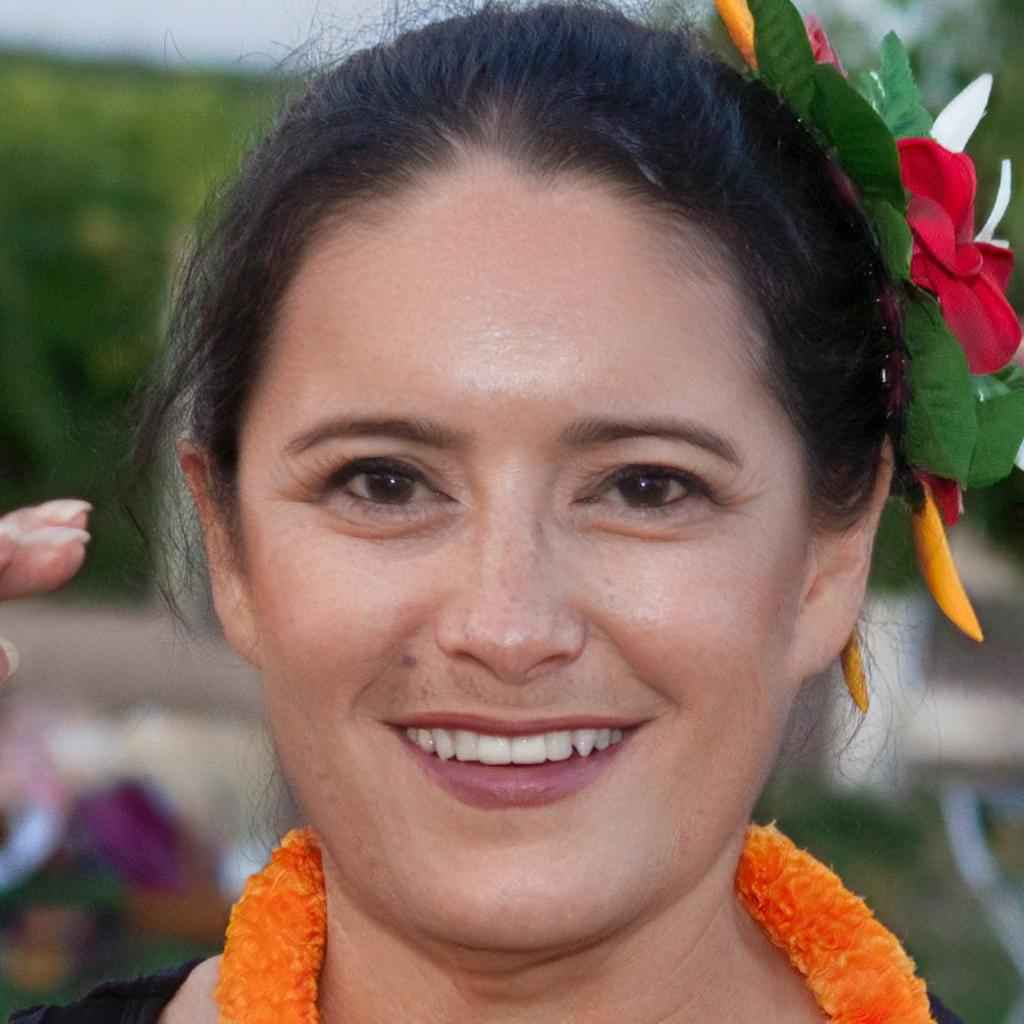} &
\includegraphics[width=0.19\linewidth]{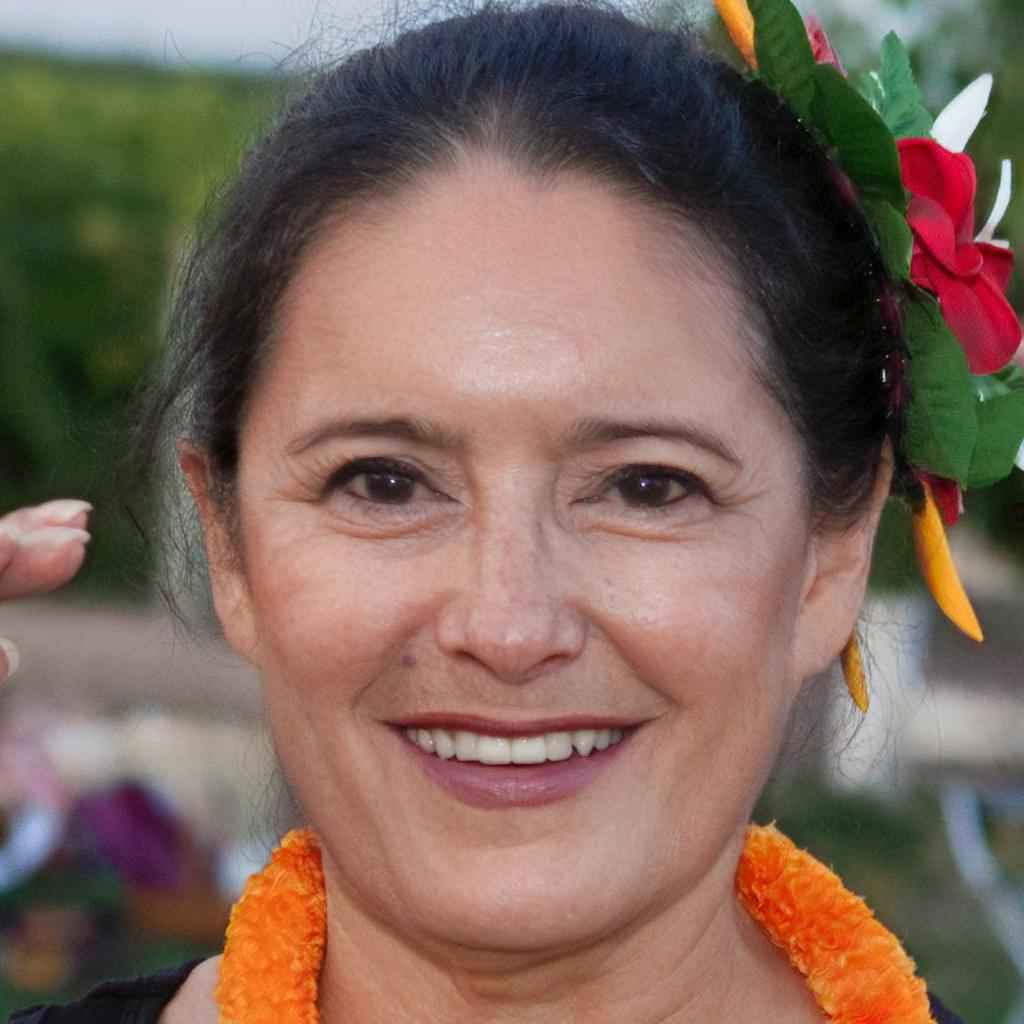} &
{\color{yellow}%
\setlength{\fboxsep}{0pt}%
\setlength{\fboxrule}{2pt}%
\fbox{\includegraphics[width=0.19\linewidth]{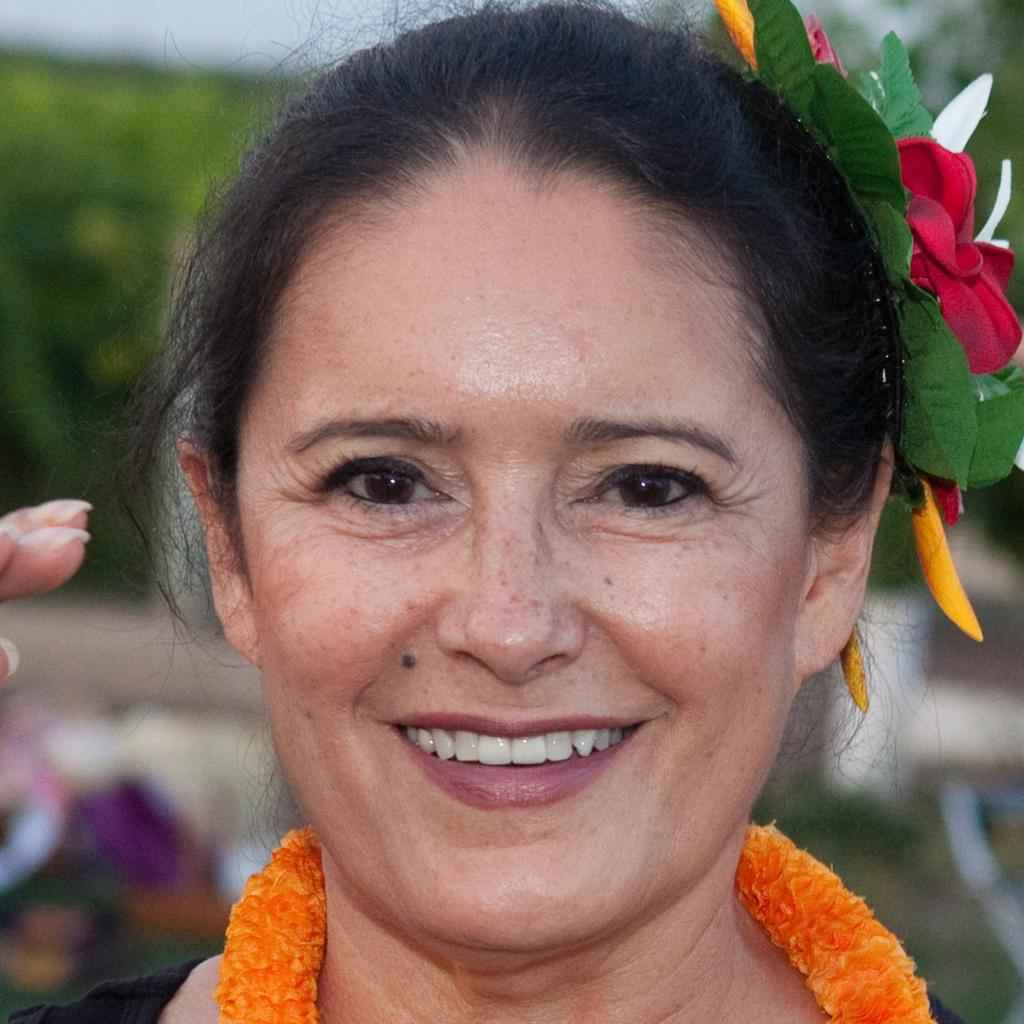}}} &  
\includegraphics[width=0.19\linewidth]{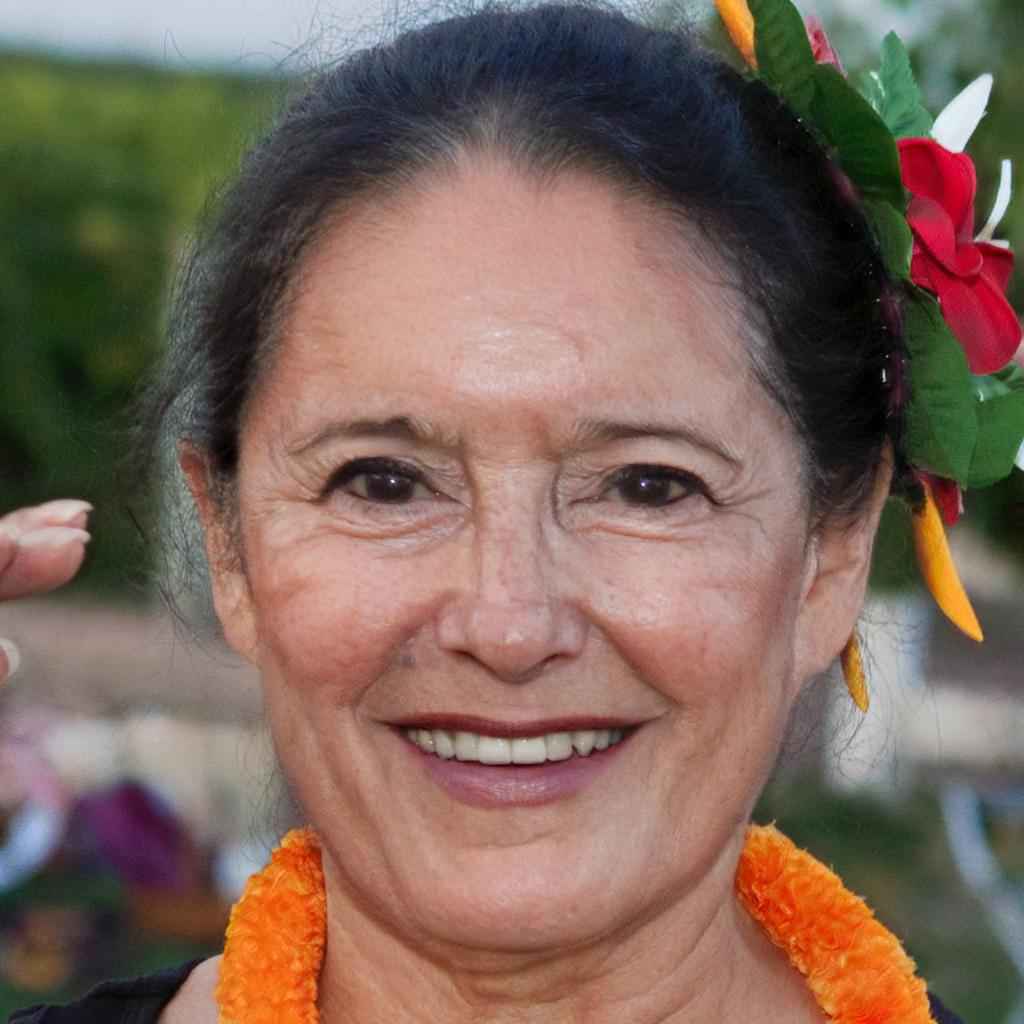} 
\\
\includegraphics[width=0.19\linewidth]{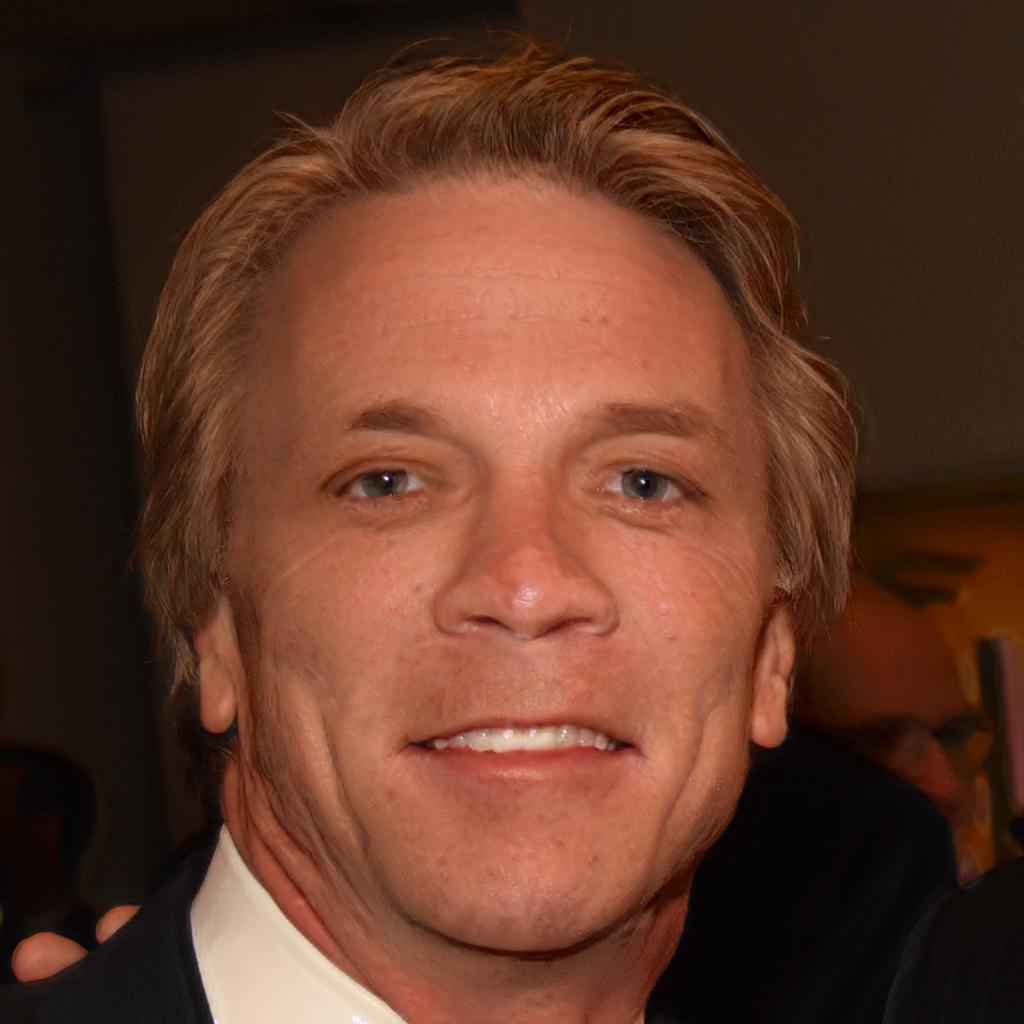} & 
\includegraphics[width=0.19\linewidth]{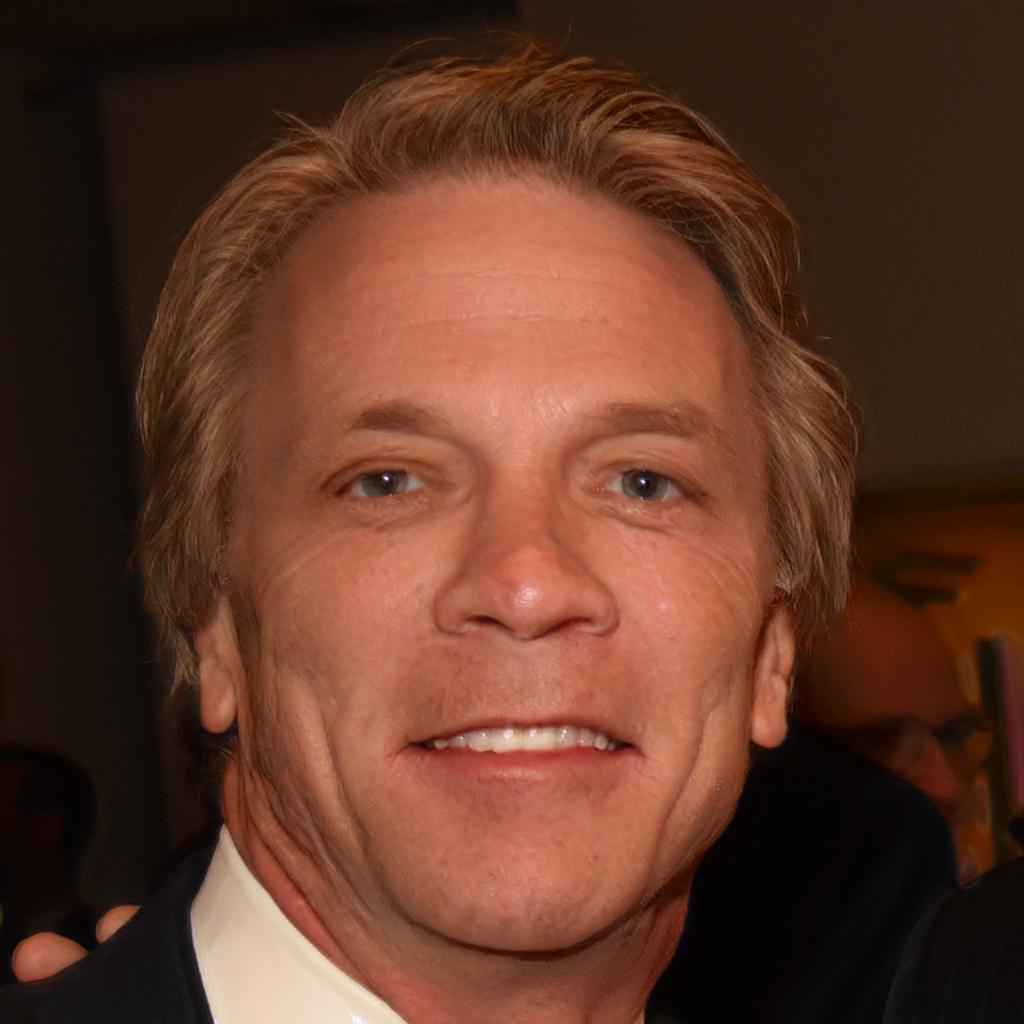} &
\includegraphics[width=0.19\linewidth]{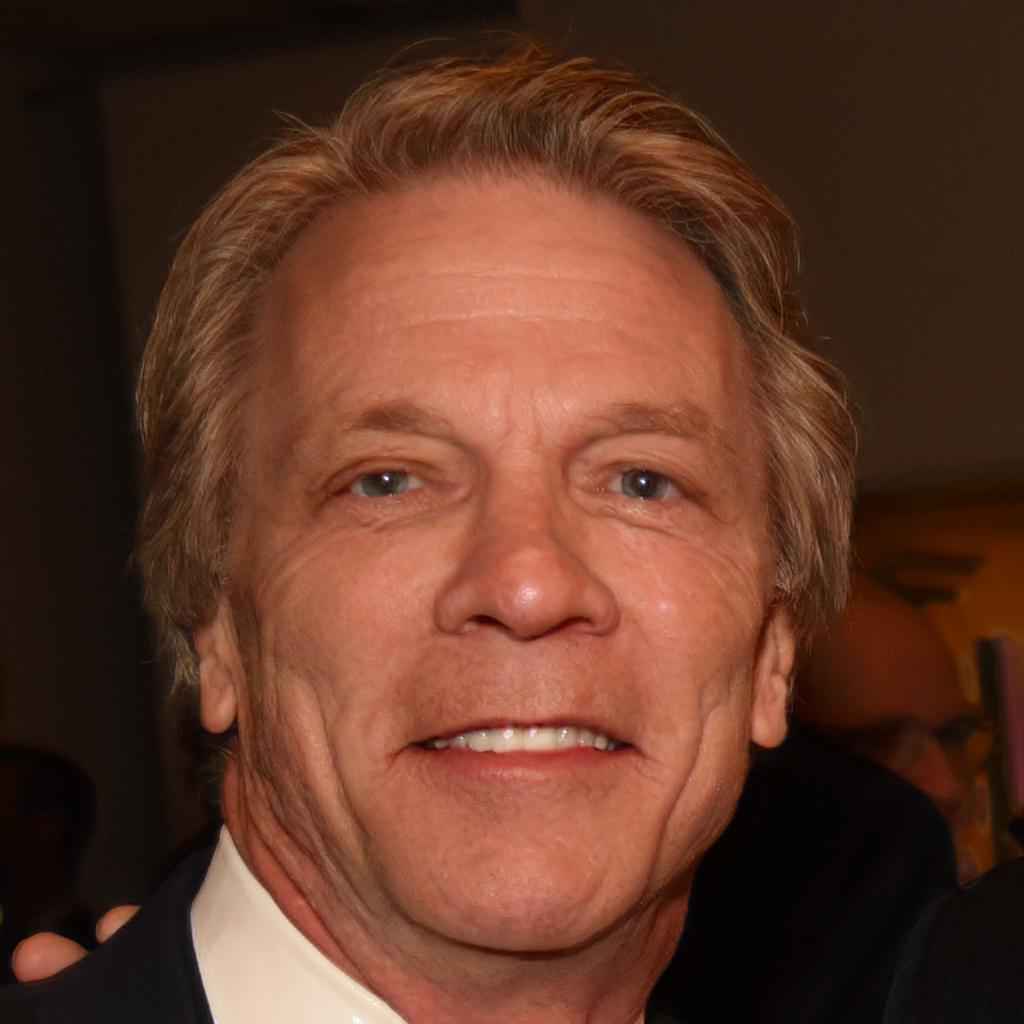} &
\includegraphics[width=0.19\linewidth]{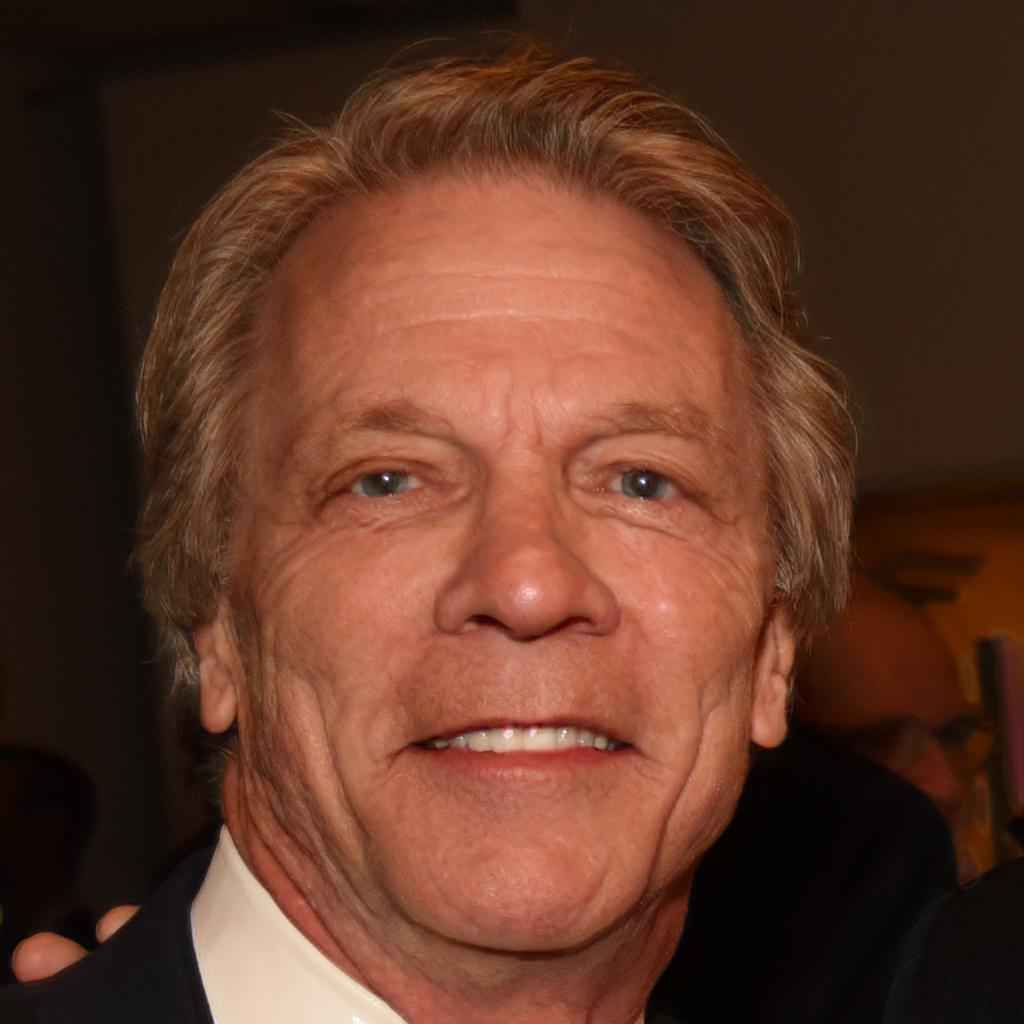} & 
{\color{yellow}%
\setlength{\fboxsep}{0pt}%
\setlength{\fboxrule}{2pt}%
\fbox{\includegraphics[width=0.19\linewidth]{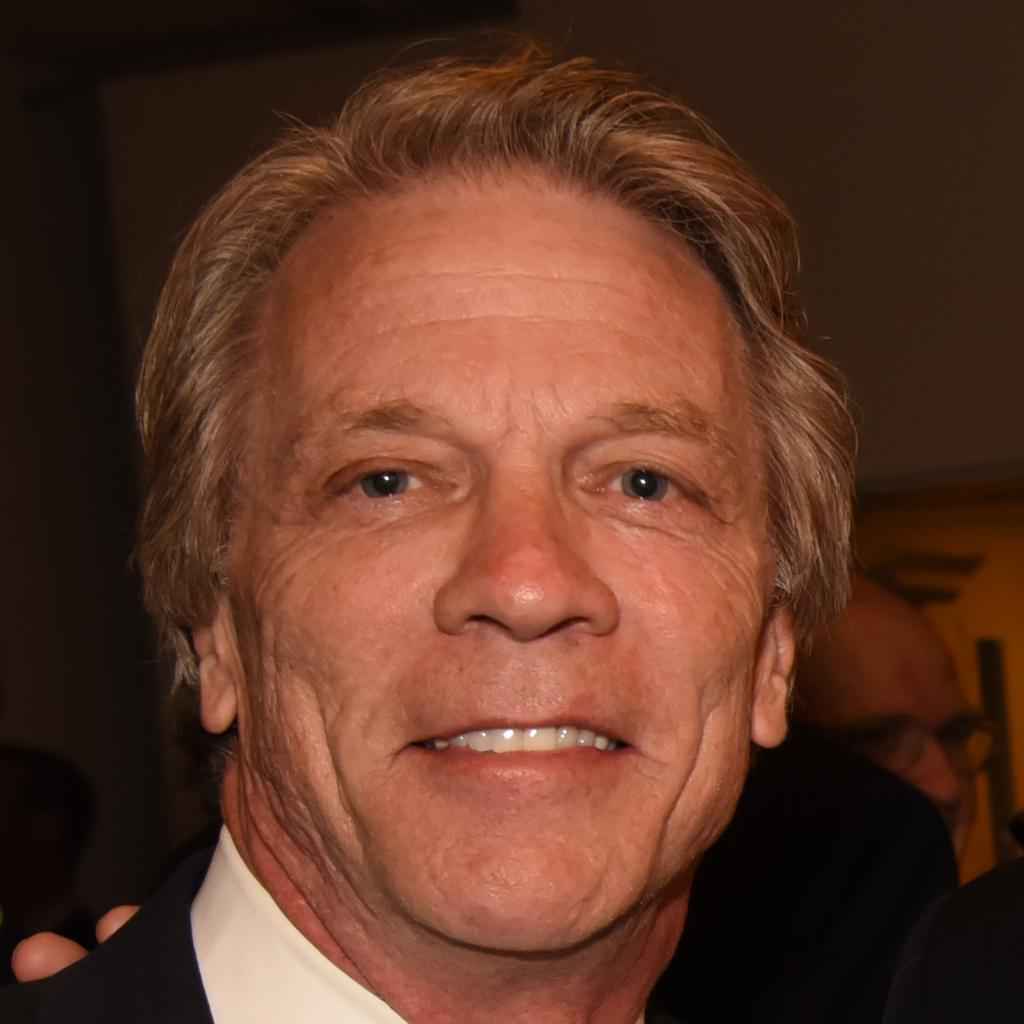}}} 
\end{tabular}
\end{center}
\caption{\textbf{Age transformation on $\bf 1024 \times 1024$ images}. On each row, the yellow frame indicates the original image. Each column corresponds to a target age of: $25$, $35$, $45$, $55$, $65$. Our approach yields visually satisfying results without introducing significant artifacts. Only age relevant features are modified, while the identity, haircut, emotion and background are perfectly preserved. 
}
\label{1024_4}
\end{figure}

\begin{figure*}
\centering
\setlength{\tabcolsep}{2pt}
\begin{tabular}{ccccc}
Input ($1024^2$)&Fader ($256^2$)&PAGGAN ($224^2$)&IPCGAN ($128^2$)&Ours ($1024^2$)\\
\includegraphics[width=0.19\linewidth]{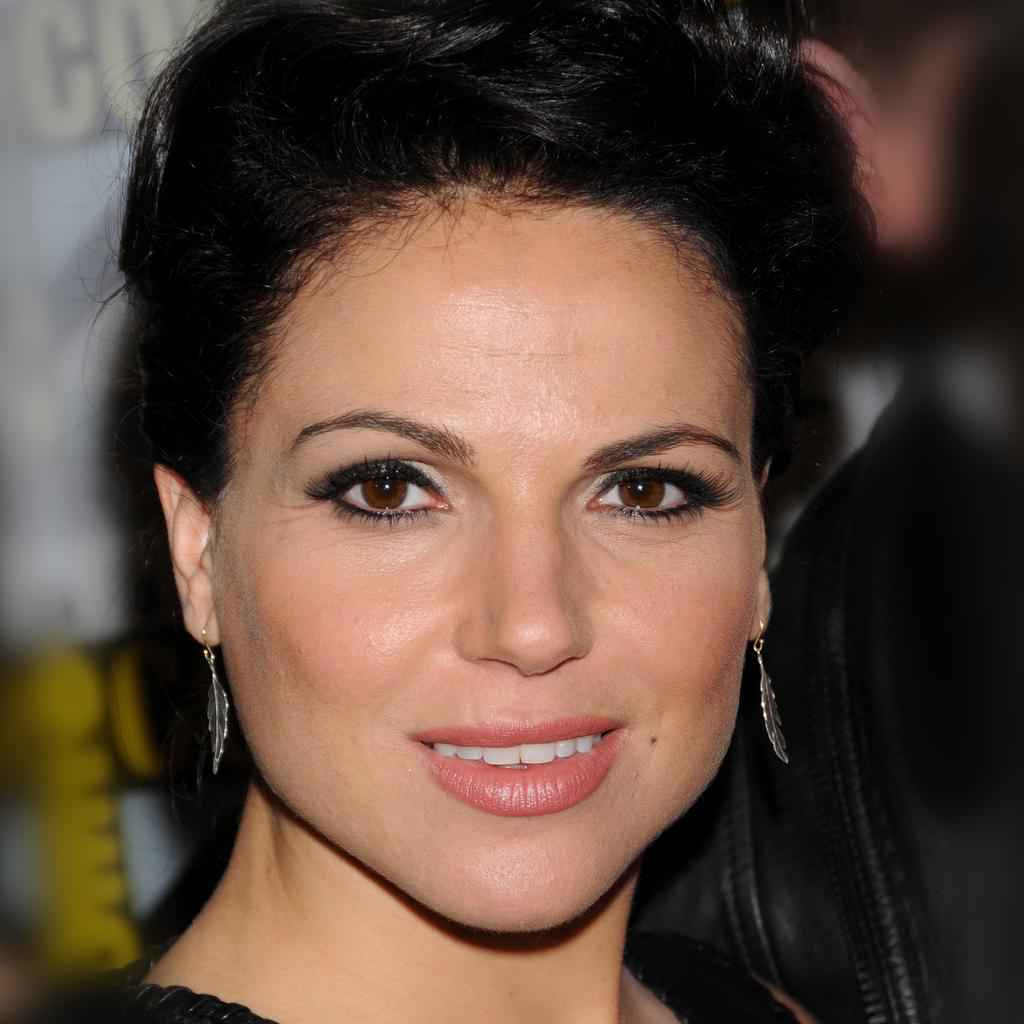} &
\includegraphics[width=0.19\linewidth]{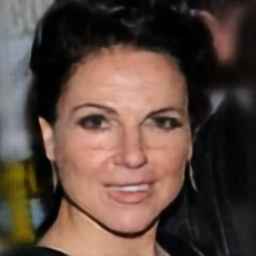} &
\includegraphics[width=0.19\linewidth]{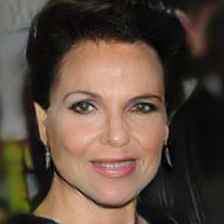} &
\includegraphics[width=0.19\linewidth]{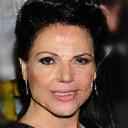} &
\includegraphics[width=0.19\linewidth]{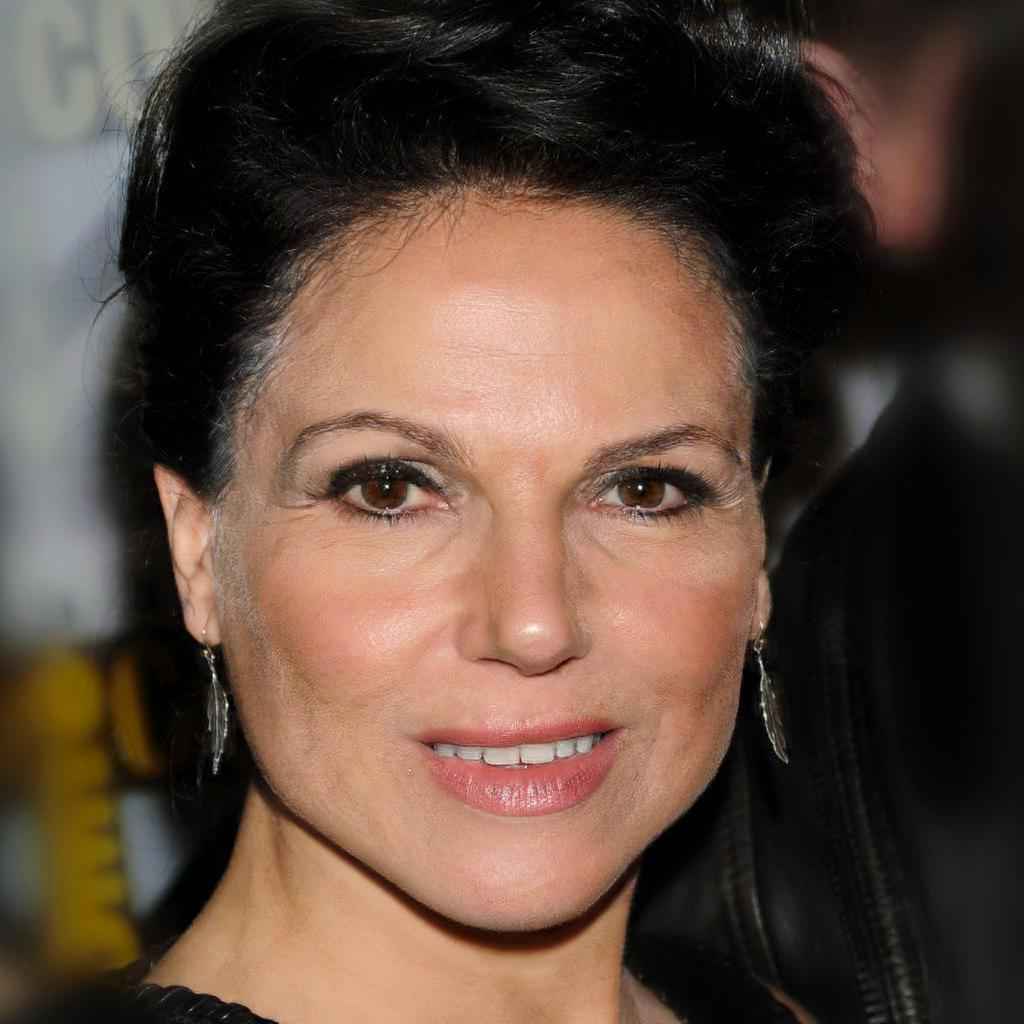} \\
\includegraphics[width=0.19\linewidth]{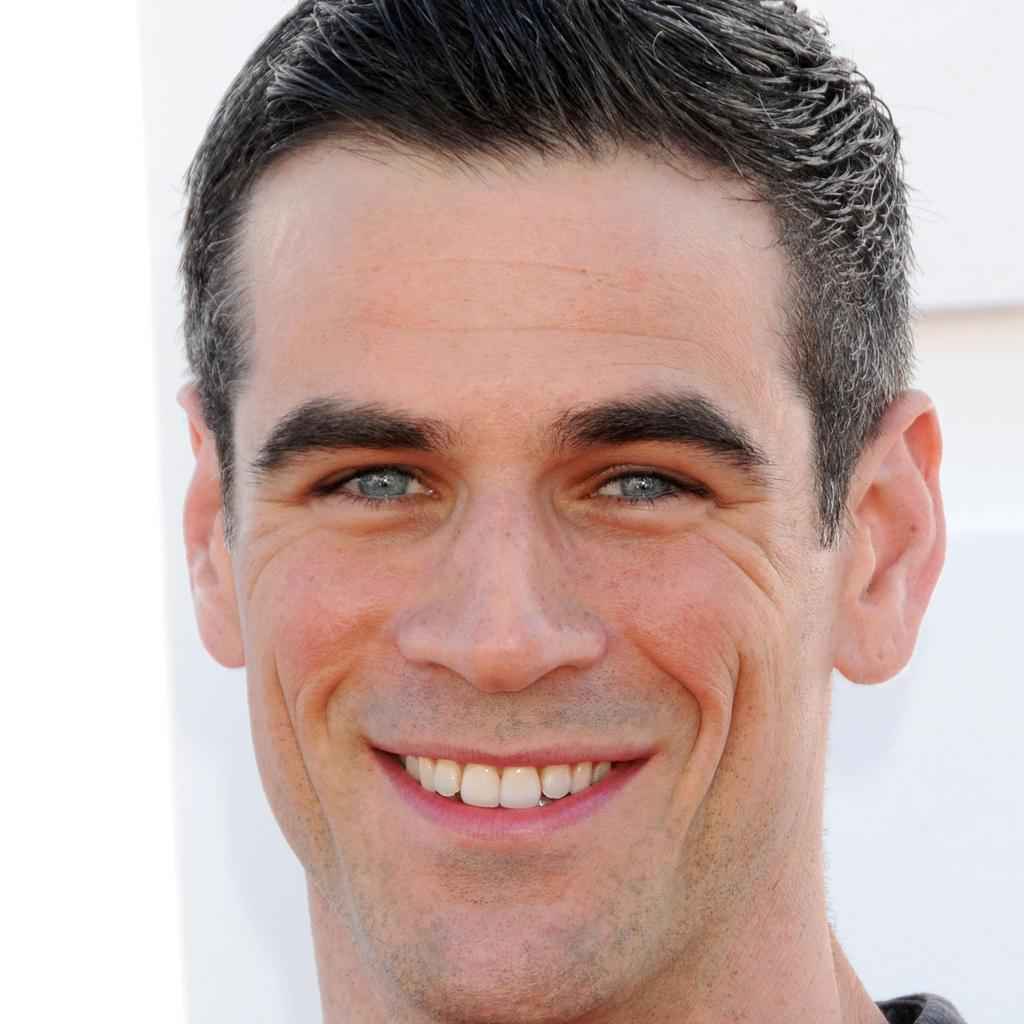} &
\includegraphics[width=0.19\linewidth]{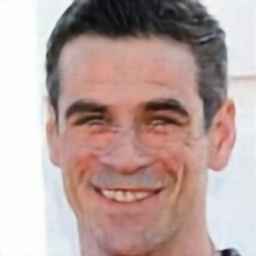} &
\includegraphics[width=0.19\linewidth]{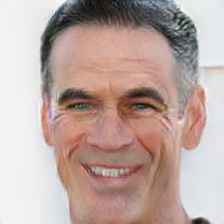} &
\includegraphics[width=0.19\linewidth]{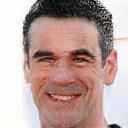} &
\includegraphics[width=0.19\linewidth]{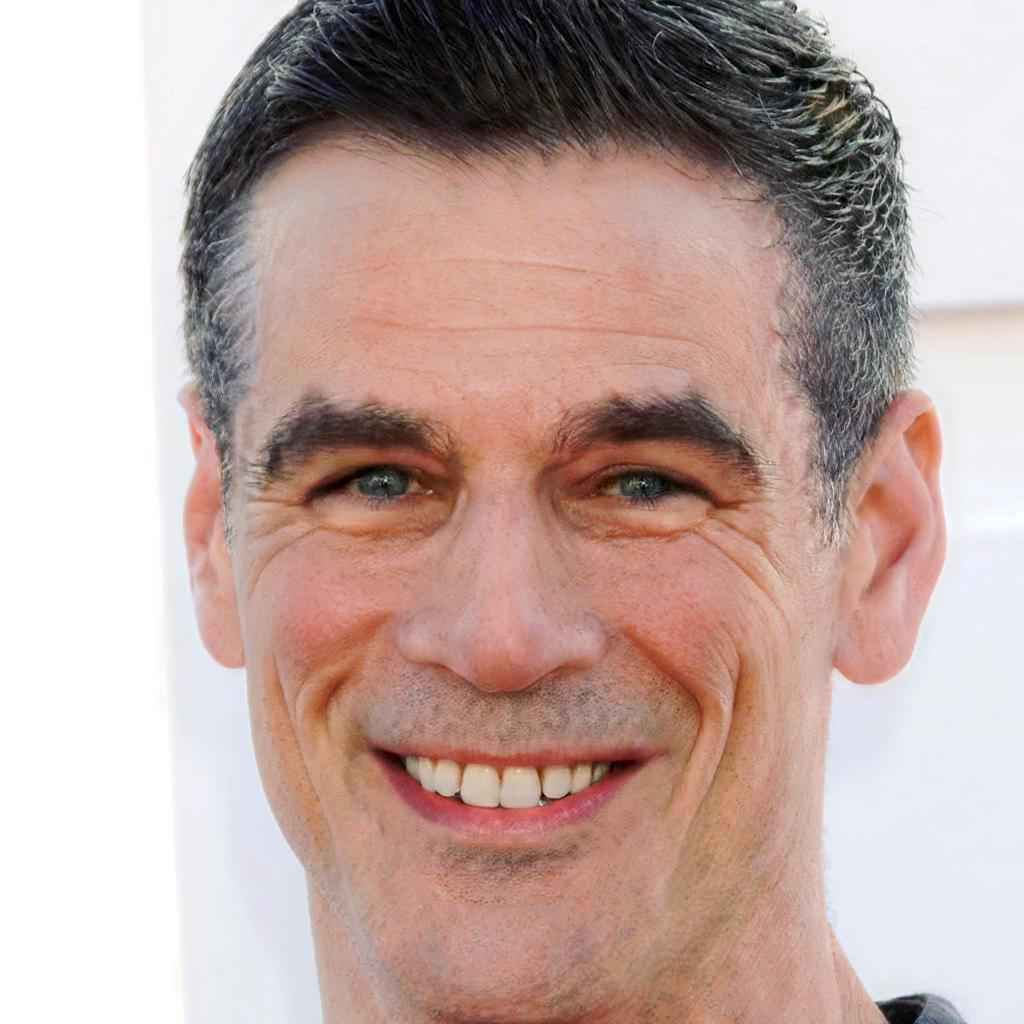} \\
\includegraphics[width=0.19\linewidth]{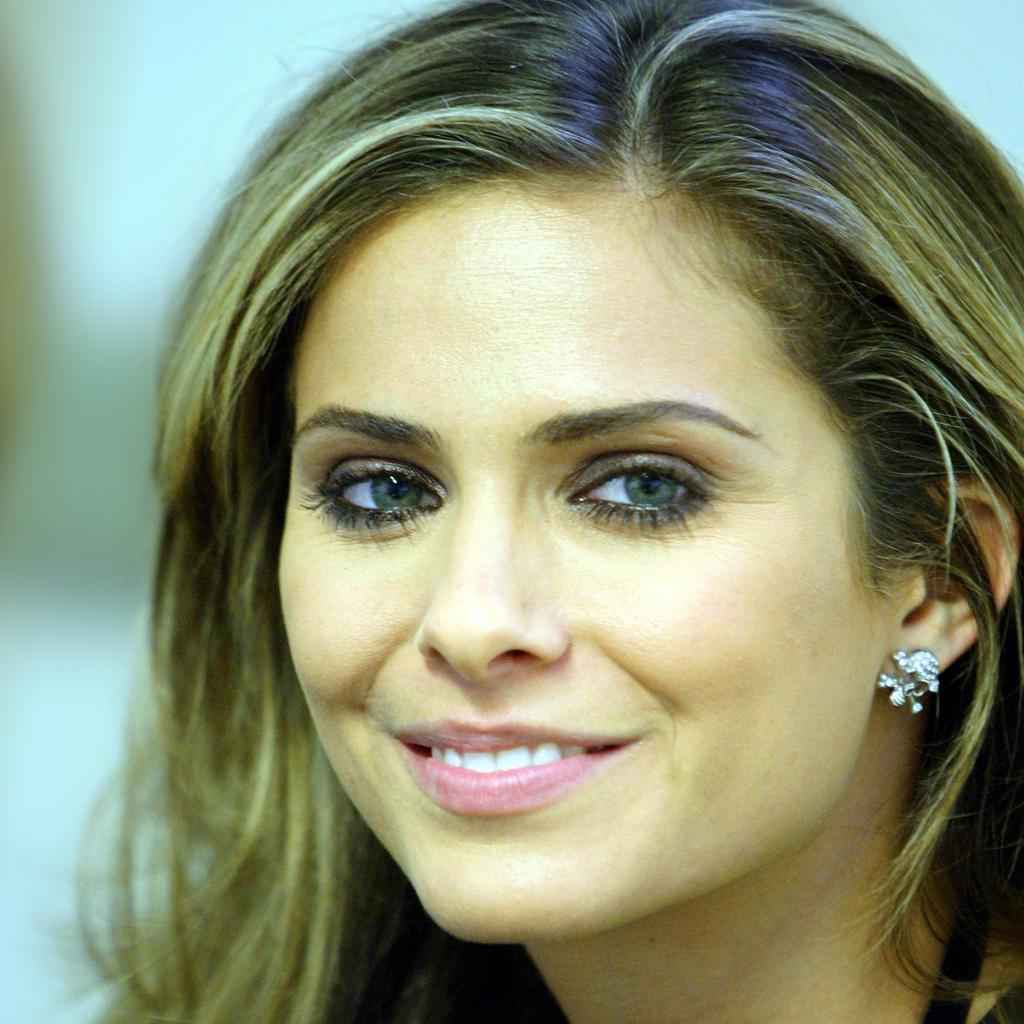} &
\includegraphics[width=0.19\linewidth]{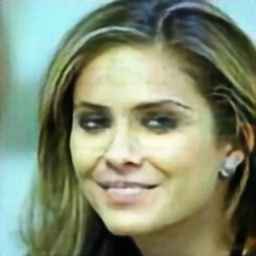} &
\includegraphics[width=0.19\linewidth]{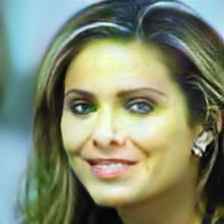} &
\includegraphics[width=0.19\linewidth]{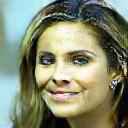} &
\includegraphics[width=0.19\linewidth]{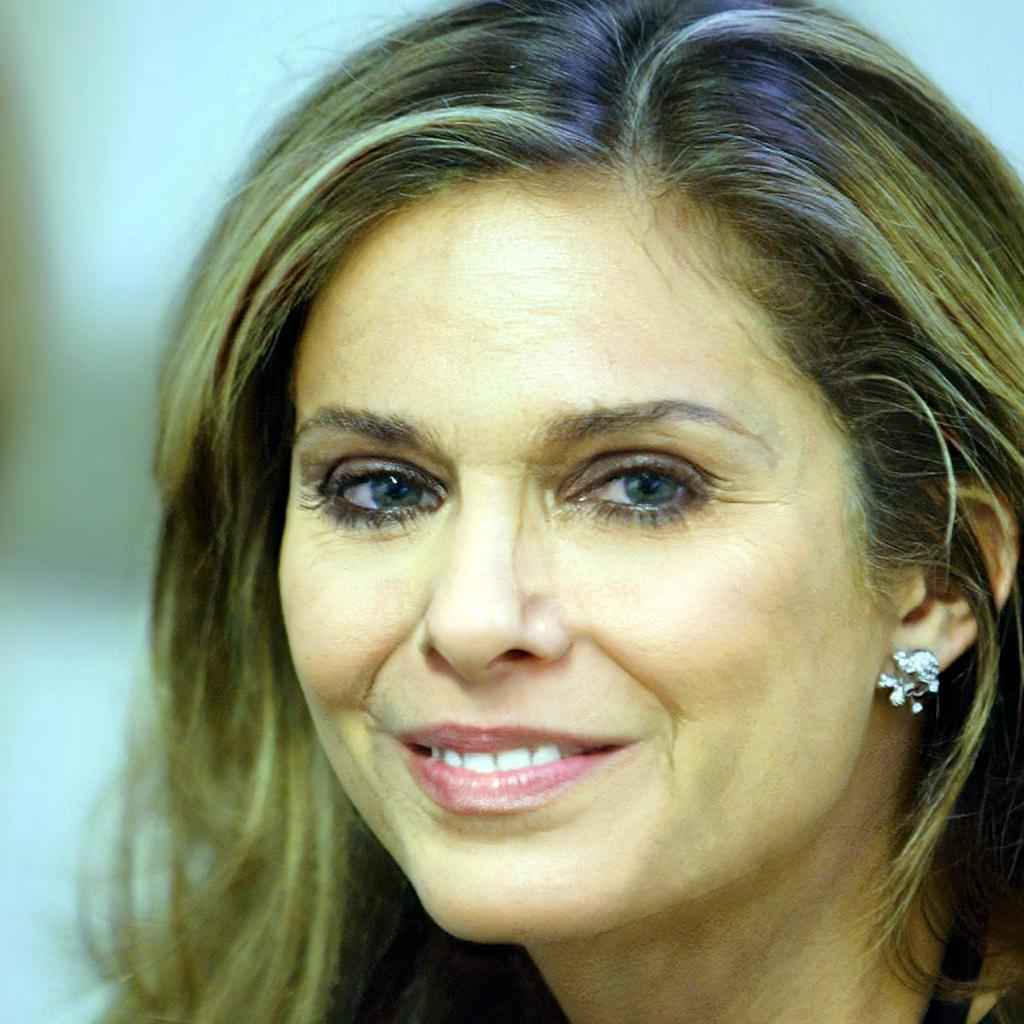} \\
\includegraphics[width=0.19\linewidth]{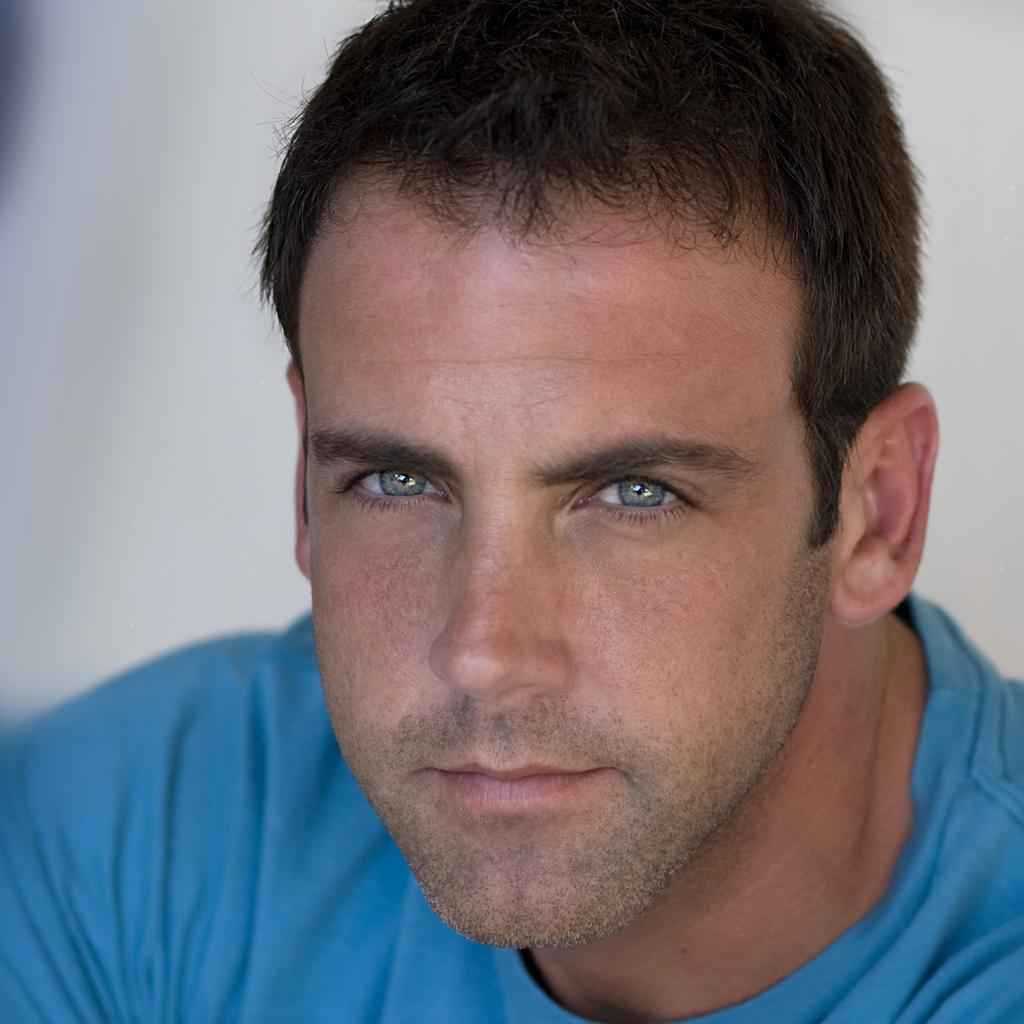} &
\includegraphics[width=0.19\linewidth]{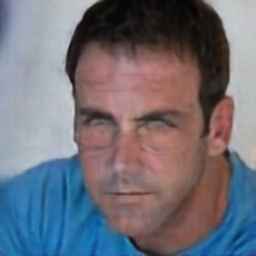} &
\includegraphics[width=0.19\linewidth]{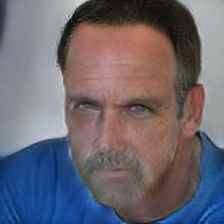} &
\includegraphics[width=0.19\linewidth]{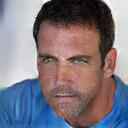} &
\includegraphics[width=0.19\linewidth]{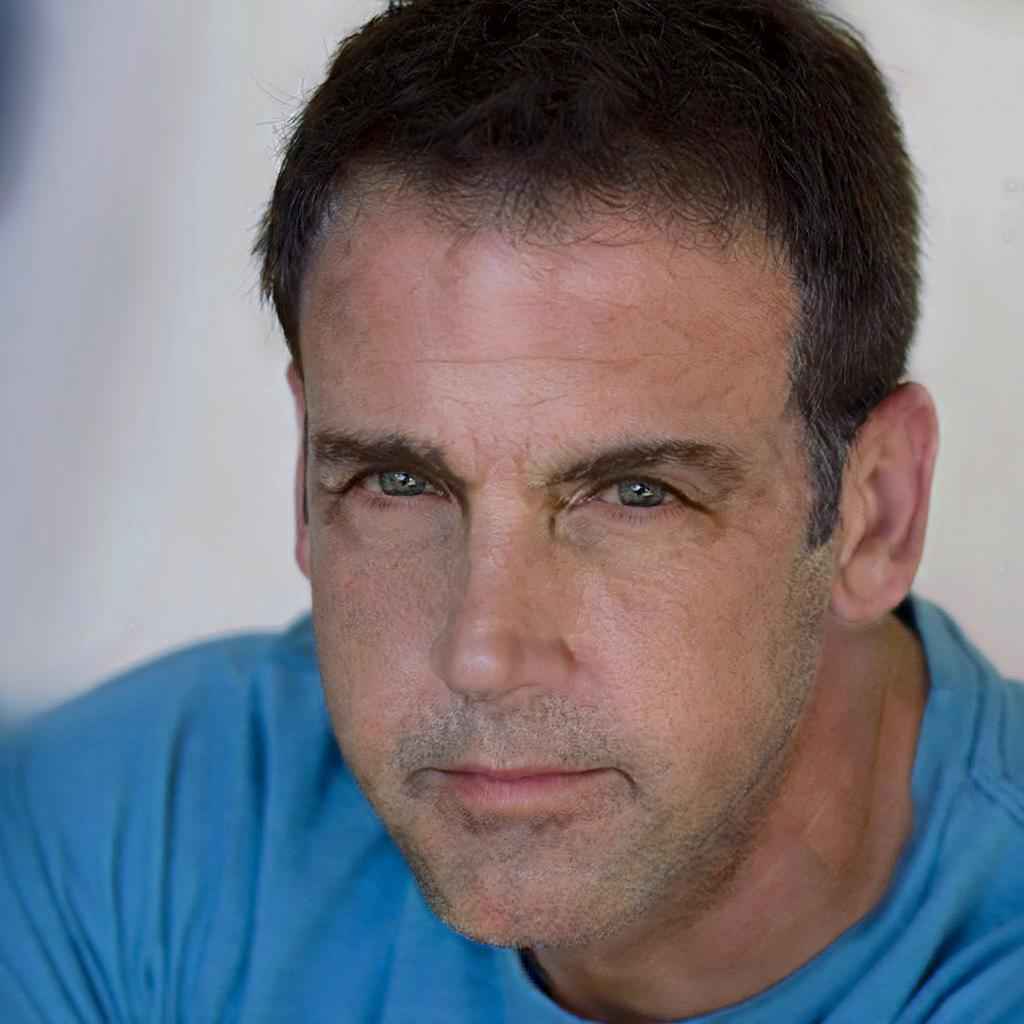} \\
\includegraphics[width=0.19\linewidth]{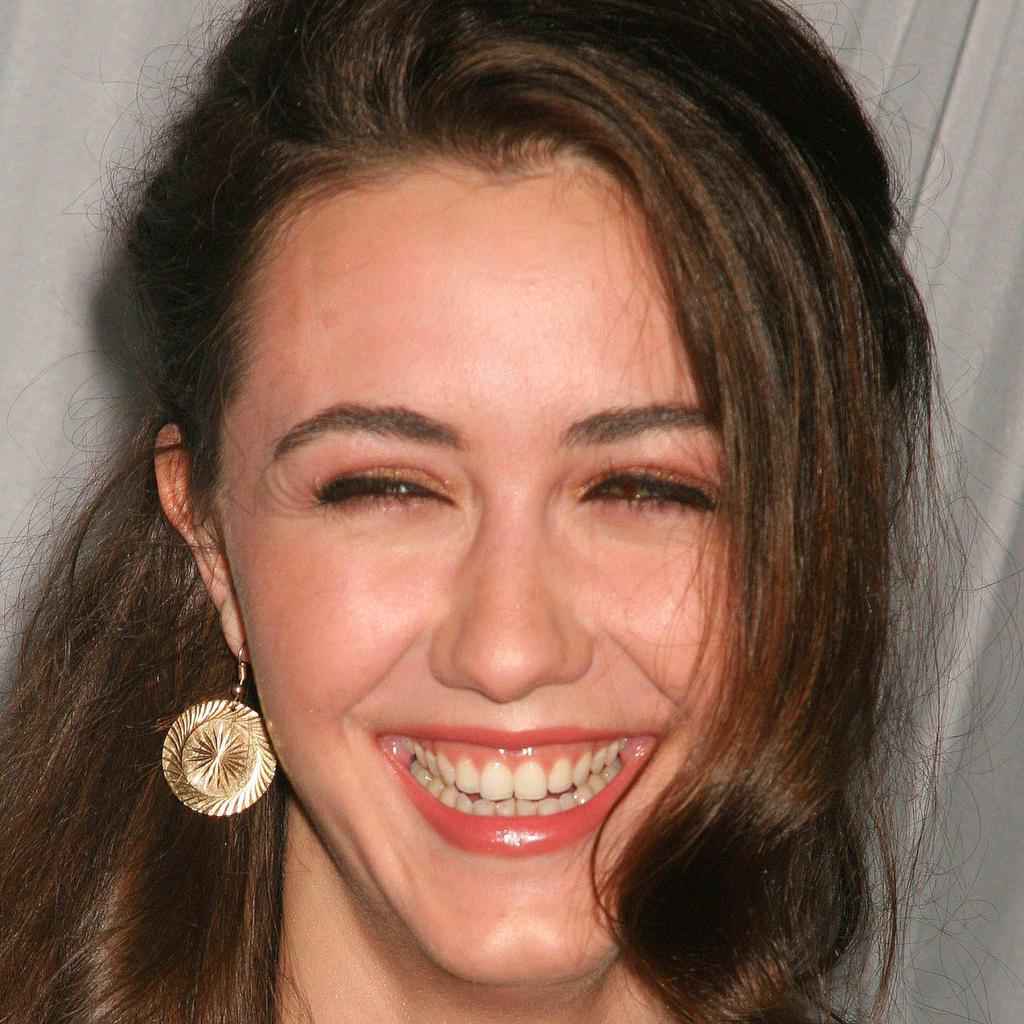} &
\includegraphics[width=0.19\linewidth]{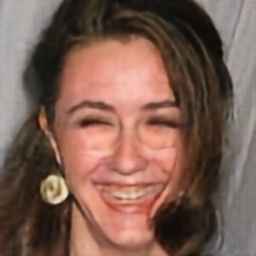} &
\includegraphics[width=0.19\linewidth]{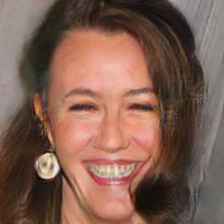} &
\includegraphics[width=0.19\linewidth]{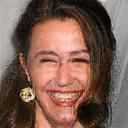} &
\includegraphics[width=0.19\linewidth]{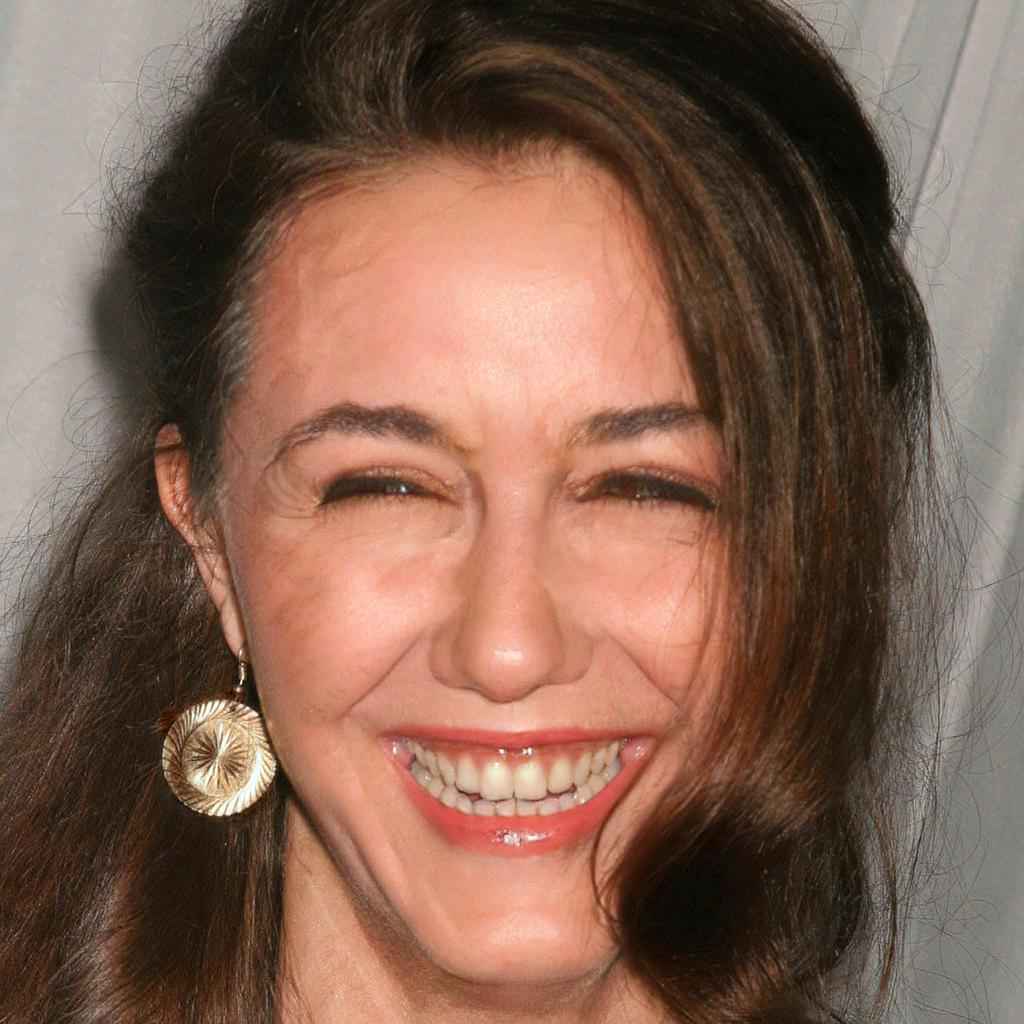} \\
\includegraphics[width=0.19\linewidth]{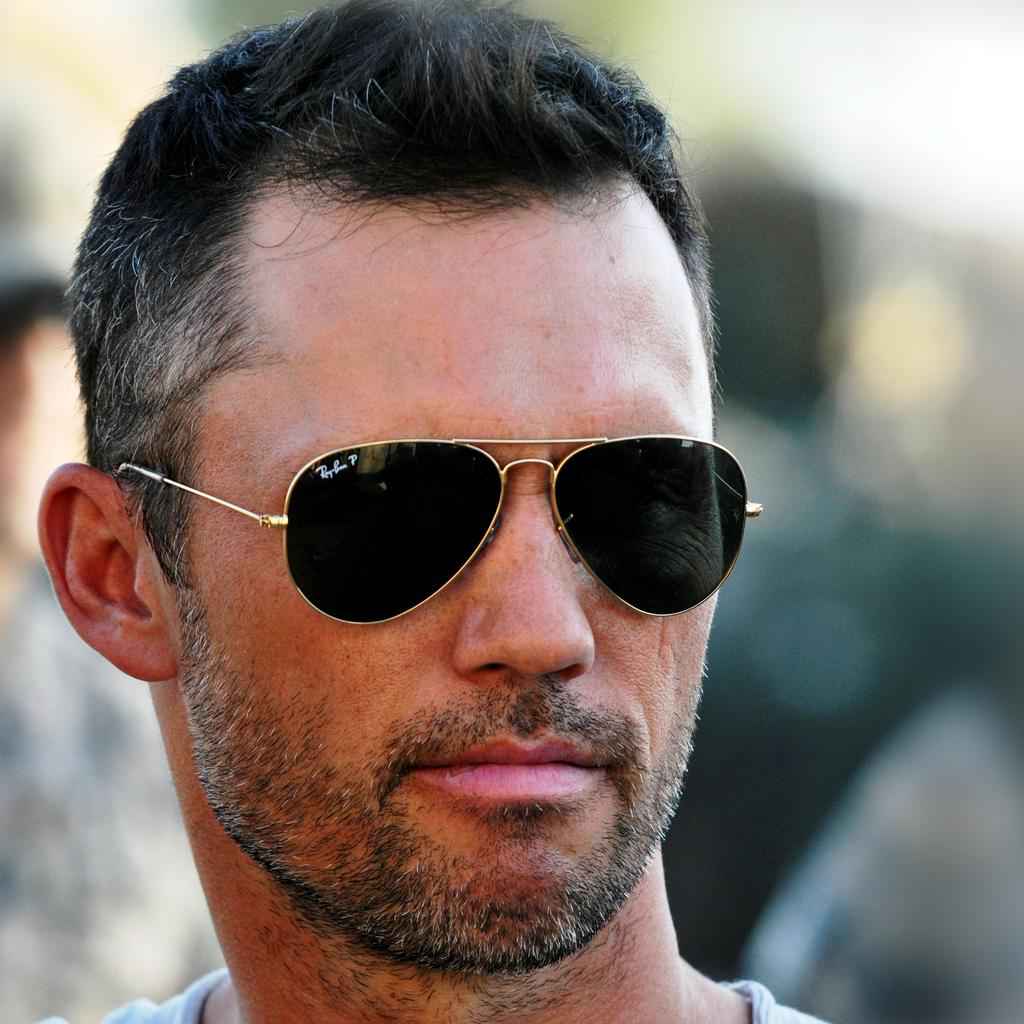} &
\includegraphics[width=0.19\linewidth]{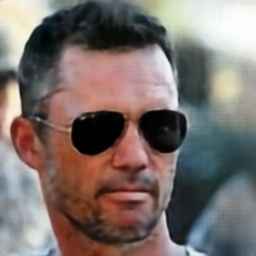} &
\includegraphics[width=0.19\linewidth]{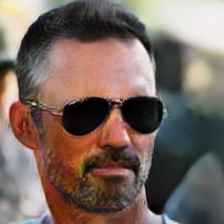} &
\includegraphics[width=0.19\linewidth]{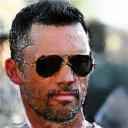} &
\includegraphics[width=0.19\linewidth]{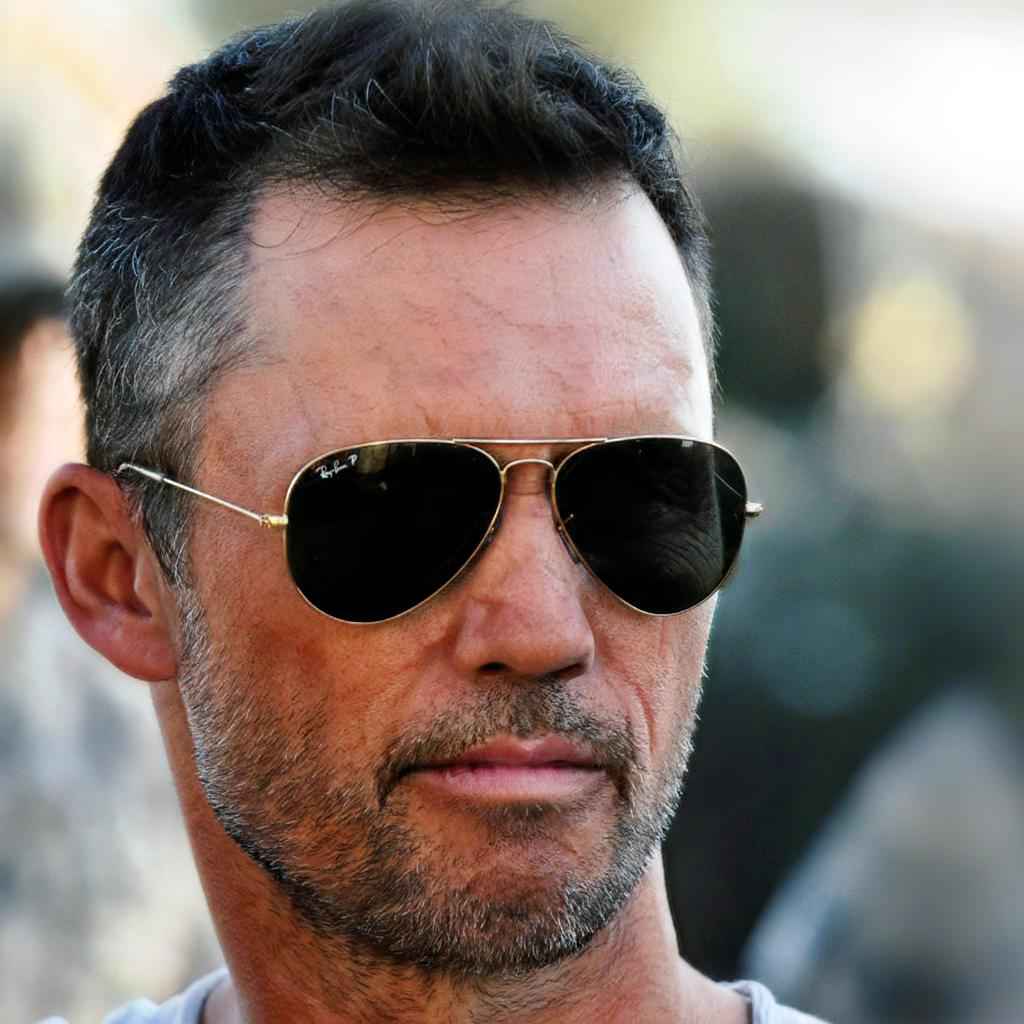} \\
\end{tabular}
\caption{
\textbf{Comparison of face aging results on CelebA HQ \cite{karras2018progressive}}. The first column are the input images. The second to fifth column are outputs from Fader Network \cite{lample2017fader}, PAG-GAN \cite{yang2018learning}, IPC-GAN \cite{wang2018face} and our method. Our results reach the highest resolution without introducing significant artifacts. Our method preserves the background better compared to other techniques, see for instance the letters on the third row. In addition, compared to other techniques, our method leads to  results that are free of artefacts and blur.
}
\label{celeba_2}
\end{figure*}

\end{document}